\documentclass[twocolumn]{svjour3}

\usepackage[latin1]{inputenc}
\usepackage{times}

\usepackage{psfrag, amsmath, amssymb, amscd, amsfonts, latexsym, epsf, graphicx}

\def\bbbr{{\mathbb R}} 
\def\bbbz{{\mathbb Z}} 

\newcommand{\htransf}{H}

\newcommand{\orth}{\bot}

\journalname{Shortened version in Journal of
  Mathematical Imaging and Vision doi:10.1007/s10851-015-0613-9
  Dec 2015}

\begin{document}

\title{\bf Time-causal and time-recursive spatio-temporal receptive fields%
\thanks{The support from the Swedish Research Council 
              (contract 2014-4083) is gratefully acknowledged. }}

\titlerunning{Time-causal and time-recursive spatio-temporal receptive fields}

\author{Tony Lindeberg}

\institute{Tony Lindeberg \at
              Department of Computational Biology,
              School of Computer Science and Communication,
              KTH Royal Institute of Technology,
             SE-100 44 Stockholm, Sweden.
              \email{tony@kth.se}}

\date{Received: date / Accepted: date}

\maketitle

\begin{abstract}
\noindent
We present an improved model and theory for time-causal and
time-recursive spatio-temporal receptive fields,
obtained by a combination of Gaussian receptive fields over the
spatial domain and first-order integrators or equivalently truncated
exponential filters coupled in cascade over the temporal domain.

Compared to previous spatio-temporal scale-space formulations in terms
of non-enhancement of local extrema or scale invariance, these
receptive fields are based on different scale-space axiomatics over
time by ensuring non-creation of new local extrema or
zero-crossings with increasing temporal scale.
Specifically, extensions are presented about: (i)~parameterizing the intermediate
temporal scale levels, (ii)~ana\-lysing the resulting temporal dynamics,
(iii)~transferring the theory to a discrete implementation in terms of 
recursive filters over time,
(iv)~computing scale-normalized spatio-temporal derivative expressions 
for spatio-temporal feature detection and
(v)~computational modelling
of receptive fields in the lateral geniculate nucleus (LGN) and the 
primary visual cortex (V1) in biological vision.

We show that by distributing the intermediate temporal scale levels
according to a logarithmic distribution, we obtain a new family of
temporal scale-space kernels with better temporal characteristics
compared to a more traditional approach of using a uniform
distribution of the intermediate temporal scale levels. 
Specifically, the new family of time-causal kernels has much faster temporal
response properties (shorter temporal delays) compared to the kernels
obtained from a uniform distribution. 
When increasing the number of temporal scale levels, the temporal scale-space kernels in the new family do also converge very rapidly to a limit kernel possessing true self-similar scale invariant properties over temporal scales. Thereby, the new representation allows for true scale invariance over variations in the temporal scale, although the underlying temporal scale-space representation is based on a discretized temporal scale parameter.

We show how scale-normalized temporal derivatives can be defined for these time-causal scale-space kernels and how the composed theory can be used for computing basic types of scale-normalized spatio-temporal derivative expressions in a computationally efficient manner.

\keywords{Scale space \and Receptive field \and Scale \and Spatial \and
  Temporal \and Spatio-temporal \and Scale-normalized derivative \and
  Scale invariance \and Differential invariant \and Natural image transformations \and
Feature detection \and Computer vision \and
Computational modelling \and Biological vision}
\end{abstract}

\section{Introduction}

Spatio-temporal receptive fields constitute an essential concept for
describing neural functions in
biological vision 
(Hubel and Wiesel \cite{HubWie59-Phys,HubWie62-Phys,HubWie05-book}; 
 DeAngelis et al.\ \cite{DeAngOhzFre95-TINS,deAngAnz04-VisNeuroSci})
and for expressing computer vision methods on video data
(Adelson and Bergen \cite{AdeBer85-JOSA}; 
 Zelnik-Manor and Irani \cite{ZelIra01-CVPR}; 
 Laptev and Lindeberg \cite{LapLin04-ECCVWS};
 Jhuang et al.\ \cite{JhuSerWolPog07-ICCV};
 Shabani et al.\ \cite{ShaClaZel12-BMVC}).

For off-line processing of pre-recorded video, non-causal Gaussian or
Gabor-based spatio-temporal receptive
fields may in some cases be sufficient.
When operating on video data in a real-time setting or when modelling
biological vision computationally, one does however need to take into explicit
account the fact that the future cannot be accessed and that the 
underlying spatio-temporal receptive fields must therefore be
{\em time-causal\/}, {\em i.e.\/}, the image operations should only
require access to image data from the
present moment and what has occurred in the past.
For computational efficiency and for keeping down memory requirements,
it is also desirable that the computations should be {\em time-recursive\/},
so that it is sufficient to keep a limited memory of the past that can
be recursively updated over time.

The subject of this article is to present an improved temporal scale-space
model for spatio-temporal receptive fields based on time-causal
temporal scale-space kernels in terms of first-order integrators
or equivalently truncated exponential filters coupled in cascade, 
which can be transferred to a discrete
implementation in terms of recursive filters over discretized time.
This temporal scale-space model will then be combined with a Gaussian
scale-space concept over continuous image space or a genuinely discrete
scale-space concept over discrete image space, resulting in both
continuous and discrete spatio-temporal scale-space concepts for
modelling time-causal and time-recursive spatio-temporal receptive
fields over both continuous and discrete spatio-temporal domains.
The model builds on previous work by 
(Fleet and Langley \cite{FleLan95-PAMI};
 Lindeberg and Fagerstr{\"o}m \cite{LF96-ECCV};
 Lindeberg \cite{Lin10-JMIV,Lin13-BICY,Lin13-ImPhys}) and is here complemented by:
(i)~a better design for the degrees of freedom in the choice of time constants for the intermediate
temporal scale levels from the original signal to any higher
temporal scale level in a cascade structure of temporal scale-space
representations over multiple temporal scales, (ii)~an analysis of the resulting temporal response dynamics,
(iii)~details for discrete implementation in a spatio-temporal visual
front-end,
(iv)~details for computing spatio-temporal image features in terms of scale-normalized spatio-temporal
differential expressions at different spatio-temporal scales and
(v) computational modelling of receptive fields in the
lateral geniculate nucleus (LGN) and the primary visual cortex (V1) in
biological vision.

In previous use of the temporal scale-space model in
(Lindeberg and Fagerstr{\"o}m \cite{LF96-ECCV}), 
a uniform distribution of the intermediate scale levels has mostly
been chosen
when coupling first-order integrators or equivalently truncated exponential kernels in cascade.
By instead using a logarithmic distribution of the intermediate scale levels,
we will here show that a new family of temporal scale-space kernels can
be obtained with much better properties in terms of:
(i)~faster temporal response dynamics and
(ii)~fast convergence towards a limit kernel that possesses true 
scale-invariant properties (self-similarity) under variations in
the temporal scale in the input data.
Thereby, the new family of kernels enables: (i)~significantly shorter temporal delays 
(as always arise for truly time-causal operations),
(ii)~much better computational approximation to true temporal scale invariance
and (iii)~computationally much more efficient numerical implementation.
Conceptually, our approach is also related to the
time-causal scale-time model by Koenderink \cite{Koe88-BC}, which is here
complemented by a truly time-recursive formulation of time-causal receptive
fields more suitable for real-time operations over a compact temporal
buffer of what has occurred in the past, including a theoretically
well-founded and computationally efficient method for discrete implementation.

Specifically, the rapid convergence of the new family of temporal scale-space kernels 
to a limit kernel when the number of intermediate temporal scale levels tends to infinity 
is theoretically very attractive, since it provides a way to define truly scale-invariant 
operations over temporal variations at different temporal scales, and to measure the 
deviation from true scale invariance when approximating the limit kernel by a finite
number of temporal scale levels.
Thereby, the proposed model allows for truly self-similar temporal operations 
over temporal scales while using a discretized temporal scale parameter, which
is a theoretically new type of construction for temporal scale spaces.

Based on a previously established analogy between scale-normalized derivatives for spatial derivative expressions 
and the interpretation of scale normalization of the corresponding Gaussian derivative kernels to 
constant $L_p$-norms over scale (Lindeberg \cite{Lin97-IJCV}), we will show how scale-invariant temporal derivative
operators can be defined for the proposed new families of temporal scale-space kernels.
Then, we will apply the resulting theory for computing basic spatio-temporal derivative
expressions of different types and describe classes of such spatio-temporal derivative
expressions that are invariant or covariant to basic types of natural image transformations,
including independent rescaling of the spatial and temporal coordinates, illumination variations
and variabilities in exposure control mechanisms.

In these ways, the proposed theory will present previously missing components for applying
scale-space theory to spatio-temporal input data (video) 
based on truly time-causal and time-recursive image operations.

A conceptual difference between the time-causal temporal scale-space
model that is developed in this paper and Koenderink's
fully continuous scale-time model \cite{Koe88-BC} or the fully
continuous time-causal semi-group derived by Fagerstr{\"o}m
\cite{Fag05-IJCV} and Lindeberg \cite{Lin10-JMIV} is that the
presented time-causal scale-space model will be semi-discrete, with a
continuous time axis and discretized temporal scale parameter.
This semi-discrete theory can then be further discretized over
time (and for spatio-temporal image data also over space) into a fully
discrete theory for digital implementation.
The reason why the temporal scale parameter has to be discrete in
this theory is that
according to theoretical results about variation-diminishing
linear transformations by Schoenberg
\cite{Sch30,Sch46,Sch47,Sch48,Sch50,Sch53,Sch88-book}
and Karlin \cite{Kar68} that we will build upon,
there is no continuous parameter semi-group
structure or continuous parameter cascade structure that guarantees
non-creation of new structures with increasing temporal scale in terms of non-creation of new
local extrema or new zero-crossings over a continuum of increasing temporal scales.

When discretizing the temporal scale parameter into a discrete set
  of temporal scale levels, we do however show that there exists such a discrete parameter 
 semi-group structure in the case of a uniform distribution of the
  temporal scale levels and a discrete parameter cascade
  structure in the case of a logarithmic distribution of the temporal
scale levels, which both guarantee non-creation of new local extrema or
zero-crossings with increasing temporal scale.
In addition, the presented semi-discrete theory allows for an
efficient time-recursive formulation for real-time implementation based on a compact temporal
buffer, which Koenderink's scale-time model \cite{Koe88-BC} does not,
and much better temporal dynamics than the time-causal
semigroup previously derived by Fagerstr{\"o}m
\cite{Fag05-IJCV} and Lindeberg \cite{Lin10-JMIV}.

Specifically, we argue that if the goal is to construct a vision
system that analyses continuous video streams in real time,
as is the main scope of this work,
a restriction of the theory to a discrete set of temporal scale
levels with the temporal scale levels determined in advance before the image data
are sampled over time is less of a practical constraint, since the
vision system anyway has to be based on a finite amount of sensors and
hardware/wetware for sampling and processing the continuous stream of image data.

\subsection{Structure of this article}

To give the contextual overview to this work, section~\ref{sec-spat-temp-RF} 
starts by presenting a previously established computational model for
spatio-temporal receptive fields in terms of spatial and temporal scale-space kernels,
based on which we will replace the temporal smoothing step.

Section~\ref{sec-time-caus-scale-spaces} starts by reviewing 
previously theoretical results for temporal scale-space models
based on the assumption of non-creation of new local extrema 
with increasing scale, showing that the canonical temporal operators
in such a model are first-order integrators or equivalently truncated
exponential kernels coupled in cascade.
Relative to previous applications of this idea based on a uniform distribution 
of the intermediate temporal scale levels, we present a conceptual extension
of this idea based on a logarithmic distribution of the intermediate temporal
scale levels, and show that this leads to a new family of kernels 
that have faster temporal response properties and correspond to more skewed distributions with
the degree of skewness determined by a distribution parameter $c$.

Section~\ref{app-temp-dyn} analyses the temporal characteristics of these kernels
and shows that they lead to faster temporal characteristics in terms of shorter temporal delays,
including how the choice of distribution parameter $c$ affects these characteristics.
In section~\ref{sec-time-caus-limit-kernel} we present a more detailed analysis
of these kernels, with emphasis on the limit case when the number of intermediate
scale levels $K$ tends to infinity, and making constructions that lead
to true self-similarity and scale invariance over a discrete set of
temporal scaling factors.

Section~\ref{sec-comp-impl} shows how these spatial and temporal kernels 
can be transferred to a discrete implementation while preserving scale-space
properties also in the discrete implementation and allowing for efficient
computations of spatio-temporal derivative approximations.
Section~\ref{sec-sc-norm-spat-temp-der} develops a model for defining
scale-normalized derivatives for the proposed temporal scale-space kernels, 
which also leads to a way of measuring
how far from the scale-invariant time-causal limit kernel a particular temporal scale-space
kernel is when using a finite number $K$ of temporal scale levels.

In section~\ref{sec-spat-temp-feat} we combine these components for
computing spatio-temporal features defined from different types of
spatio-temporal differential invariants, including an analysis of their
invariance or covariance properties under natural image transformations,
with specific emphasis on independent scalings of the spatial and 
temporal dimensions, illumination variations and variations in
exposure control mechanisms.
Finally, section~\ref{sec-sum-disc} concludes with a summary and discussion,
including a description about relations and differences to other temporal scale-space models.

To simplify the presentation, we have put some of the theoretical analysis in the appendix.
Appendix~\ref{app-freq-anal-time-caus-kern} presents a frequency analysis of the 
proposed time-causal scale-space kernels, including a detailed characterization of
the limit case when the number of temporal scale levels $K$ tends to infinity
and explicit expressions their moment (cumulant) descriptors up to order four.
Appendix~\ref{app-comp-koe-model} presents a comparison with the
temporal kernels in Koenderink's scale-time model, including a minor
modification of Koenderink's model to make the temporal kernels normalized to 
unit $L_1$-norm and a mapping between the parameters in his model
(a temporal offset $\delta$ and a dimensionless amount of smoothing $\sigma$
relative to a logarithmic time scale) and the parameters in our model
(the temporal variance $\tau$, a distribution parameter $c$ and the number of temporal scale levels $K$)
including graphs of similarities {\em vs.\/}\ differences between
these models.
Appendix~\ref{app-sc-inv-sc-norm-temp-der-limit-kern} shows that for the
temporal scale-space representation given by convolution with the
scale-invariant time-causal limit kernel, the corresponding scale-normalized
derivatives become fully scale covariant/invariant for temporal scaling
transformations that correspond to exact mappings between the discrete
temporal scale levels.

This paper is a much further developed version of a conference paper
\cite{Lin15-SSVM} presented
at the SSVM 2015, with substantial additions
concerning:
\begin{itemize}
\item
  the theory that implies that the temporal scales are implied to be discrete (sections~\ref{sec-time-caus-scps-pure-temp-domain}-\ref{sec-class-cont-scsp-kernels}),
\item
  more detailed modelling of biological receptive fields (section~\ref{sec-comp-model-biol-RF}),
\item
  the construction of a truly self-similar and scale-invariant
  time-causal limit kernel (section~\ref{sec-time-caus-limit-kernel}),
\item
  theory for implementation in terms of discrete time-causal scale-space kernels (section~\ref{sec-class-disc-scsp-kern}),
\item
  details concerning more rotationally symmetric implementation over spatial domain (section~\ref{app-disc-gauss-smooth}),
\item
  definition of scale-normalized temporal derivatives for the
  resulting time-causal scale-space (section~\ref{sec-sc-norm-spat-temp-der}),
\item
  a framework for spatio-temporal feature detection based on time-causal and time-recursive spatiotemporal
scale space, including scale normalization as well as covariance and invariance
properties under natural image transformations and experimental results (section~\ref{sec-spat-temp-feat}),
\item
  a frequency analysis of the time-causal and time-recursive
  scale-space kernels (appendix~\ref{app-freq-anal-time-caus-kern}),
\item
  a comparison between the presented semi-discrete model and
  Koenderink's fully continuous model, including comparisons between
  the temporal kernels in the two models and a mapping
between the parameters in our model and Koenderink's model
(appendix~\ref{app-comp-koe-model}) and
\item
  a theoretical analysis of the evolution properties over scales of
  temporal derivatives obtained from the time-causal limit kernel,
  including the scaling properties of the scale normalization factors 
  under $L_p$-normalization and a proof that the resulting
  scale-normalized derivatives become scale invariant/covariant
  (appendix~\ref{app-sc-inv-sc-norm-temp-der-limit-kern}).
\end{itemize}
In relation to the SSVM 2015 paper, this paper therefore first shows how the presented framework
applies to spatio-temporal feature detection and computational modelling of biological vision, which
could not be fully described because of space limitations, and then presents important
theoretical extensions in terms of theoretical properties (scale
invariance) and theoretical analysis as well as other technical details that could not be
included in the conference paper because of space limitations.

\section{Spatio-temporal receptive fields}
\label{sec-spat-temp-RF}

The theoretical structure that we start from is a general result 
from axiomatic derivations of a spatio-temporal scale-space
based on assumptions of non-enhancement of local extrema
and the existence of a continuous temporal scale parameter,
which states that the spatio-temporal receptive fields should be based on
spatio-temporal smoothing kernels of the form
(see overviews in Lindeberg \cite{Lin10-JMIV,Lin13-BICY}):
\begin{equation}
      \label{eq-spat-temp-RF-model}
       T(x_1, x_2, t;\; s, \tau;\; v, \Sigma)  
       = g(x_1 - v_1 t, x_2 - v_2 t;\; s, \Sigma) \, h(t;\; \tau)
\end{equation}
  where
  \begin{itemize}
  \item
     $x = (x_1, x_2)^T$ denotes the image coordinates,
  \item
     $t$ denotes time,
  \item
    $s$ denotes the spatial scale,
\item
    $\tau$ denotes the temporal scale,
  \item
   $v = (v_1, v_2)^T$ denotes a local image velocity,
  \item
    $\Sigma$ denotes a spatial covariance matrix determining the
    spatial shape of an affine Gaussian kernel
   $g(x;\; s, \Sigma)  = \frac{1}{2 \pi s \sqrt{\det\Sigma}} e^{-x^T \Sigma^{-1} x/2s}$,
  \item
     $g(x_1 - v_1 t, x_2 - v_2 t;\; s, \Sigma)$ denotes a spatial affine Gaussian kernel
     that moves with image velocity $v = (v_1, v_2)$ in space-time and
\item
   $h(t;\; \tau)$ is a temporal smoothing kernel over time.
\end{itemize}
A biological motivation for this form of separability between the
smoothing operations over space and time can also be obtained from 
the facts that 
(i)~most receptive
fields in the retina and the LGN are to a first approximation space-time separable and
(ii)~the receptive fields of simple cells in V1 can be either space-time
separable or inseparable, where the simple cells with inseparable
receptive fields exhibit receptive fields subregions that are tilted
in the space-time domain and the tilt is an excellent predictor 
of the preferred direction and speed of motion
(DeAngelis {\em et al.\/}\
\protect\cite{DeAngOhzFre95-TINS,deAngAnz04-VisNeuroSci}).

For simplicity, we shall here restrict the above family of affine Gaussian
kernels over the spatial domain to rotationally symmetric Gaussians of
different size $s$, by setting the covariance matrix $\Sigma$ to a unit
matrix.
We shall also mainly restrict ourselves to space-time separable
receptive fields by setting the image velocity $v$ to zero.

A conceptual difference that we shall pursue is by relaxing the
requirement of a semi-group structure over a continuous temporal scale parameter in the above
axiomatic derivations by a weaker Markov property over a discrete temporal scale parameter.
We shall also replace the previous axiom about non-creation of new
image structures with increasing scale in terms of non-enhancement of
local extrema (which requires a continuous scale parameter) by the
requirement that the temporal smoothing process, when seen as an operation
along a one-dimensional temporal axis only, must not increase the number of
local extrema or zero-crossings in the signal. Then, another family of
time-causal scale-space kernels becomes permissible and uniquely determined, in terms of
first-order integrators or truncated exponential filters
coupled in cascade. 

The main topics of this paper are to handle the remaining degrees of freedom
resulting from this construction about: (i)~choosing and parameterizing
the distribution of temporal scale levels, (ii)~analysing the resulting temporal
dynamics, (iii)~describing how this model can be transferred to a
discrete implementation over discretized time, space or both while retaining discrete scale-space
properties, (iv)~using the resulting theory for
computing scale-normalized spatio-temporal derivative expressions for
purposes in computer vision and (v)~computational modelling of biological vision.

\section{Time-causal temporal scale-space}
\label{sec-time-caus-scale-spaces}

When constructing a system for real-time processing of sensor data,
a fundamental constraint on the temporal smoothing
kernels is that they have to be {\em time-causal\/}. 
The ad hoc solution of using a truncated
symmetric filter of finite temporal extent in combination with a
temporal delay is not appropriate in a time-critical context.
Because of computational and memory efficiency, the computations
should furthermore be based on a compact temporal buffer that contains
sufficient information for representing the sensor information at multiple
temporal scales and computing features therefrom.
Corresponding requirements are necessary in computational
modelling of biological perception.

\subsection{Time-causal scale-space kernels for pure temporal domain}
\label{sec-time-caus-scps-pure-temp-domain}

To model the temporal component of the smoothing operation in equation~(\ref{eq-spat-temp-RF-model}), 
let us initially consider a signal $f(t)$ defined over a one-dimensional continuous temporal axis $t \in \bbbr$.
To define a one-parameter family of temporal scale-space representation from
this signal, we consider a one-parameter family of smoothing kernels $h(t;\, \tau)$
where $\tau \geq 0$ is the temporal scale parameter
\begin{equation}
  L(t;\; \tau) = (h(\cdot;\; \tau) * f(\cdot))(t;\; \tau) = \int_{u = 0}^{\infty} h(u;\ \tau) \, f(t-u) \, du
\end{equation}
and $L(t;\; 0) = f(t)$.
To formalize the requirement that this transformation must not introduce new structures from 
a finer to a coarser temporal scale, let us following Lindeberg \cite{Lin90-PAMI} require that
between any pair of temporal scale levels $\tau_2 > \tau_1 \geq 0$
 the number of local extrema at scale $\tau_2$ must not exceed the number 
of local extrema at scale $\tau_1$.
Let us additionally require the family of temporal smoothing kernels $h(u;\ \tau)$ to obey the
following cascade relation 
\begin{equation}
  \label{eq-casc-rel-temp-scsp}
   h(\cdot;\; \tau_2) = (\Delta h)(\cdot;\; \tau_1 \mapsto \tau_2) * h(\cdot;\; \tau_1)
\end{equation}
between any pair of temporal scales $(\tau_1, \tau_2)$ with $\tau_2 > \tau_1$
for some family of transformation kernels $(\Delta h)(t;\; \tau_1 \mapsto \tau_2)$.
Note that in contrast to most other axiomatic scale-space definitions, we
do, however, not impose a strict semi-group property on the kernels.
The motivation for this is to make it possible to take larger scale steps 
at coarser temporal scales, which will give higher flexibility and enable the
construction of more efficient temporal scale-space representations.

Following Lindeberg \cite{Lin90-PAMI}, let us further define a scale-space kernel
as a kernel that guarantees that the number of local extrema in the convolved signal
can never exceed the number of local extrema in the input signal.
Equivalently, this condition can be expressed in terms of the number of zero-crossings in the signal.
Following Lindeberg and Fagerstr{\"o}m \cite{LF96-ECCV}, let us
additionally define a {\em temporal scale-space kernel\/} as a kernel
that both satisfies the temporal causality requirement $h(t;\; \tau) = 0$ if $t< 0$
and guarantees that the number of local extrema does not increase
under convolution.
If both the raw transformation kernels $h(u;\ \tau)$ and the cascade kernels
$(\Delta h)(t;\; \tau_1 \mapsto \tau_2)$ are scale-space kernels, we do hence
guarantee that the number of local extrema in $L(t;\; \tau_2)$ can
never exceed the number of local extrema in $L(t;\; \tau_1)$.
If the kernels $h(u;\ \tau)$ and additionally the cascade kernels
$(\Delta h)(t;\; \tau_1 \mapsto \tau_2)$ are temporal
scale-space kernels, these kernels do hence constitute natural kernels for defining a temporal scale-space representation.

\subsection{Classification of scale-space kernels for continuous signals}
\label{sec-class-cont-scsp-kernels}

Interestingly, the classes of scale-space kernels and temporal scale-space kernels 
can be completely classified based on classical results by Schoenberg and Karlin 
regarding the theory of variation-diminishing linear transformations.
Schoenberg studied this topic in a series of papers over about 20 years
(Schoenberg \cite{Sch30,Sch46,Sch47,Sch48,Sch50,Sch53,Sch88-book}) and Karlin \cite{Kar68} then wrote an
excellent monograph on the topic of total positivity.

\paragraph{Variation diminishing 
transformations.}

Summarizing main results from this theory in a form relevant to the construction
of the scale-space concept for one-dimensional continuous signals
(Lindeberg \cite[section~3.5.1]{Lin93-Dis}), let
$S^-(f)$ denote the number of sign changes in a function $f$ 
\begin{equation}
  S^-(f) = \sup V^- (f(t_1), f(t_2), \dots, f(t_m)),
\end{equation}
where the supremum is extended over all sets $t_1 < t_2 < \dots < t_J$
($t_j \in \bbbr$),
$J$ is arbitrary but finite, and $V^-(v)$ denotes the number of
sign changes in a vector $v$. Then, the transformation
\begin{equation}
  \label{eq-cont-scsp-conv-transf}
  f_{\mbox{\scriptsize\em out}}(\eta) =
  \int_{\xi = -\infty}^{\infty} f_{\mbox{\scriptsize\em in}}(\eta - \xi) \, dG(\xi),
\end{equation}
where $G$ is a distribution function (essentially the primitive function of a convolution kernel),
is said to be {\em variation-diminishing\/} if
\begin{equation}
  S^-(f_{\mbox{\scriptsize\em out}}) \leq S^-(f_{\mbox{\scriptsize\em in}})
\end{equation}
holds for all continuous and bounded $f_{\mbox{\scriptsize\em in}}$.
Specifically, the transformation (\ref{eq-cont-scsp-conv-transf})
is variation diminishing
if and only if $G$ has a bilateral Laplace-Stieltjes transform
  of the form (Schoenberg \cite{Sch50})
  \begin{equation}
    \label{eq-char-var-dim-kernels-cont-case-Laplace}
    \int_{\xi = - \infty}^{\infty} e^{-s \xi} \, dG(\xi) =
    C \, e^{\gamma s^2 + \delta s}
    \prod_{i = 1}^{\infty} \frac{e^{a_i s}}{1 + a_i s}
      \quad
  \end{equation}
  for $-c < \mbox{Re}(s) < c$ and some $c > 0$,
  where $C \neq 0$, $\gamma \geq 0$, $\delta$ and $a_i$
  are real, and $\sum_{i=1}^{\infty} a_i^2$ is convergent.

\paragraph{Classes of continuous scale-space kernels.}

Interpreted in the temporal domain, this result implies that for
continuous signals there are four primitive types of linear and
shift-invariant smoothing transformations;
convolution with the {\em Gaussian kernel,\/}
\begin{equation}
  \label{eq-Gauss-polya-comp}
  h(\xi) = e^{-\gamma \xi^2},
\end{equation}
convolution with the {\em {truncated exponential function}s,}
\begin{equation}
  \label{eq-truncexp-polya-comp}
  h(\xi) =
  \left\{
    \begin{array}{lcl}
      e^{- |\lambda| \xi} & & \xi \geq 0, \\
      0                 & & \xi < 0,
    \end{array}
  \right.
    \quad\quad
  h(\xi) =
  \left\{
    \begin{array}{lcl}
      e^{|\lambda| \xi} & & \xi \leq 0, \\
      0                 & & \xi > 0,
    \end{array}
  \right.
\end{equation}
as well as trivial {\em translation\/} and {\em rescaling}.
Moreover, it means that a shift-invariant linear transformation is
variation diminishing if and only if it
can be decomposed into these primitive operations.

\subsection{Temporal scale-space kernels over continuous temporal domain}
\label{sec-cont-temp-scsp-kern}

In the above expressions, the first class of scale-space kernels (\ref{eq-Gauss-polya-comp})
corresponds to using a non-causal Gaussian scale-space concept over time, 
which may constitute a straightforward model for analysing pre-recorded temporal
data in an offline setting where temporal causality is not critical and can be disregarded
by the possibility of accessing the virtual future in relation to any pre-recorded time moment.

Adding temporal causality as a necessary requirement, and with additional normalization
of the kernels to unit $L_1$-norm to leave a constant signal unchanged,
it follows that the following family of truncated exponential kernels
  \begin{equation}
    h_{exp}(t;\; \mu_k) 
    = \left\{
        \begin{array}{ll}
          \frac{1}{\mu_k} e^{-t/\mu_k} & t \geq 0 \\
          0         & t < 0
        \end{array}
      \right.
  \end{equation}
constitutes the only
class of time-causal scale-space kernels over a continuous temporal domain in
the sense of guaranteeing both temporal causality and non-creation of
new local extrema (or equivalently zero-crossings) with increasing scale
(Lindeberg \cite{Lin90-PAMI}; Lindeberg and Fagerstr{\"o}m \cite{LF96-ECCV}).
The Laplace transform of such a kernel is given by
\begin{equation}
  \label{eq-FT-composed-kern-casc-truncexp}
    H_{exp}(q;\; \mu_k) 
    = \int_{t = - \infty}^{\infty} h_{exp}(t;\; \mu_k) \, e^{-qt} \, dt
    = \frac{1}{1 + \mu_k q}
  \end{equation}
  and coupling $K$ such kernels in cascade leads to a composed kernel
  \begin{equation}
    \label{eq-comp-trunc-exp-cascade}
    h_{composed}(\cdot;\; \mu) 
    = *_{k=1}^{K} h_{exp}(\cdot;\; \mu_k)
  \end{equation}
  having a Laplace transform of the form
  \begin{align}
    \begin{split}
       \label{eq-expr-comp-kern-trunc-exp-filters}
       H_{composed}(q;\; \mu) 
       & = \int_{t = - \infty}^{\infty} 
              *_{k=1}^{K} h_{exp}(\cdot;\; \mu_k)(t) 
                 \, e^{-qt} \, dt
   \end{split}\nonumber\\
   \begin{split}
       & =  \prod_{k=1}^{K} \frac{1}{1 + \mu_k q}.
    \end{split}
  \end{align}
  The composed kernel has temporal mean and variance
  \begin{equation}
     \label{eq-mean-var-trunc-exp-filters}
     m_K = \sum_{k=1}^{K} \mu_k \quad\quad \tau_K = \sum_{k=1}^{K} \mu_k^2.
  \end{equation}

\begin{figure*}[hbtp]
  \begin{center}
    \begin{tabular}{ccc}
      {\small $h(t;\; \mu, K=7)$} 
      & {\small $h_{t}(t;\; \mu, K=7)$} 
      & {\small $h_{tt}(t;\; \mu, K=7)$} \\
      \includegraphics[width=0.23\textwidth]{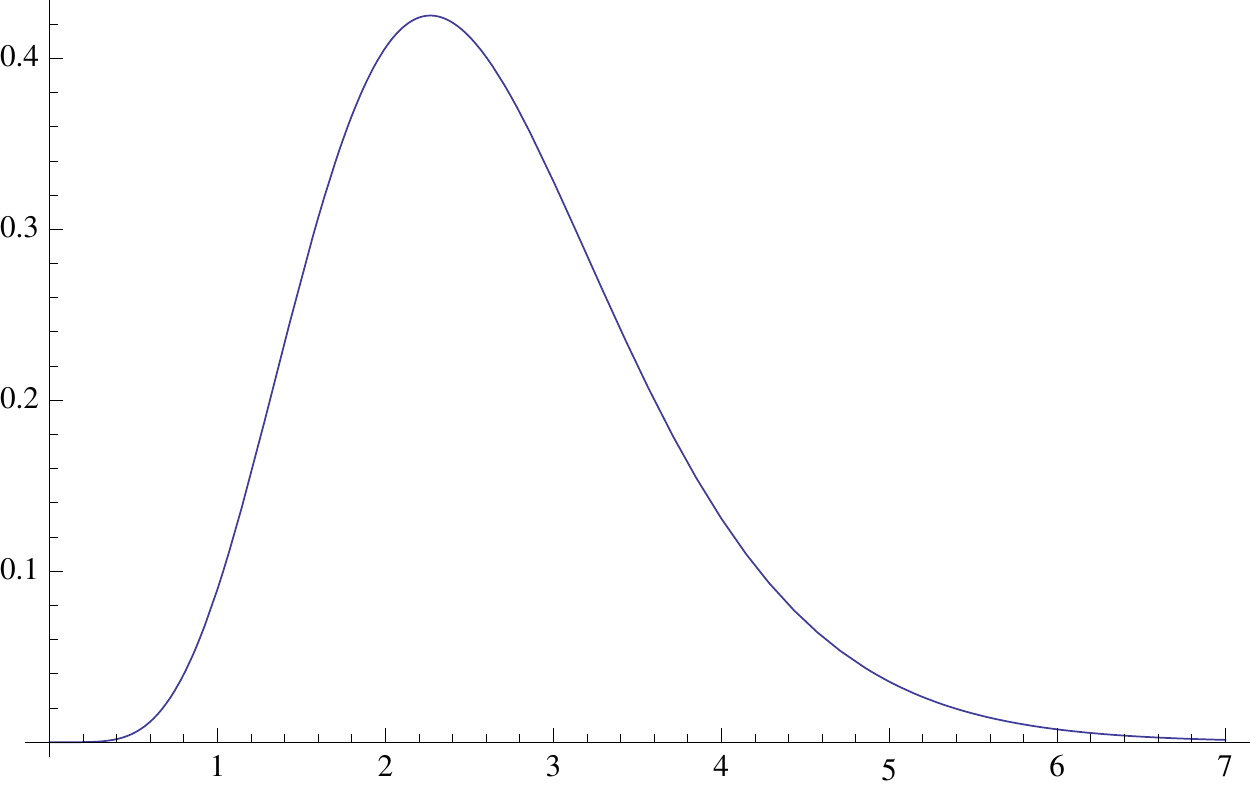} &
      \includegraphics[width=0.23\textwidth]{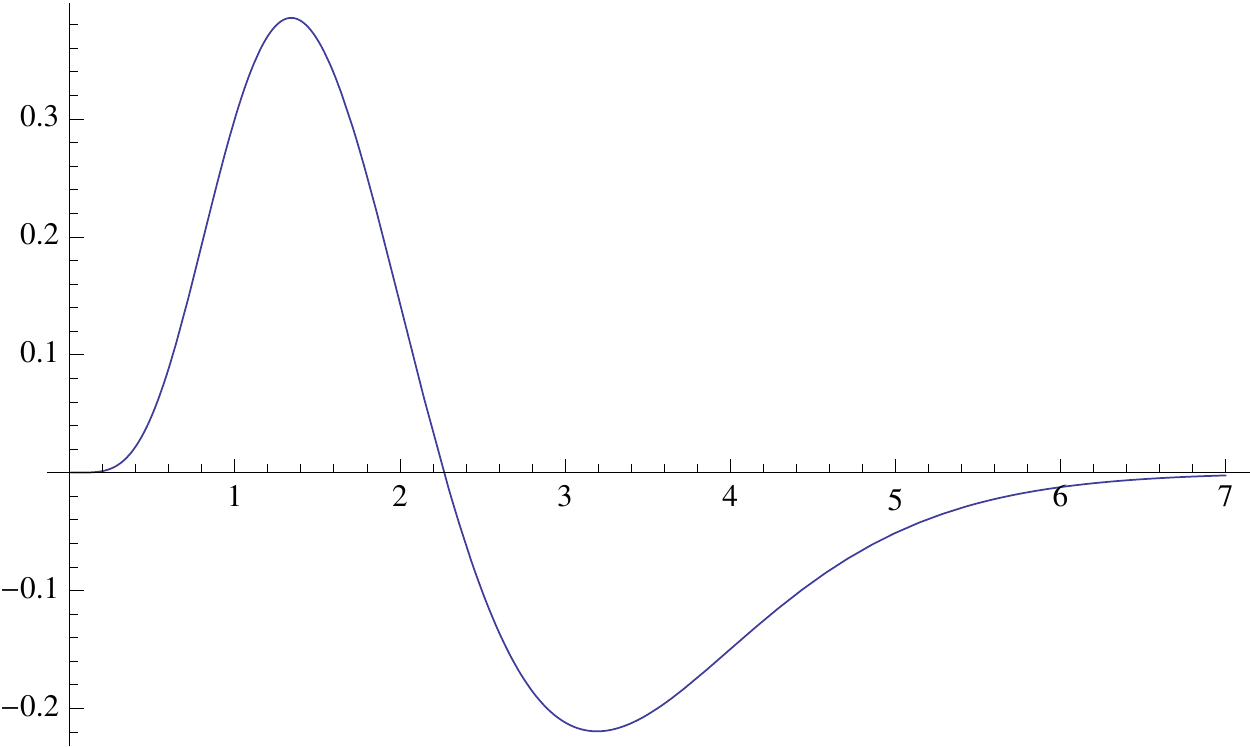} &
      \includegraphics[width=0.23\textwidth]{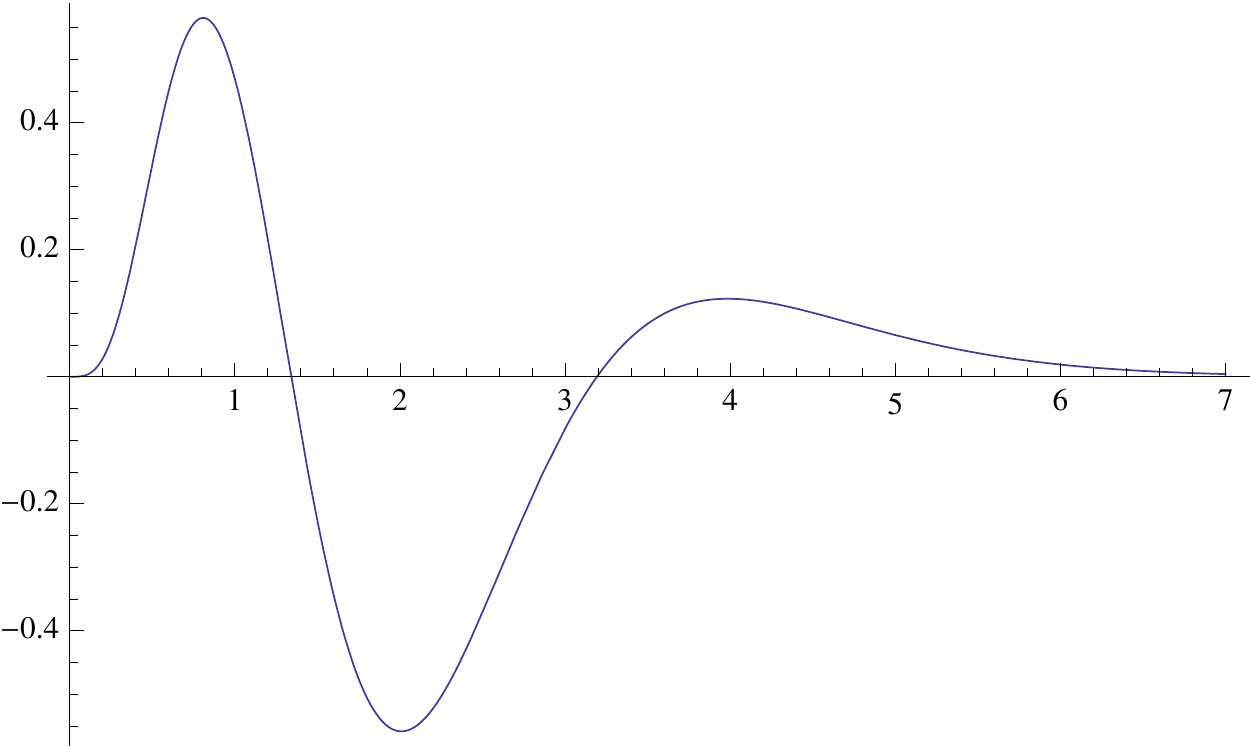} \\
  \\
      {\small $h(t;\; K=7, c = \sqrt{2})$} 
      & {\small $h_{t}(t;\; K=7, c = \sqrt{2}))$} 
      & {\small $h_{tt}(t;\; K=7, c = \sqrt{2}))$} \\
      \includegraphics[width=0.23\textwidth]{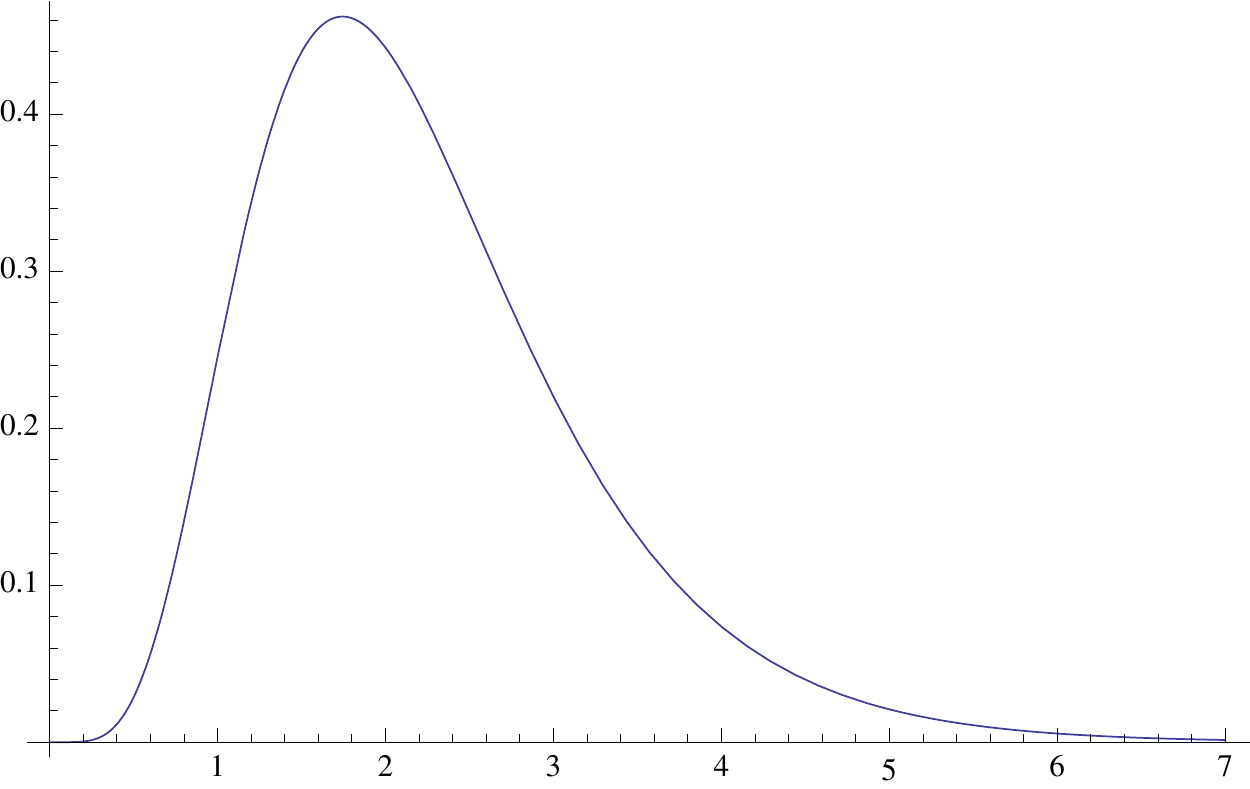} &
      \includegraphics[width=0.23\textwidth]{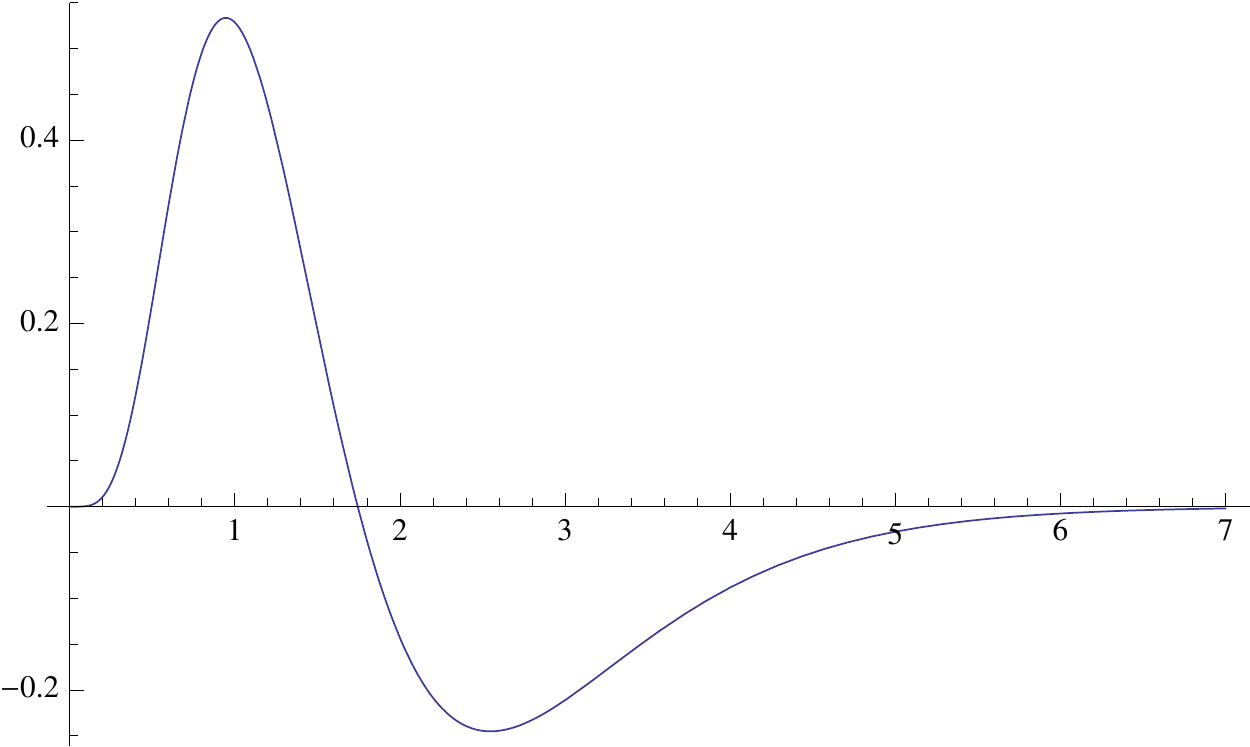} &
      \includegraphics[width=0.23\textwidth]{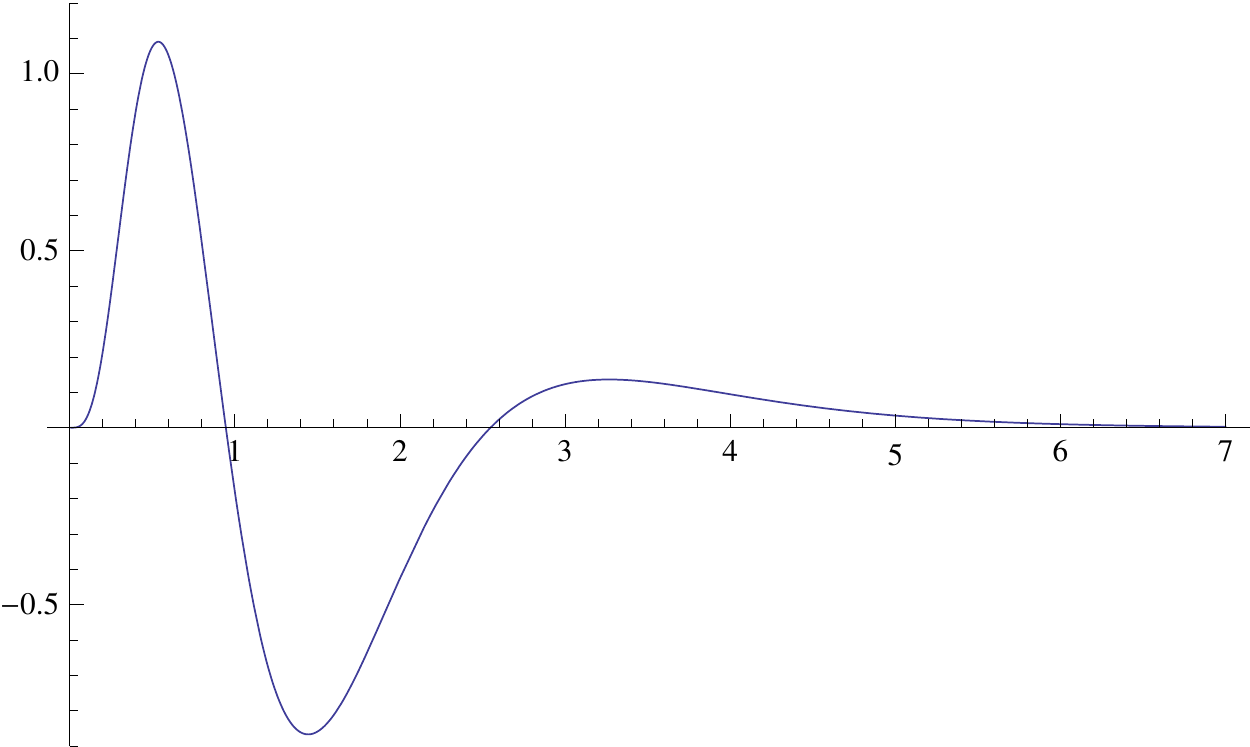} \\
\\
      {\small $h(t;\; K=7, c = 2^{3/4})$} 
      & {\small $h_{t}(t;\; K=7, c = 2^{3/4}))$} 
      & {\small $h_{tt}(t;\; K=7, c = 2^{3/4}))$} \\
      \includegraphics[width=0.23\textwidth]{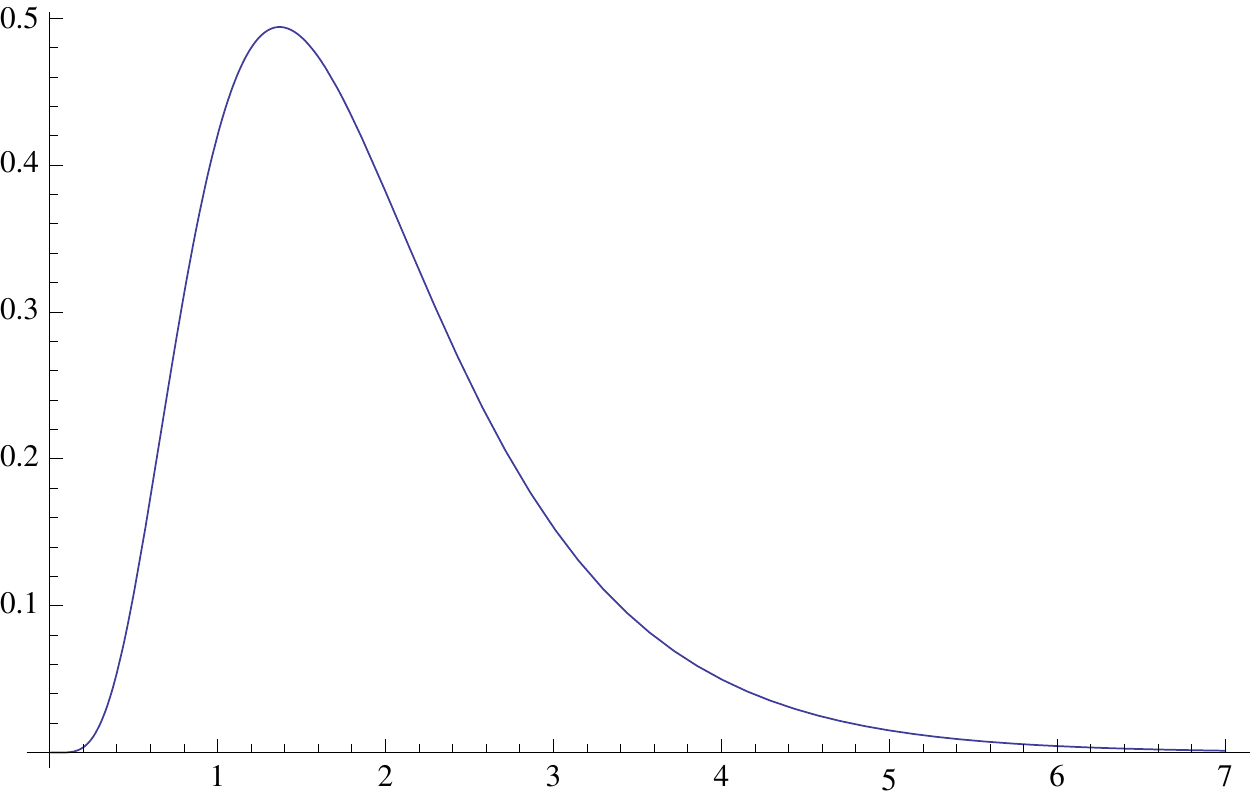} &
      \includegraphics[width=0.23\textwidth]{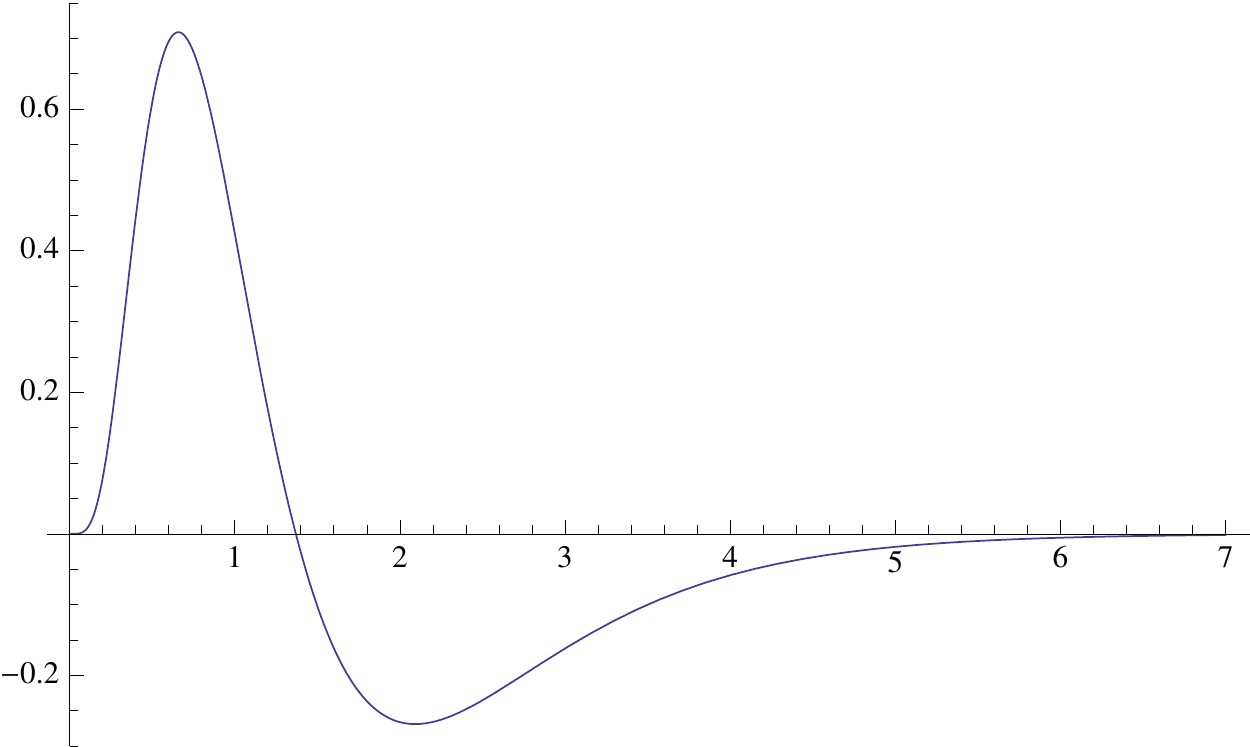} &
      \includegraphics[width=0.23\textwidth]{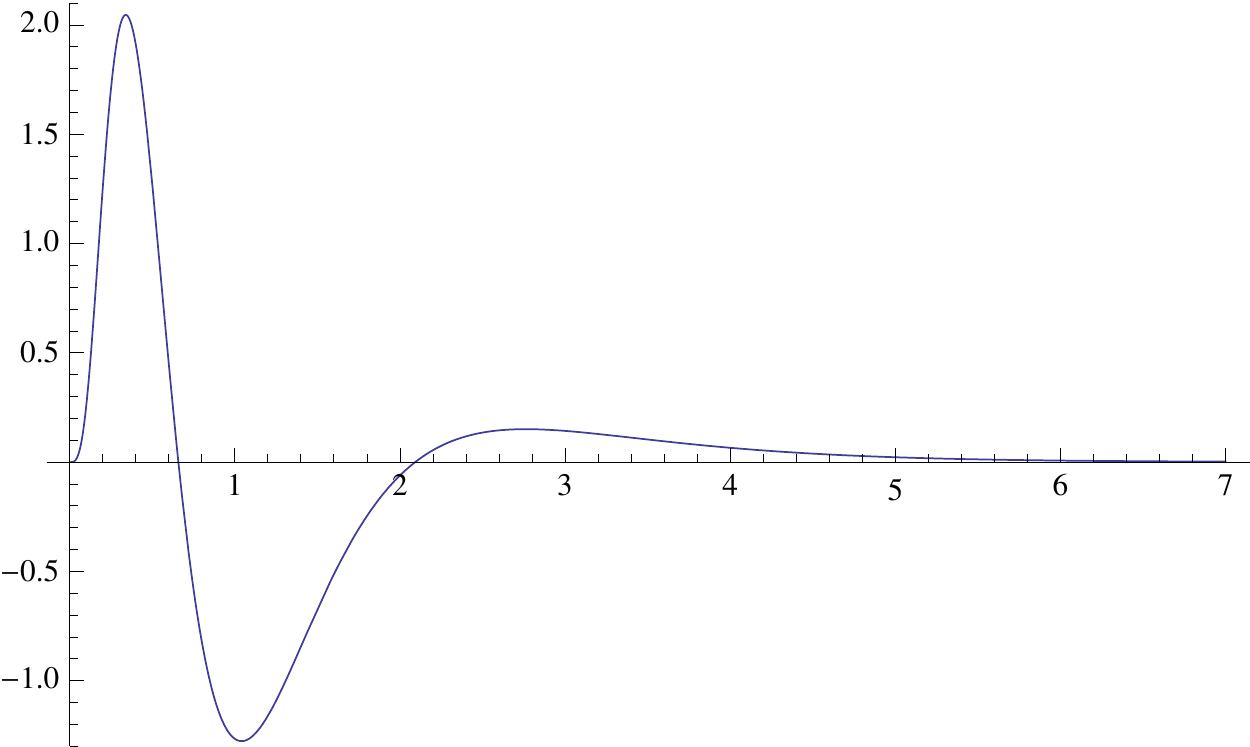} \\
\\
      {\small $h(t;\; K=7, c = 2$} 
      & {\small $h_{t}(t;\; K=7, c = 2))$} 
      & {\small $h_{tt}(t;\; K=7, c = 2))$} \\
      \includegraphics[width=0.23\textwidth]{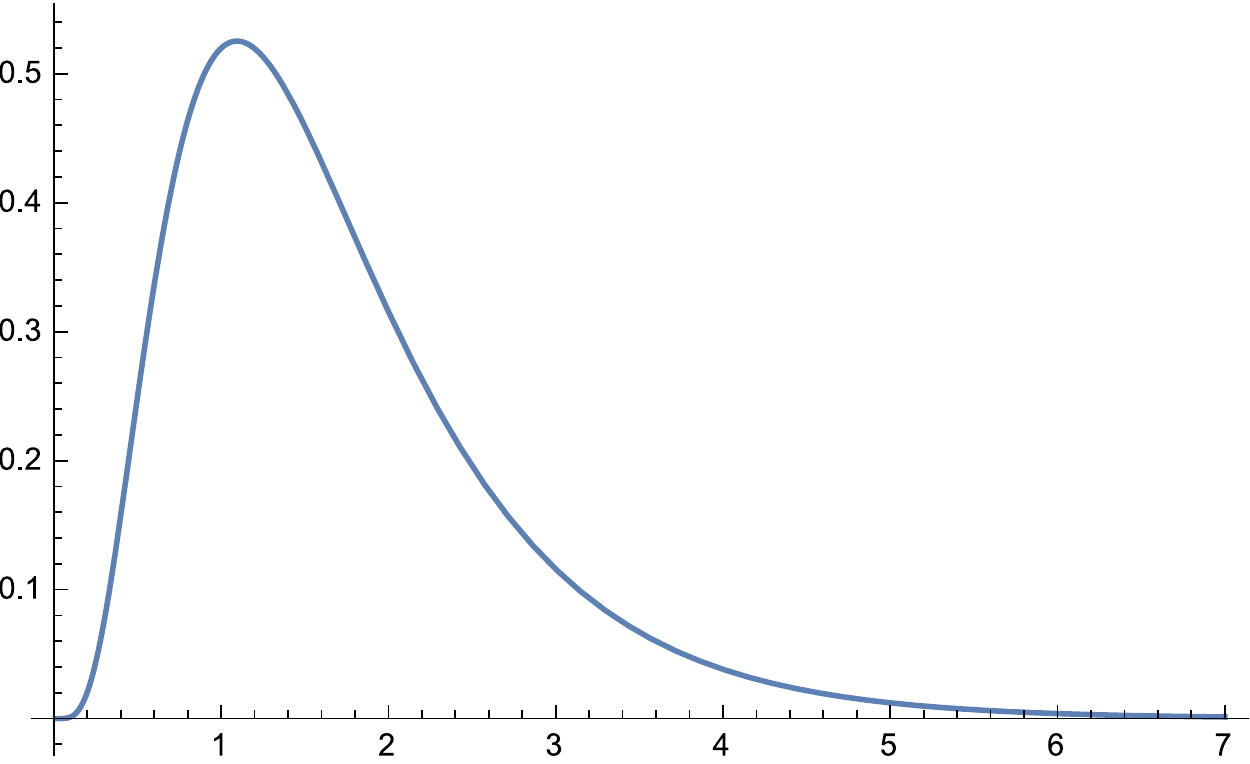} &
      \includegraphics[width=0.23\textwidth]{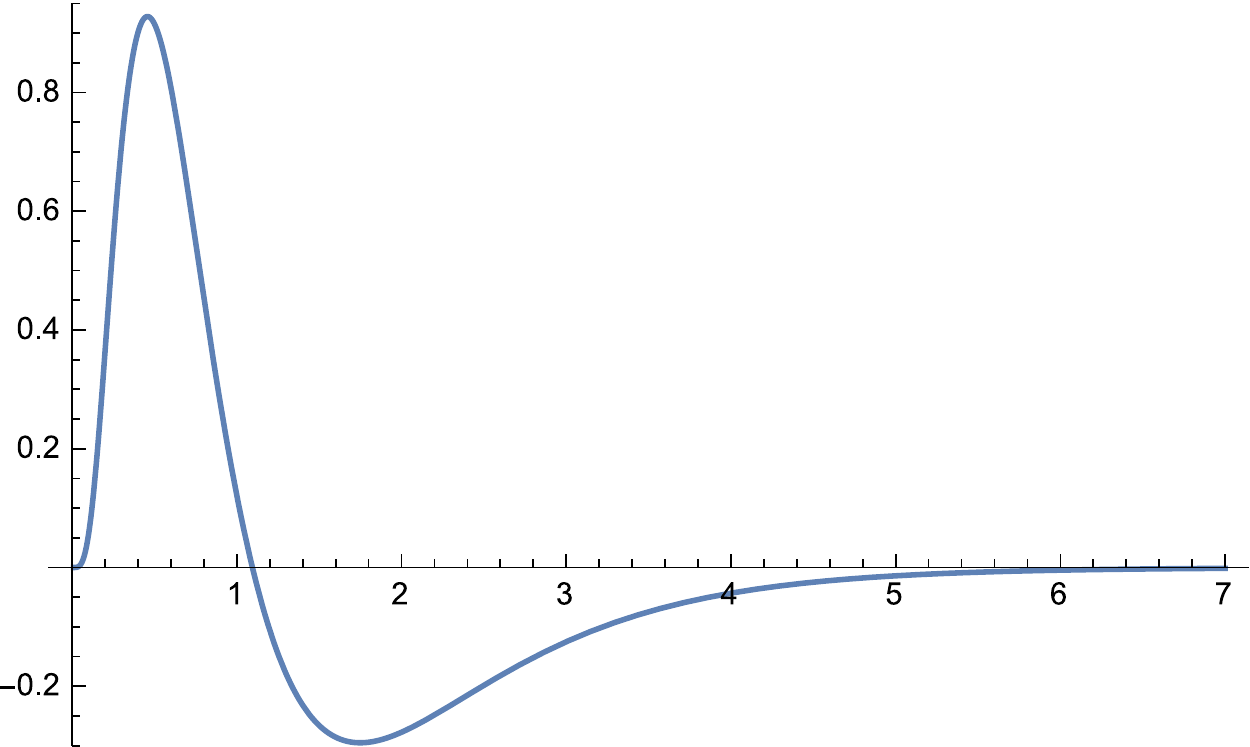} &
      \includegraphics[width=0.23\textwidth]{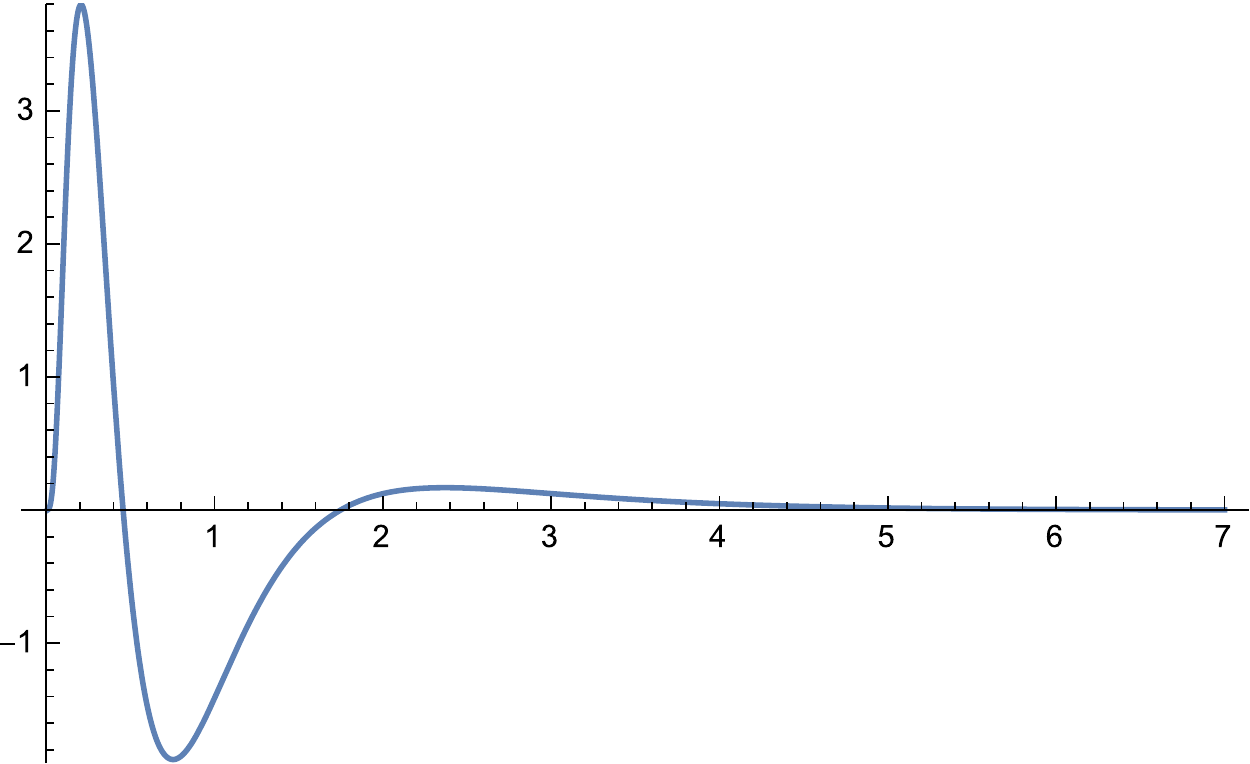} \\
     \end{tabular} 
  \end{center}
   \caption{Equivalent kernels with temporal variance $\tau = 1$ corresponding to the composition of
           $K = 7$ {\em truncated exponential kernels\/} in cascade and
           their first- and second-order derivatives.
           (top row) Equal time constants $\mu$.
           (second row) Logarithmic distribution of the scale levels
           for $c = \sqrt{2}$.
          (third row) Logarithmic distribution 
           for $c = 2^{3/4}$.
          (bottom row) Logarithmic distribution 
           for $c = 2$.}
  \label{fig-trunc-exp-kernels-1D}
\end{figure*}

\begin{figure}[!h]
   \begin{center}
      \includegraphics[width=0.45\textwidth]{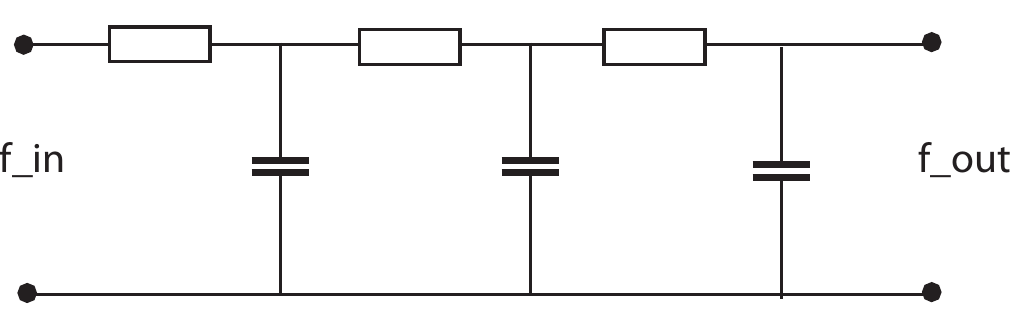}
   \end{center}
\caption{Electric wiring diagram consisting of a set of resistors and
  capacitors that emulate a series of first-order integrators coupled in
  cascade, if we regard the time-varying voltage $f_{in}$ as
  representing the time varying input signal and the resulting output
  voltage $f_{out}$ as representing the time varying output signal at a
  coarser temporal scale.
  According to the presented theory, the corresponding truncated exponential kernels of time are the only
  primitive temporal smoothing kernels that guarantee both temporal
  causality and non-creation of local extrema
  (alternatively zero-crossings) with increasing temporal scale.
  Such first-order temporal integration can be used as a straightforward
  computational model for temporal processing in biological neurons
  (see also Koch \protect\cite[Chapters~11--12]{Koch99-book} regarding 
  physical modelling of the information transfer in dendrites of neurons).}
  \label{fig-first-order-integrators-electric}
\end{figure}

\noindent
In terms of physical models, repeated convolution with such kernels
corresponds to coupling a series
of {\em first-order integrators\/} with time constants $\mu_k$ in cascade
\begin{equation}
  \label{eq-first-ord-int}
   \partial_t L(t;\; \tau_k) 
   = \frac{1}{\mu_k} \left( L(t;\; \tau_{k-1}) - L(t;\; \tau_k) \right)
\end{equation}
with $L(t;\; 0) = f(t)$.
  In the sense of guaranteeing non-creation of new local extrema or zero-crossings over time,
these kernels have a desirable and well-founded smoothing
  property that can be used for defining multi-scale observations over time.
  A constraint on this type of temporal scale-space representation,
  however, is that the {\em scale levels are required to be discrete\/}
  and that the scale-space representation does hence not admit
  a continuous scale parameter.
 Computationally, however, the scale-space representation based
  on truncated exponential kernels can be highly efficient and admits
  for direct implementation in terms of hardware (or wetware) that emulates
  first-order integration over time, and where the temporal scale
  levels together also serve as a sufficient time-recursive memory of
  the past (see figure~\ref{fig-first-order-integrators-electric}).

\begin{figure}[hbt]
  \begin{center}
     \begin{tabular}{c}
        \includegraphics[width=0.47\textwidth]{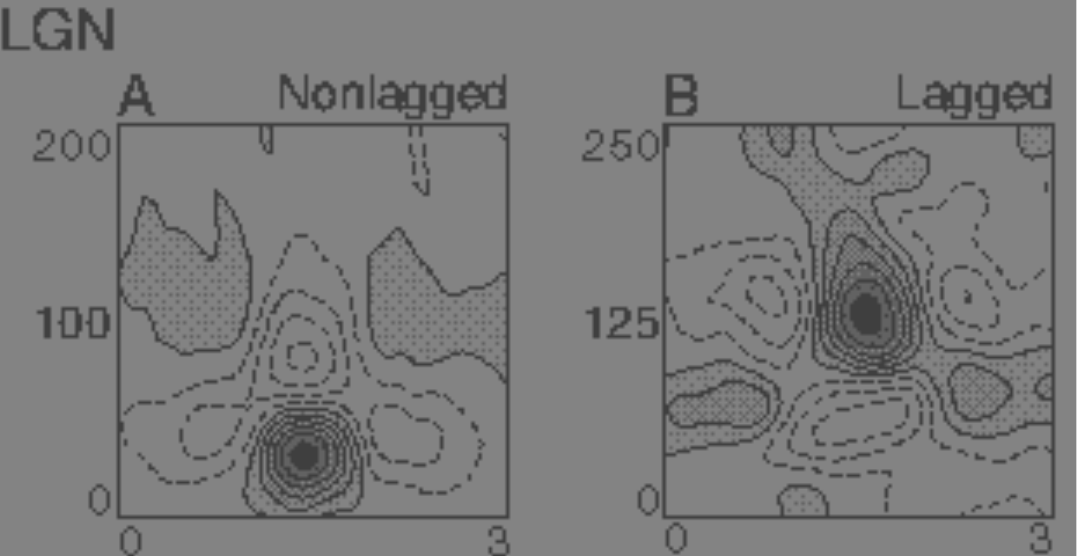}
    \end{tabular}
  \end{center}
  \begin{center}
    \begin{tabular}{cc}
      \hspace{4mm} {\footnotesize $h_{xxt}(x, t;\; s, \tau)$}
      & \hspace{2mm} {\footnotesize $-h_{xxtt}(x, t;\; s, \tau)$} \\
\hspace{4mm} \includegraphics[width=0.20\textwidth]{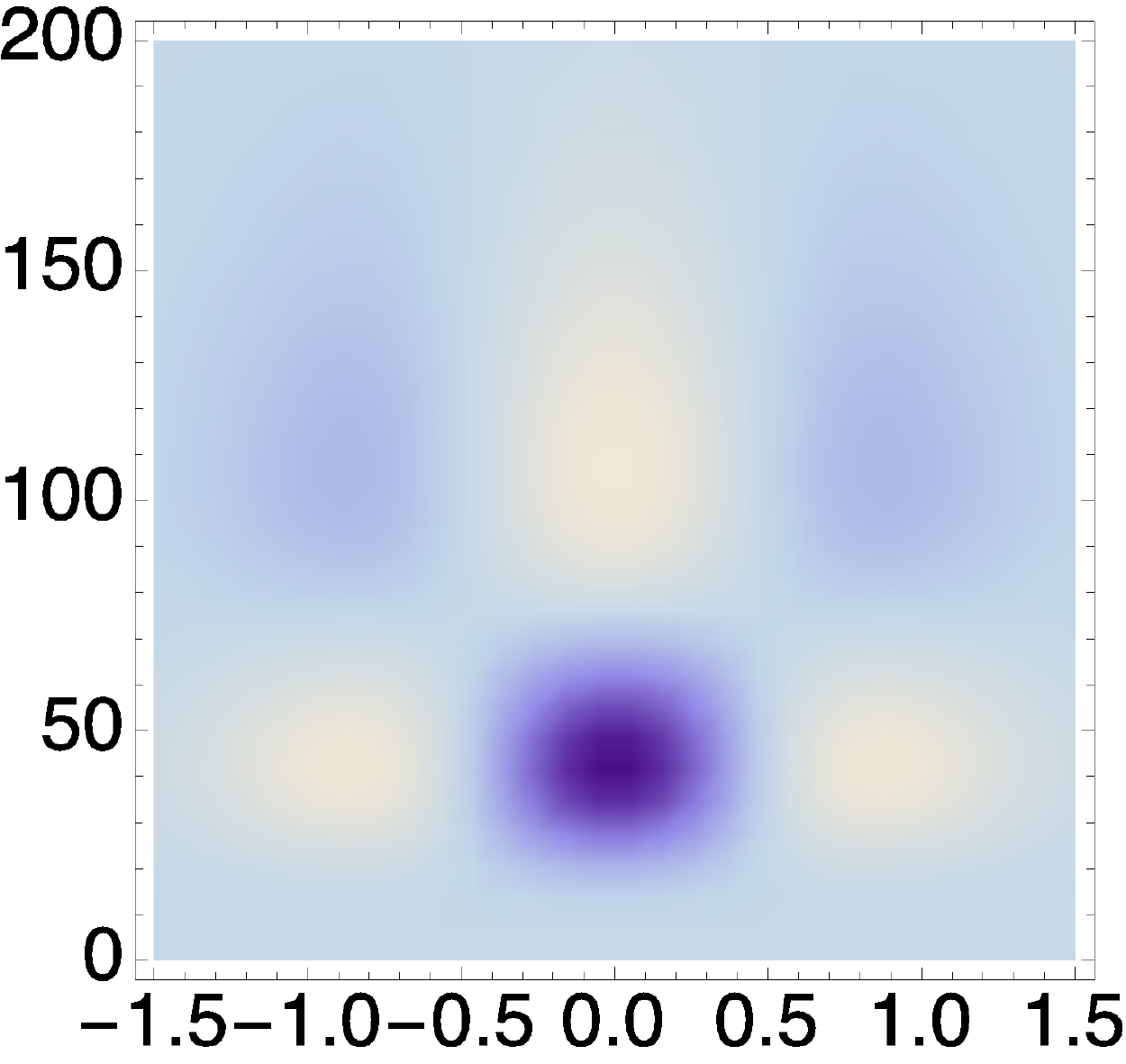}
      &
\hspace{2mm}  \includegraphics[width=0.20\textwidth]{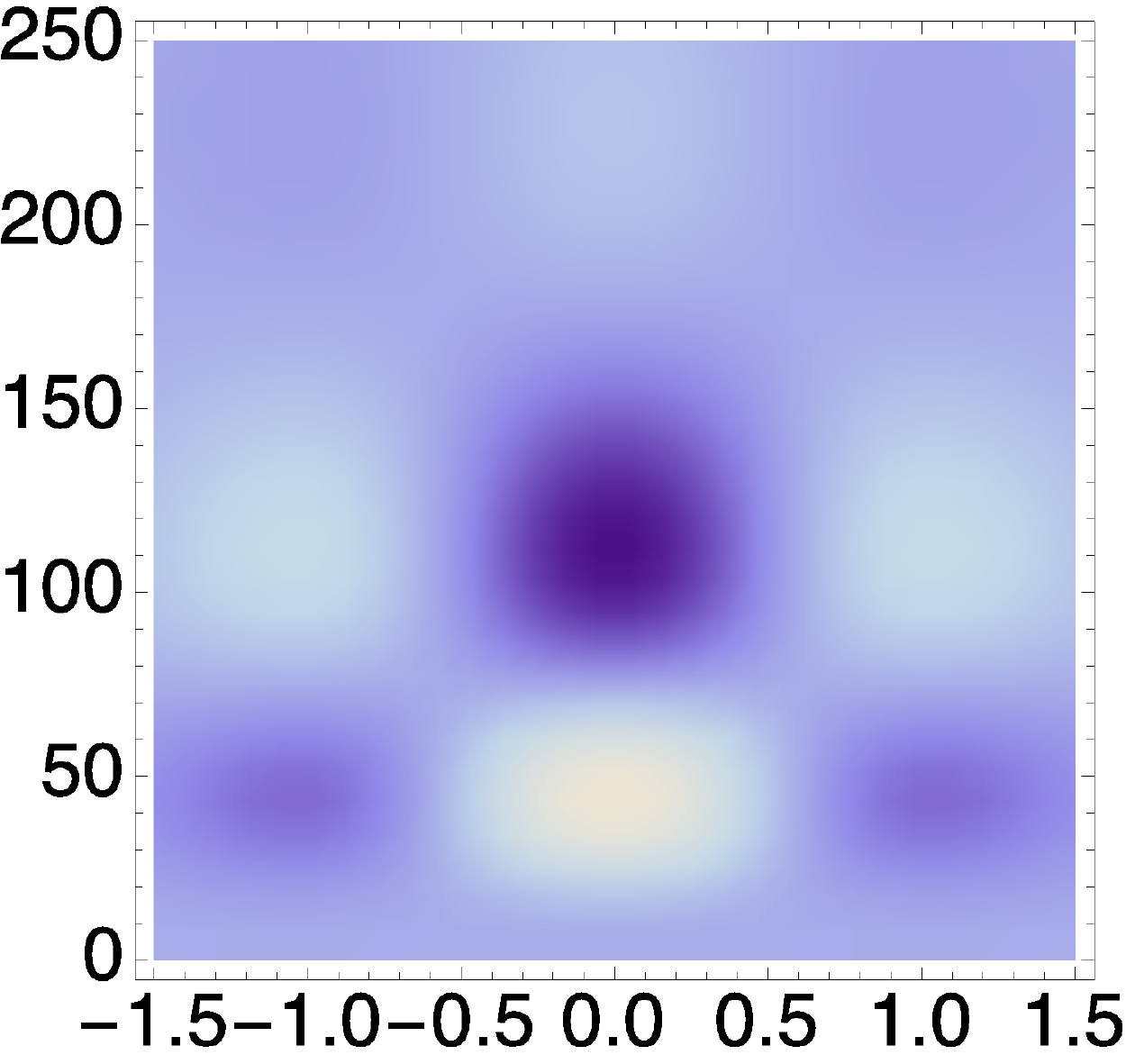} 
    \end{tabular}
  \end{center}
\caption{Computational modelling of space-time separable receptive
  field profiles in the lateral geniculate nucleus (LGN)
           as reported by DeAngelis {\em et al.\/}\ \protect\cite{DeAngOhzFre95-TINS} using
    idealized spatio-temporal receptive fields of the form
     $T(x, t;\; s, \tau) = \partial_{x^{\alpha}} \partial_{t^{\beta}} g(;\; s) \, h(t;\; \tau)$
   according to equation~(\protect\ref{eq-spat-temp-RF-model}) and with
    the temporal smoothing function $h(t;\; \tau)$
    modelled as a cascade of first-order integrators/truncated exponential
    kernels of the form (\protect\ref{eq-comp-trunc-exp-cascade}).
    (left) a ``non-lagged cell'' modelled using first-order temporal derivatives
    (right) a ``lagged cell'' modelled using second-order temporal
    derivatives.
    Parameter values: 
    (a) $h_{xxt}$: $\sigma_x = 0.5$~degrees, $\sigma_t = 40$~ms.
    (b) $h_{xxtt}$: $\sigma_x = 0.6$~degrees, $\sigma_t = 60$~ms.
           (Horizontal dimension: space $x$. Vertical dimension: time~$t$.)}
  \label{fig-deang-tins-temp-resp-prof-lagged-nonlagged}
\end{figure}

\begin{figure*}[hbtp]
  \begin{center}
     \begin{tabular}{c}
        \includegraphics[width=0.78\textwidth]{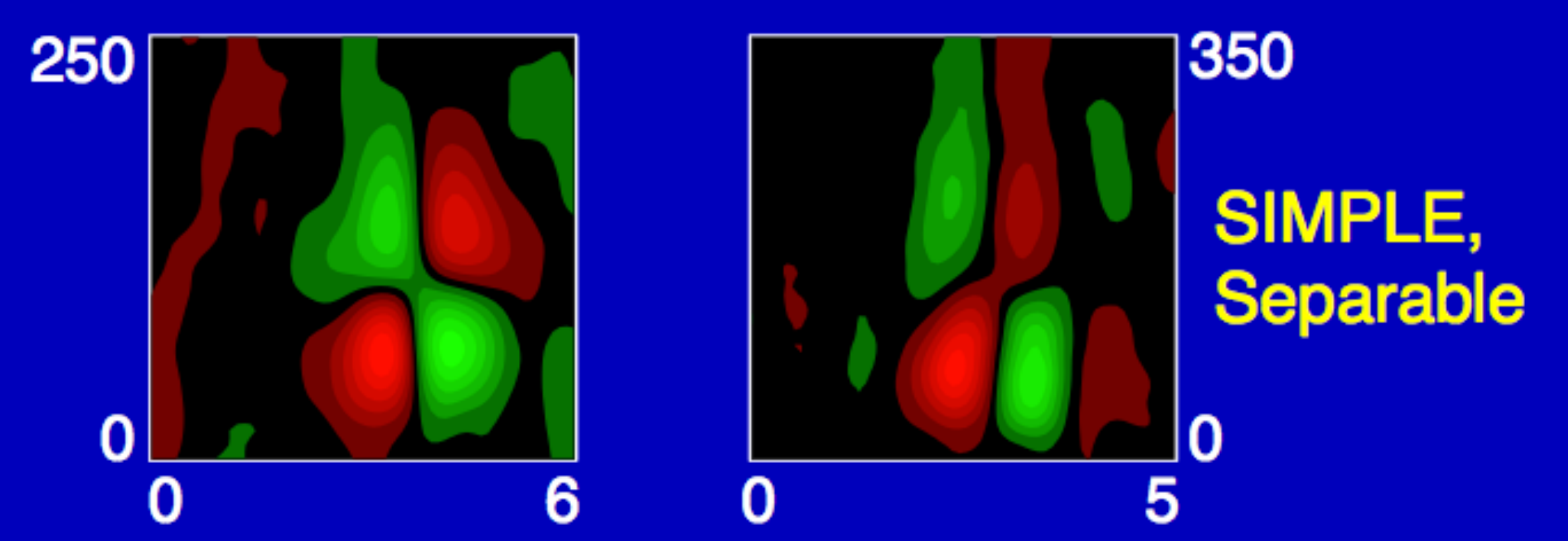} 
    \end{tabular}
  \end{center}
  \begin{center}
    \begin{tabular}{cc}
      \hspace{-26mm} {\footnotesize $h_{xt}(x, t;\; s, \tau)$}
      & \hspace{4mm} {\footnotesize $-h_{xxt}(x, t;\; s, \tau)$} \\
\hspace{-26mm} \includegraphics[width=0.24\textwidth]{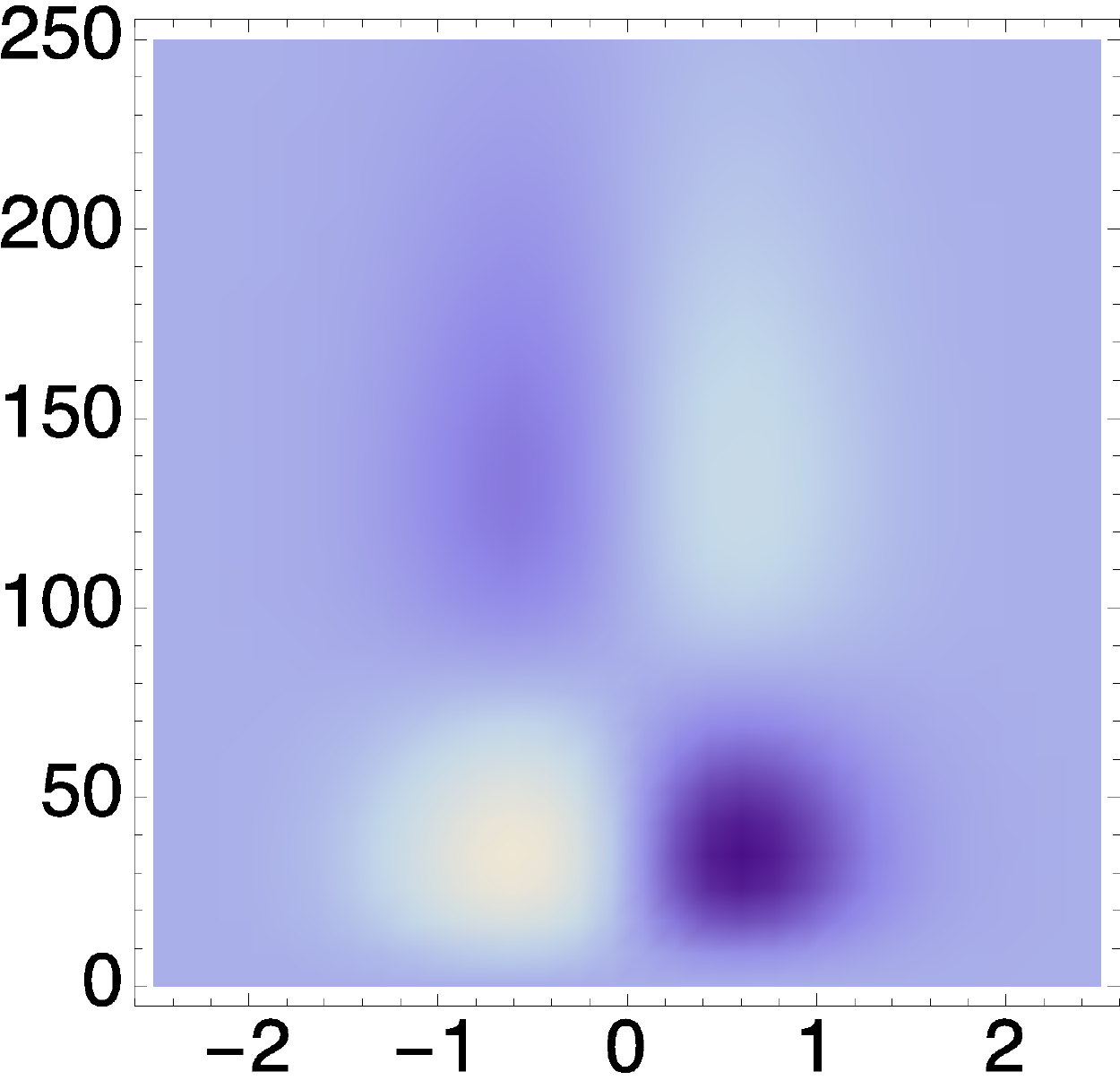}
      &
\hspace{4mm}  \includegraphics[width=0.24\textwidth]{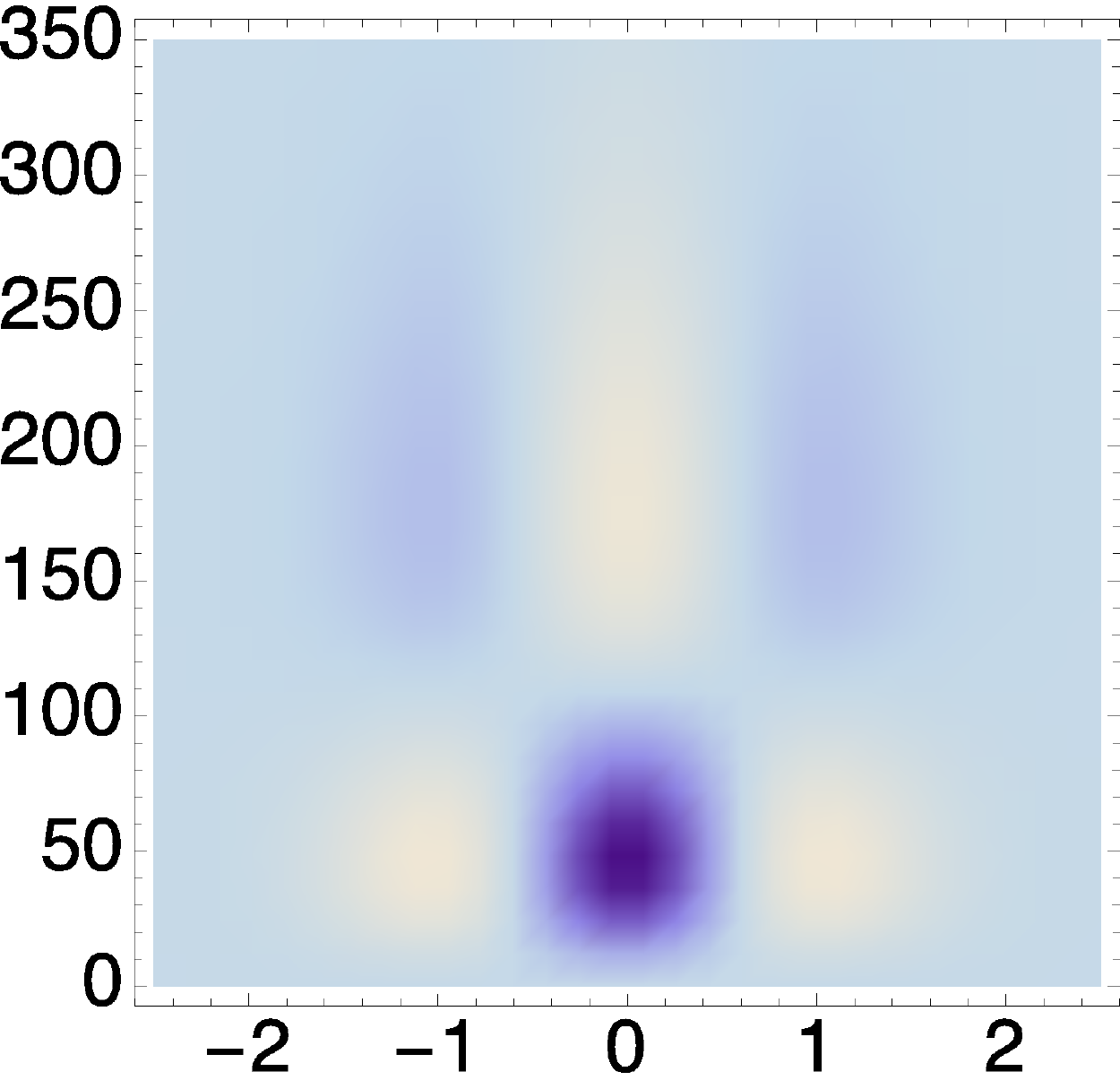} 
    \end{tabular}
  \end{center}

\smallskip

  \begin{center}
     \begin{tabular}{c}
        \includegraphics[width=0.78\textwidth]{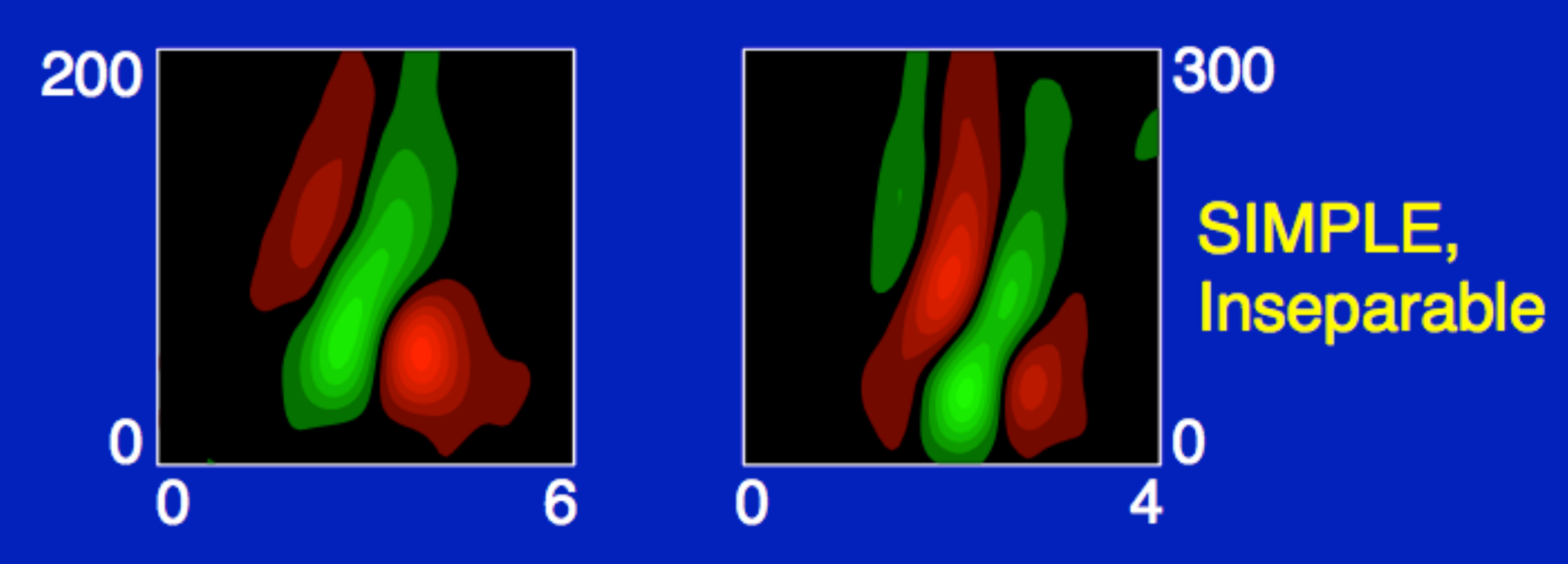}
    \end{tabular}
  \end{center}
  \begin{center}
    \begin{tabular}{cc}
      \hspace{-26mm} {\footnotesize $h_{xx}(x, t;\; s, \tau, v)$}
      & \hspace{4mm} {\footnotesize $-h_{xxx}(x, t;\; s, \tau, v)$} \\
\hspace{-26mm}
\includegraphics[width=0.24\textwidth]{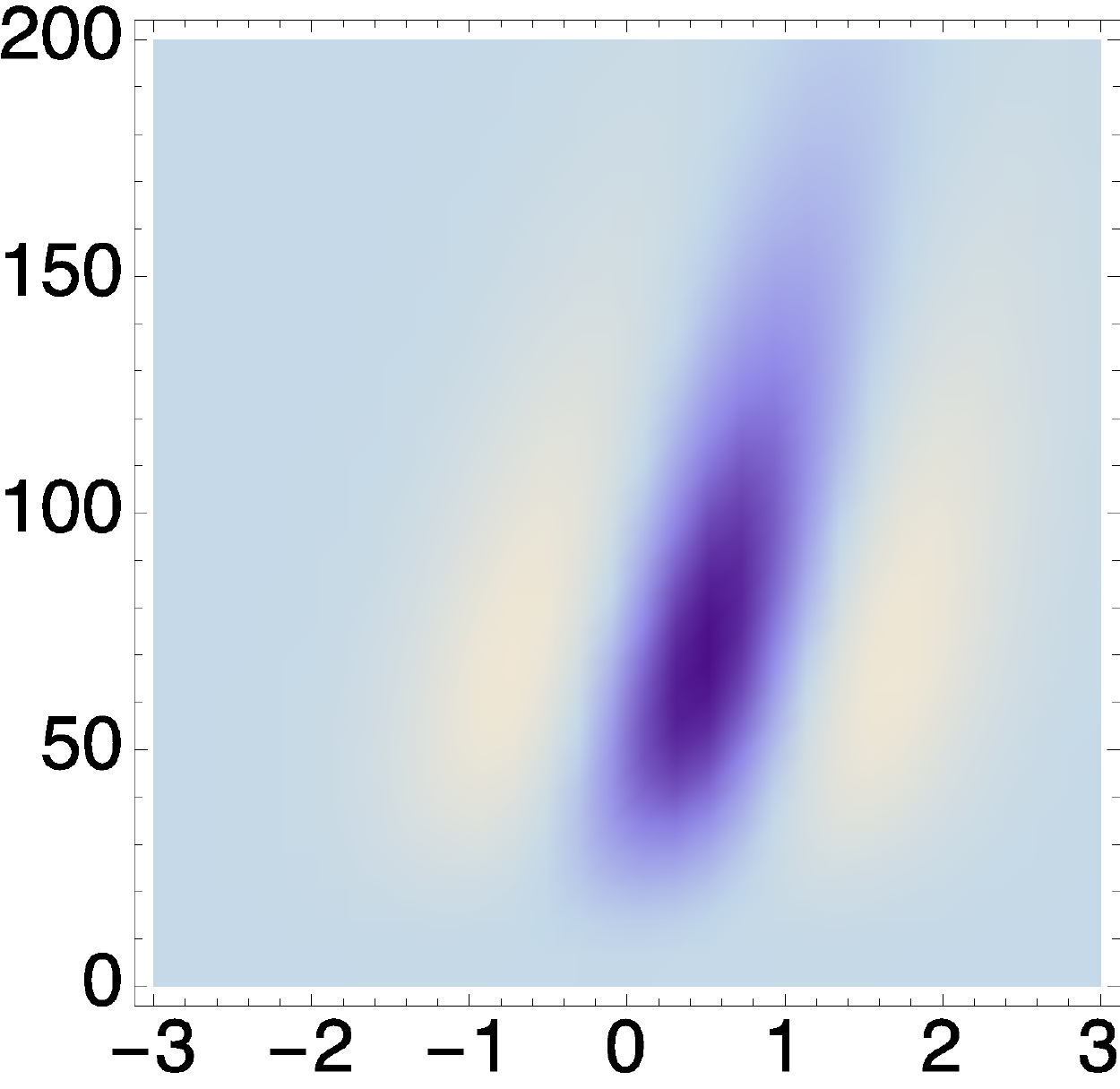}
      &
\hspace{4mm}  \includegraphics[width=0.24\textwidth]{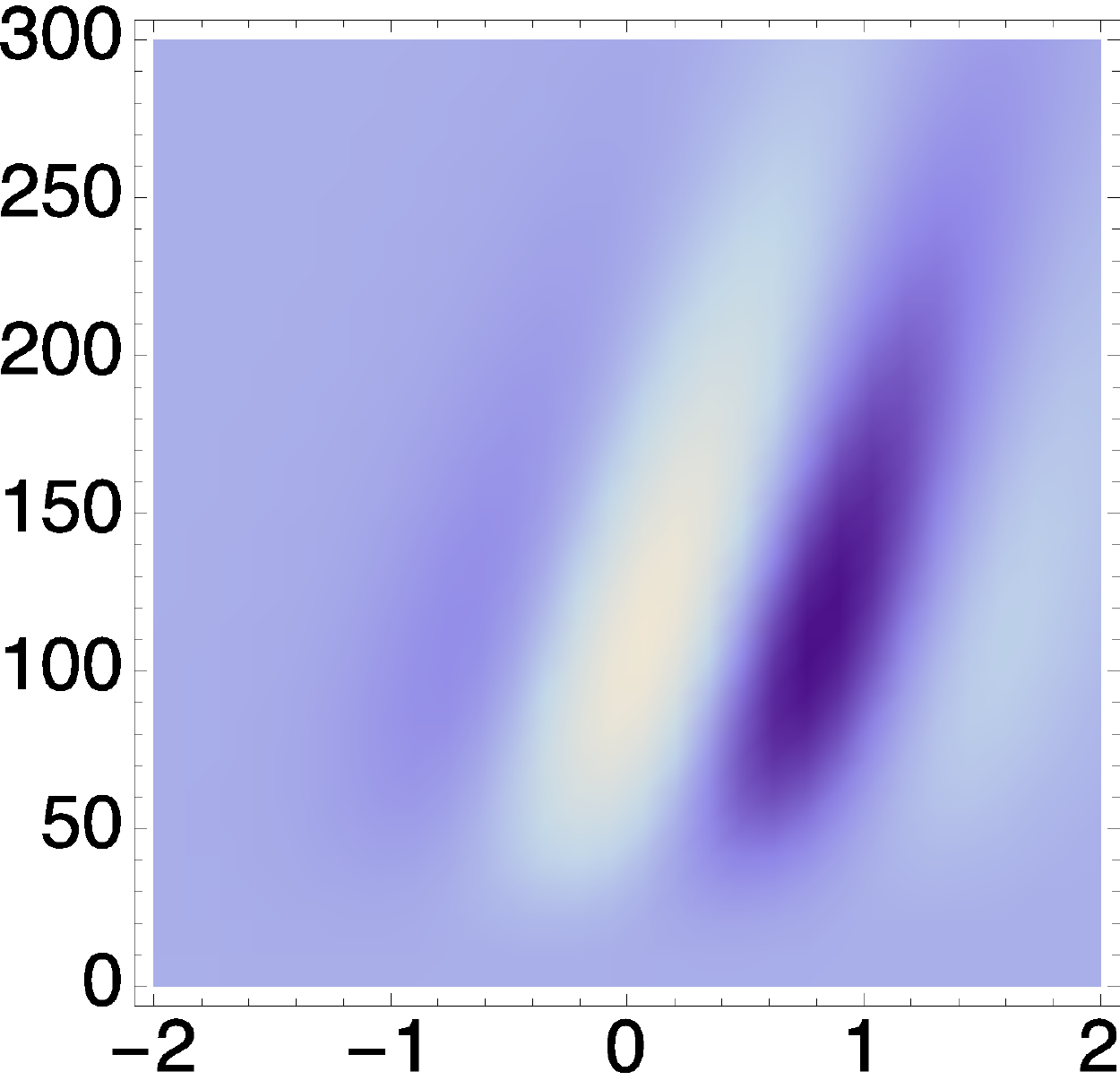}
    \end{tabular}
  \end{center}

  \caption{Computational modelling of simple cells in the primary
    visual cortex (V1) as reported
    by DeAngelis et al.\ \cite{DeAngOhzFre95-TINS} using
    idealized spatio-temporal receptive fields of the form
     $T(x, t;\; s, \tau, v)
           = \partial_{x^{\alpha}} \partial_{t^{\beta}} g(x - v t;\; s) \, h(t;\; \tau)$
    according to equation~(\protect\ref{eq-spat-temp-RF-model}) and
    with the temporal smoothing function $h(t;\; \tau)$
    modelled as a cascade of first-order integrators/truncated exponential
    kernels of the form (\protect\ref{eq-comp-trunc-exp-cascade}).
    (left column) Separable receptive fields corresponding to mixed
    derivatives of first- or second-order derivatives over space with
    first-order derivatives over time.
    (right column) Inseparable velocity-adapted receptive fields
    corresponding to second- or third-order derivatives over space.
    Parameter values: 
    (a) $h_{xt}$: $\sigma_x = 0.6$~degrees, $\sigma_t = 60$~ms.
    (b) $h_{xxt}$: $\sigma_x = 0.6$~degrees, $\sigma_t = 80$~ms.
    (c) $h_{xx}$: $\sigma_x = 0.7$~degrees, $\sigma_t = 50$~ms, $v = 0.007$~degrees/ms.
    (d) $h_{xxx}$: $\sigma_x = 0.5$~degrees, $\sigma_t = 80$~ms, $v = 0.004$~degrees/ms.
    (Horizontal axis: Space $x$ in degrees of visual angle. Vertical axis: Time $t$ in ms.)}
   \label{fig-biol-model-simple-cells-rec-filters-over-time}
\end{figure*}

\subsection{Distributions of the temporal scale levels}

When implementing this temporal scale-space concept, a set of
intermediate scale levels $\tau_k$ has to be distributed between some minimum
and maximum scale levels $\tau_{min} = \tau_1$ and $\tau_{max} =
\tau_K$. Next, we will present three ways of discretizing the temporal
scale parameter over $K$ temporal scale levels.

\paragraph{Uniform distribution of the temporal scales.}

If one chooses a uniform distribution of the
intermediate temporal scales
\begin{equation}
  \label{eq-distr-tau-values-uni}
  \tau_k = \frac{k}{K} \, \tau_{max}
\end{equation}
then the time constants of all the individual smoothing steps are given by
\begin{equation}
  \label{eq-mu-k-uni}
   \mu_k = \sqrt{\frac{\tau_{max}}{K}}.
\end{equation}

\paragraph{Logarithmic distribution of the temporal scales with free
  minimum scale.}

More natural is to distribute the temporal scale levels according to a geometric series,
corresponding to a uniform distribution in units of
effective temporal scale $\tau_{eff} = \log \tau$ (Lindeberg \cite{Lin92-PAMI}).
If we have a free choice of what minimum temporal
scale level $\tau_{min}$ to use, a natural way
of parameterizing these temporal scale levels is by using a distribution
parameter $c > 1$ 
\begin{equation}
  \label{eq-distr-tau-values}
  \tau_k = c^{2(k-K)} \tau_{max} \quad\quad (1 \leq k \leq K)
\end{equation}
which by equation~(\ref{eq-mean-var-trunc-exp-filters}) implies that time
constants of the individual first-order integrators should be given by
\begin{align}
  \begin{split}
     \label{eq-mu1-log-distr}
     \mu_1 & = c^{1-K} \sqrt{\tau_{max}}
  \end{split}\\
  \begin{split}
     \label{eq-muk-log-distr}
     \mu_k & = \sqrt{\tau_k - \tau_{k-1}} = c^{k-K-1} \sqrt{c^2-1} \sqrt{\tau_{max}} \quad (2 \leq k \leq K)
  \end{split}
\end{align}

\paragraph{Logarithmic distribution of the temporal scales with given
  minimum scale.}

If the temporal signal is on the other hand given at some minimum
temporal scale $\tau_{min}$, 
we can instead determine $c = \left( \frac{\tau_{max}}{\tau_{min}} \right)^{\frac{1}{2(K-1)}}$
 in (\ref{eq-distr-tau-values})
such that $\tau_1 = \tau_{min}$ 
and add $K - 1$ temporal scales with $\mu_k$ according to (\ref{eq-muk-log-distr}).

\paragraph{Logarithmic memory of the past.}

When using a logarithmic distribution of the temporal scale levels
according to either of the last two methods,
the different levels in the temporal scale-space representation at
increasing temporal scales will serve as a logarithmic memory of the
past, with qualitative similarity to the mapping of the
past onto a logarithmic time axis in the scale-time model by
Koenderink \cite{Koe88-BC}.
Such a logarithmic memory of the past can also be extended to later
stages in the visual hierarchy.

\subsection{Temporal receptive fields}

Figure~\ref{fig-trunc-exp-kernels-1D} shows graphs of such temporal scale-space kernels
that correspond to the same value of the composed variance,
using either a uniform distribution or a logarithmic distribution of
the intermediate scale levels.

In general, these kernels are all highly asymmetric for small values
of $K$, whereas the kernels based on a uniform distribution of the
intermediate temporal scale levels become gradually more symmetric
around the temporal maximum as $K$ increases.
The degree of continuity at the origin and the smoothness of
transition phenomena increase with $K$ such that
coupling of $K \geq 2$ kernels in cascade implies a
$C^{K-2}$-continuity of the temporal scale-space kernel.
To guarantee at least $C^1$-continuity of the temporal derivative computation
kernel at the origin, the order $n$ of differentiation of a temporal scale-space kernel
should therefore not exceed $K - 2$.
Specifically, the kernels based on a logarithmic
distribution of the intermediate scale levels (i)~have a higher degree of
temporal asymmetry which increases with the distribution parameter $c$
and (ii)~allow for faster
temporal dynamics compared to the kernels based on a uniform
distribution. 

In the case of a logarithmic distribution of the intermediate temporal
scale levels, the choice of the distribution parameter $c$ leads to a
trade-off issue in that smaller values of $c$ allow for a denser
sampling of the temporal scale levels, whereas larger values of $c$ 
lead to faster temporal dynamics and a more skewed shape of the temporal
receptive fields with larger deviations from the shape of Gaussian
derivatives of the same order.

\subsection{Computational modelling of biological receptive fields}
\label{sec-comp-model-biol-RF}

\paragraph{Receptive fields in the LGN.}

Regarding visual receptive fields in the lateral geniculate nucleus (LGN), 
DeAngelis {\em et al.\/}\ \cite{DeAngOhzFre95-TINS,deAngAnz04-VisNeuroSci}
report that most neurons
(i)~have approximately circular center-surround organization in the
 spatial domain and that 
(ii)~most of the receptive fields are separable in space-time.
There are two main classes of temporal responses for such cells:
(i)~a ``non-lagged cell'' is defined as a cell for which the first temporal lobe is the largest one
  (figure~\ref{fig-deang-tins-temp-resp-prof-lagged-nonlagged}(left)), whereas
(ii)~a ``lagged cell'' is defined as a cell for which the second lobe dominates
(figure~\ref{fig-deang-tins-temp-resp-prof-lagged-nonlagged}(right)).

Such temporal response properties are typical for {\em first- and
second-order temporal derivatives\/} of a time-causal temporal scale-space
representation.
For the first-order temporal derivative of a time-causal temporal
scale-space kernel, the first peak is strongest, 
whereas the second peak may be the most dominant one for
second-order temporal derivatives.
The spatial response, on the other hand, shows a high similarity
to a Laplacian of a Gaussian, leading to an idealized receptive field
model of the form (Lindeberg \cite[equation~(108)]{Lin13-BICY})
\begin{equation}
  \label{eq-lgn-model-1}
  h_{LGN}(x, y, t;\; s, \tau) 
  = \pm (\partial_{xx} + \partial_{yy}) \, g(x, y;\; s) \, \partial_{t^n} \, h(t;\; \tau).
\end{equation}
Figure~\ref{fig-deang-tins-temp-resp-prof-lagged-nonlagged} shows
results of modelling separable receptive fields in the LGN in this way,
using a cascade of first-order integrators/truncated exponential
kernels of the form (\protect\ref{eq-comp-trunc-exp-cascade}) for
modelling the temporal smoothing function $h(t;\; \tau)$.

\paragraph{Receptive fields in V1.}

Concerning the neurons in the primary visual cortex (V1), 
DeAngelis {\em et al.\/}
\cite{DeAngOhzFre95-TINS,deAngAnz04-VisNeuroSci}
describe that their receptive fields are generally different 
from the receptive fields in the LGN in the sense that they are
(i)~oriented in the spatial domain and 
(ii)~sensitive to specific stimulus velocities.
Cells (iii)~for which there are precisely localized ``on'' and ``off''
subregions with (iv)~spatial summation within each subregion,
(v)~spatial antagonism between on- and off-subregions and
(vi)~whose visual responses to stationary or moving spots can be
predicted from the spatial subregions are referred to as 
  simple cells (Hubel and Wiesel \cite{HubWie59-Phys,HubWie62-Phys}).
In Lindeberg \cite{Lin13-BICY}, an idealized model of such receptive
fields was proposed of the form
\begin{align}
  \begin{split}
       & h_{simple-cell}(x_1, x_2, t;\; s, \tau, v, \Sigma)  =
  \end{split}\nonumber\\
  \begin{split}
       & \quad (\cos \varphi \, \partial_{x_1} + \sin \varphi \, \partial_{x_2})^{m_1}
              (\sin \varphi \, \partial_{x_1} - \cos \varphi \, \partial_{x_2})^{m_2}
  \end{split}\nonumber\\
  \begin{split}
         & \quad  (v_1 \, \partial_{x_1} + v_2 \, \partial_{x_2} + \partial_t)^n
    \end{split}\nonumber\\
  \begin{split}
       \label{eq-simple-cell}
        & \quad g(x_1 - v_1 t, x_2 - v_2 t;\; s \, \Sigma) \, h(t;\; \tau)
   \end{split}
\end{align}
  where
  \begin{itemize}
  \item
      $\partial_{\varphi} = \cos \varphi \, \partial_{x_1} + \sin \varphi \, \partial_{x_2}$ 
     and
     $\partial_{\orth \varphi} = \sin \varphi \, \partial_{x_1} - \cos \varphi \, \partial_{x_2}$
     denote spatial directional derivative operators in two orthogonal directions $\varphi$
     and $\orth \varphi$,
  \item
     $m_1 \geq 0$ and $m_2 \geq 0$ denote the orders of differentiation in the two
     orthogonal directions in the spatial domain with the overall
     spatial order of differentiation $m = m_1 + m_2$,
  \item 
    $v_1 \, \partial_{x_1} + v_2 \, \partial_{x_2} + \partial_t$ denotes a
    velocity-adapted temporal derivative operator
\end{itemize}
and the meanings of the other symbols are similar as explained in
connection with equation~(\ref{eq-spat-temp-RF-model}).

Figure~\ref{fig-biol-model-simple-cells-rec-filters-over-time} shows
the result of modelling the spatio-temporal receptive fields of simple
cells in V1 in this way, using the general idealized model of
spatio-temporal receptive fields in equation~(\protect\ref{eq-spat-temp-RF-model})
in combination with a temporal smoothing kernel obtained by coupling a
set of first-order integrators or truncated exponential kernels in
cascade. As can be seen from the figures, the proposed idealized receptive
field models do well reproduce the qualitative shape of the
neurophysiologically recorded biological receptive fields.

These results complement the general theoretical model for visual
receptive fields in Lindeberg \cite{Lin13-BICY} by (i)~temporal kernels
that have better temporal dynamics than the time-causal semi-group 
derived in Lindeberg \cite{Lin10-JMIV} by decreasing faster with time
(decreasing exponentially instead of polynomially) and with (ii)~explicit modelling results
and a theory (developed in more detail in following sections)%
\footnote{The theoretical results following in section~\ref{sec-time-caus-limit-kernel} 
  state that temporal scale covariance becomes possible using a logarithmic
  distribution of the temporal scale levels. 
  Section~\ref{app-temp-dyn} states that the temporal response
  properties are faster for a logarithmic distribution of the
  intermediate temporal scale levels compared to a uniform
  distribution.  If one has requirements about how fine the temporal
  scale sampling needs to be or maximally allowed temporal delays, 
  then table~\ref{tab-tmax-uni-log-compare} in
  section~\ref{app-temp-dyn} provides constraints on permissable values of 
  the distribution parameter $c$. 
  Finally, the quantitative criterion in 
  section~\ref{sec-meas-dev-from-scinv-lim-kern} (see
  table~\ref{tab-rel-dev-from-limit-n1-n2}) states how many
  intermediate temporal scale levels 
  are needed to approximate temporal scale invariance up to a given accuracy.}
for choosing and parameterizing the intermediate 
discrete temporal scale levels in the time-causal model.

With regard to a possible biological implementation of this theory, the evolution
  properties of the presented scale-space models over scale and time are governed
  by diffusion and difference equations 
(see
equations~(\ref{eq-scale-evol-spat-scsp})--(\ref{eq-time-evol-veladapt-firstordint})
in the next section),
  which can be implemented by
  {\em operations over neighbourhoods\/} in combination with
  first-order integration over time.
  Hence, the computations can naturally be implemented in terms of
  {\em connections between different cells\/}.
  Diffusion equations are also used in mean field theory for
  approximating the computations that are performed by populations of neurons
  (Omurtag {\em et al.\/}\ \cite{OmuKniSir00-CompNeuro};
  Mattia and Guidice \cite{MatGui02-PhysRevE};
  Faugeras {\em et al.\/}\ \cite{FauTouCes09-FrontCompNeuroSci}).

By combination of the theoretical properties of these
kernels regarding scale-space properties between receptive field
responses at different spatial and temporal scales as well as their
covariance properties under natural image transformations
(described in more detail in the next section),
the proposed theory can be seen as a both theoretically well-founded
and biologically plausible model for time-causal and time-recursive
spatio-temporal receptive fields.

\subsection{Theoretical properties of time-causal spatio-temporal scale-space}

Under evolution of time and with increasing spatial scale, the
corresponding time-causal spatio-temporal scale-space representation 
generated by convolution with kernels of the form
(\ref{eq-spat-temp-RF-model})
with specifically the temporal smoothing kernel $h(t;\; \tau)$
defined as a set of truncated exponential kernels/first-order
integrators in cascade (\ref{eq-comp-trunc-exp-cascade}) obeys the
following system of differential/difference equations
\begin{align}
  \begin{split}
    \label{eq-scale-evol-spat-scsp}
    \partial_s L
      & = \frac{1}{2} \nabla_x^T (\Sigma \, \nabla_x L),
  \end{split}\\
  \begin{split}
    \label{eq-time-evol-veladapt-firstordint}
    \partial_t L
      & = - v^T (\nabla_x L) 
               - \frac{1}{\mu_k} \delta_{\tau} L,
  \end{split}
\end{align}
with the difference operator $\delta_{\tau}$ over temporal scale 
\begin{multline}
  (\delta_{\tau} L)(x, t;\; s, \tau_k;\; \Sigma, v) = \\
  L(x, t;\; s, \tau_{k};\; \Sigma, v) - L (x, t;\; s, \tau_{k-1};\; \Sigma, v).
\end{multline}
Theoretically, the resulting spatio-temporal scale-space
representation obeys similar
scale-space properties over the {\em spatial domain\/} as the two other
spatio-temporal scale-space models derived in Lindeberg
\cite{Lin10-JMIV,Lin13-BICY,Lin13-ImPhys} regarding 
(i)~linearity over the spatial domain,
(ii)~shift invariance over space, 
(iii)~semi-group and cascade properties over spatial scales,
(iv)~self-similarity and scale covariance over spatial scales so
that for any uniform scaling transformation $(x', t')^T = (S x, t)^T$ the
spatio-temporal scale-space representations are related by 
$L'(x', t';\; s', \tau_k;\; \Sigma, v') = L(x, t;\; s, \tau_k;\; \Sigma, v)$
with $s' = S^2 s$ and $v' = S v$ and
(v)~non-enhancement of local extrema with increasing spatial scale.

If the family of receptive fields in equation~(\ref{eq-spat-temp-RF-model})
is defined over the full group of positive definite spatial covariance matrices $\Sigma$ 
in the spatial affine Gaussian scale-space \cite{Lin93-Dis,LG96-IVC,Lin10-JMIV},
then the receptive field family also obeys (vi)~closedness and covariance under
time-independent affine transformations of the spatial image domain,
$(x', t')^T = (A x, t)^T$ implying
$L'(x', t';\; s, \tau_k;\; \Sigma', v') = L(x, t;\; s, \tau_k;\; \Sigma, v)$
with $\Sigma' = A\Sigma A^T$ and $v' = Av$,
and as resulting from {\em e.g.\/}\ local linearizations of the
perspective mapping (with locality defined as over the support region
of the receptive field).
When using rotationally symmetric Gaussian kernels for smoothing, the
corresponding spatio-temporal scale-space representation does instead
obey (vii)~rotational invariance.

Over the {\em temporal domain\/}, convolution with these kernels obeys
(viii)~linearity over the temporal domain,
(ix)~shift invariance over the temporal domain,
(x)~temporal causality,
(xi)~cascade property over temporal scales,
(xii)~non-creation of local extrema for any purely temporal signal.
If using a uniform distribution of the intermediate temporal scale
levels, the spatio-temporal scale-space representation obeys a
(xiii)~semi-group property
\footnote{When using a uniform distribution of the intermediate
  temporal scale levels, with temporal scale increment $\Delta \tau$ 
  between adjacent temporal scale levels, we can equivalently
  parameterize the temporal scale parameter by its temporal scale
  index $k$. Then, any temporal scale level $k$ corresponding to the
  composed temporal variance $\tau = k \, \Delta \tau$ is given by
  $L(t;\; k) = \left( *_{i = 1}^K h_{exp}(\cdot;\; \Delta \tau) \right) * f(\cdot))(t)$
  with the composed convolution kernel
  $h(\cdot;\; k) = *_{i = 1}^k h_{exp}(\cdot;\; \Delta \tau)$
  obeying the discrete semi-group property
  $h(\cdot;\; k_1) * h(\cdot;\; k_2) = h(\cdot;\; k_1 + k_2)$. 
  Parameterized over the temporal scale parameter $\tau$, the
  semi-group property does instead read 
$h(\cdot;\; k_1 \Delta \tau) * h(\cdot;\; k_2 \Delta \tau) 
  = h(\cdot;\; (k_1 + k_2) \Delta \tau)$, where $\Delta \tau = \mu^2$
  and $\mu$ is the time-constant of the first-order integrator.}
over discrete temporal scales.
Due to the finite number of discrete temporal scale levels,
the corresponding spatio-temporal scale-space representation cannot
however for general values of the time constants $\mu_k$ obey
full self-similarity and scale covariance over temporal scales.
Using a logarithmic distribution of the temporal scale levels and
an additional limit case construction to the infinity, we will however show in 
section~\ref{sec-time-caus-limit-kernel} that it is possible
to achieve (xiv)~self-similarity (\ref{eq-recur-rel-limit-kernel}) and scale covariance 
(\ref{eq-closedness-spattemp-temp-scaling-limit-kernel}) over
the discrete set of temporal
scaling transformations $(x', t')^T = (x, c^j t)^T$ that precisely corresponds to mappings between any
pair of discretized temporal scale levels as implied by the logarithmically
distributed temporal scale parameter with distribution parameter $c$.

Over the {\em composed spatio-temporal domain\/}, these kernels obey
(xv)~positivity and (xvi)~unit normalization in $L_1$-norm.
The spatio-temporal scale-space representation also obeys
(xvii)~closedness and covariance under local Galilean transformations in
space-time, in the sense that for any Gali\-lean transformation 
$(x', t')^T = (x - ut, t)^T$ with two video sequences
related by $f'(x', t') = f(x, t)$ their corresponding
spatio-temporal scale-space representations will be equal for
corresponding parameter values 
$L'(x', t';\; s, \tau_k;\; \Sigma, v') = L(x, t;\; s, \tau_k;\; \Sigma, v)$
with $v' = v-u$.

If additionally the velocity value $v$ and/or the spatial covariance matrix 
$\Sigma$ can be adapted to the local image structures in terms of
Galilean and/or affine invariant fixed point properties
\cite{Lin10-JMIV,LinAkbLap04-ICPR,Lin93-Dis,LG96-IVC},
then the spatio-temporal receptive field responses can additionally be made
(xviii)~Galilean invariant and/or (xix)~affine invariant.

\section{Temporal dynamics of the time-causal kernels}
\label{app-temp-dyn}

For the time-causal filters obtained by coupling truncated exponential
kernels in cascade, there will be an inevitable temporal delay 
depending on the time constants $\mu_k$ of the individual filters.
A straightforward way of estimating this delay is by using the
additive property of mean values under convolution
  $m_K = \sum_{k=1}^K \mu_k$ 
according to (\ref{eq-mean-var-trunc-exp-filters}).
In the special case when all the time constants are equal $\mu_k =
\sqrt{\tau/K}$, this measure is given by
\begin{equation}
  \label{eq-delta-recfilt-uni}
  m_{uni} = \sqrt{K \tau} 
\end{equation}
showing that the temporal delay increases if the temporal smoothing
operation is divided into a larger number of smaller individual smoothing steps.

In the special case when the intermediate temporal scale levels are
instead distributed logarithmically according to (\ref{eq-distr-tau-values}), 
with the individual time constants given by
(\ref{eq-mu1-log-distr}) and (\ref{eq-muk-log-distr}),
this measure for the temporal delay is given by
\begin{align}
  \begin{split}
  \label{eq-delta-recfilt-log}
  m_{log} 
   & = \frac{c^{-K} \left(c^2-\left(\sqrt{c^2-1}+1\right) c+\sqrt{c^2-1} \, c^K\right)}{c-1} \, \sqrt{\tau }
  \end{split}
\end{align}
with the limit value
\begin{equation}
  m_{log-limit} = \lim_{K \rightarrow \infty} m_{log} 
   = \sqrt{\frac{c+1}{c-1}} \sqrt{\tau}
\end{equation}
when the number of filters tends to infinity.

By comparing equations (\ref{eq-delta-recfilt-uni}) and (\ref{eq-delta-recfilt-log}),
we can specifically note that with increasing number of intermediate
temporal scale levels a logarithmic distribution of the intermediate
scales implies shorter temporal delays than a uniform
distribution of the intermediate scales.

Table~\ref{tab-tmean-uni-log-compare} shows numerical values
of these measures for different values of $K$ and $c$.
As can be seen, the logarithmic
distribution of the intermediate scales allows for significantly
faster temporal dynamics than a uniform distribution.

\begin{table}[!hbt]
  \begin{center}
   \footnotesize
  \begin{tabular}{ccccc}
  \hline
   \multicolumn{5}{c}{Temporal mean values $m$ of time-causal kernels} \\
  \hline
    $K$ & $m_{uni}$  & $m_{log}$  ($c = \sqrt{2}$) & $m_{log}$  ($c = 2^{3/4}$) & $m_{log}$ ($c = 2$) \\
  \hline
    2 & 1.414 & 1.414 & 1.399 & 1.366 \\
    3 & 1.732 & 1.707 & 1.636 & 1.549 \\
    4 & 2.000 & 1.914 & 1.777 & 1.641 \\
    5 & 2.236 & 2.061 & 1.860 & 1.686 \\
    6 & 2.449 & 2.164 & 1.910 & 1.709 \\
    7 & 2.646 & 2.237 & 1.940 & 1.721 \\
    8 & 2.828 & 2.289 & 1.957 & 1.726 \\
    9 & 3.000 & 2.326 & 1.968 & 1.729 \\
    10 & 3.162 & 2.352 & 1.974 & 1.730 \\
    11 & 3.317 & 2.370 & 1.978 & 1.731 \\
    12 & 3.464 & 2.383 & 1.980 & 1.732 \\

  \hline
  \end{tabular}
\end{center}
\caption{Numerical values of the {\em temporal delay
    in terms of the temporal mean\/} $m = \sum_{k=1}^K \mu_k$
    in units of $\sigma = \sqrt{\tau}$ for time-causal kernels obtained by coupling $K$ truncated
    exponential kernels in cascade in the cases of a uniform
    distribution of the intermediate temporal scale levels $\tau_k = k \tau/K$ or a
    logarithmic distribution $\tau_k = c^{2(k-K)} \tau$.}
  \label{tab-tmean-uni-log-compare}

\medskip

\begin{center}
   \footnotesize
  \begin{tabular}{ccccc}
  \hline
   \multicolumn{5}{c}{Temporal delays $t_{max}$ from the maxima of time-causal kernels} \\
  \hline
    $K$ & $t_{uni}$  & $t_{log}$ ($c = \sqrt{2}$) & $t_{log}$ ($c = 2^{3/4}$) & $t_{log}$ ($c = 2$) \\
  \hline
    2 & 0.707 & 0.707 & 0.688 & 0.640 \\
    3 & 1.154 & 1.122 & 1.027 & 0.909 \\
    4 & 1.500 & 1.385 & 1.199 & 1.014 \\
    5 & 1.789 & 1.556 & 1.289 & 1.060 \\
    6 & 2.041 & 1.669 & 1.340 & 1.083 \\
    7 & 2.268 & 1.745 & 1.370 & 1.095 \\
    8 & 2.475 & 1.797 & 1.388 & 1.100 \\
    9 & 2.667 & 1.834 & 1.398 &  1.103 \\
 10 & 2.846 & 1.860 & 1.404 &  1.104 \\
 11 & 3.015 & 1.879 & 1.408 &  1.105 \\
 12 & 3.175 & 1.892 & 1.410 &  1.106 \\
  \hline
  \end{tabular}
\end{center}
\caption{Numerical values for the {\em temporal delay of the local maximum\/} in
  units of $\sigma = \sqrt{\tau}$
    for time-causal kernels obtained by coupling $K$ truncated
    exponential kernels in cascade in the cases of a uniform
    distribution of the intermediate temporal scale levels $\tau_k = k \tau/K$ or a
    logarithmic distribution $\tau_k = c^{2(k-K)} \tau$ with $c > 1$.}
  \label{tab-tmax-uni-log-compare}
\end{table}

\paragraph{Additional temporal characteristics.}

Because of the asymmetric tails of the time-causal temporal smoothing
kernels, temporal delay estimation by the mean value may however lead
to substantial overestimates compared to {\em e.g.\/} the position of the local maximum.
To provide more precise characteristics, let us first consider the case of a uniform
distribution of the intermediate temporal scales, for which
a compact closed form expression is available for the composed kernel
and corresponding to the probability density function of the Gamma distribution
\begin{equation}
 \label{eq-composed-all-mu-equal}
  h_{composed}(t;\; \mu, K) = \frac{t^{K-1} \, e^{-t/\mu}}{\mu^K \, \Gamma(K)}.
\end{equation}
The temporal derivatives of these kernels relate to Laguerre
functions (Laguerre polynomials $p_n^{\alpha}(t)$ multiplied by a truncated
exponential kernel) according to Rodrigues formula:
\begin{equation}
  p_n^{\alpha}(t) \, e^{-t} = \frac{t^{-\alpha}}{n!} \, \partial_t^n (t^{n+\alpha} e^{-t}).
\end{equation}
Let us differentiate the temporal smoothing kernel
\begin{equation}
    \partial_t \left( h_{composed}(t;\; \mu, K) \right)
    = \frac{e^{-\frac{t}{\mu }} ((K-1) \mu -t) \left(\frac{t}{\mu
        }\right)^{K+1}}{t^3 \, \Gamma(K)}
\end{equation}
and solve for the position of the local maximum
\begin{align}
  \begin{split}
     \label{eq-tmax-recfilt-uni}
     t_{max,uni} & = (K-1) \, \mu 
                = \frac{(K-1) }{\sqrt{K}} \sqrt{\tau}.
  \end{split}
\end{align}
Table~\ref{tab-tmax-uni-log-compare} shows numerical values for the
position of the local maximum for both types of time-causal kernels.
As can be seen from the data, the temporal response properties are
significantly faster for a logarithmic distribution of the
intermediate scale levels compared to a uniform distribution
and the difference increases rapidly with $K$.
These temporal delay estimates are also significantly shorter than the
temporal mean values, in particular for the logarithmic distribution.

If we consider a temporal event that occurs as a step function over
time ({\em e.g.\/} a new object appearing in the field of view) and 
if the time of this event is estimated from the local maximum over
time in the first-order temporal derivative response, 
then the temporal variation in the response over time will be given by
the shape of the temporal
smoothing kernel. The local maximum over time will occur at a time
delay equal to the time at which the temporal kernel has its maximum
over time. Thus, the position of the maximum over time of the
temporal smoothing kernel is highly relevant for quantifying the
temporal response dynamics. 

\section{The scale-invariant time-causal limit kernel}
\label{sec-time-caus-limit-kernel}

In this section, we will show that in the case of a logarithmic
distribution of the intermediate temporal scale levels it is possible
to extend the previous temporal scale-space concept into a limit case
that permits for covariance under temporal scaling transformations,
corresponding to closedness of the temporal scale-space representation
to a compression or stretching of the temporal scale axis by
any integer power of the distribution parameter $c$.

Concerning the need for temporal scale invariance of a temporal
scale-space representation, let us first note that one could
possibly first argue that the need for temporal scale invariance in a 
temporal scale-space representation is different from the need for
spatial scale invariance in a spatial scale-space representation.
Spatial scaling transformations always occur because of 
perspective scaling effects caused by variations in the distances
between objects in the world and the observer and do therefore 
always need to be handled by a vision system, whereas the temporal scale remains 
unaffected by the perspective mapping from the scene to the image.

Temporal scaling transformations are, however, nevertheless important
because of physical phenomena or spatio-temporal events occurring faster or
slower. This is analogous to another source of scale variability over the spatial domain,
caused by objects in the world having different physical size.
To handle such scale variabilities over the temporal domain,
it is therefore desirable to develop temporal scale-space
concepts that allow for temporal scale invariance.

\paragraph{Fourier transform of temporal scale-space kernel.}

When using a logarithmic distribution of the intermediate scale levels (\ref{eq-distr-tau-values}),
the time constants of the individual first-order integrators are given by (\ref{eq-mu1-log-distr})
and (\ref{eq-muk-log-distr}).
Thus, the explicit expression for the Fourier transform obtained by setting $q = i \omega$ in
the expression (\ref{eq-FT-composed-kern-casc-truncexp}) 
is of the form 
\begin{multline}
  \label{eq-FT-comp-kern-log-distr}
    \hat{h}_{exp}(\omega;\; \tau, c, K) 
       = \\ \frac{1}{1 + i \, c^{1-K} \sqrt{\tau} \, \omega}
               \prod_{k=2}^{K} \frac{1}{1 + i \, c^{k-K-1} \sqrt{c^2-1} \sqrt{\tau} \, \omega}.
\end{multline}

\paragraph{Characterization in terms of temporal moments.}

Although the explicit expression for the composed time-causal kernel may be somewhat
cumbersome to handle for any finite number of $K$,
in appendix~\ref{app-freq-anal-time-caus-kern-log-distr} we show 
how one based on a Taylor expansion of the Fourier transform can derive
compact closed-form moment or cumulant descriptors of these time-causal 
scale-space kernels.
Specifically, the limit values of the first-order moment $M_1$ and the higher-order central
moments up to order four when the number of temporal scale levels $K$
tends to infinity are given by
\begin{align}
  \begin{split}
      \lim_{K \rightarrow \infty} M_1 & = \sqrt{\frac{c+1}{c-1}}\, \tau^{1/2}
  \end{split}\\
  \begin{split}
      \lim_{K \rightarrow \infty} M_2 & = \tau
  \end{split}\\
  \begin{split}
    \lim_{K \rightarrow \infty} M_3 & = \frac{2 (c+1) \sqrt{c^2-1} \, \tau^{3/2}}{\left(c^2+c+1\right)}
  \end{split}\\
 \begin{split}
      \lim_{K \rightarrow \infty} M_4 & =\frac{3 \left(3 c^2-1\right) \tau ^2}{c^2+1}
  \end{split}
\end{align}
and give a coarse characterization of the limit behaviour of these kernels 
essentially corresponding to the terms in a Taylor expansion of the Fourier transform up to order four.
Following a similar methodology, explicit expressions for higher-order moment 
descriptors can also be derived in an analogous fashion, from the Taylor coefficients
of higher order, if needed for special purposes.

In figure~\ref{fig-skew-kurt-explogdistr} in
appendix~\ref{app-freq-anal-time-caus-kern-log-distr} we show graphs
of the corresponding skewness and kurtosis measures as function
of the distribution parameter $c$, showing that both these measures
increase with the distribution parameter $c$.
In figure~\ref{fig-skew-kurt-compare-truncexpcasc-scaletime} in
appendix~\ref{app-comp-koe-model} we provide a comparison between
the behaviour of this limit kernel and the temporal kernel in
Koenderink's scale-time model showing that although the 
temporal kernels in these two models to a first approximation
share qualitatively coarsely similar properties in terms of their overall shape
(see figure~\ref{fig-trunc-exp-kernels-1D+scaletime} in appendix~\ref{app-comp-koe-model}), 
the temporal kernels in these two models differ significantly in terms 
of their skewness and kurtosis measures.

\paragraph{The limit kernel.}

By letting the number of temporal scale levels $K$ tend to infinity, we can define
a limit kernel $\Psi(t;\; \tau, c)$ via the limit of the Fourier transform (\ref{eq-FT-comp-kern-log-distr}) according to (and with the indices relabelled to better fit the limit case):
\begin{align}
  \begin{split}
     \hat{\Psi}(\omega;\; \tau, c) 
     & = \lim_{K \rightarrow \infty} \hat{h}_{exp}(\omega;\; \tau, c, K) 
  \end{split}\nonumber\\
  \begin{split}
     \label{eq-FT-comp-kern-log-distr-limit}
     & = \prod_{k=1}^{\infty} \frac{1}{1 + i \, c^{-k} \sqrt{c^2-1} \sqrt{\tau} \, \omega}.
  \end{split}
\end{align}
By treating this limit kernel as an object by itself, which will be well-defined because of the rapid convergence by the summation of variances according to a geometric series, interesting relations can be expressed
between the temporal scale-space representations
\footnote{Concerning the definition of the
  temporal scale-space representation
  (\ref{eq-temp-scsp-conv-limit-kernel}) obtained by convolution with
  the limit kernel (\ref{eq-FT-comp-kern-log-distr-limit}), it should
be noted that although these definitions formally hold for any values
of $\tau$ and $c$, the information reducing property in terms
non-creation of new local extrema or zero-crossings from finer to
coarser scales is only guaranteed
to hold if the transformation between two temporal scale levels
$\tau_2 > \tau_1$ can be written on the form 
$L(\cdot;\; \tau_2, c_2) = h(\cdot;\; (\tau_1, c_1) \mapsto (\tau_2, c_2)) * L(\cdot;\; \tau_1, c_1)$ 
with $h(\cdot;\; (\tau_1, c_1) \mapsto (\tau_2, c_2))$ being a
temporal scale-space kernel of the form
(\ref{eq-char-var-dim-kernels-cont-case-Laplace}).
Such an information reducing property is always guaranteed to hold for
temporal scale levels of the form
(\ref{eq-temp-scale-levels-limit-kernel}) with $c_1 = c_2 = c$, but
does in general not hold for arbitrary combinations of $(\tau_1, c_1)$
and $(\tau_2, c_2)$. Therefore, the definitions
(\ref{eq-temp-scsp-conv-limit-kernel}) and
(\ref{eq-FT-comp-kern-log-distr-limit}) are primarily intended to be
applied over a discrete set of temporal scale levels of the form (\ref{eq-temp-scale-levels-limit-kernel}).}
\begin{equation}
  \label{eq-temp-scsp-conv-limit-kernel}
  L(t;\; \tau, c) = \int_{u = 0}^{\infty} \Psi(u;\; \tau, c) \, f(t-u) \, du
\end{equation}
obtained by convolution with this limit kernel.

\paragraph{Self-similar recurrence relation for the limit kernel over temporal scales.}

Using the limit kernel, an infinite number of discrete temporal scale levels
is implicitly defined given the specific choice of one temporal scale $\tau = \tau_0$:
\begin{equation}
  \label{eq-temp-scale-levels-limit-kernel}
  \dots \frac{\tau_0}{c^6}, \frac{\tau_0}{c^4}, \frac{\tau_0}{c^2}, \tau_0,
  c^2 \tau_0, c^4 \tau_0, c^6 \tau_0, \dots 
\end{equation}
Directly from the definition of the limit kernel, we obtain the following recurrence relation
between adjacent scales:
\begin{equation}
  \label{eq-recur-rel-limit-kernel}
   \Psi(\cdot;\; \tau, c) = h_{exp}(\cdot;\; \tfrac{\sqrt{c^2-1}}{c} \sqrt{\tau}) * \Psi(\cdot;\; \tfrac{\tau}{c^2}, c)
\end{equation}
and in terms of the Fourier transform:
\begin{equation}
  \label{eq-recur-rel-limit-kernel-FT}
   \hat{\Psi}(\omega;\; \tau, c) 
   = \frac{1}{1 + i \, \tfrac{\sqrt{c^2-1}}{c} \sqrt{\tau} \, \omega} \,
       \hat{\Psi}(\omega;\; \tfrac{\tau}{c^2}, c).
\end{equation}

\paragraph{Behaviour under temporal rescaling transformations.}

From the Fourier transform of the limit kernel (\ref{eq-FT-comp-kern-log-distr-limit}),
we can observe that for any temporal scaling factor $S$ it holds that
\begin{equation}
  \label{eq-sc-transf-limit-kernel-FT}
   \hat{\Psi}(\tfrac{\omega}{S};\; S^2 \tau, c) = \hat{\Psi}(\omega;\; \tau, c).
\end{equation}
Thus, the limit kernel transforms as follows under a scaling
transformation of the temporal domain:
\begin{equation}
  \label{eq-sc-transf-limit-kernel}
   S \, \Psi(S \, t;\; S^2 \tau, c) = \Psi(t;\; \tau, c).
\end{equation}
If we for a given choice of distribution parameter $c$ rescale the 
input signal $f$ by a scaling factor $S = 1/c$ such that $t' = t/c$, it then follows that 
the scale-space representation of $f'$ at temporal scale $\tau' = \tau/c^2$
\begin{equation}
  L'(t';\; \tfrac{\tau}{c^2}, c) = (\Psi(\cdot;\; \tfrac{\tau}{c^2}, c) * f'(\cdot))(t';\; \tfrac{\tau}{c^2}, c)
\end{equation}
will be equal to the temporal scale-space representation of the
original signal $f$ at scale $\tau$
\begin{equation}
  \label{eq-scale-cov-limit-kernel-scsp}
  L'(t';\; \tau', c) = L(t;\; \tau, c).
\end{equation}
Hence, under a rescaling of the original signal by a scaling factor $c$,
a rescaled copy of the temporal scale-space representation of the 
original signal can be found at the next lower discrete temporal scale relative 
to the temporal scale-space representation of the original signal.

Applied recursively, this result implies that 
the temporal scale-space representation
obtained by convolution with the limit kernel {\em obeys a 
closedness property over all temporal scaling transformations $t' = c^j t$ 
with temporal rescaling factors $S = c^{j}$ ($j \in \bbbz$) that are 
integer powers of the distribution parameter $c$\/},
\begin{equation}
  L'(t';\; \tau', c) = L(t;\; \tau, c) \quad\mbox{for}\quad t' = 
c^j t \quad\mbox{and} \quad \tau' = c^{2j} \tau,
\end{equation} 
allowing for perfect scale invariance over the restricted subset of scaling factors
that precisely matches the specific set
of discrete temporal scale levels that is defined by a specific choice of the
distribution parameter $c$.
Based on this desirable and highly useful property, it is natural to refer to the limit kernel 
as {\em the scale invariant time-causal limit kernel\/}.

Applied to the spatio-temporal scale-space representation defined by convolution
with a velocity-adapted affine Gaussian kernel $g(x-vt;\; s, \Sigma)$ over space and
the limit kernel $\Psi(t;\; \tau, c)$ over time  
\begin{multline}
  L(x, t;\; s, \tau, c;\; \Sigma, v)
  = \\ \int_{\eta \in \bbbr^2} \int_{\zeta = 0}^{\infty} 
      g(\eta - v \zeta;\; s, \Sigma) \, \Psi(\zeta ;\; \tau, c) \, 
     f(x - \eta, t - \zeta) \, d\eta \, d\zeta,
\end{multline}
the corresponding
spatio-temporal scale-space representation will then under a scaling 
transformation of time $(x', t')^T = (x, c^j t)^T$ obey the closedness
property
\begin{equation}
  \label{eq-closedness-spattemp-temp-scaling-limit-kernel}
  L'(x', t';\; s, \tau', c;\; \Sigma, v') = L(x, t;\; s, \tau, c;\; \Sigma, v)
\end{equation}
with $\tau' = c^{2j} \tau$ and $v' = v/c^j$.

\paragraph{Self-similarity and scale invariance of the limit kernel.}

Combining the recurrence relations of the limit kernel with
its transformation property under scaling transformations,
it follows that the limit kernel can be regarded as truly self-similar
over scale in the sense that: (i)~the scale-space representation at a coarser temporal 
scale (here $\tau$) can be recursively computed from the scale-space representation
at a finer temporal scale (here $\tau/c^2$) according to (\ref{eq-recur-rel-limit-kernel}),
(ii)~the representation at the coarser temporal scale is derived from the
input in a functionally similar way as the representation at the finer temporal scale and
(iii)~the limit kernel and its Fourier transform are transformed in a self-similar way 
(\ref{eq-sc-transf-limit-kernel}) and (\ref{eq-sc-transf-limit-kernel-FT})  under scaling transformations.

In these respects, the temporal receptive fields arising from temporal
derivatives of the limit kernel share structurally similar mathematical
properties as continuous wavelets 
(Daubechies \cite{Dau92-book}; Heil and Walnut \cite{HeiWal99-SIAM};
Mallat \cite{Mal99-book}; 
Misiti {\em et al.\/} \cite{MisMisOppPog07-book})
and fractals (Mandelbrot \cite{Man82-book}; Barnsley \cite{Bar88-BOOK}; Barnsley and Rising \cite{BarRis93-book}),
while with the here conceptually novel extension that the scaling behaviour and self-similarity over scale is
achieved over a time-causal and time-recursive temporal domain.

\section{Computational implementation}
\label{sec-comp-impl}

The computational model for spatio-temporal receptive fields presented here
is based on spatio-temporal image data that are assumed to be
continuous over time.
When implementing this model on sampled video data,
the continuous theory must be transferred to discrete space and
discrete time.

In this section we describe how the temporal and spatio-temporal
receptive fields can be implemented in terms of corresponding discrete
scale-space kernels that possess scale-space properties over discrete
spatio-temporal domains.

\subsection{Classification of scale-space kernels for discrete
  signals}
\label{sec-class-disc-scsp-kern}

In section~\ref{sec-class-cont-scsp-kernels}, we described how
the class of continuous scale-space kernels over a one-dimensional
domain can be classified based on classical results by Schoenberg 
regarding the theory of variation-diminishing transformations
as applied to the construction of discrete scale-space theory in 
Lindeberg \cite{Lin90-PAMI} \cite[section~3.3]{Lin93-Dis}.
To later map the temporal smoothing operation to theoretically
well-founded discrete scale-space kernels, we shall in this section
describe corresponding classification result regarding scale-space
kernels over a discrete temporal domain.

\paragraph{Variation diminishing 
transformations.}

Let $v = (v_1, v_2, \dots, v_n)$ be a vector of $n$ real numbers and
let $V^-(v)$ denote the (minimum)
number of sign changes obtained in the sequence
$v_1, v_2, \dots, v_n$ if all zero terms are deleted.
Then, based on a 
result by Schoenberg \cite{Sch48} the convolution transformation
  \begin{equation}
    f_{\mbox{\scriptsize\em out}}(t) =
     \sum_{n = -\infty}^{\infty} c_{n} f_{\mbox{\scriptsize\em in}}(t-n)
  \end{equation}
  is variation-diminishing i.e.
  \begin{equation}
    V^-(f_{\mbox{\scriptsize\em out}}) \leq V^-(f_{\mbox{\scriptsize\em in}})
  \end{equation}
  holds for all $f_{\mbox{\scriptsize\em in}}$
  if and only if the generating function of the sequence of filter coefficients
  $\varphi(z) = \sum_{n=-\infty}^{\infty} c_n z^n$ is of the form
  \begin{equation}
    \label{eq-char-pf}
    \varphi(z) = c \; z^k \; e^{(q_{-1}z^{-1} + q_1z)}
                 \prod_{i=1}^{\infty} \frac{(1+\alpha_i z)(1+\delta_i
    z^{-1})} {(1-\beta_i z)(1-\gamma_i z^{-1})}
  \end{equation}
where
$c > 0$, $k \in \bbbz$, 
$q_{-1}, q_1, \alpha_i, \beta_i, \gamma_i, \delta_i \geq 0$ and
$\sum_{i=1}^{\infty}(\alpha_i + \beta_i + \gamma_i + \delta_i) < \infty$.
Interpreted over the temporal domain, this means that besides trivial
rescaling and translation, there are three basic classes of discrete
smoothing transformations: 
  \begin{itemize}
  \item
    {two-point weighted average} or {generalized binomial smoothing}
    \begin{equation}
      \begin{split}
        f_{\mbox{\scriptsize\em out}}(x)
        & =
          f_{\mbox{\scriptsize\em in}}(x) +
          \alpha_i \, f_{\mbox{\scriptsize\em in}}(x - 1)
          \quad (\alpha_i \geq 0),\\
        f_{out}(x)
        & =
          f_{\mbox{\scriptsize\em in}}(x) +
          \delta_i \, f_{\mbox{\scriptsize\em in}}(x + 1)
          \quad (\delta_i \geq 0),
      \end{split}
    \end{equation}
  \item
    moving average or first-order recursive filtering
    \begin{equation}
      \begin{split}
        f_{\mbox{\scriptsize\em out}}(x)
        & =
        f_{\mbox{\scriptsize\em in}}(x) +
          \beta_i \, f_{\mbox{\scriptsize\em out}}(x - 1) \quad
          (0 \leq \beta_i < 1), \\
        f_{\mbox{\scriptsize\em out}}(x)
        & =
          f_{\mbox{\scriptsize\em in}}(x) +
            \gamma_i \, f_{\mbox{\scriptsize\em out}}(x + 1) \quad
            (0 \leq \gamma_i < 1),
      \end{split}
    \end{equation}
  \item
   infinitesimal smoothing%
\footnote{These kernels correspond to infinitely divisible
  distributions as can be described with the theory of L{\'e}vy processes
  \cite{Sat99-Book}, where specifically the case $q_{-1} = q_1$ 
corresponds to convolution with the non-causal discrete analogue of the Gaussian
kernel \cite{Lin90-PAMI} and the case $q_{-1} = 0$ to convolution with time-causal Poisson kernel
\cite{LF96-ECCV}.}
 or diffusion as arising from the
   continuous semi-groups made possible by the factor\newline
   $e^{(q_{-1}z^{-1} + q_1z)}$.
\end{itemize}
To transfer the continuous first-order integrators derived in
section~\ref{sec-cont-temp-scsp-kern} to a discrete implementation, 
we shall in this treatment focus on the first-order recursive
filters, which by additional normalization constitute both the discrete
correspondence and a numerical approximation of time-causal and time-recursive first-order
temporal integration (\ref{eq-first-ord-int}).

\subsection{Discrete temporal scale-space kernels based on
  recursive filters}
\label{app-disc-temp-smooth}

Given video data that has been sampled by some temporal
frame rate $r$, the temporal
scale $\sigma_t$ in the continuous model in units of seconds is first 
transformed to a variance $\tau$ relative to a unit time sampling
\begin{equation}
  \label{eq-transf-tau-sampl}
  \tau = r^2 \, \sigma_t^2
\end{equation}
where $r$ may typically be either 25 fps or 50 fps.
Then, a discrete set of intermediate temporal scale levels $\tau_k$ is defined by
(\ref{eq-distr-tau-values})
or (\ref{eq-distr-tau-values-uni})
with the difference between successive scale levels according to 
  $\Delta \tau_k = \tau_k - \tau_{k-1}$ (with $\tau_0 = 0$).

For implementing the temporal smoothing operation between two such
adjacent scale levels (with the lower level in each pair of adjacent
scales referred to as $f_{in}$ and
the upper level as $f_{out}$), we make use of a {\em first-order
  recursive filter\/}
normalized to the form
\begin{equation}
  \label{eq-norm-update}
  f_{out}(t) - f_{out}(t-1)
  = \frac{1}{1 + \mu_k} \,
    (f_{in}(t) - f_{out}(t-1))
\end{equation}
and having a generating function of the form
\begin{equation}
  \label{eq-gen-fcn-first-order-rec-filt}
  \htransf_{geom}(z) = \frac{1}{1 - \mu_k \, (z - 1)}
\end{equation}
which is a time-causal kernel and satisfies discrete
scale-space properties of guaranteeing that the number of local extrema
or zero-crossings in the signal will not increase with increasing scale
(Lindeberg \cite{Lin90-PAMI}; Lindeberg and Fagerstr{\"o}m
\cite{LF96-ECCV}).
These recursive filters are the discrete analogue of the continuous 
first-order integrators~(\ref{eq-first-ord-int}).
Each primitive recursive filter (\ref{eq-norm-update}) has temporal mean value $m_k = \mu_k$ and temporal variance
$\Delta \tau_k = \mu_k^2 + \mu_k$, and we compute $\mu_k$ from 
$\Delta \tau_k$ according to
\begin{equation}
  \mu_k = \frac{\sqrt{1 + 4 \Delta \tau_k}-1}{2}.
\end{equation}
By the additive property of variances under convolution,
the discrete variances of
the discrete temporal scale-space kernels will perfectly match those
of the continuous model, whereas the mean values and the temporal
delays may differ somewhat. 
If the temporal scale $\tau_k$ is large relative to the
temporal sampling density, the discrete model should be a good approximation in this respect.

By the time-recursive formulation of this temporal scale-space
concept, the computations can be performed based on a
compact temporal buffer over time, which contains the temporal
scale-space representations at temporal scales $\tau_k$ and with
no need for storing any additional temporal buffer of what
has occurred in the past to perform the corresponding temporal operations.

Concerning the actual implementation of these operations
computationally on signal processing hardware
of software with built-in support for higher order recursive
filtering, one can specifically note the following: If one is only interested in the receptive field response
at a single temporal scale, then one can combine a set of $K'$
first-order recursive filters (\ref{eq-norm-update}) into a higher
order recursive filter by multiplying their generating functions (\ref{eq-gen-fcn-first-order-rec-filt})
\begin{align}
  \begin{split}
     \htransf_{composed}(z) 
     & = \prod_{k=1}^{K'} \frac{1}{1 - \mu_k \, (z - 1)}
  \end{split}\nonumber\\
  \begin{split}
     \label{eq-gen-fcn-composed-rec-filt}
     & = \frac{1}{a_0 + a_1 \, z + a_2 \, z^2 + \dots + a_{K'} \, z^{K'}}
  \end{split}
\end{align}
thus performing $K'$ recursive filtering steps by a single call to the
signal processing hardware or software.
If using such an approach, it should be noted, however, that depending on the
internal implementation of this functionality in the signal processing
hardware/software, the
composed call (\ref{eq-gen-fcn-composed-rec-filt}) may not be as
numerically well-conditioned as the individual smoothing steps (\ref{eq-norm-update}) which
are guaranteed to dampen any local perturbations.
In our Matlab implementation for offline processing of this receptive
field model, we have therefore limited the number of compositions to
$K' = 4$.

\subsection{Discrete implementation of spatial Gaussian smoothing}
\label{app-disc-gauss-smooth}

To implement the spatial Gaussian operation on discrete sampled data, we do first
transform a spatial scale parameter $\sigma_x$ in units of {\em e.g.\/} degrees 
of visual angle to a spatial variance $s$ relative to 
a unit sampling density according to
\begin{equation}
  \label{eq-transf-tau-sampl-s}
  s = p^2 \sigma_x^2
\end{equation}
where $p$ is the number of pixels per spatial unit {\em e.g.\/} in
terms of degrees of visual angle at the image center. Then, we
convolve the image data with the separable two-dimensional {\em discrete analogue of the
Gaussian kernel\/} (Lindeberg \cite{Lin90-PAMI}) 
\begin{equation}
  \label{eq-disc-analog-Gauss-2D-sep}
  T(n_1, n_2;\; s) = e^{-2s} I_{n_1}(s)  \, I_{n_2}(s),
\end{equation}
where $I_n$ denotes the modified Bessel functions of integer order
and which corresponds to the solution of the semi-discrete diffusion
equation
\begin{equation}
  \partial_s L(n_1, n_2;\; s) =
  \frac{1}{2} (\nabla_5^2  L) (n_1, n_2;\; s),
\end{equation}
where $\nabla_5^2$ denotes the five-point discrete Laplacian
operator defined by
$(\nabla_5^2 f)(n_1, n_2) = 
f(n_1-1, n_2) + f(n_1+1, n_2) +f(n_1, n_2-1) + f(n_1, n_2+1)- 4 f(n_1, n_2)$.
These kernels constitute the natural way to define
a scale-space concept for discrete signals corresponding to the
Gaussian scale-space over a symmetric domain.

This operation can be implemented either by explicit spatial
convolution with spatially truncated kernels 
\begin{equation}
   \sum_{n_1=-N}^{N} \sum_{n_2=-N}^{N}  T(n_1, n_2;\; s) > 1 -\varepsilon
\end{equation}
for small $\varepsilon$ of the order $10^{-8}$ to $10^{-6}$ with
mirroring at the image boundaries (adiabatic boundary conditions
corresponding to no heat transfer across the image boundaries)
or using the closed-form expression of the Fourier transform
\begin{align}
  \begin{split}
    \varphi_T(\theta_1, \theta_2) 
    & = \sum_{n_1=-\infty}^{\infty} \sum_{n_1=-\infty}^{\infty} T(n_1, n_2;\; s) \, e^{-i (n_1 \theta_1 + n_2 \theta_2)}
  \end{split}\nonumber\\
  \begin{split}
     = e^{-2 t(\sin^2(\frac{\theta_1}{2}) +\sin^2(\frac{\theta_2}{2}))}.
   \end{split}
\end{align}
Alternatively, to approximate rotational symmetry by higher degree of
accuracy, one can define the 2-D spatial discrete scale-space 
from the solution of (Lindeberg \cite[section~4.3]{Lin93-Dis})
\begin{equation}
  \label{eq-2D-scsp-1undet-par}
  \partial_s L =
  \frac{1}{2}
  \left(
   (1 - \gamma ) \nabla_5^2 L + \gamma \nabla_{\times^2}^2 L
  \right),
\end{equation}
where 
$(\nabla_{\times}^2 f)(n_1, n_2) = 
\tfrac{1}{2} 
(f(n_1+1, n_2+1) + f(n_1+1, n_2-1) +f(n_1-1, n_2+1) + f(n_1-1, n_2-1)- 4 f(n_1, n_2))$
and specifically the choice $\gamma = 1/3$ gives the best
approximation of rotational symmetry.
In practice, this operation can be implemented
\footnote{This four step combined diagonal and Cartesian separability property
can be understood by writing the discrete Laplacian operator 
$\nabla_{\gamma}^2 
 = (1 - \gamma ) \nabla_5^2 + \gamma \nabla_{\times^2}^2$ 
in (\protect\ref{eq-2D-scsp-1undet-par})
for $\gamma = 1/3$ as 
$\nabla_{\gamma}^2  
  = \frac{2}{3} (\delta_{xx} + \delta_{yy}) 
  + \frac{1}{3} (\delta_{\nearrow\nearrow} + \delta_{\nwarrow\nwarrow})$, 
  where $\delta_{xx}$ + $\delta_{yy}$ are the horizontal and vertical 
  difference operators with coefficients $(1, -2, 1)$ while 
 $\delta_{\nearrow\nearrow}$ and $\delta_{\nwarrow\nwarrow}$ are the
 two possible diagonal difference operators with coefficients $(1/2, -1, 1/2)$.
 The generating function of the convolution kernel is obtained by
 exponentiating (\protect\ref{eq-2D-scsp-1undet-par}) leading to
 $\varphi(z, w) = \exp(s \tilde{\nabla}_{\gamma}^2) 
   = \exp(\frac{2}{3} s \tilde{\delta}_{xx}) \, 
       \exp(\frac{2}{3} s \tilde{\delta}_{yy}) \,
       \exp(\frac{1}{3} s \tilde{\delta}_{\nearrow\nearrow}) \,
      \exp(\frac{1}{3} s \tilde{\delta}_{\nwarrow\nwarrow})$
 with $\tilde{\delta}_{xx} = z + z^{-1} - 2$, $\tilde{\delta}_{yy} = w + w^{-1} - 2$,
        $\tilde{\delta}_{\nearrow\nearrow} = (zw + z^{-1}w^{-1})/2 - 1$ and
        $\tilde{\delta}_{\nwarrow\nwarrow} = (z w^{-1} + z^{-1} w)/2 - 1$.
 The expressions $\exp(s \tilde{\delta}_{xx})$ and $\exp(s \tilde{\delta}_{yy})$
 are the generating functions of the regular one-dimensional discrete analogue
 of the Gaussian kernel $T(n;\; s) = e^{-s} I_n(s)$ along the
 horizontal and vertical Cartesian directions, whereas the expressions
 $\exp(s \tilde{\delta}_{\nearrow\nearrow})$ and $\exp(s \tilde{\delta}_{\nwarrow\nwarrow})$
 are the generating functions corresponding to applying the discrete
 analogue of the Gaussian kernel with a different scale parameter
 $T(n;\; s/2) = e^{-s/2} I_n(s/2)$ in the two possible diagonal
 directions.
 The reason why the scale parameter is different in the
 diagonal directions is because of the larger grid spacing in the
 diagonal {\em vs.\/} the Cartesian directions.} 
by first one step of
diagonal separable discrete smoothing at scale $s_{\times} = s/6$ followed by
a Cartesian separable discrete smoothing at scale $s_5 = 2s/3$ or
using a closed form expression for the Fourier transform derived
from the difference operators
  \begin{equation}
    \label{eq-2D-FS}
      \varphi_T(\theta_1, \theta_2)
       = e^{- (2 - \gamma )t 
                + (1 - \gamma) (\cos \theta_1 + \cos \theta_2 ) t 
                + (\gamma \cos \theta_1 \cos \theta_2) t)}.
  \end{equation}

\subsection{Discrete implementation of spatio-temporal receptive fields}
\label{sec-disc-spat-temp-RF}

For separable spatio-temporal receptive fields, we implement the
spatio-temporal smoothing operation by separable combination of the
spatial and temporal 
scale-space concepts in sections~\ref{app-disc-temp-smooth} and 
\ref{app-disc-gauss-smooth}.
From this representation, spatio-temporal derivative
approximations are then computed from {\em difference operators\/}
\footnote{Note that the below purely one-dimensional spatial derivative
  approximation operators are primarily intended to be used in
  connection with the separable discrete spatial scale-space concept
  (\protect\ref{eq-disc-analog-Gauss-2D-sep}).
  When using the non-separable discrete spatial scale-space concept
  (\protect\ref{eq-2D-scsp-1undet-par}) that enables better numerical
  approximation to rotational invariance, it can be motivated to also use
  two-dimensional discrete derivative approximation operators
  (Lindeberg \protect\cite[section~5.3.3.2]{Lin93-Dis}).}
\begin{align}
  \begin{split}
     \delta_t & = (-1, +1) \quad\quad\quad
     \delta_{tt} = (1, -2, 1)
  \end{split}\\
 \begin{split}
     \delta_{x} & = (-\frac{1}{2}, 0, +\frac{1}{2}) \quad\quad
     \delta_{xx} = (1, -2, 1)
  \end{split}\\
  \begin{split}
     \delta_{y} & = (-\frac{1}{2}, 0, +\frac{1}{2}) \quad\quad
     \delta_{yy} = (1, -2, 1)
  \end{split}
\end{align}
expressed over the appropriate dimensions and with higher order derivative
approximations constructed as combinations of these primitives,
{\em e.g.\/} $\delta_{xy} = \delta_x \, \delta_y$, 
$\delta_{xxx} = \delta_x \, \delta_{xx}$, $\delta_{xxt} = \delta_{xx} \, \delta_t$, etc.
From the general theory in (Lindeberg \cite{Lin93-JMIV,Lin93-Dis}) it follows that
the scale-space properties for the original 
zero-order signal will be transferred to such derivative
approximations, including a true cascade smoothing property for the
spatio-temporal discrete derivative approximations
\begin{align}
  \begin{split}
   L_{x_1^{m_1} x_2^{m_2} t^n}(x_1, x_2, t;\; s_2, \tau_{k_2}) = 
  \end{split}\nonumber\\
  \begin{split}
  & = \left( \vphantom{L_{x_1^{m_1} x_2^{m_2} t^n}} \left( 
             T(\cdot, \cdot ;\; s_2 - s_1) \, (\Delta h)(\cdot;\; \tau_{k_1} \mapsto \tau_{k_2}) 
         \right) \right. *
  \end{split}\nonumber\\
  \begin{split}
     & \phantom{= \left( \right.} 
         \left. \,\,
               L_{x_1^{m_1} x_2^{m_2} t^n}(\cdot, \cdot, \cdot;\; s_1,
               \tau_{k_1})
         \right) (x_1, x_2, t;\; s_2, \tau_{k_2}).
  \end{split}
\end{align}
The motivation for using symmetric%
\footnote{It should be noted, however, that as a side effect of this
  choice of a symmetric first-order derivative approximation,
                 the second order difference operator $\delta_{xx}$
                 will not be equal to the first-order difference
                 operator applied twice $\delta_x \, \delta_x \neq \delta_{xx}$.
                 Since the symmetric first-order difference operator 
                $(-1/2, 0, 1/2)$ corresponds to result of smoothing
                the tighter and non-symmetric difference operator
                $(-1, 1)$ with the binomial kernel $(1/2, 1/2)$, 
                the symmetric first-order derivative approximation
                could be seen as computed at a slightly coarser scale
                $\Delta s = 1/4$ compared to second-order derivative
                approximation obtained by the second-order difference
                operator $(1, -2, 1)$ corresponding to the tighter
                first-order difference $(-1, 1)$ applied twice. If that would be regarded
                as a problem, one could try to compensate for this
                effect by smoothing the second-order derivative
                kernels with the symmetric generalized binomial kernel
                $(\Delta s/2, 1 - \Delta s, \Delta s/2)$ for
                $\Delta s = 1/4$, however, then at the cost of
                destroying the relation
                (\protect\ref{eq-2der-disc-gauss-kern}) between the
                second-order derivative approximations and derivatives
                with respect to scale. The generalized binomial kernel $(\Delta s/2, 1 -
                \Delta s, \Delta s/2)$ is for $0 \leq \Delta s \leq 1/2$ 
                also a discrete scale-space
                kernel and has variance $\Delta s$ (Lindeberg
                \cite[sections~3.2.2 and 3.6.2]{Lin93-Dis}).} differences for the first order
spatial derivative approximations $\delta_x$ and $\delta_y$ is to 
have the derivative approximations maximally accurate at the grid
points to enable straightforward combination into higher order 
differential invariants over image space. 
With this choice also certain algebraic relations that hold for
derivatives of continuous Gaussian kernels will be transferred to corresponding
algebraic relations for difference approximations applied to the
discrete Gaussian kernel
(Lindeberg \cite[equations (5.34) and (5.36) at page 133]{Lin93-Dis}):
\begin{align}
  \begin{split}
     \label{eq-1der-disc-gauss-kern}
     (\delta_x T)(x;\; t) 
      & = - \frac{x}{t} \, T(x;\; t)
  \end{split}\\
  \begin{split}
     \label{eq-2der-disc-gauss-kern}
     (\delta_{xx} T)(x;\; t) 
      & = 2 (\partial_t T)(x;\; t)
  \end{split}
\end{align}
The motivation for using non-symmetric first-order derivative approximations 
$(-1, 1)$ over time is because of the temporal causality that implies
the impossibility of having access to data from the future and then
within this constraint minimize the temporal delay as much as possible
using temporal difference operators of minimum support.
Because of the non-causal temporal smoothing operation, one anyway
gets an additional and much larger temporal delay that implies that
all filter responses are computed with a certain and non-neglible temporal
delay.

For non-separable spatio-temporal receptive fields corresponding
to a non-zero image velocity $v = (v_1, v_2)^T$,
we implement the spatio-temporal smoothing operation by first warping 
the video data 
$(x_1', x_2')^T = (x_1 - v_1 t, x_2 - v_2 t)^T$
using spline interpolation. Then, we apply separable spatio-temporal
smoothing in the transformed domain and unwarp the result back to the
original domain.
Over a continuous domain, such an operation is equivalent to
convolution with corresponding velocity-adapted spatio-temporal
receptive fields, while being significantly faster in a discrete
implementation than explicit convolution with
non-separable receptive fields over three dimensions.

\section{Scale normalization for spatio-temporal derivatives}
\label{sec-sc-norm-spat-temp-der}

When computing spatio-temporal derivatives at different scales, some
mechanism is needed for normalizing the derivatives with respect to
the spatial and temporal scales, to make derivatives at different
spatial and temporal scales comparable and to enable spatial and temporal scale selection.

\subsection{Scale normalization of spatial derivatives}

For the Gaussian scale-space concept defined over a purely spatial
domain, it can be shown that the canonical way of defining scale-normalized
derivatives at different spatial scales $s$ is according to 
(Lindeberg \cite{Lin97-IJCV})
\begin{equation}
  \partial_{\xi_1} = s^{\gamma_s/2} \, \partial_{x_1},
  \quad\quad
  \partial_{\xi_2} = s^{\gamma_s/2} \, \partial_{x_2},
\end{equation}
where $\gamma_s$ is a free parameter. Specifically, it can be shown 
(Lindeberg \cite[section~9.1]{Lin97-IJCV}) that
this notion of $\gamma$-normalized derivatives corresponds to 
normalizing the $m$:th order Gaussian derivatives 
$g_{\xi^m} = g_{\xi_1^{m_1} \xi_2^{m_2}}$ in $N$-dimensional
image space to constant $L_p$-norms over scale 
\begin{equation}
  \| g_{\xi^m}(\cdot;\; s) \|_p 
  = \left( 
        \int_{x \in \bbbr^N} |g_{\xi^m}(x;\; s)|^p \, dx
      \right)^{1/p} 
   = G_{m,\gamma_s}
\end{equation}
with
\begin{equation}
 \label{eq-sc-norm-p-from-gamma}
  p = \frac{1}{1 + \frac{|m|}{N} (1 - \gamma_s)}
\end{equation}
where the perfectly scale invariant case $\gamma_s = 1$ corresponds to
$L_1$-normalization for all orders $|m| = m_1 + \dots + m_N$.
In this paper, we will throughout use this approach for normalizing
spatial differentiation operators with respect to the spatial scale
parameter $s$.

\subsection{Scale normalization of temporal derivatives}

If using a non-causal Gaussian temporal scale-space concept,
scale-normalized temporal derivatives can be defined in an analogous
way as scale-normalized spatial derivatives as described in the previous section.

For the time-causal temporal scale-space concept based on first-order
temporal integrators coupled in cascade, we can also define a
corresponding notion of scale-normalized temporal derivatives
\begin{equation}
  \label{eq-sc-norm-der-var-norm}
  \partial_{\zeta^n} = \tau^{n \gamma_{\tau}/2} \, \partial_{t^n}
\end{equation}
which will be referred to as {\em variance-based normalization\/}
reflecting the fact the parameter $\tau$ corresponds to variance of
the composed temporal smoothing kernel.
Alternatively, we can determine a temporal scale normalization factor $\alpha_{n,\gamma_{\tau}}(\tau)$
\begin{equation}
  \label{eq-sc-norm-der-Lp-norm-1}
  \partial_{\zeta^n} = \alpha_{n,\gamma_{\tau}}(\tau) \, \partial_{t^n}
\end{equation}
such that the $L_p$-norm (with $p$ determined as function of $\gamma$ according to
(\ref{eq-sc-norm-p-from-gamma})) of the corresponding composed scale-normalized
temporal derivative computation kernel $\alpha_{n,\gamma_{\tau}}(\tau) \, h_{t^n}$
equals the $L_p$-norm of some other reference kernel, where we
here initially take the $L_p$-norm of the corresponding Gaussian
derivative kernels
\begin{align}
  \begin{split}
     \| \alpha_{n,\gamma_{\tau}}(\tau) \, h_{t^n}(\cdot;\; \tau) \|_p 
     & = \alpha_{n,\gamma_{\tau}}(\tau) \, \| h_{t^n}(\cdot;\; \tau) \|_p 
  \end{split}\nonumber\\
  \begin{split}
     \label{eq-sc-norm-der-Lp-norm-2}
     & = \| g_{\xi^n}(\cdot;\; \tau) \|_p = G_{n,\gamma_{\tau}}.
  \end{split}
\end{align}
This latter approach will be referred to as {\em $L_p$-normalization\/}.%
\footnote{These definitions generalize the previously defined notions of $L_p$-normalization and
variance-based normalization over discrete scale-space representation
in (Lindeberg \cite{Lin97-IJCV}) and pyramids in (Lindeberg and Bretzner 
\cite{LinBre03-ScSp}) to temporal scale-space representations.}

For the discrete temporal scale-space concept over discrete time,
scale normalization factors for discrete $l_p$-normal\-ization are
defined in an analogous way with the only difference that the 
continuous $L_p$-norm is replaced by a discrete $l_p$-norm.

In the specific case when the temporal scale-space representation is
defined by convolution with the scale-invariant time-causal limit
kernel according to (\ref{eq-temp-scsp-conv-limit-kernel}) and
(\ref{eq-FT-comp-kern-log-distr-limit}), it is shown in
appendix~\ref{app-sc-inv-sc-norm-temp-der-limit-kern} that the
corresponding scale-normalized derivatives become truly scale covariant under temporal
scaling transformations $t' = c^j t$ with scaling factors $S = c^j$ that are
integer powers of the distribution parameter $c$
\begin{align}
  \begin{split}
     L'_{\zeta'^n}(t';\, \tau', c) 
     & = c^{j n (\gamma-1)} \, L_{\zeta^n}(t;\, \tau, c) 
  \end{split}\nonumber\\
  \begin{split}
     & = c^{j (1 - 1/p)} \, L_{\zeta^n}(t;\, \tau, c)
  \end{split}
\end{align}
between matching temporal scale levels $\tau' = c^{2j} \tau$.
Specifically, for $\gamma = 1$ corresponding to $p = 1$ the
scale-normalized temporal derivatives become fully scale invariant
\begin{equation}
  L'_{\zeta'^n}(t';\, \tau', c) = L_{\zeta^n}(t;\, \tau, c) .
\end{equation}

\begin{table*}[hbtp]
  \addtolength{\tabcolsep}{1pt}
  \begin{center}
   \footnotesize
  \begin{tabular}{cccccc}
  \hline
   \multicolumn{6}{c}{Temporal scale normalization factors for 
                                 $n = 1$ at $\tau = 1$} \\
  \hline
    $K$ & $\tau^{n/2}$  
       & $\alpha_{n,\gamma_{\tau}}(\tau)$  (uni) 
       & $\alpha_{n,\gamma_{\tau}}(\tau)$  ($c = \sqrt{2}$) 
       & $\alpha_{n,\gamma_{\tau}}(\tau)$  ($c = 2^{3/4}$) 
       &$\alpha_{n,\gamma_{\tau}}(\tau)$ ($c = 2$) \\
  \hline
    2 & 1.000 & 0.744 & 0.744 & 0.737 & 0.723 \\
    3 & 1.000 & 0.805 & 0.794 & 0.765 & 0.736 \\
    4 & 1.000 & 0.847 & 0.814 & 0.771 & 0.737 \\
    5 & 1.000 & 0.877 & 0.821 & 0.772 & 0.738 \\
    6 & 1.000 & 0.901 & 0.823 & 0.772 & 0.738 \\
    7 & 1.000 & 0.920 & 0.823 & 0.772 & 0.738 \\
    8 & 1.000 & 0.935 & 0.823 & 0.772 & 0.738 \\
\vdots \\
   16 & 1.000 & 0.998 & 0.823 & 0.772 & 0.738 \\
  \hline
  \end{tabular}
\end{center}

\smallskip

  \begin{center}
   \footnotesize
  \begin{tabular}{cccccc}
  \hline
   \multicolumn{6}{c}{Temporal scale normalization factors for 
                                 $n = 1$ at $\tau = 16$} \\
  \hline
    $K$ & $\tau^{n/2}$  
       & $\alpha_{n,\gamma_{\tau}}(\tau)$  (uni) 
       & $\alpha_{n,\gamma_{\tau}}(\tau)$  ($c = \sqrt{2}$) 
       & $\alpha_{n,\gamma_{\tau}}(\tau)$  ($c = 2^{3/4}$) 
       &$\alpha_{n,\gamma_{\tau}}(\tau)$ ($c = 2$) \\
  \hline
    2 & 4.000 & 3.056 & 3.056 & 3.016 & 2.938 \\
    3 & 4.000 & 3.398 & 3.341 & 3.210 & 3.041 \\
    4 & 4.000 & 3.553 & 3.432 & 3.223 & 3.068 \\
    5 & 4.000 & 3.642 & 3.442 & 3.227 & 3.071 \\
    6 & 4.000 & 3.731 & 3.452 & 3.228 & 3.071 \\
    7 & 4.000 & 3.744 & 3.457 & 3.228 & 3.071 \\
    8 & 4.000 & 3.809 & 3.459 & 3.228 & 3.071 \\
  \vdots \\
  16 & 4.000 & 3.891 & 3.460 & 3.338 & 3.071  \\
\hline
  \end{tabular}
\end{center}

\smallskip

  \begin{center}
   \footnotesize
  \begin{tabular}{cccccc}
  \hline
   \multicolumn{6}{c}{Temporal scale normalization factors for 
                                 $n = 1$ at $\tau = 256$} \\
  \hline
    $K$ & $\tau^{n/2}$  
       & $\alpha_{n,\gamma_{\tau}}(\tau)$  (uni) 
       & $\alpha_{n,\gamma_{\tau}}(\tau)$  ($c = \sqrt{2}$) 
       & $\alpha_{n,\gamma_{\tau}}(\tau)$  ($c = 2^{3/4}$) 
       &$\alpha_{n,\gamma_{\tau}}(\tau)$ ($c = 2$) \\
  \hline
    2 & 16.000 & 12.270 & 12.270 & 12.084 & 11.711 \\
    3 & 16.000 & 13.612 & 13.420 & 12.835 & 12.147 \\
    4 & 16.000 & 14.242 & 13.732 & 12.932 & 12.162 \\
    5 & 16.000 & 14.610 & 13.815 & 12.930 & 12.155 \\
    6 & 16.000 & 14.850 & 13.816 & 12.927 & 12.152 \\
    7 & 16.000 & 15.018 & 13.817 & 12.922 & 12.151 \\
    8 & 16.000 & 15.145 & 13.817 & 12.922 & 12.151 \\
   \vdots \\
   16 & 16.000 & 15.583 & 13.816 & 12.922 & 12.151 \\
\hline
  \end{tabular}
\end{center}

\caption{Numerical values of {\em scale normalization factors for
    discrete temporal derivative approximations\/},
    using either variance-based normalization $\tau^{n/2}$ or
    $l_p$-normalization $\alpha_{n,\gamma_{\tau}}(\tau)$, for temporal
    derivatives of order $n = 1$ and at temporal scales $\tau = 1$,
    $\tau = 16$ and $\tau = 256$
    relative to a unit temporal sampling rate with $\Delta t = 1$ 
    and with $\gamma_{\tau} = 1$,
    for time-causal kernels obtained by coupling $K$ first-order 
    recursive filters in cascade with either a uniform distribution of
    the intermediate scale levels 
    or a  logarithmic distribution 
    for $c = \sqrt{2}$, $c = 2^{3/4}$ and $c = 2$.}
  \label{tab-alpha-Lp-m1}
\end{table*}

\begin{table*}[hbtp]
  \addtolength{\tabcolsep}{1pt}
  \begin{center}
   \footnotesize
  \begin{tabular}{cccccc}
  \hline
   \multicolumn{6}{c}{Temporal scale normalization factors for 
                                 $n = 2$ at $\tau = 1$} \\
  \hline
    $K$ & $\tau^{n/2}$  
       & $\alpha_{n,\gamma_{\tau}}(\tau)$  (uni) 
       & $\alpha_{n,\gamma_{\tau}}(\tau)$  ($c = \sqrt{2}$) 
       & $\alpha_{n,\gamma_{\tau}}(\tau)$  ($c = 2^{3/4}$) 
       &$\alpha_{n,\gamma_{\tau}}(\tau)$ ($c = 2$) \\
  \hline
    2 & 1.000 & 0.617 & 0.617 & 0.606 & 0.586 \\
    3 & 1.000 & 0.711 & 0.694 & 0.649 & 0.607 \\
    4 & 1.000 & 0.738 & 0.718 & 0.659 & 0.609 \\
    5 & 1.000 & 0.755 & 0.721 & 0.660 & 0.609 \\
    6 & 1.000 & 0.768 & 0.722 & 0.660 & 0.609 \\
    7 & 1.000 & 0.779 & 0.722 & 0.660 & 0.609 \\
    8 & 1.000 & 0.787 & 0.722 & 0.660 & 0.609 \\
\vdots \\
   16 & 1.000 & 0.824 & 0.722 & 0.660 & 0.609 \\
  \hline
  \end{tabular}
  \end{center}

\smallskip

  \begin{center}
   \footnotesize
  \begin{tabular}{cccccc}
  \hline
   \multicolumn{6}{c}{Temporal scale normalization factors for 
                                 $n = 2$ at $\tau = 16$} \\
  \hline
    $K$ & $\tau^{n/2}$  
       & $\alpha_{n,\gamma_{\tau}}(\tau)$  (uni) 
       & $\alpha_{n,\gamma_{\tau}}(\tau)$  ($c = \sqrt{2}$) 
       & $\alpha_{n,\gamma_{\tau}}(\tau)$  ($c = 2^{3/4}$) 
       &$\alpha_{n,\gamma_{\tau}}(\tau)$ ($c = 2$) \\
 \hline
    2 & 16.000 &    4.622 & 4.622 & 4.472 & 4.172 \\
    3 & 16.000 &    8.429 & 8.017 & 6.897 & 5.701 \\
    4 & 16.000 &  10.184 & 9.160 & 7.885 & 6.208 \\
    5 & 16.000 &  11.363 & 9.698 & 7.871 & 6.296 \\
    6 & 16.000 & 12.241 & 10.022 & 7.864 & 6.305 \\
    7 & 16.000 & 12.690 & 10.088 & 7.862 & 6.305 \\
    8 & 16.000 & 13.106 & 10.068 & 7.862 & 6.305 \\
  \vdots \\
  16 & 16.000 & 14.575 & 10.058 & 7.862 & 6.305 \\
  \hline
  \end{tabular}
\end{center}

\smallskip

  \begin{center}
   \footnotesize
  \begin{tabular}{cccccc}
  \hline
   \multicolumn{6}{c}{Temporal scale normalization factors for 
                                 $n = 2$ at $\tau = 256$} \\
  \hline
    $K$ & $\tau^{n/2}$  
       & $\alpha_{n,\gamma_{\tau}}(\tau)$  (uni) 
       & $\alpha_{n,\gamma_{\tau}}(\tau)$  ($c = \sqrt{2}$) 
       & $\alpha_{n,\gamma_{\tau}}(\tau)$  ($c = 2^{3/4}$) 
       &$\alpha_{n,\gamma_{\tau}}(\tau)$ ($c = 2$) \\
 \hline
    2 & 256.00 &   58.95 &   58.95 &   56.63 &   51.84 \\
    3 & 256.00 & 133.37 & 127.68 & 112.66 &   94.71 \\
    4 & 256.00 & 165.14 & 148.96 & 124.04 & 101.16 \\
    5 & 256.00 & 183.75 & 156.04 & 126.42 & 101.13 \\
    6 & 256.00 & 195.99 & 158.69 & 126.65 & 101.12 \\
    7 & 256.00 & 204.71 & 159.17 & 126.56 & 101.12 \\
    8 & 256.00 & 211.10 & 159.23 & 126.55 & 101.12 \\
  \vdots \\
   16 & 256.00 & 233.78 & 159.28 & 126.55 & 101.12 \\
  \hline
  \end{tabular}
\end{center}
\caption{Numerical values of {\em scale normalization factors for
    discrete temporal derivative approximations\/},
    for either variance-based normalization $\tau^{n/2}$ or
    $l_p$-normalization $\alpha_{n,\gamma_{\tau}}(\tau)$, for temporal
    derivatives of order $n = 2$ and at temporal scales $\tau = 1$,
    $\tau = 16$ and $\tau = 256$
    relative to a unit temporal sampling rate with $\Delta t = 1$ 
    and with $\gamma_{\tau} = 1$,
    for time-causal kernels obtained by coupling $K$ first-order
    recursive filters in cascade with either a uniform distribution of
    the intermediate scale levels 
    or a logarithmic distribution 
    for $c = \sqrt{2}$, $c = 2^{3/4}$ and $c = 2$.}
  \label{tab-alpha-Lp-m2}
\end{table*}

\begin{table*}[!hbt]
 \addtolength{\tabcolsep}{1pt}
  \begin{center}
   \footnotesize
  \begin{tabular}{ccccc}
  \hline
   \multicolumn{5}{c}{Relative deviation from limit of scale normalization factors for 
                                 $n = 1$ at $\tau = 256$} \\
  \hline
    $K$ & $\varepsilon_n$ (uni) 
       &  $\varepsilon_n$  ($c = \sqrt{2}$)
       &  $\varepsilon_n$  ($c = 2^{3/4}$)
       &  $\varepsilon_n$  ($c = 2$) \\
  \hline
      2 & 0.233 & $1.1 \cdot 10^{-1}$ & $6.5 \cdot 10^{-2}$ & $3.6 \cdot 10^{-2}$ \\
      4 & 0.110 & $6.1 \cdot 10^{-3}$ & $8.5 \cdot 10^{-4}$ & $8.6 \cdot 10^{-4}$ \\
      8 & 0.053 & $4.9 \cdot 10^{-4}$ & $1.1 \cdot 10^{-5}$ & $2.0 \cdot 10^{-7}$ \\
    16 & 0.026 & $1.2 \cdot 10^{-7}$ & $9.0 \cdot 10^{-13}$ & $1.5 \cdot 10^{-15}$ \\
    32 & 0.013 & $3.1 \cdot 10^{-14}$ & $2.9 \cdot 10^{-14}$ & $3.4 \cdot 10^{-14}$ \\
  \hline
  \end{tabular}
\end{center}

\smallskip

  \begin{center}
   \footnotesize
  \begin{tabular}{ccccc}
  \hline
   \multicolumn{5}{c}{Relative deviation from limit of scale normalization factors for 
                                 $n = 2$ at $\tau = 256$} \\
  \hline
    $K$ & $\varepsilon_n$ (uni) 
       &  $\varepsilon_n$  ($c = \sqrt{2}$)
       &  $\varepsilon_n$  ($c = 2^{3/4}$)
       &  $\varepsilon_n$  ($c = 2$) \\
  \hline
      2 & 0.770 & $6.3 \cdot 10^{-1}$ & $5.5 \cdot 10^{-1}$ & $4.9 \cdot 10^{-1}$ \\
      4 & 0.354 & $6.5 \cdot 10^{-2}$ & $2.0 \cdot 10^{-2}$ & $4.1 \cdot 10^{-2}$ \\
      8 & 0.174 & $3.2 \cdot 10^{-4}$ & $1.3 \cdot 10^{-5}$ & $1.6 \cdot 10^{-8}$ \\
    16 & 0.085 & $1.8 \cdot 10^{-7}$ & $1.0 \cdot 10^{-12}$ & $9.6 \cdot 10^{-15}$ \\
    32 & 0.042 & $1.2 \cdot 10^{-13}$ & $6.2 \cdot 10^{-14}$ & $4.0 \cdot 10^{-14}$ \\
  \hline
  \end{tabular}
\end{center}

\caption{Numerical estimates of the relative deviation from the limit
  case when using different numbers $K$ of temporal scale levels for a
  uniform {\em vs.\/}\ a logarithmic distribution of the intermediate
  scale levels. The deviation measure $\varepsilon_n$ according to
  equation~(\ref{eq-eps-n-rel-dev-limit-case-alpha-n}) measures the
  relative deviation of the scale normalization factors when using a
  finite number $K$ of temporal scale levels compared to the limit case
  when the number of temporal scale levels $K$ tends to
  infinity. (These estimates have been computed at a coarse temporal scale
  $\tau = 256$ relative to a unit grid spacing so that the influence
  of discretization effects should be small. The limit case has been
  approximated by $K = 1000$ for the uniform distribution and $K =
  500$ for the logarithmic distribution.)}
  \label{tab-rel-dev-from-limit-n1-n2}
\end{table*}

\subsection{Computation of temporal scale normalization factors}

For computing the temporal scale normalization factors 
\begin{equation}
  \alpha_{n,\gamma_{\tau}}(\tau) 
  = \frac{\| g_{\xi^n}(\cdot;\; \tau) \|_p}{\| h_{t^n}(\cdot;\; \tau) \|_p}
\end{equation}
in (\ref{eq-sc-norm-der-Lp-norm-1}) for $L_p$-normalization 
according to (\ref{eq-sc-norm-der-Lp-norm-2}),
we compute the $L_p$-norms of the scale-normalized Gaussian
derivatives, from closed-form expressions if $\gamma = 1$
(corresponding to $p = 1$)
\begin{align}
    \begin{split}
      \label{eq-L1-norm-gauss-der1}
      G_{1,1} 
      & = \left. \int_{-\infty}^{\infty} |g_{\xi}(u;\;t)| \, du \right|_{\gamma=1}
        = \sqrt{\frac{2}{\pi}} \approx 0.797885,
    \end{split}\\
    \begin{split}
      \label{eq-L1-norm-gauss-der2}
      G_{2,1} 
      & = \left. \int_{-\infty}^{\infty} |g_{\xi^2}(u;\;t)| \, du \right|_{\gamma=1}
        = \sqrt{\frac{8}{\pi \, e}} \approx 0.967883,
    \end{split}\\
    \begin{split}
      \label{eq-L1-norm-gauss-der3}
      G_{3,1} 
      & = \left. \int_{-\infty}^{\infty} |g_{\xi^3}(u;\;t)| \, du \right|_{\gamma=1}
    \end{split}\nonumber\\
    \begin{split}
      &
        = \sqrt{\frac{2}{\pi}}
          \left(
            1 + \frac{4}{e^{3/2}} 
          \right) 
          \approx 1.51003,
    \end{split}\\
\end{align}
\begin{align}
    \begin{split}
      \label{eq-L1-norm-gauss-der4}
      G_{4,1} 
      & = \left. \int_{-\infty}^{\infty} |g_{\xi^4}(u;\;t)| \, du \right|_{\gamma=1}
    \end{split}\nonumber\\
    \begin{split}
      & = \frac{4 \sqrt{3}}
               {e^{3/2 + \sqrt{3/2}} \, \sqrt{\pi}}
         (\sqrt{3 - \sqrt{6}} \, e^{\sqrt{6}}
          + \sqrt{3 + \sqrt{6}}) 
    \end{split}\nonumber\\
    \begin{split}
      &
      \approx 2.8006.
    \end{split}
  \end{align}
or for values of $\gamma \neq 1$ by numerical integration.
For computing the discrete $l_p$-norm of discrete temporal derivative
approximations, 
we first (i)~filter a discrete delta function by the corresponding
cascade of first-order integrators to obtain the temporal smoothing
kernel and then (ii)~apply discrete derivative approximation operators to
this kernel to obtain the corresponding equivalent temporal
derivative kernel, (iii)~from which the discrete $l_p$-norm is
computed
by straightforward summation.

To illustrate how the choice of temporal scale normalization method
may affect the results in a discrete implementation, 
tables~\ref{tab-alpha-Lp-m1}--\ref{tab-alpha-Lp-m2}
show examples of temporal scale normalization factors computed in
these ways by either
(i)~variance-based normalization $\tau^{n/2}$ according to (\ref{eq-sc-norm-der-var-norm}) or
(ii)~$L_p$-norm\-alization $\alpha_{n,\gamma_{\tau}}(\tau)$ according to
(\ref{eq-sc-norm-der-Lp-norm-1})--(\ref{eq-sc-norm-der-Lp-norm-2})
for different orders of
temporal temporal differentiation $n$, different distribution
parameters $c$ and at different temporal scales
$\tau$, relative to a unit temporal sampling rate.
The value $c = \sqrt{2}$ corresponds to a natural minimum value of the
distribution parameter from the constraint $\mu_2 \geq \mu_1$, 
the value $c = 2$ to a doubling scale sampling strategy as used in a regular
spatial pyramids and $c = 2^{3/4}$ to a 
natural intermediate value between these two.
The temporal scale level $\tau = 1$ is near the discrete
temporal sampling rate where temporal discretization effects are strong,
$\tau = 16$ is a higher temporal scale where temporal sampling effects
are moderate and $\tau = 256$ corresponds to a temporal scale much higher than
discrete temporal sampling distance and the temporal sampling effects
therefore can be expected to be small.

Notably, the numerical values of the
resulting scale normalization factors may differ substantially
depending on the type of scale normalization method and the underlying
number of first-order recursive filters that are coupled in cascade.
Therefore, the choice of temporal scale normalization method warrants
specific attention in applications where the relations between
numerical values of temporal derivatives at different temporal scales may have
critical influence.

Specifically, we can note that the temporal scale
normalization factors based on $L_p$-normalization differ more from
the scale normalization factors from variance-based normalization 
(i)~in the case of a logarithmic distribution of the intermediate
temporal scale levels compared to a uniform distribution,
(ii)~when the distribution parameter $c$ increases within the family
of temporal receptive fields based on a logarithmic distribution of
the intermediate scale levels or 
(iii)~a very low number of
recursive filters are coupled in cascade.
In all three cases, the resulting temporal smoothing kernels become
more asymmetric and do hence differ more from the symmetric
Gaussian model. 

On the other hand, with increasing values of $K$ the numerical values of the scale
normalization factors converge much faster to their limit values when
using a logarithmic distribution of the intermediate scale levels
compared to using a uniform distribution. Depending on the value of
the distribution parameter $c$, the scale normalization factors do
reasonably well approach their limit values after $K = 4$ to $K = 8$
scale levels, whereas much larger values of $K$ would be needed if
using a uniform distribution. The convergence rate is faster for
larger values of $c$.

\subsection{Measuring the deviation from the scale-invariant
  time-causal limit kernel}
\label{sec-meas-dev-from-scinv-lim-kern}

To quantify how good an approximation a time-causal kernel with a finite
number of $K$ scale levels is to the limit case when the number of
scale levels $K$ tends to infinity, let us 
measure the relative deviation of the scale normalization factors from
the limit kernel according to
\begin{equation}
  \label{eq-eps-n-rel-dev-limit-case-alpha-n}
  \varepsilon_n(\tau)
  = \frac{\left| 
                \left. \alpha_{n}(\tau) \right|_{K} -
                \left. \alpha_{n}(\tau) \right|_{K \rightarrow \infty}
              \right|}
              {\left. \alpha_{n}(\tau) \right|_{K \rightarrow \infty}}.
\end{equation}
Table~\ref{tab-rel-dev-from-limit-n1-n2} shows numerical estimates of
this relative deviation measure for different values of $K$ from 
$K = 2$ to $K = 32$ for the time-causal kernels obtained from a uniform
{\em vs.\/}\ a logarithmic distribution of the scale values.
From the table, we can first note that the
convergence rate with increasing values of $K$ is significantly faster
when using a logarithmic {\em vs.\/}\ a uniform distribution of the
intermediate scale levels.

Not even $K = 32$ scale levels is sufficient to drive the relative
deviation measure below $1~\%$ for a uniform distribution, whereas the
corresponding deviation measures are down to machine precision when using
$K = 32$ levels for a logarithmic distribution.
When using $K = 4$ scale levels, the relative derivation measure 
is down to $10^{-2}$ to $10^{-4}$ for a logarithmic distribution.
If using $K = 8$ scale levels, the relative deviation measure is down
to $10^{-4}$ to $10^{-8}$ depending on the value of the distribution
parameter $c$ and the order $n$ of differentiation.

From these results, we can conclude that one should not use a too low number of recursive filters
that are coupled in cascade when computing temporal derivatives. Our
recommendation is to use a logarithmic distribution with a minimum of four recursive filters
for derivatives up to order two at finer scales and a larger number of
recursive filters at coarser scales. When performing computations at a
single temporal scale, we often use $K = 7$ or $K = 8$ as default.

\section{Spatio-temporal feature detection}
\label{sec-spat-temp-feat}

In the following, we shall apply the above theoretical framework for
separable time-causal spatio-temporal receptive fields for computing 
different types of spatio-temporal feature, defined from spatio-temporal
derivatives of different spatial and temporal orders, which may additionally
be combined into composed (linear or non-linear) differential expressions.

\begin{figure}[!p]
  \begin{center}
    \begin{tabular}{c}
     \hspace{-2mm} {\footnotesize $-T(x, t;\; s, \tau)$} \hspace{-2mm} \\
      \hspace{-2mm} \includegraphics[width=0.13\textwidth]{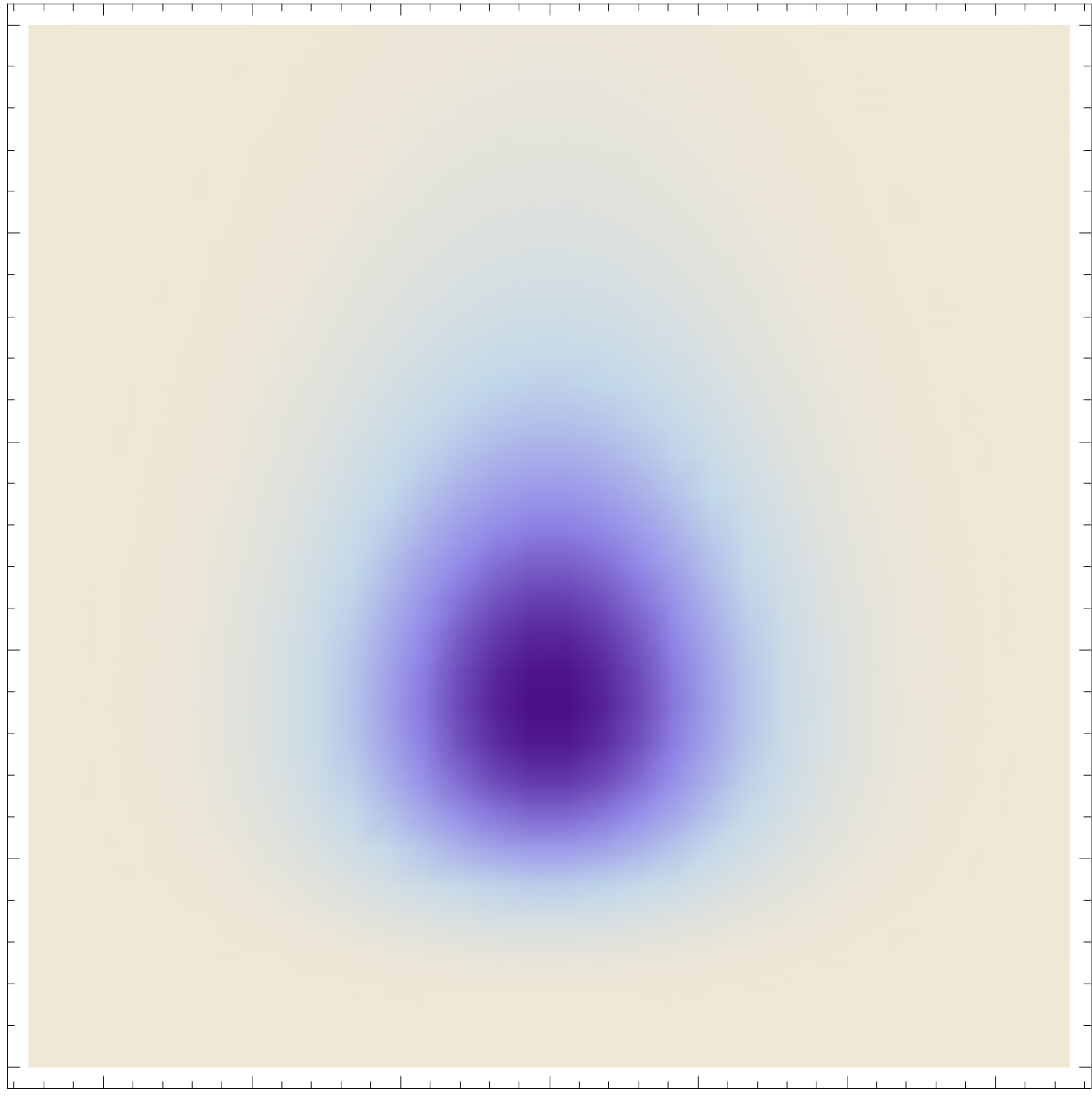} \hspace{-2mm} \\
    \end{tabular} 
  \end{center}
  \vspace{-6mm}
  \begin{center}
    \begin{tabular}{cc}
      \hspace{-2mm} {\footnotesize $T_x(x, t;\; s, \tau)$} \hspace{-2mm} 
      & \hspace{-2mm} {\footnotesize $T_t(x, t;\; s, \tau, \delta)$} \hspace{-2mm} \\
      \hspace{-2mm} \includegraphics[width=0.13\textwidth]{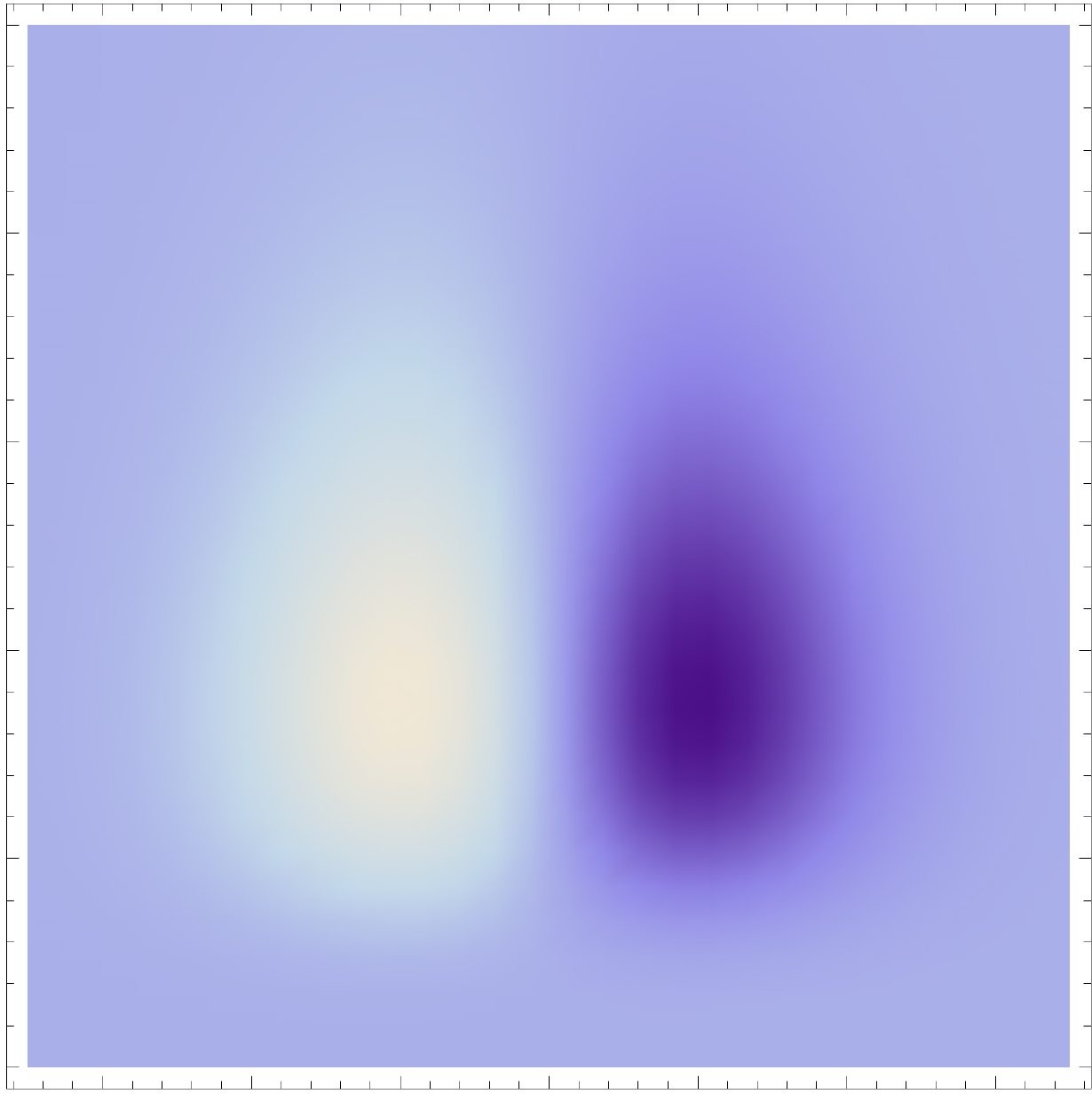} \hspace{-2mm} &
      \hspace{-2mm} \includegraphics[width=0.13\textwidth]{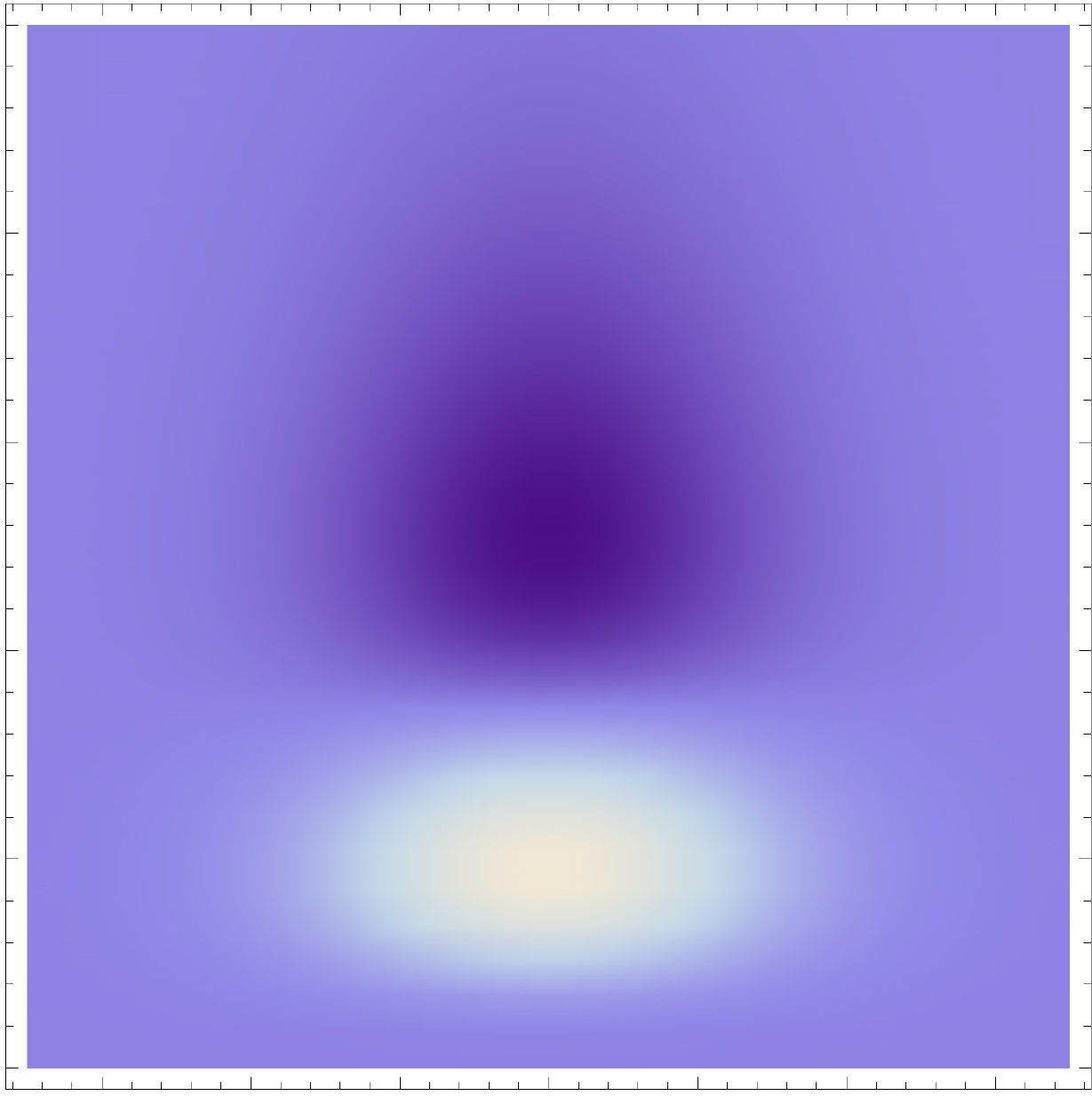} \hspace{-2mm} \\
    \end{tabular} 
  \end{center}
  \vspace{-6mm}
  \begin{center}
    \begin{tabular}{ccc}
      \hspace{-2mm} {\footnotesize $T_{xx}(x, t;\; s, \tau)$} \hspace{-2mm} 
      & \hspace{-2mm} {\footnotesize $T_{xt}(x, t;\; s, \tau)$} \hspace{-2mm} 
      & \hspace{-2mm} {\footnotesize $T_{tt}(x, t;\; s, \tau)$} \hspace{-2mm} \\
      \hspace{-2mm} \includegraphics[width=0.13\textwidth]{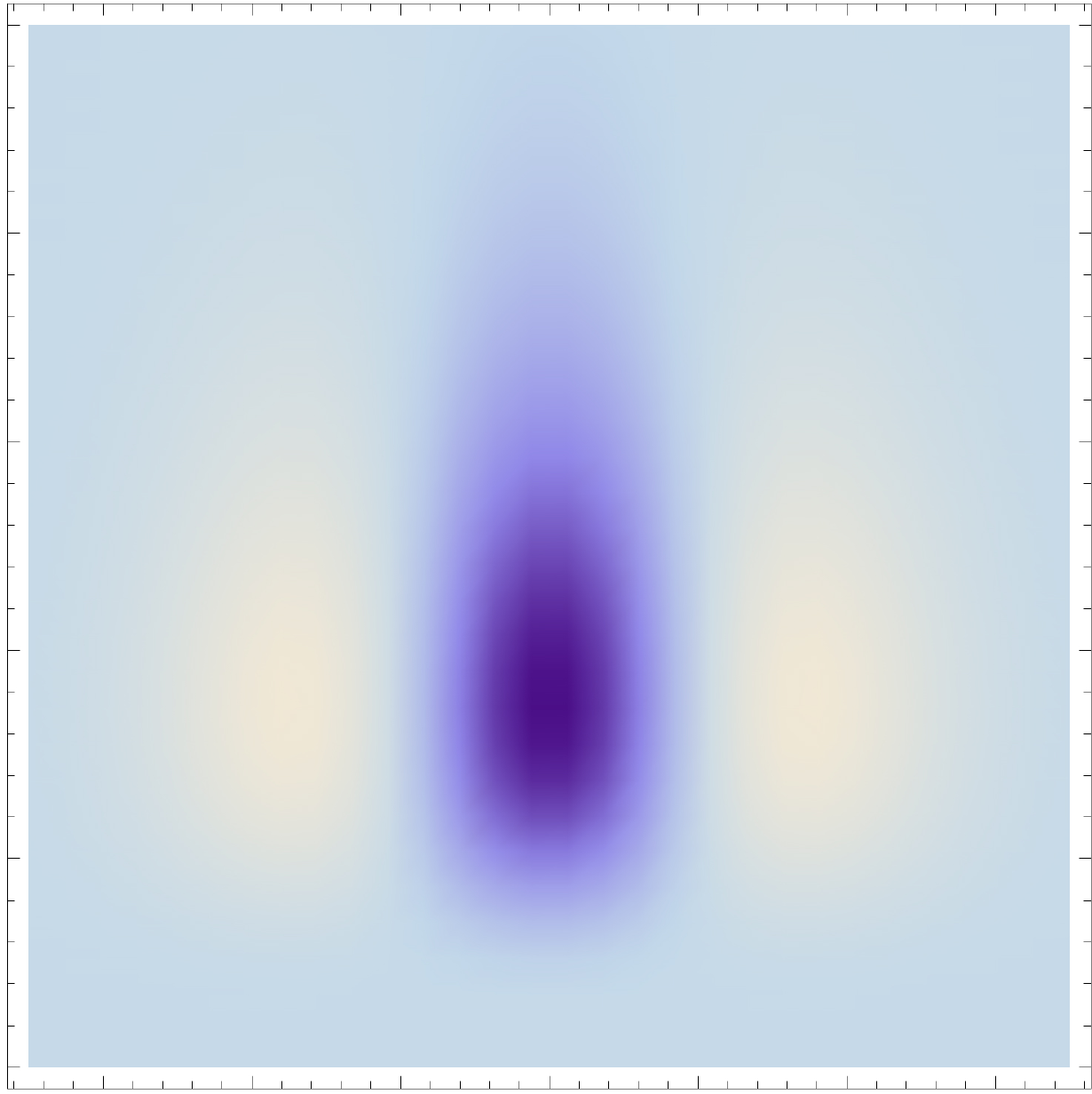} \hspace{-2mm} &
      \hspace{-2mm} \includegraphics[width=0.13\textwidth]{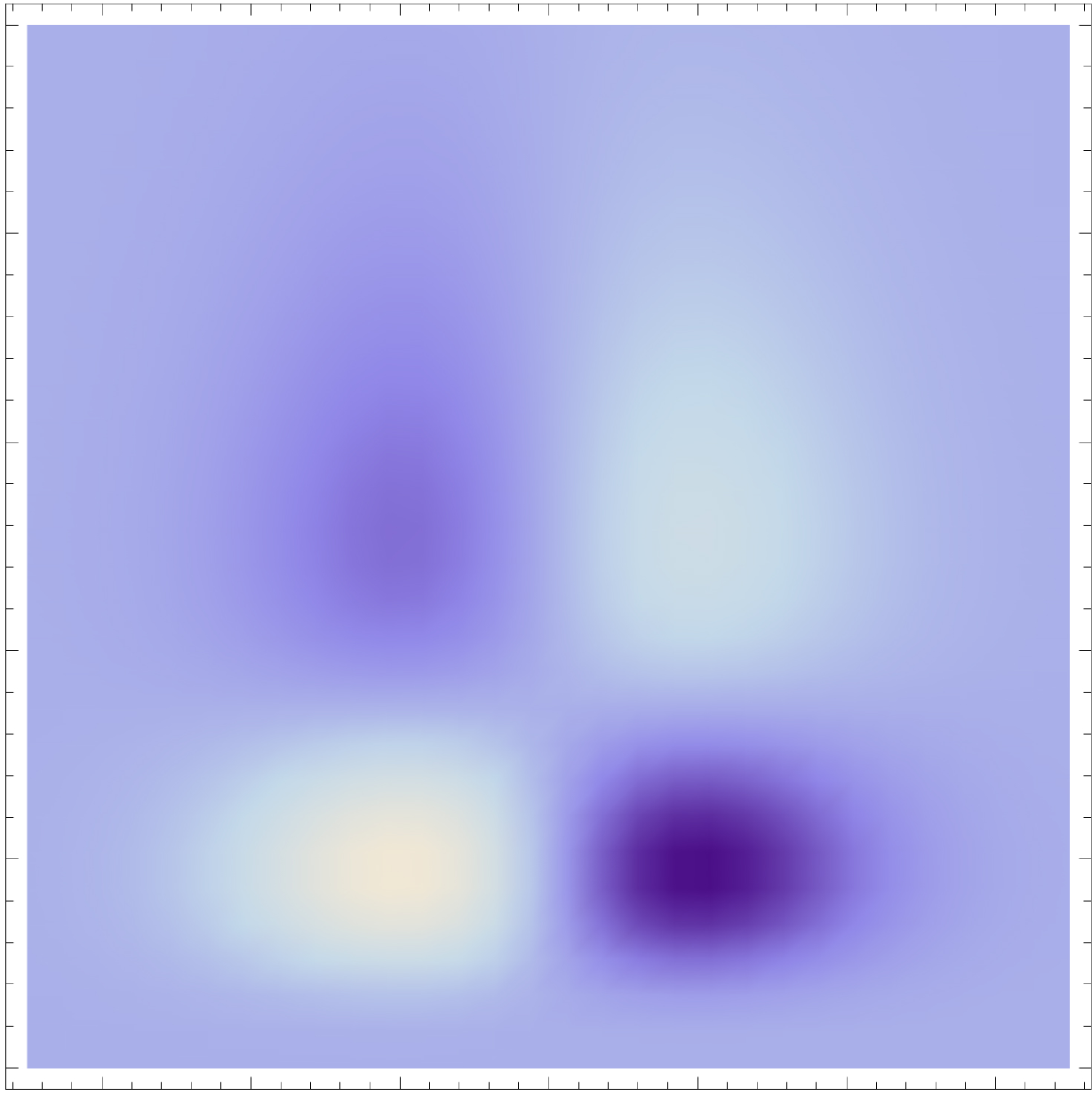} \hspace{-2mm} &
      \hspace{-2mm} \includegraphics[width=0.13\textwidth]{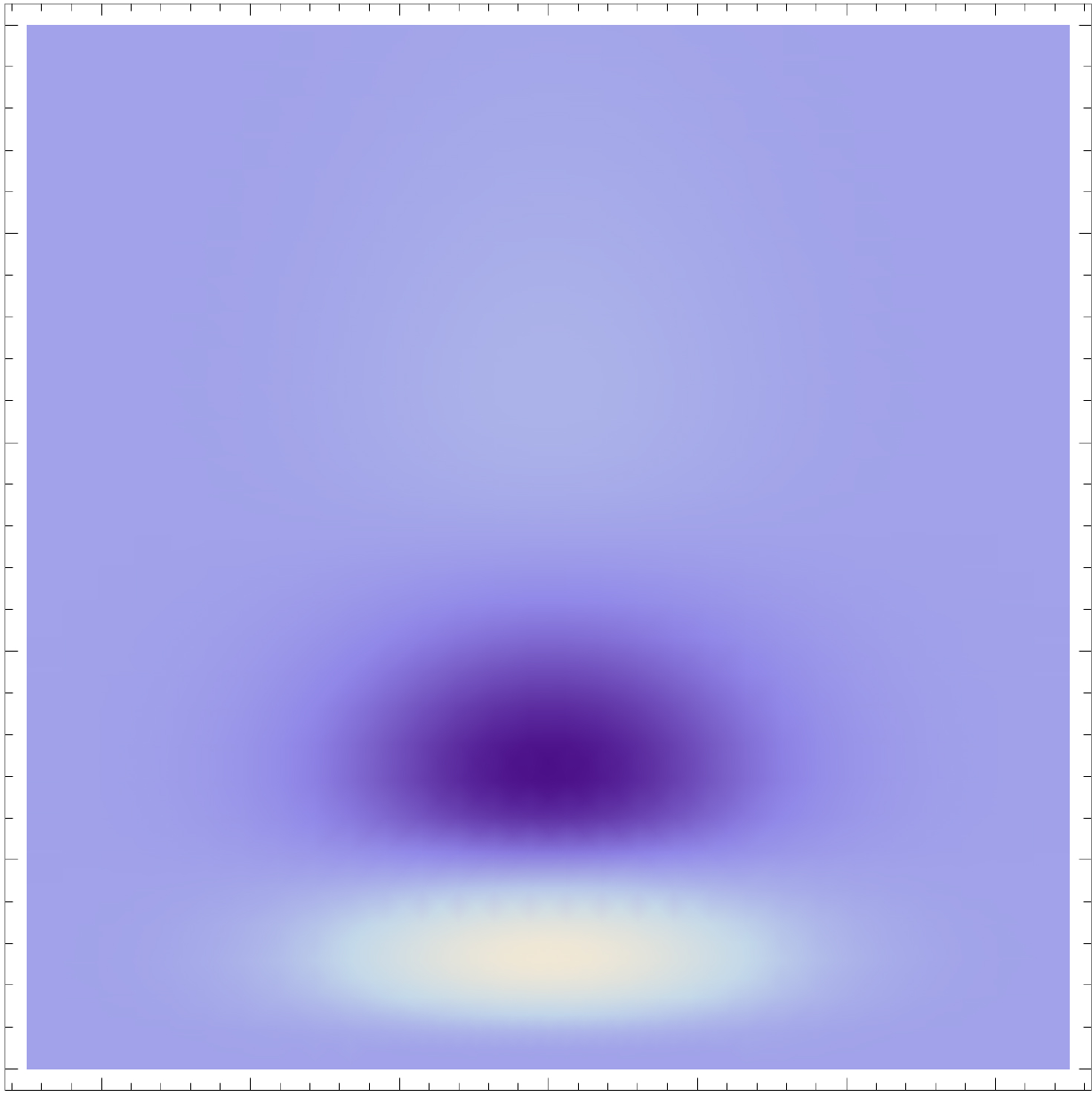} \hspace{-2mm} \\
    \end{tabular} 
  \end{center}
  \caption{{\em Space-time separable kernels\/}
           $T_{x^{m}t^{n}}(x, t;\; s, \tau) = \partial_{x^m t^n} (g(x;\; s) \, h(t;\; \tau))$ 
           up to order two obtained as the composition of Gaussian
           kernels over the spatial domain $x$ and a cascade of
           truncated exponential kernels over the temporal domain $t$
           with a logarithmic distribution of the intermediate
           temporal scale levels ($s = 1$, $\tau = 1$, $K = 7$, $c = \sqrt{2}$).
           (Horizontal axis: space $x$. Vertical axis: time $t$.)}
  \label{fig-non-caus-sep-spat-temp-rec-fields}

   \medskip

  \begin{center}
    \begin{tabular}{c}
     \hspace{-2mm} {\footnotesize $-T(x, t;\; s, \tau, v)$} \hspace{-2mm} \\
      \hspace{-2mm} \includegraphics[width=0.13\textwidth]{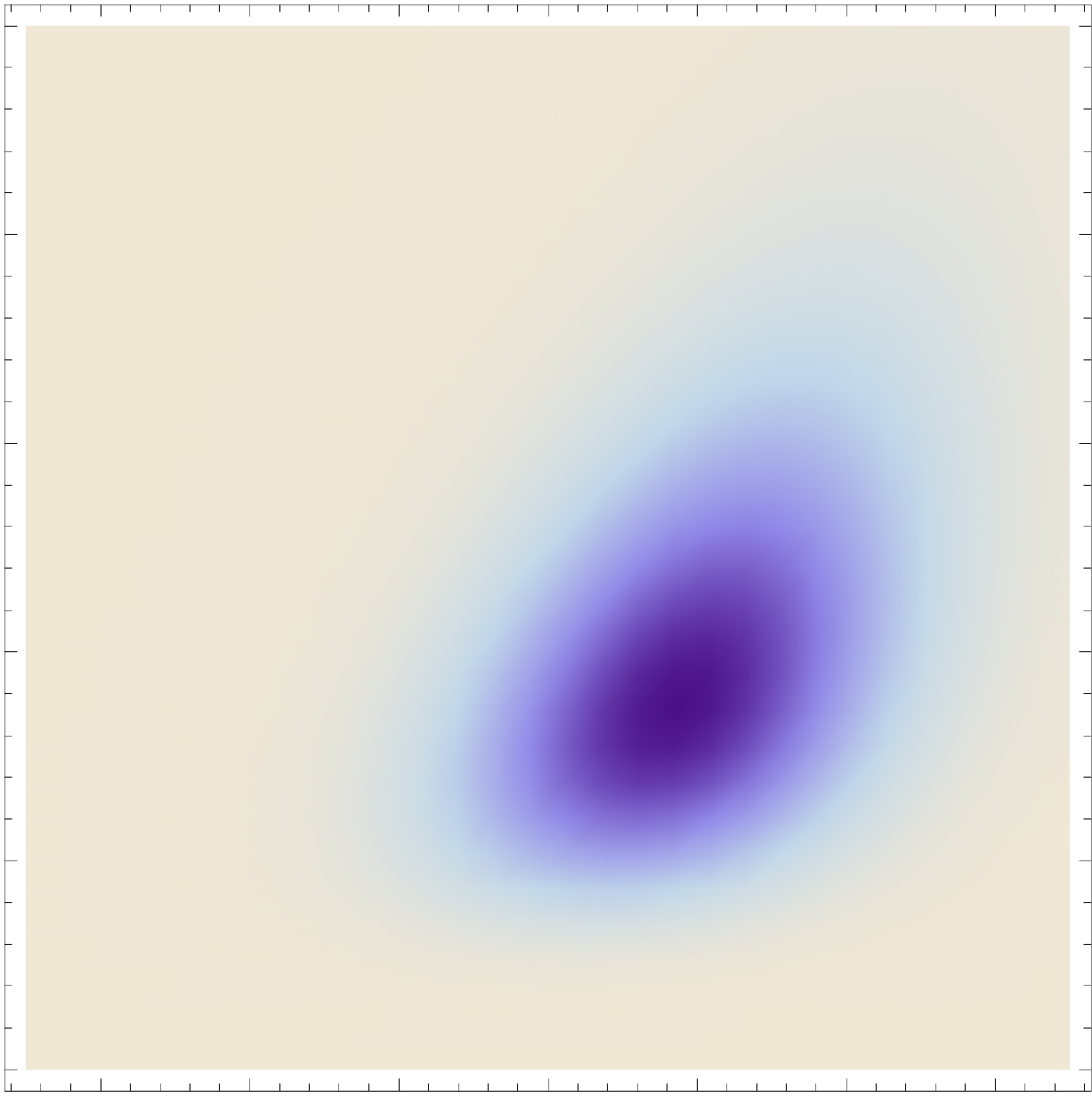} \hspace{-2mm} \\
    \end{tabular} 
  \end{center}
  \vspace{-6mm}
  \begin{center}
    \begin{tabular}{cc}
      \hspace{-2mm} {\footnotesize $T_x(x, t;\; s, \tau, v)$} \hspace{-2mm} 
      & \hspace{-2mm} {\footnotesize $T_t(x, t;\; s, \tau, \delta)$} \hspace{-2mm} \\
      \hspace{-2mm} \includegraphics[width=0.13\textwidth]{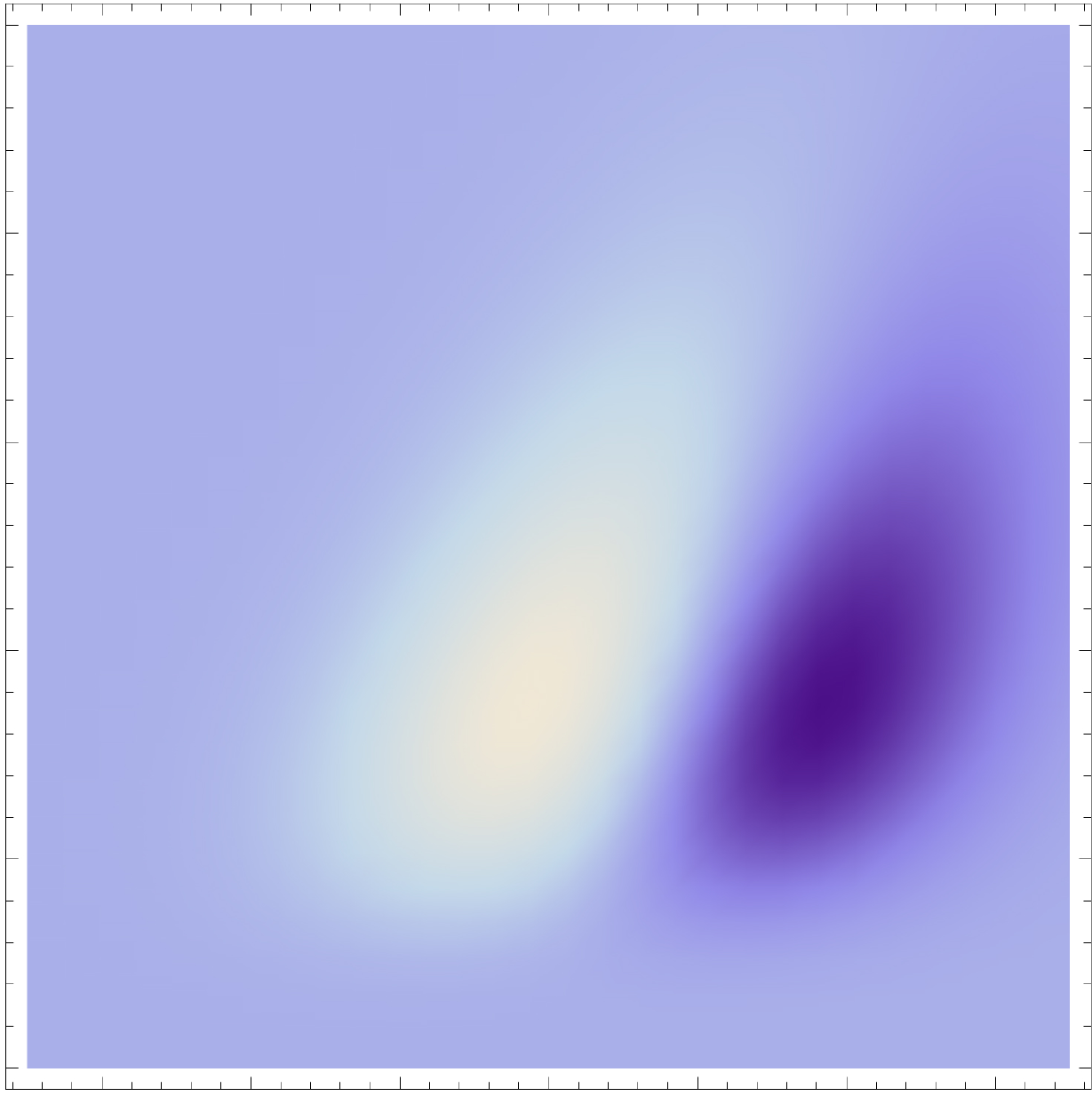} \hspace{-2mm} &
      \hspace{-2mm} \includegraphics[width=0.13\textwidth]{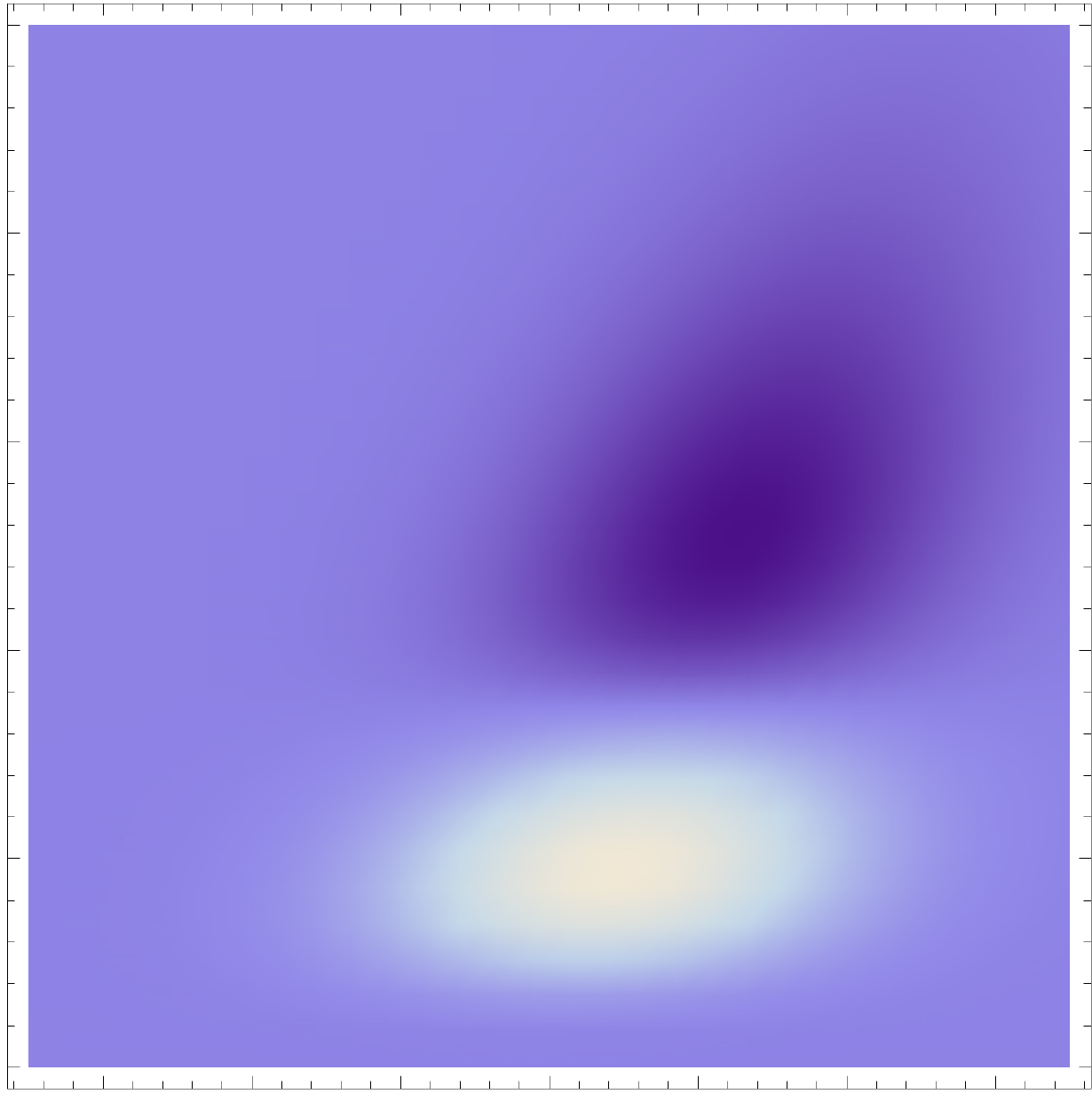} \hspace{-2mm} \\
    \end{tabular} 
  \end{center}
  \vspace{-6mm}
  \begin{center}
    \begin{tabular}{ccc}
      \hspace{-2mm} {\footnotesize $T_{xx}(x, t;\; s, \tau, v)$} \hspace{-2mm} 
      & \hspace{-2mm} {\footnotesize $T_{xt}(x, t;\; s, \tau, v)$} \hspace{-2mm} 
      & \hspace{-2mm} {\footnotesize $T_{tt}(x, t;\; s, \tau, v)$} \hspace{-2mm} \\
      \hspace{-2mm} \includegraphics[width=0.13\textwidth]{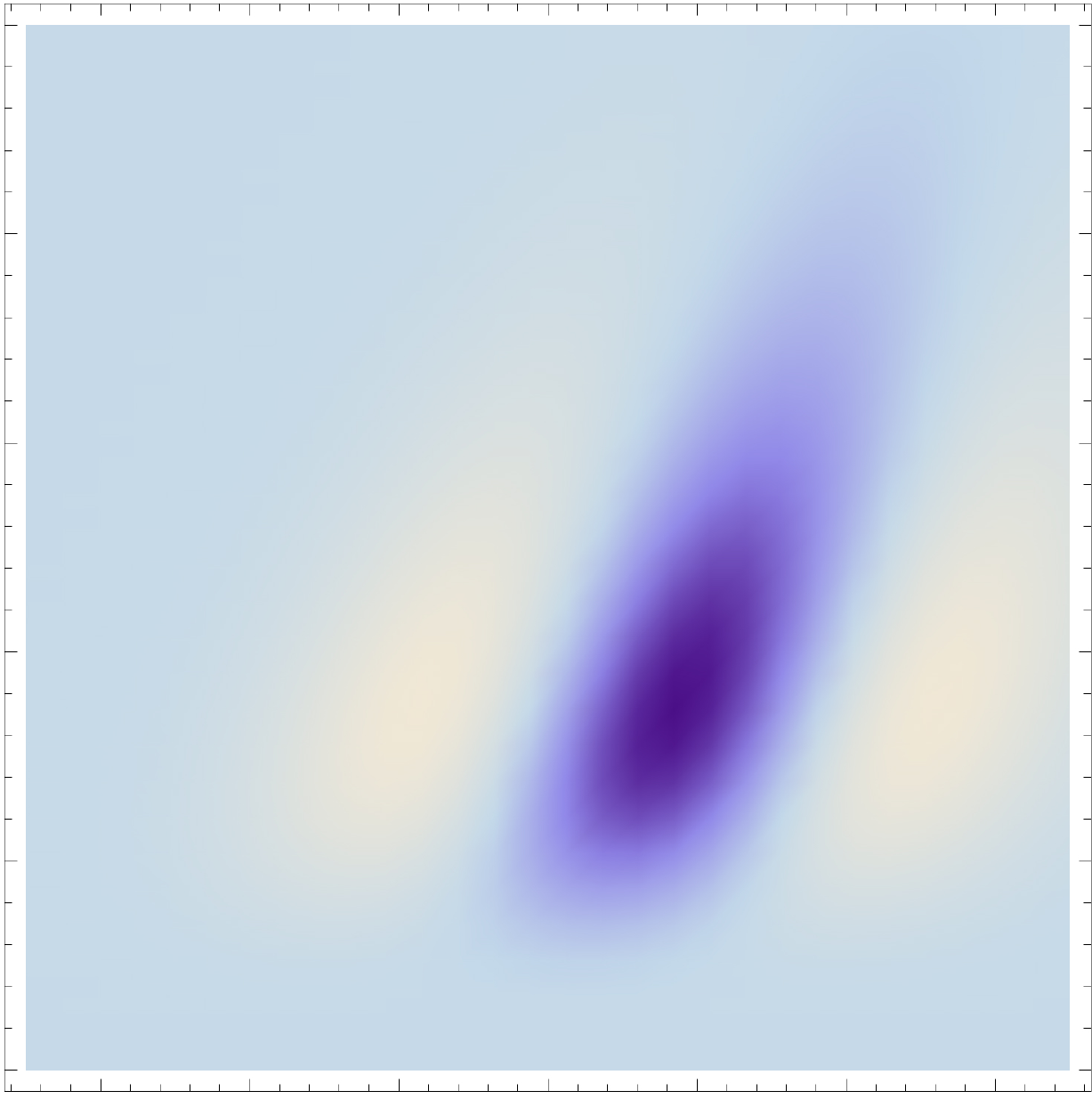} \hspace{-2mm} &
      \hspace{-2mm} \includegraphics[width=0.13\textwidth]{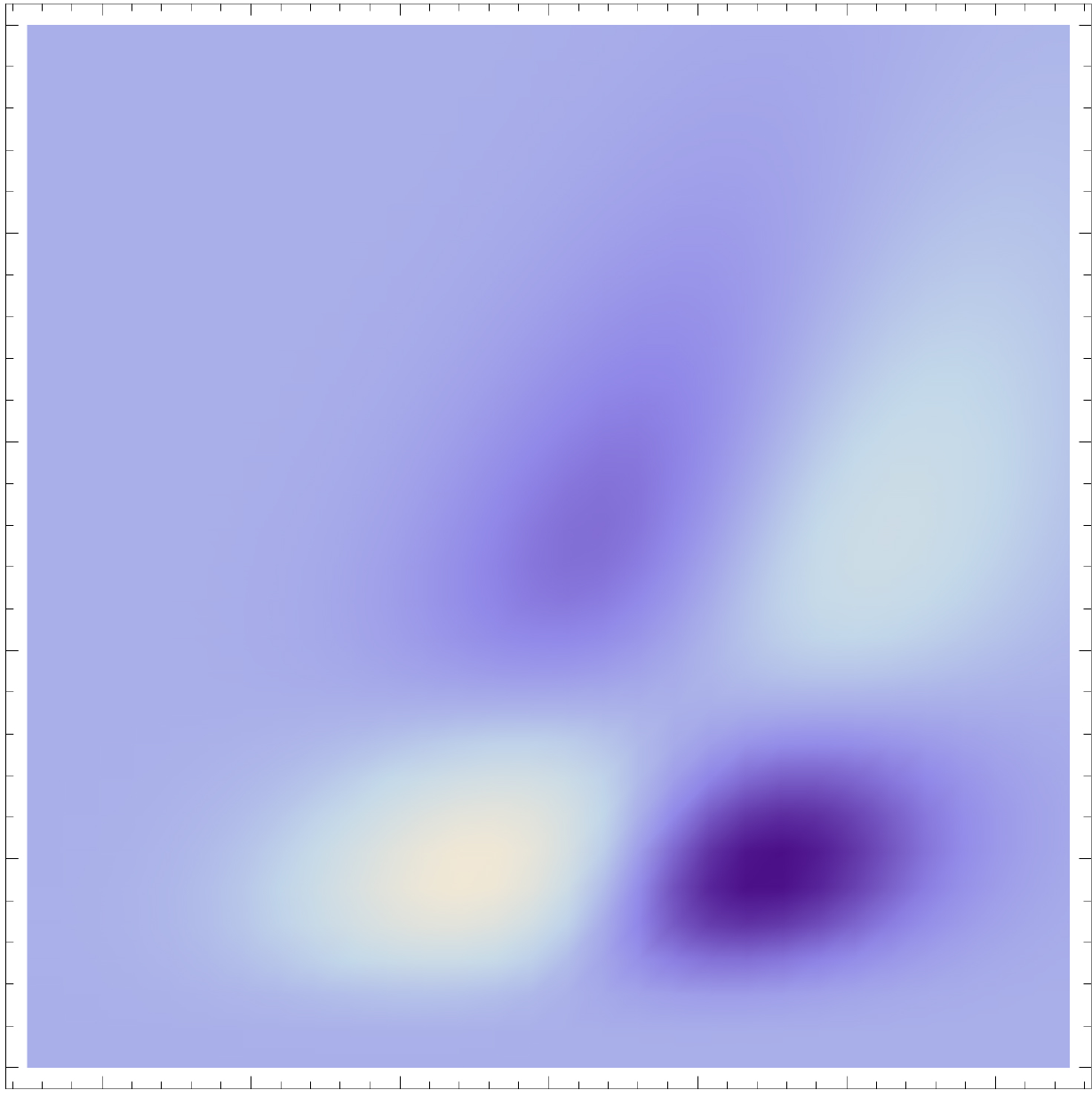} \hspace{-2mm} &
      \hspace{-2mm} \includegraphics[width=0.13\textwidth]{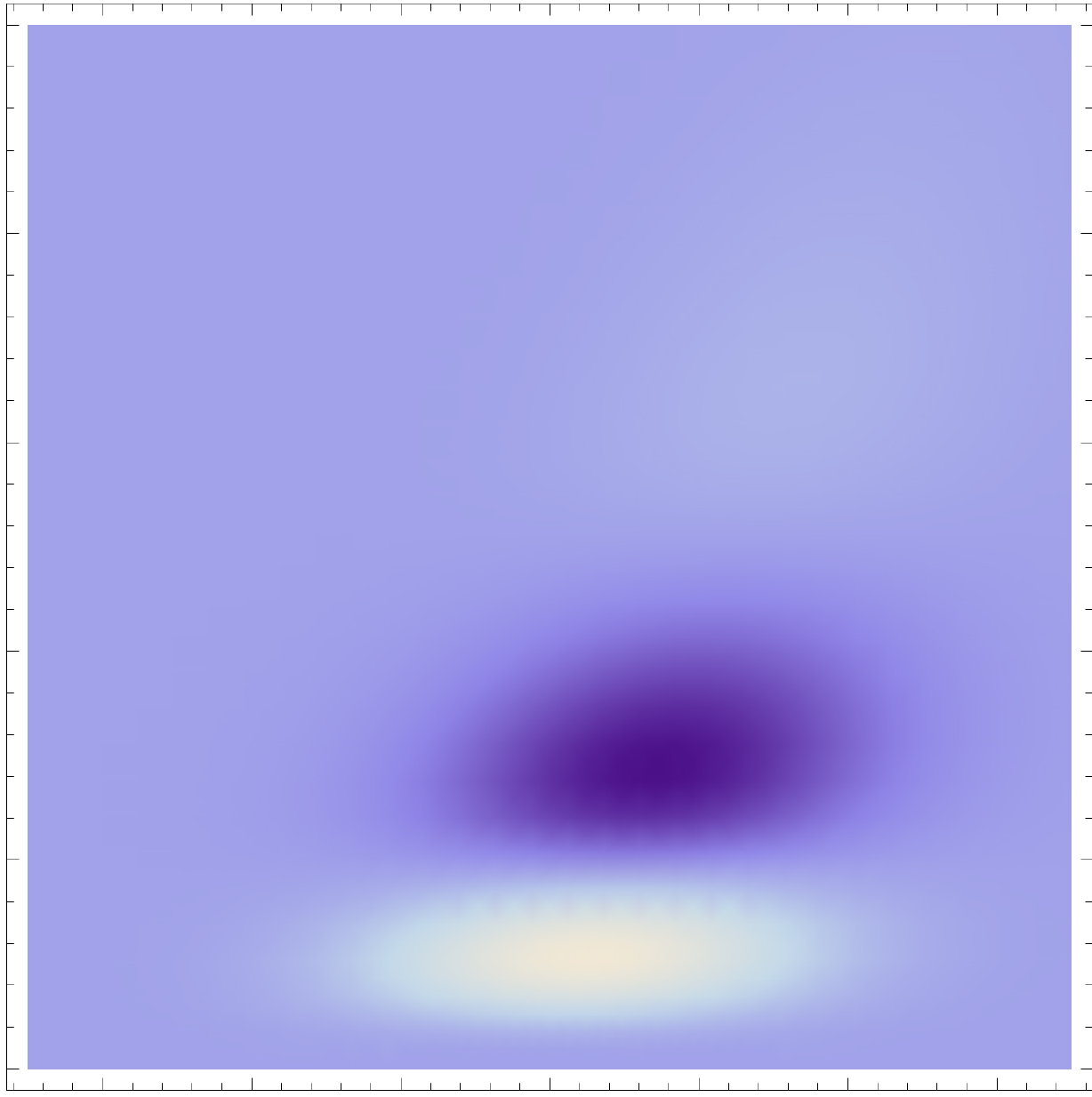} \hspace{-2mm} \\
    \end{tabular} 
  \end{center}
  \caption{{\em Velocity-adapted spatio-temporal kernels\/}
               $T_{x^{m}t^{n}}(x, t;\; s, \tau, v) = \partial_{x^m t^n} (g(x - vt;\; s) \, h(t;\; \tau))$ 
           up to order two obtained as the composition of Gaussian
           kernels over the spatial domain $x$ and a cascade of
           truncated exponential kernels over the temporal domain $t$
           with a logarithmic distribution of the intermediate
           temporal scale levels ($s = 1$, $\tau = 1$, $K = 7$, $c =
           \sqrt{2}$, $v = 0.5$).
           (Horizontal axis: space $x$. Vertical axis: time $t$.)}
  \label{fig-non-caus-vel-adapt-spat-temp-rec-fields}
\end{figure}

\subsection{Partial derivatives}

A most basic approach is to first define a spatio-temporal
scale-space representation 
$L \colon \bbbr^2 \times \bbbr \times \bbbr_+ \times \bbbr_+$
from any video data $f \colon \bbbr^2 \times \bbbr$ and
then defining partial derivatives of any spatial and temporal orders
$m = (m_1, m_2)$ and $n$ at any spatial and
temporal scales $s$ and $\tau$ according to
\begin{multline}
  \label{eq-part-der-spat-temp-scsp}
   L_{x_1^{m_1} x_2^{m_1} t^n}(x_1, x_2, t;\; s, \tau) 
  = \\ \partial_{x_1^{m_1} x_2^{m_2} t^n}
           \left( \left( g(\cdot, \cdot;\; s) \, h(\cdot;\; \tau) \right) *
           f(\cdot, \cdot, \cdot) \right)(x_1, x_2, t;\; s, \tau) 
\end{multline}
leading to a spatio-temporal $N$-jet representation of any order
\begin{equation}
  \{ L_x, L_y, L_t, L_{xx}, L_{xy}, L_{yy}, L_{xt}, L_{yt}, L_{tt}, \dots \}.
\end{equation}
Figure~\ref{fig-non-caus-sep-spat-temp-rec-fields} shows such kernels
up to order two in the case of a 1+1-D space-time.

\subsection{Directional derivatives}

By combining spatial directional
derivative operators over any pair of ortogonal
directions $\partial_{\varphi} = \cos \varphi \, \partial_x + \sin \varphi \, \partial_y$
and
  $\partial_{\orth \varphi} = \sin \varphi \, \partial_x - \cos \varphi \, \partial_y$
and velocity-adapted temporal derivatives $\partial_{t_v} = \partial_t + v_x \, \partial_x + v_y \, \partial_y$
over any motion direction $v = (v_x, v_y, 1)$,
a filter bank of spatio-temporal derivative responses can be created
\begin{equation}
  \label{eq-dir-der-spattemp}
  L_{\varphi^{m_1} \orth \varphi^{m_2} t_v^n} 
   = \partial_{\varphi}^{m_1} \partial_{\orth \varphi}^{m_2} \partial_{t_v}^n L
\end{equation}
for different sampling strategies over image orientations
$\varphi$ and $\orth \varphi$ in image space and over motion directions
$v$ in space-time (see
figure~\ref{fig-non-caus-vel-adapt-spat-temp-rec-fields} for
illustrations of such kernels up to order two 
in the case of a 1+1-D space-time).

Note that as long as the spatio-temporal smoothing operations are
performed based on rotationally symmetric Gaussians over the
spatial domain and using space-time separable kernels over space-time,
the responses to these directional derivative
operators can be directly related to corresponding partial
derivative operators by mere linear combinations.
If extending the rotationally symmetric Gaussian scale-space concept
is to an anisotropic affine Gaussian scale-space and/or if we
make use of non-separable velocity-adapted receptive fields over
space-time in a spatio-temporal scale space, to enable true affine and/or Galilean invariances,
such linear relationships will, however, no longer hold on a similar form.

For the image orientations $\varphi$ and $\orth \varphi$, it is for
purely spatial derivative operations, in the case of rotationally
symmetric smoothing over the spatial domain, in principle sufficient to 
to sample the image orientation according to a uniform distribution on 
the semi-circle using at least $|m|+1$ directional derivative filters 
for 
 derivatives of order $|m|$.

For temporal directional derivative operators to make fully sense in a
geometrically meaningful manner (covariance under Galilean transformations
of space-time), they should however also be combined with
Galilean velocity adaptation of the spatio-temporal smoothing operation
in a corresponding direction $v$ according to
(\ref{eq-spat-temp-RF-model})
(Lindeberg \cite{Lin97-ICSSTCV,Lin10-JMIV}; Laptev and Lindeberg
\cite{LapLin03-IVC,LapCapSchLin07-CVIU}).
Regarding the distribution of such motion directions $v = (v_x, v_y)$, 
it is natural to distribute the magnitudes $|v| = \sqrt{v_x^2 + v_y^2}$ 
according to a self-similar distribution 
\begin{equation}
   |v|_j = |v|_1 \, \varrho^{j} \quad\quad j = 1 \dots J
\end{equation}
for some suitably selected constant $\rho > 1$ and using a uniform distribution 
of the motion directions $e_v = v/|v|$ on the full circle.

\subsection{Differential invariants over spatial derivative operators}

Over the spatial domain, we will in this treatment make use of the
gradient magnitude $|\nabla_{(x, y)} L|$,
the Laplacian $\nabla_{(x, y)}^2 L$,
the determinant of the Hessian $\det {\cal H}_{(x, y)} L$,
the rescaled level curve curvature $\tilde{\kappa}(L)$
and the quasi quadrature energy measure ${\cal Q} _{(x, y)} L$,
which are transformed to
scale-normalized differential expressions with $\gamma = 1$
(Lindeberg \cite{Lin93-Dis,Lin97-IJCV,Lin08-EncCompSci}):
\begin{align}
   \begin{split}
     |\nabla_{(x, y),norm} L| 
    & = \sqrt{s L_x^2 + s L_y^2} 
    = \sqrt{s} \, |\nabla_{(x, y)} L|,
  \end{split}\\
  \begin{split}
   \label{eq-spat-lapl-norm}
     \nabla_{(x, y),norm}^2 L 
    & = s \, (L_{xx} + L_{yy}) = s \, \nabla_{(x, y)}^2 L,
  \end{split}\\
  \begin{split}
     \det {\cal H}_{(x, y),norm} L 
        & = s^2 (L_{xx}  L_{yy} - L_{xy}^2) 
 \end{split}\nonumber\\
  \begin{split}
     \label{eq-spat-dethess-norm}
         &   = s^2 \det {\cal H} _{(x, y)} L,
  \end{split}\\
  \begin{split}
     \tilde{\kappa}_{norm}(L) 
     & = s^2 (L_x^2  L_{yy} + L_y^2  L_{xx} - 2 L_x  L_y  L_{xy})
 \end{split}\nonumber\\
  \begin{split}
     \label{eq-spat-kappa-norm}
      & = s^2 \, \tilde{\kappa}(L) ,
  \end{split}\\
  \begin{split}
     {\cal Q} _{(x, y),norm} L 
     & = s \, (L_x^2 + L_y^2) 
 \end{split}\nonumber\\
  \begin{split}
     \label{eq-spat-quasi-norm}
    & \phantom{=} \,
      + C s^2 \left( L_{xx}^2 + 2 L_{xy}^2 +  L_{yy}^2 \right),
  \end{split}
\end{align}
(and the corresponding unnormalized expressions are obtained
by replacing $s$ by 1).%
\footnote{When using the Laplacian operator in this paper, the
  notation $\nabla_{(x, y)}^2$ should be understood as the covariant expression
$\nabla_{(x, y)}^2 =\nabla_{(x, y)}^T \nabla_{(x, y)}$ with
$\nabla_{(x, y)} = (\partial_x, \partial_y)^T$, etc.}
For mixing first- and second-order
derivatives in the quasi quadrature entity ${\cal Q} _{(x, y),norm}
L$, we use $C = 2/3$ or 
$C = e/4$ according to (Lindeberg \cite{Lin97-AFPAC}).

\subsection{Space-time coupled spatio-temporal derivative expressions}
\label{sec-spat-temp-der-expr}

A more general approach to spatio-temporal feature detection than
partial derivatives or directional derivatives consists of
defining spatio-temporal derivative operators that combine spatial and
temporal derivative operators in an integrated manner.

\paragraph{Temporal derivatives of the spatial Laplacian.}

Inspired by the way neurons in the lateral geniculate nucleus (LGN)
respond to visual input (DeAngelis et al \cite{DeAngOhzFre95-TINS,deAngAnz04-VisNeuroSci}),
which for many LGN cells can be modelled by idealized operations of
the form (Lindeberg \cite[equation~(108)]{Lin13-BICY})
\begin{equation}
  \label{eq-lgn-model-2}
  h_{LGN}(x, y, t;\; s, \tau) 
  = \pm (\partial_{xx} + \partial_{yy}) \, g(x, y;\; s) \, \partial_{t^n} \, h(t;\; \tau),
\end{equation}
we can define the following differential entities
\begin{align}
  \begin{split}
    \partial_t (\nabla_{(x,y)}^2 L) & = L_{xxt} + L_{yyt}
  \end{split}\\
  \begin{split}
    \partial_{tt} (\nabla_{(x,y)}^2 L) & = L_{xxtt} + L_{yytt}
  \end{split}
\end{align}
and combine these entities into a quasi quadrature measure over
time of the form
\begin{equation}
  \label{eq-quasi-t-spat-lapl}
  {\cal Q}_t(\nabla_{(x,y)}^2 L) 
  = \left( \partial_t (\nabla_{(x,y)}^2 L) \right)^2 + C \left( \partial_{tt} (\nabla_{(x,y)}^2 L) \right)^2,
\end{equation}
where $C$ again may be set to $C = 2/3$ or $C = e/4$.
The first entity $\partial_t (\nabla_{(x,y)}^2 L)$ can be expected to give
strong respondes to spatial blob responses whose intensity values 
vary over time, whereas the second entity $\partial_{tt} (\nabla_{(x,y)}^2 L)$
can be expected to give strong responses to spatial blob responses
whose intensity values vary strongly around local minima or local
maxima over time. 

By combining these two entities into a quasi quadrature measure 
${\cal Q}_t(\nabla_{(x,y)}^2 L)$ over time, we obtain a differential entity that can be 
expected to give strong responses when then the intensity varies 
strongly over both image space and over time, while giving no response
if there are no intensity variations over space or time.
Hence, these three differential operators could be regarded as a primitive spatio-temporal
interest operators that can be seen as compatible with existing knowledge about neural
processes in the LGN.

\paragraph{Temporal derivatives of the determinant of the spatial
  Hessian.}

Inspired by the way local extrema of the determinant of the spatial
Hessian (\ref{eq-spat-dethess-norm}) can be shown to constitute a better 
interest point detector than local extrema of the spatial Laplacian
(\ref{eq-spat-lapl-norm}) (Lindeberg \cite{Lin12-JMIV,Lin15-JMIV}), 
we can compute corresponding first- and
second-order derivatives over time of the determinant of the
spatial Hessian
\begin{align}
  \begin{split}
    \partial_t (\det {\cal H}_{(x,y)} L) 
    & = L_{xxt} L_{yy} + L_{xx} L_{yyt} - 2 L_{xy} L_{xyt}
  \end{split}\\
  \begin{split}
    \partial_{tt} (\det {\cal H}_{(x,y)} L)
    & = L_{xxtt} L_{yy} + 2 L_{xxt} L_{yyt}  + L_{xx} L_{yytt}
  \end{split}\nonumber\\
  \begin{split}
     & \phantom{=} \quad
           - 2 L_{xyt}^2 - 2 L_{xy} L_{xytt}
  \end{split}
\end{align}
and combine these entities into a quasi quadrature measure over time
\begin{multline}
   {\cal Q}_t (\det {\cal H}_{(x,y)} L)
  = \\ \left( \partial_t (\det {\cal H}_{(x,y)} L) \right)^2 + C \left( \partial_{tt} (\det {\cal H}_{(x,y)} L) \right)^2.
\end{multline}
As the determinant of the spatial Hessian can be expected to give
strong responses when there are strong intensity variations in two
spatial directions, the corresponding spatio-temporal operator 
${\cal Q}_t (\det {\cal H}_{(x,y)} L)$ can be expected to give strong
responses at such spatial points at which there are additionally
strong intensity variations over time as well.

\paragraph{Genuinely spatio-temporal interest operators.}

A less temporal slice oriented and more genuine 3-D
spatio-temporal approach to defining interest point detectors 
from second-order spatio-temporal derivatives is by considering
feature detectors such as {\em the determinant of the spatio-temporal Hessian
  matrix\/}
\begin{align}
  \begin{split}
    \det {\cal H}_{(x, y, t)} L 
    = \, &  L_{xx} L_{yy} L_{tt} + 2 L_{xy} L_{xt} L_{yt} 
  \end{split}\nonumber\\
  \begin{split}
     \label{eq-spat-temp-det-hess}
      - L_{xx} L_{yt}^2 - L_{yy} L_{xt}^2 - L_{tt} L_{xy}^2,
  \end{split}
\end{align}
the {\em rescaled spatio-temporal Gaussian curvature\/}
\begin{align}
  \begin{split}
    & {\cal G}_{(x, y, t)}(L)
  \end{split}\nonumber\\
  \begin{split}
    & = \left( 
              (L_t (L_{xx} L_t - 2 L_x L_{xt}) + L_x^2 L_{tt}) \times
           \right.  
  \end{split}\nonumber\\
  \begin{split}
     & \phantom{= \left( \right.} \,
            \left.    
              (L_t (L_{yy} L_t - 2 L_y L_{yt}) +L_y^2 L_{tt}) 
            \right. 
  \end{split}\nonumber\\
  \begin{split} 
     \label{eq-spat-temp-gauss-curv}
     & \phantom{= \left( \right.} \,
             \left. 
               -(L_t (-L_x L_{yt} + L_{xy} L_t - L_{xt} L_y) + L_x L_y L_{tt})^2 
             \right)/L_t^2,
  \end{split}
\end{align}
which can be seen as a 3-D correspondence of the 2-D rescaled
level curve curvature operator $\tilde{\kappa}_{norm}(L)$ 
in equation~(\ref{eq-spat-kappa-norm}),
or possibly trying to define a {\em spatio-temporal Laplacian\/}
\begin{equation}
  \label{eq-spat-temp-lapl}
  \nabla_{(x, y, t)}^2 L = L_{xx} + L_{yy} + \varkappa^2 L_{tt}.
\end{equation}
Detection of local extrema of the determinant of the spatio-temporal
Hessian has been proposed as a spatio-temporal interest point detector
by (Willems {\em et al.\/}\ \cite{WilTuyGoo08-ECCV}). Properties of
the 3-D rescaled Gaussian curvature
have been studied in (Lindeberg \cite{Lin12-JMIV}).

If aiming at defining a spatio-temporal analogue of the Laplacian
operator, one does, however, need to consider that the most
straightforward way of defining such an operator 
$\nabla_{(x, y, t)}^2 L = L_{xx} + L_{yy} + L_{tt}$
is not covariant under independent scaling of the spatial and temporal
coordinates as occurs if observing the same scene with cameras having
independently different spatial and temporal sampling
rates. Therefore, the choice of the relative weighting factor 
$\varkappa^2$ between temporal {\em vs.\/}\ spatial derivatives
introduced in equation~(\ref{eq-spat-temp-lapl}) is in
principle arbitrary. 
By the homogeneity of the determinant of the Hessian
(\ref{eq-spat-temp-det-hess}) and the spatio-temporal Gaussian
curvature (\ref{eq-spat-temp-gauss-curv}) in terms of the orders of spatial {\em vs.\/}\ temporal
differentiation that are multiplied in each term, these expressions
are on the other hand truly covariant under independent rescalings of the spatial and
temporal coordinates and therefore better candidates for being used as
spatio-temporal interest operators, unless the relative scaling and weighting of
temporal {\em vs.\/}\ spatial coordinates can be handled by some
complementary mechanism.

\paragraph{Spatio-temporal quasi quadrature entities.}

Inspired by the way the spatial quasi quadrature measure ${\cal Q} _{(x, y)} L$
in (\ref{eq-spat-quasi-norm}) is defined as a measure of the amount of
information in first- and second-order spatial derivatives, we may
consider different types of spatio-temporal extensions of this entity
\begin{align}
  \begin{split}
      {\cal Q} _{1,(x, y, t)} L 
     & = L_x^2 + L_y^2 + \varkappa^2 L_t^2 +
  \end{split}\nonumber\\
  \begin{split}
      \label{eq-Q1-not-scalenorm-ders}
      & \phantom{=} \,
            + C \left( L_{xx}^2 + 2 L_{xy}^2 + L_{yy}^2 \right.
  \end{split}\nonumber\\
  \begin{split}
      & \phantom{= + C ()} \, \left.
           + \varkappa^2 (L_{xt}^2 + L_{yt}^2) + \varkappa^4 L_{tt}^2\right),
  \end{split}\\
  \begin{split}
      {\cal Q} _{2,(x, y, t)} L 
       & = {\cal Q}_t L \times {\cal Q}_{(x, y)} L 
  \end{split}\nonumber\\
  \begin{split}
      \label{eq-Q2-not-scalenorm-ders}
      & = \left( L_t^2 + C L_{tt}^2\right) \times
  \end{split}\nonumber\\
  \begin{split}
      & \phantom{=} \,\,\,
            \left( L_x^2 + L_y^2 + C \left( L_{xx}^2 + 2 L_{xy}^2 + L_{yy}^2 \right) \right),
  \end{split}\\
  \begin{split}
      {\cal Q} _{3,(x, y, t)} L 
     & = {\cal Q}_{(x, y)} L_t + C \, {\cal Q}_{(x, y)} L_{tt} 
 \end{split}\nonumber\\
  \begin{split}
      & = L_{xt}^2 + L_{yt}^2 + C \left( L_{xxt}^2 + 2 L_{xyt}^2 + L_{yyt}^2 \right)
  \end{split}\nonumber\\
  \begin{split}
      & \phantom{=}  \,
           + C \, \left( L_{xtt}^2 + L_{ytt}^2 
          \right.
 \end{split}\nonumber\\
 \begin{split}
      \label{eq-Q3-not-scalenorm-ders}
      & \phantom{=  \, + C \, \left( \right.}
      \left.
    + C \left( L_{xxtt}^2 + 2 L_{xytt}^2 + L_{yytt}^2 \right) \right),
    \end{split}
\end{align}
where in the first expression when needed because of different dimensionalities
in terms of spatial {\em vs.\/}\ temporal derivatives, a free
parameter $\varkappa$ has been included to adapt the differential
expressions to unknown relative scaling and thus weighting
between the temporal {\em vs.\/}\ spatial dimensions.%
\footnote{To make the differential entities in
  equations~(\ref{eq-Q1-not-scalenorm-ders}),
  (\ref{eq-Q2-not-scalenorm-ders})  and
  (\ref{eq-Q3-not-scalenorm-ders}) fully consistent and meaningful,
  they do additionally have to be transformed into scale-normalized
  derivatives as later done in equations~(\ref{eq-Q1-scalenorm-ders}),
  (\ref{eq-Q2-scalenorm-ders})  and
  (\ref{eq-Q3-scalenorm-ders}).
 With scale-normalized derivatives
 for $\gamma = 1$, the resulting scale-normalized
 derivatives then become dimensionless, which makes it possible to add 
   first- and second-order derivatives of the same variable
 (over either space or time) in a scale-invariant manner. Then, similar arguments as are used
 for deriving the blending parameter $C$ between first- and
 second-order temporal derivatives in (Lindeberg \cite{Lin97-AFPAC}) can be used for deriving a similar
 blending parameter between first- and second-order spatial derivatives.}

The formulation of these quasi quadrature entities is
inspired by the existence of non-linear complex cells in the
primary visual cortex that: (i)~do not obey the superposition principle,
(ii)~have response properties independent of the
polarity of the stimuli and (iii)~are rather insensitive to the phase of
the visual stimuli (Hubel and Wiesel \cite{HubWie59-Phys,HubWie62-Phys}).
Specifically, De~Valois {\em et al.\/}\ \cite{ValCotMahElfWil00-VR}
show that first- and second-order receptive fields
typically occur in pairs that can be modelled as approximate Hilbert
pairs.

Within the framework of the presented spatio-temporal scale-space
concept, it is interesting to note that non-linear receptive 
fields with qualitatively similar properties can be constructed
by squaring first- and second-order derivative responses and summing up
these components (Koenderink and van Doorn \cite{KoeDoo90-BC}).
The use of quasi quadrature model can therefore be interpreted as a
Gaussian derivative based
analogue of energy models as proposed by
Adelson and Bergen \cite{AdeBer85-JOSA} and Heeger \cite{Hee92-VisNeuroSci}.
To obtain local phase independence over variations over both space and
time simultaneously, we do here additionally extend the notion of quasi quadrature to composed space-time,
by simultaneously summing up squares of odd and even filter responses
over both space and time, leading to quadruples or octuples of filter
responses, complemented by additional terms to achieve rotational invariance over the
spatial domain.

For the first quasi quadrature entity ${\cal Q} _{1,(x, y, t)} L$ to
respond, it is sufficient if there are intensity
variations in the image data either over space or over time.
For the second quasi quadrature entity ${\cal Q} _{2,(x, y, t)} L$ to
respond, it is on the other hand necessary that there are intensity
variations in the image data over both space and time.
For the third quasi quadrature entity ${\cal Q} _{3,(x, y, t)} L$ to
respond, it is also necessary that there are intensity variations in
the image data over both space and time.
Additionally, the third quasi quadrature entity ${\cal Q} _{3,(x, y, t)} L$ 
requires there to be intensity variations over both space and time for
each primitive receptive field in terms of plain partial derivatives
that contributes to the output of the composed quadrature entity.
Conceptually, the third quasi quadrature entity can therefore be seen
as more related to the form of temporal quasi quadrature entity applied to the
idealized model of LGN cells in (\ref{eq-quasi-t-spat-lapl})
\begin{equation}
  \label{eq-quasi-t-spat-lapl-2}
  {\cal Q}_t(\nabla_{(x,y)}^2 L) 
  = \left( \nabla_{(x,y)}^2 L_t \right)^2 + C \left( \nabla_{(x,y)}^2 L_{tt} \right)^2
\end{equation}
with the difference that the spatial Laplacian operator $\nabla_{(x,y)}^2$
followed by squaring in (\ref{eq-quasi-t-spat-lapl-2}) is here replaced 
by the spatial quasi quadrature operator ${\cal Q} _{(x, y)}$.

These feature detectors can therefore be seen as biologically inspired
change detectors or as ways of measuring the combined strength of a
set of receptive fields at any point, as possibly combined with
variabilities over other parameters in the family of receptive fields.

\begin{figure*}[hbtp]
  \begin{center}
    \begin{tabular}{ccc}
      {\small original video $f$} 
      & {\small spat-temp smoothing $L$} 
      & {\small $\nabla_{(x, y),norm}^2 L$} \\
      \includegraphics[width=0.30\textwidth]{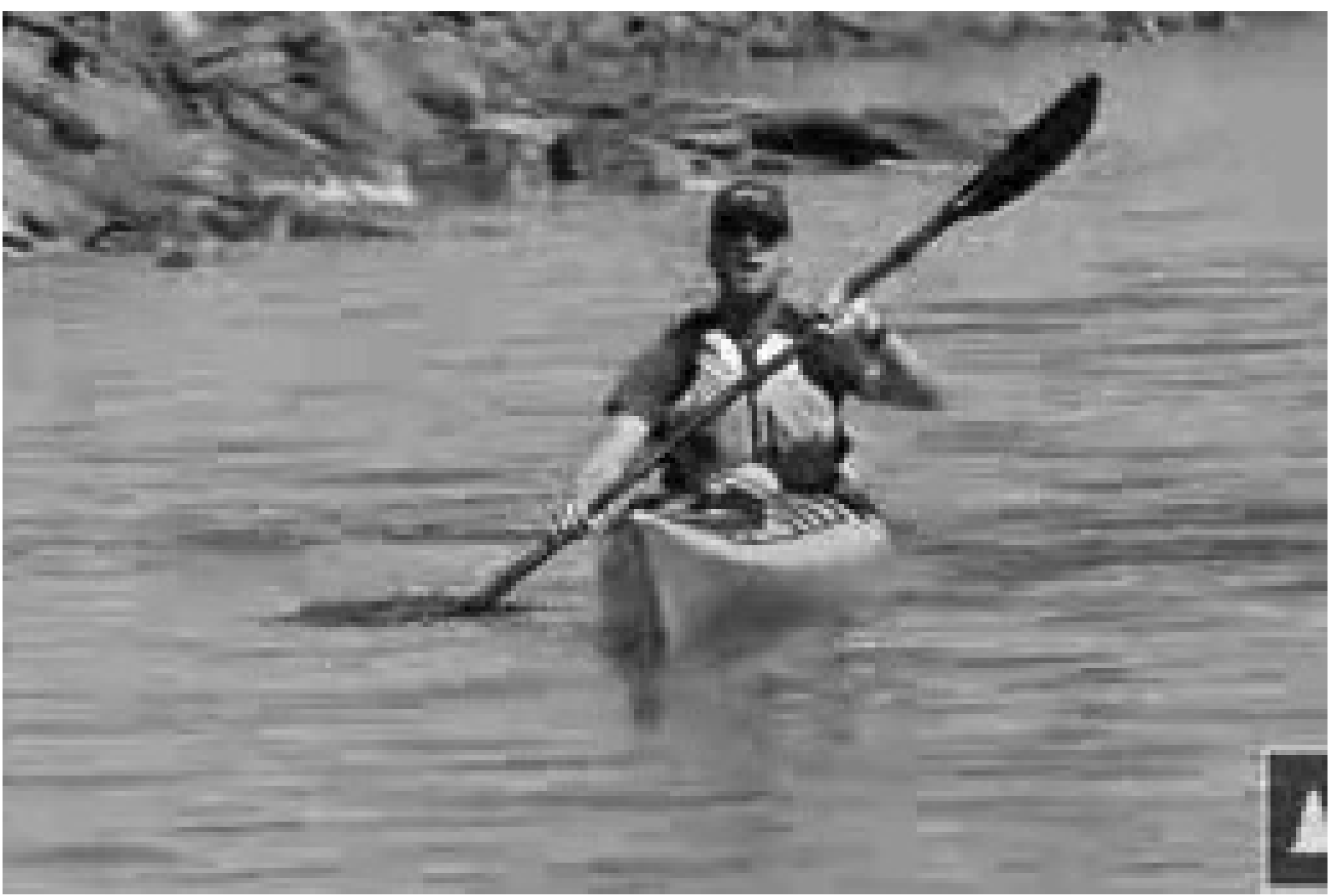} &
      \includegraphics[width=0.30\textwidth]{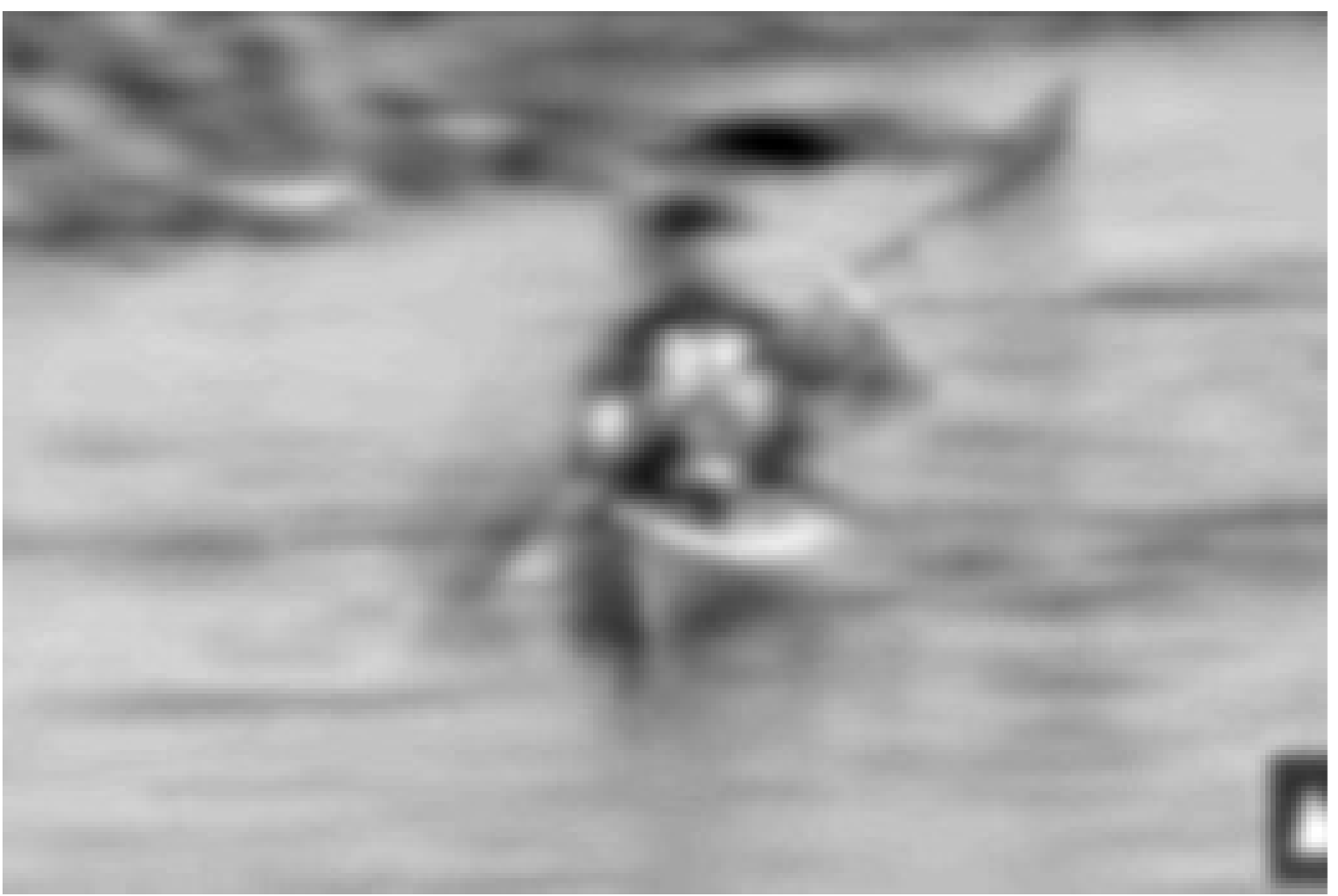} &
      \includegraphics[width=0.30\textwidth]{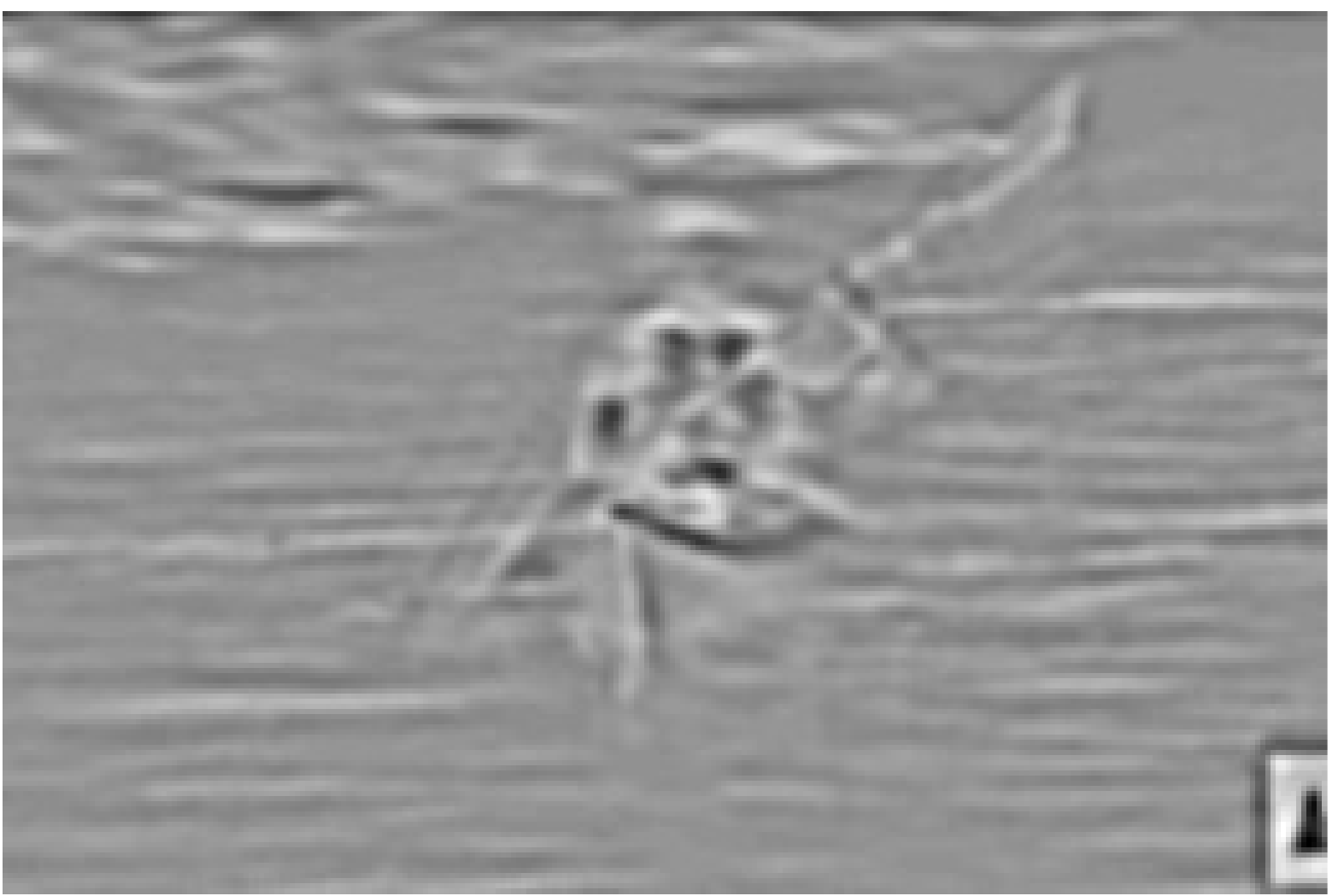} \\
    \end{tabular} 
  \end{center}

  \begin{center}
    \begin{tabular}{ccc}
      {\small $L_{t,norm}$}
      & {\small $L_{tt,norm}$}
      & {\small $\det {\cal H}_{(x, y),norm} L$} \\
      \includegraphics[width=0.30\textwidth]{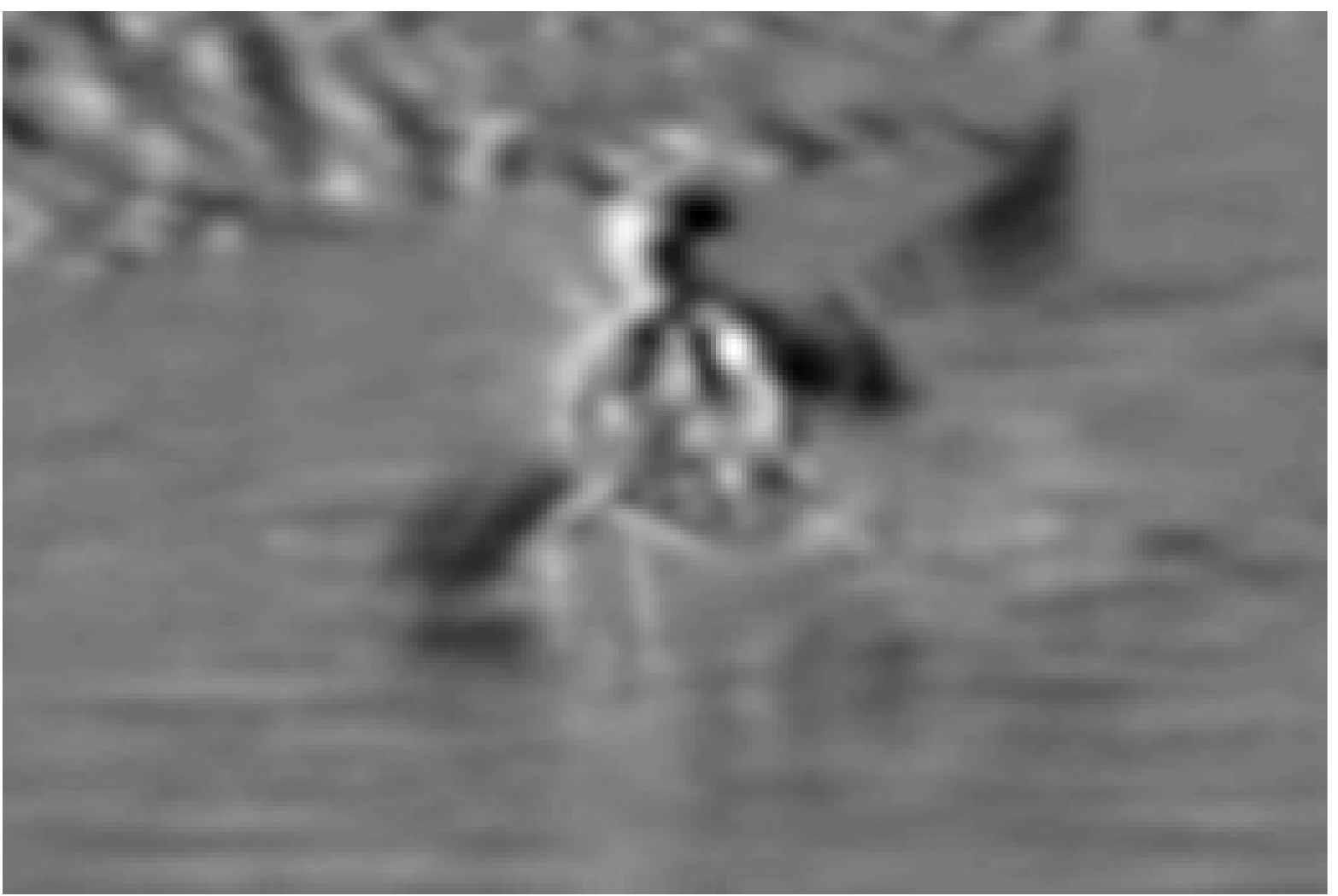} &
      \includegraphics[width=0.30\textwidth]{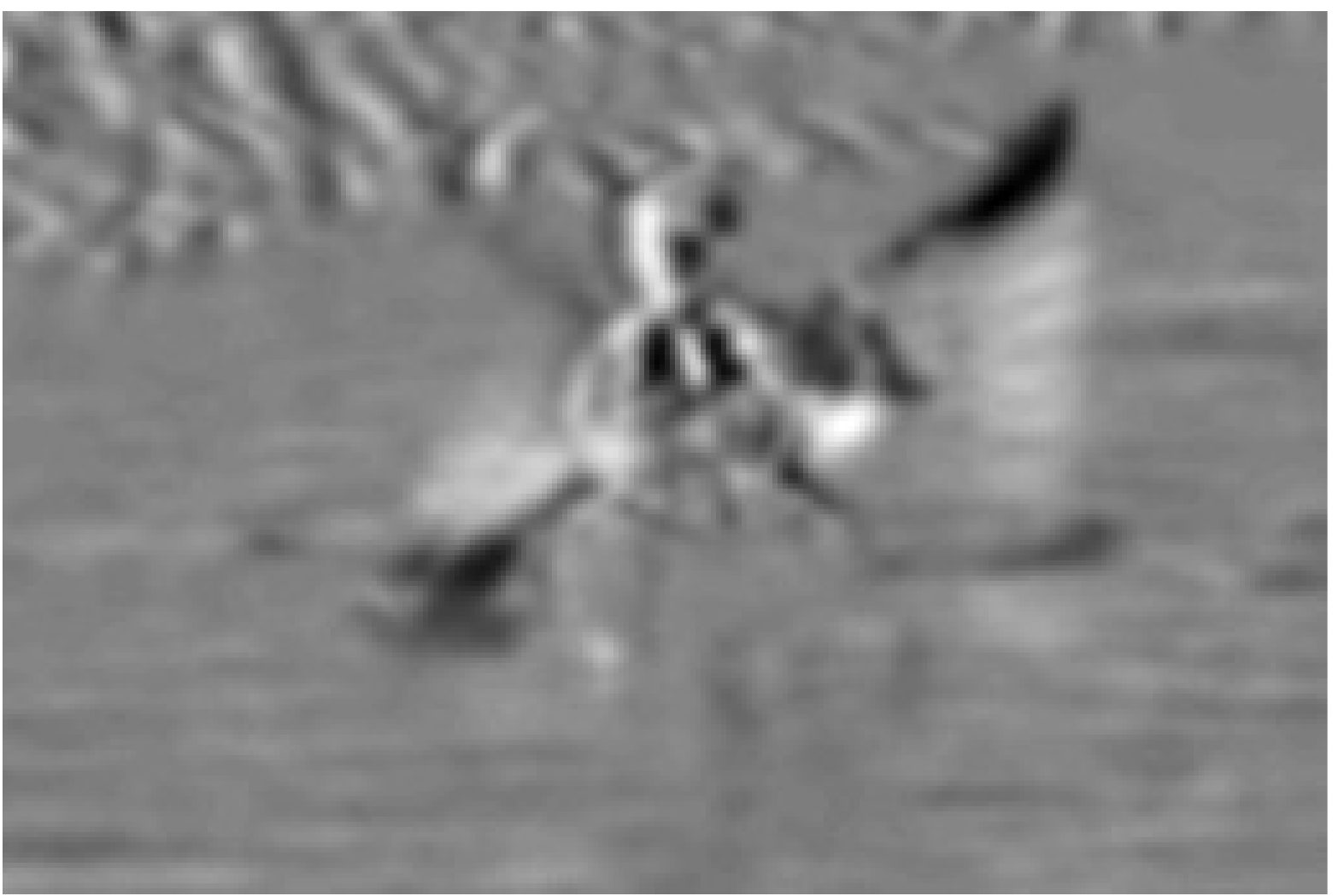} &
      \includegraphics[width=0.30\textwidth]{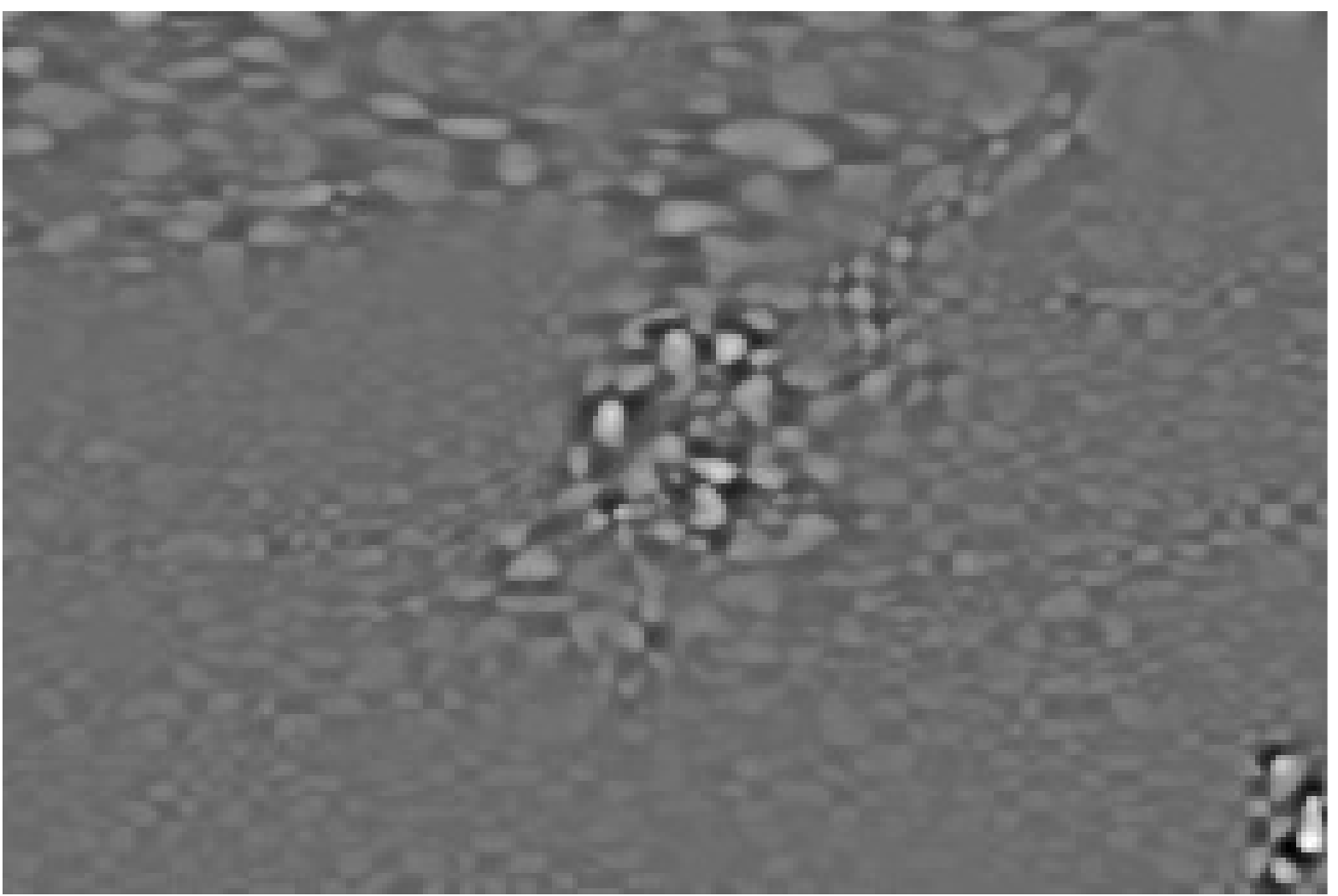} \\
    \end{tabular} 
  \end{center}

  \begin{center}
    \begin{tabular}{ccc}
      {\small $\partial_t (\nabla_{(x,y),norm}^2 L)$} 
      & {\small $\partial_{tt} (\nabla_{(x,y),norm}^2 L)$} 
      & {\small $-\sqrt{{\cal Q}_t(\nabla_{(x,y),norm}^2 L)}$} \\
      \includegraphics[width=0.30\textwidth]{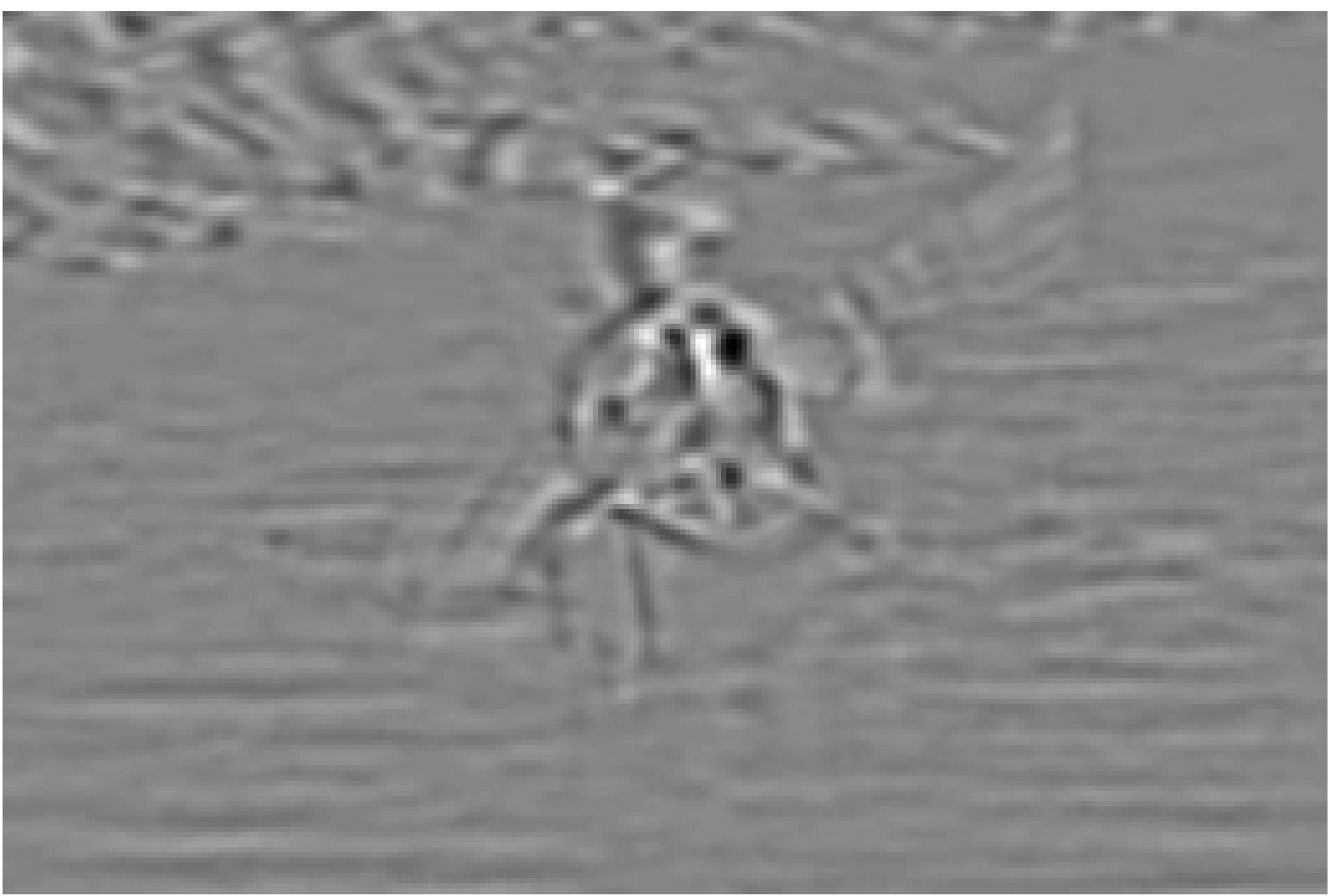} &
      \includegraphics[width=0.30\textwidth]{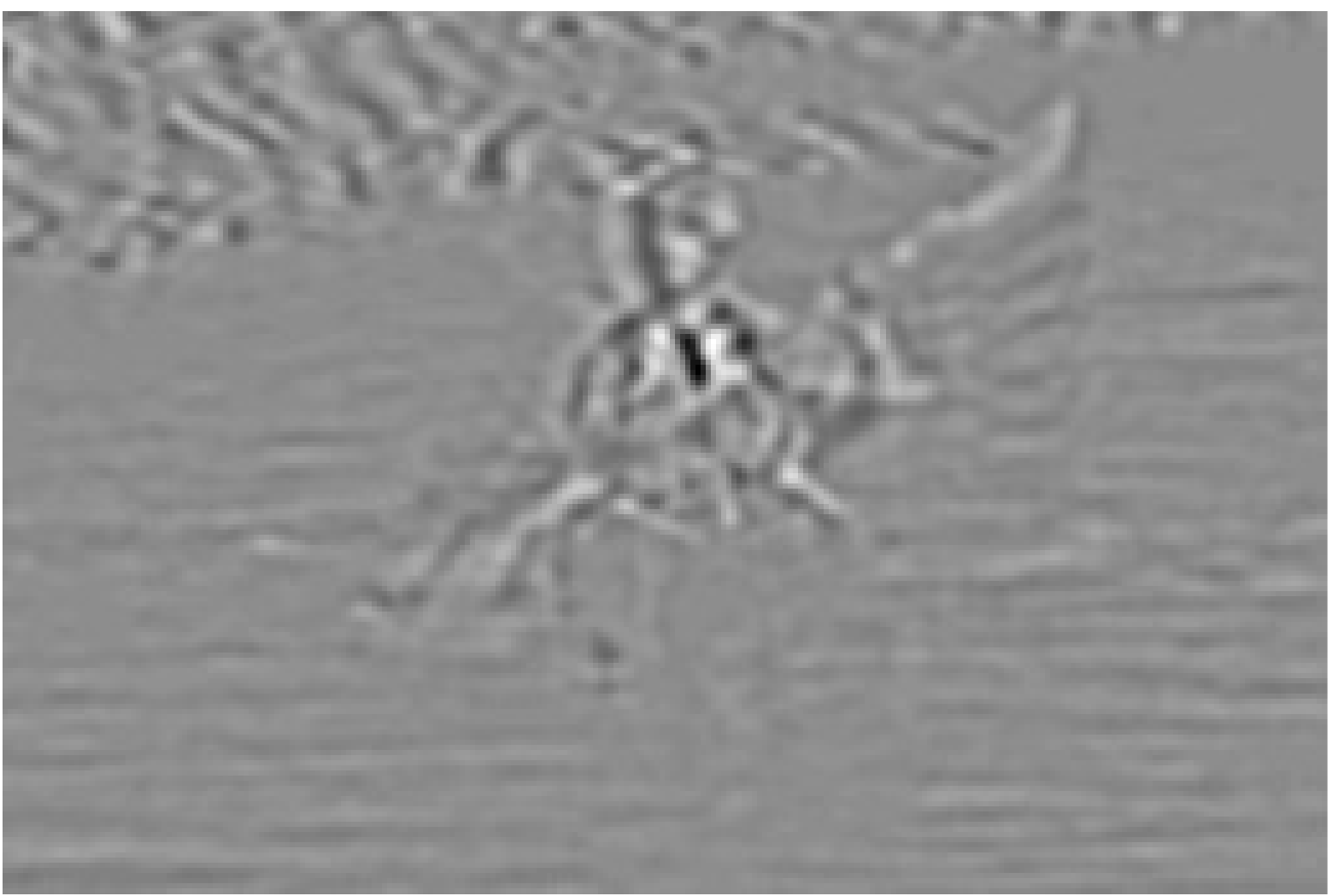} &
      \includegraphics[width=0.30\textwidth]{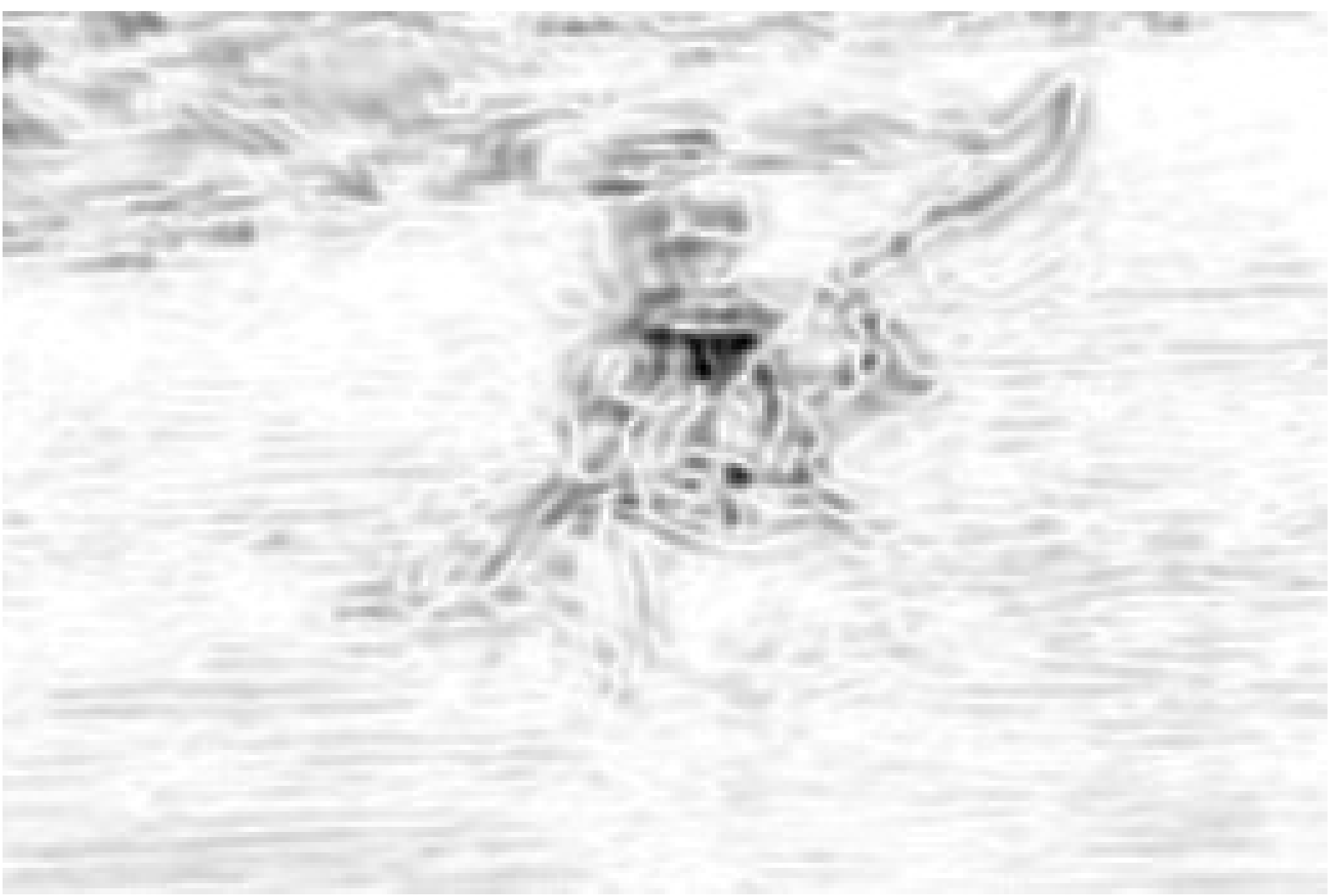} \\
    \end{tabular} 
  \end{center}

  \begin{center}
    \begin{tabular}{ccc}
      {\small $\det {\cal H}_{(x,y,t),norm} L$} 
      & {\small ${\cal G}_{(x,y,t),norm} L$} 
      & {\small $-\sqrt[3]{{\cal Q}_t(\det {\cal H}_{(x, y),norm} L)}$} \\
      \includegraphics[width=0.30\textwidth]{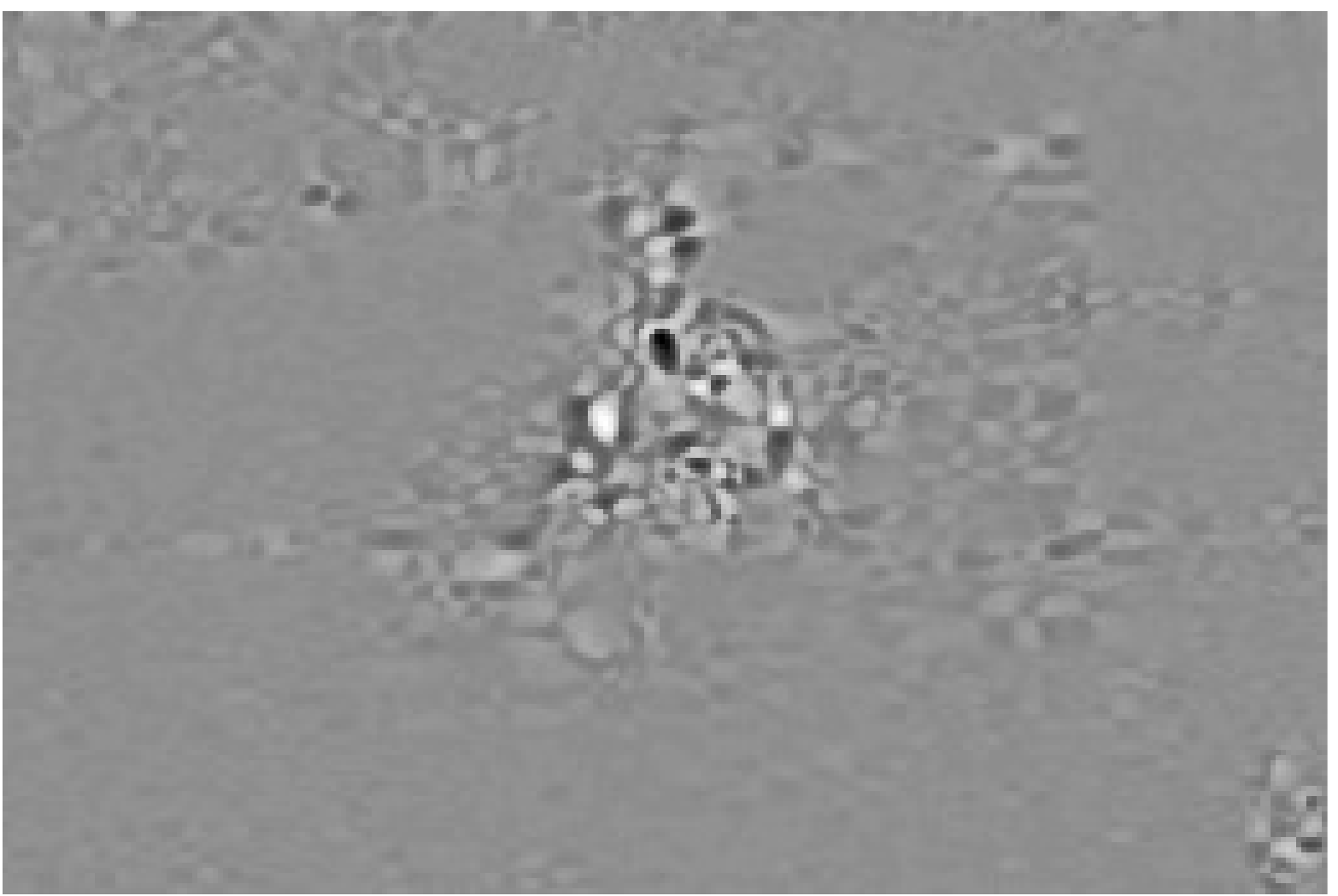} &
      \includegraphics[width=0.30\textwidth]{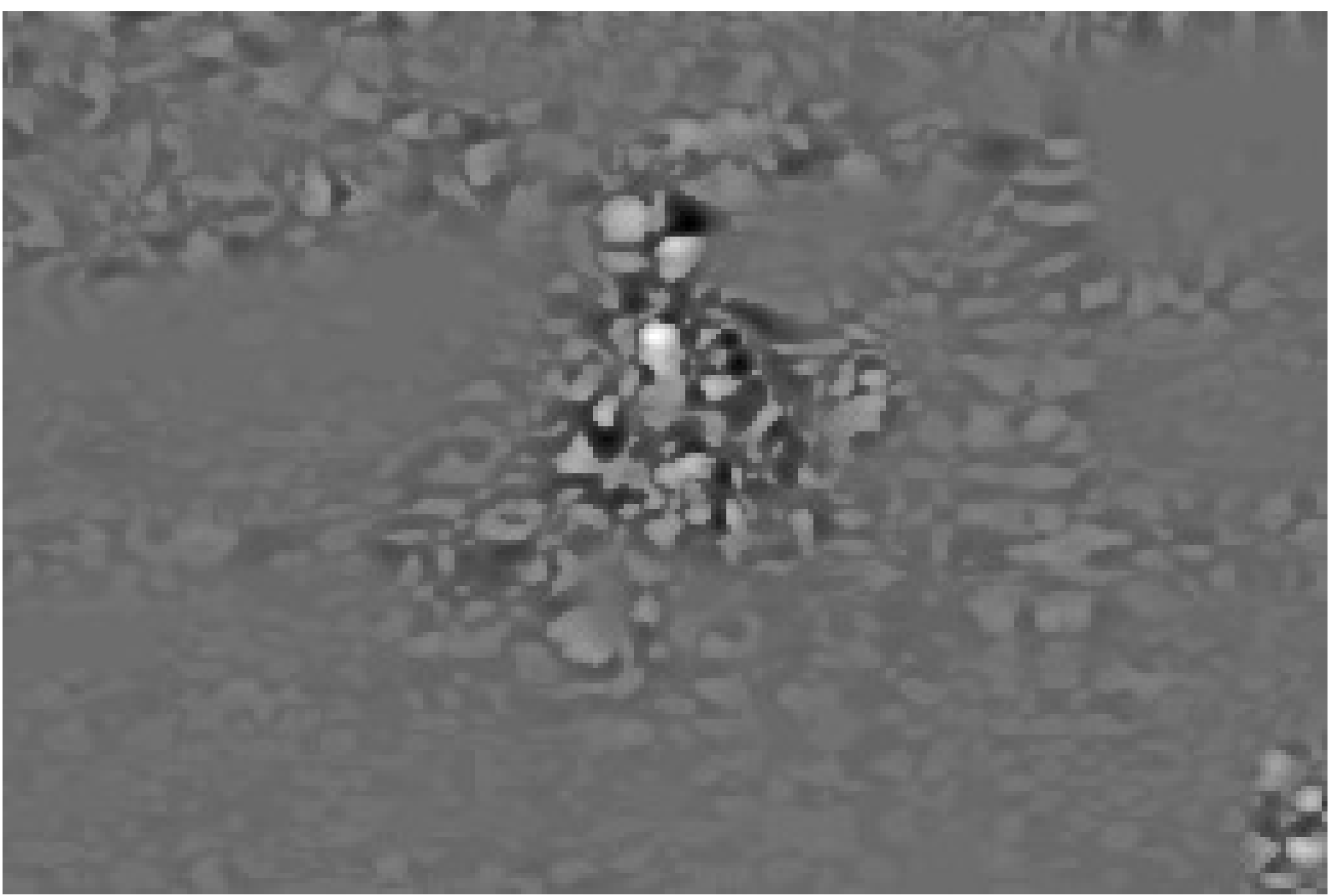} &
      \includegraphics[width=0.30\textwidth]{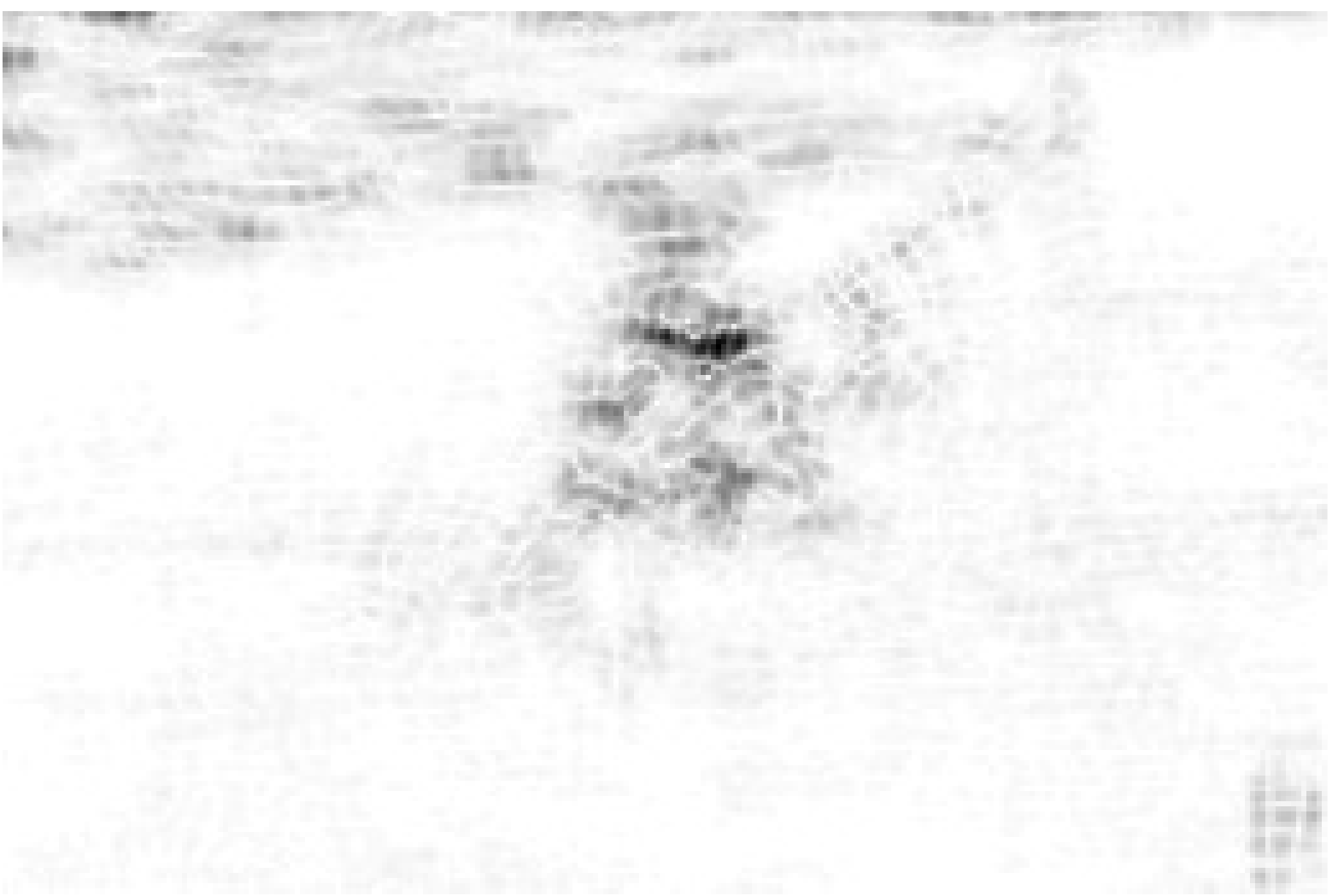} \\
    \end{tabular} 
  \end{center}

  \begin{center}
    \begin{tabular}{ccc}
      {\small $-\sqrt{{\cal Q}_{1,(x,y,t),norm} L}$}
      & {\small $-\sqrt{{\cal Q}_{2,(x,y,t),norm} L}$}
      & {\small $-\sqrt{{\cal Q}_{3,(x,y,t),norm} L}$}\\
      \includegraphics[width=0.30\textwidth]{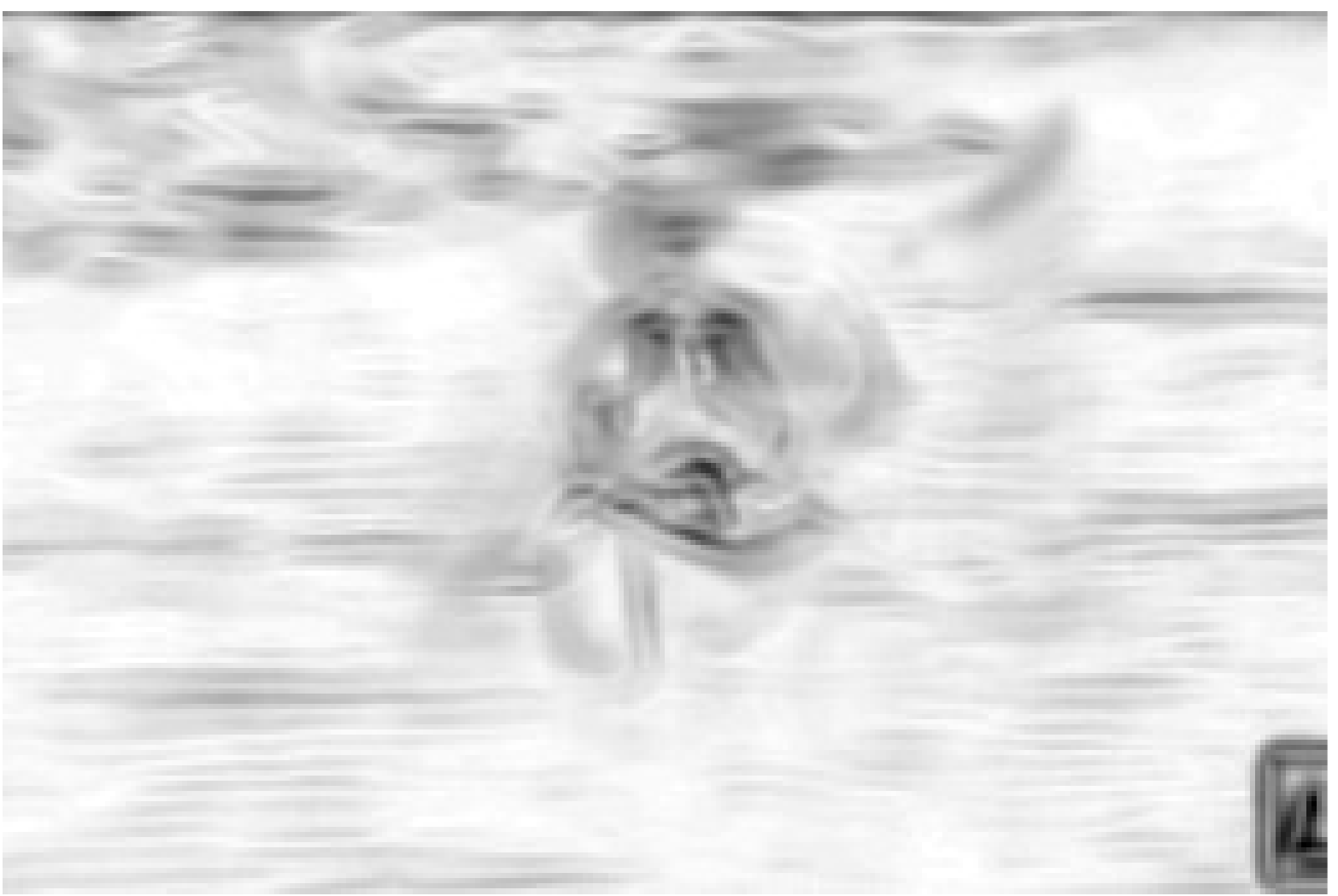} &
      \includegraphics[width=0.30\textwidth]{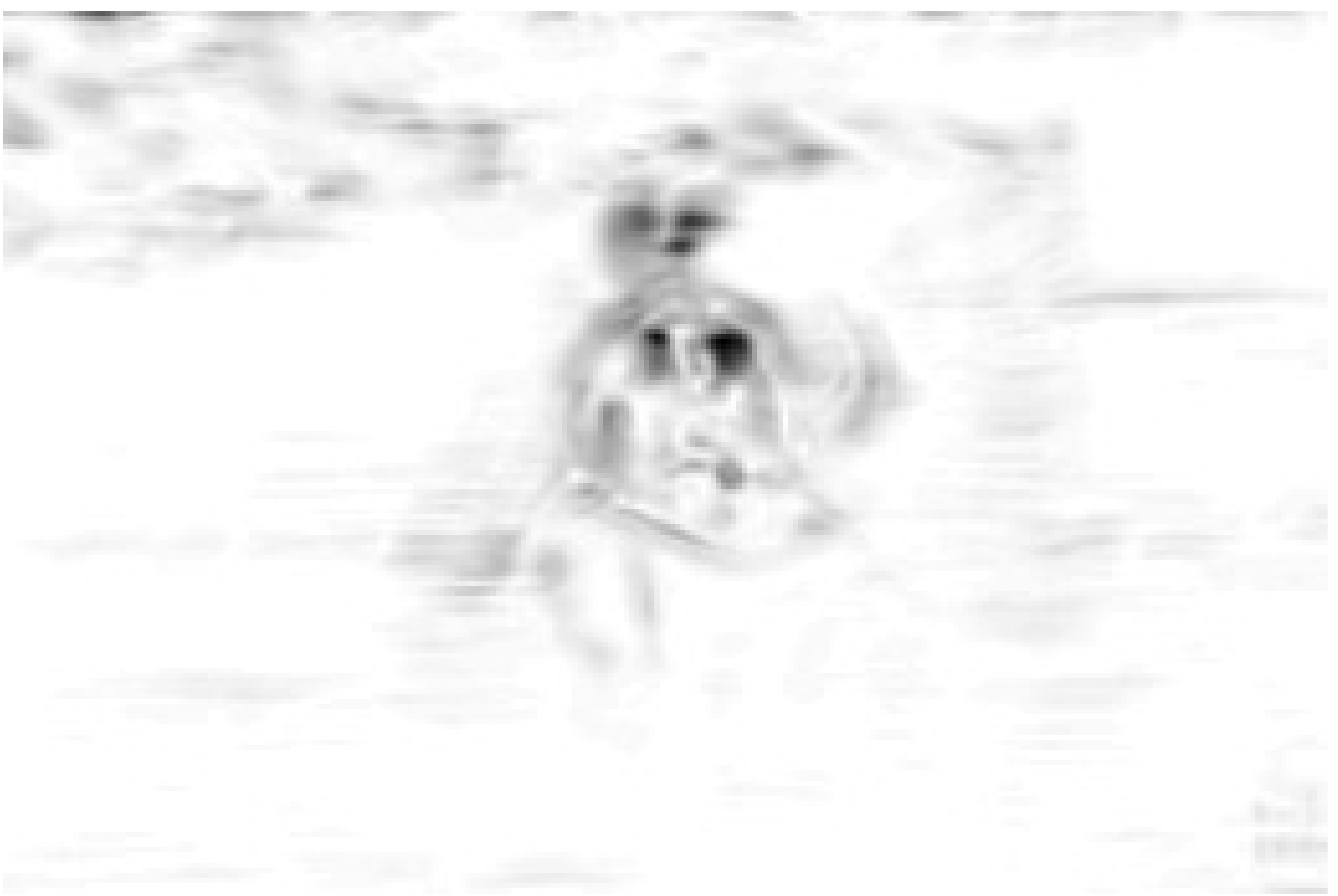} &
      \includegraphics[width=0.30\textwidth]{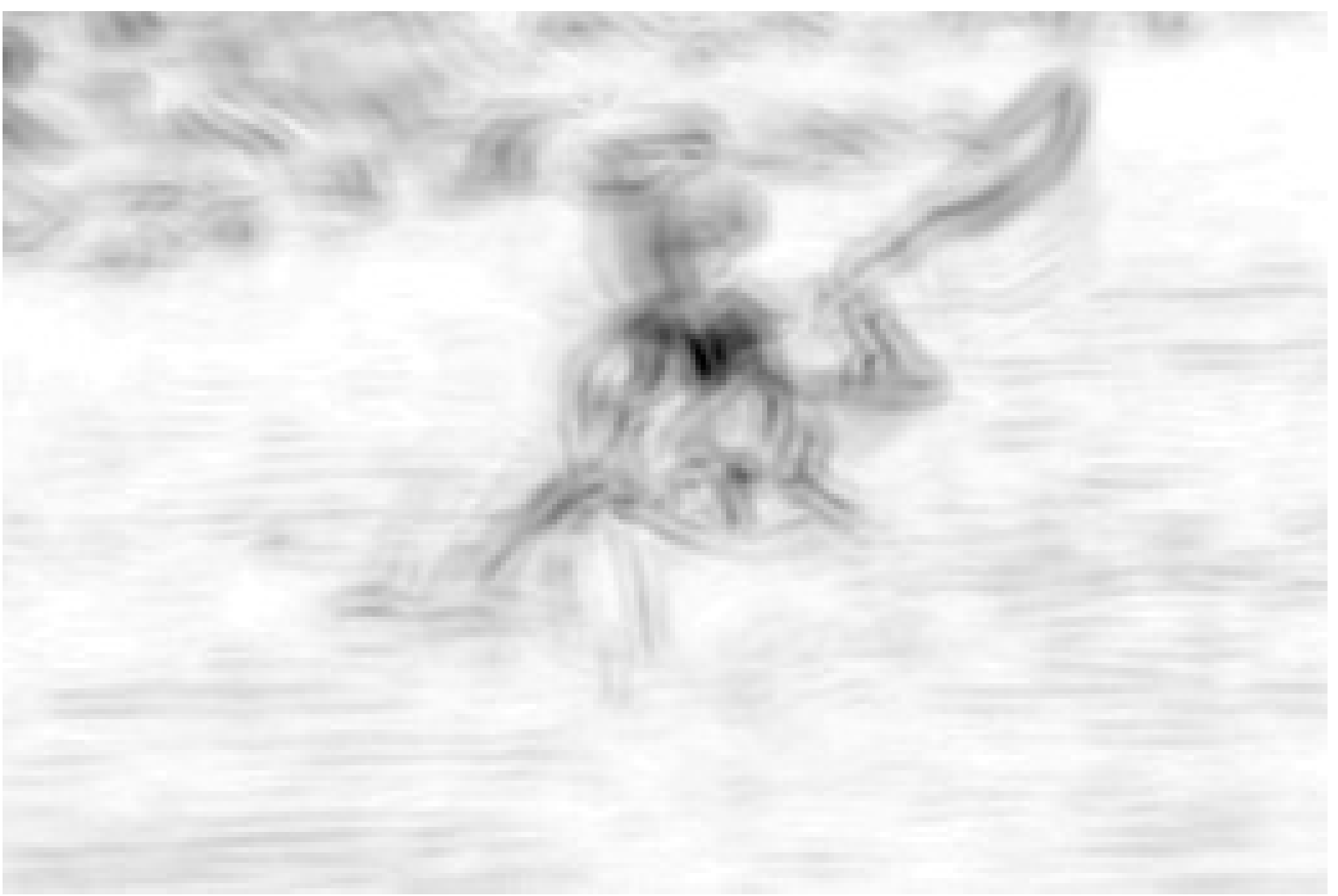} \\
    \end{tabular} 
  \end{center}

   \caption{Spatio-temporal features computed from a video
     sequence in the UCF-101 dataset (Kayaking\_g01\_c01.avi, cropped) at spatial
     scale $\sigma_x = 2~\mbox{pixels}$ and temporal scale $\sigma_t = 0.2~\mbox{seconds}$ 
    using the proposed separable spatio-temporal receptive field model
    with Gaussian filtering over the spatial
     domain and here a cascade of 7 recursive filters over the temporal
     domain with a logarithmic distribution of the intermediate scale
     levels for $c = \sqrt{2}$ and with $L_p$-normalization of both the
     spatial and temporal derivative operators. Each figure shows a snapshot around frames 90-97 for the
     spatial or spatio-temporal differential expression shown above
     the figure with in some cases additional monotone stretching of
     the magnitude values to simplify visual interpretation.
     (Image
     size: $258 \times 172$ pixels of original $320 \times 240$ pixels
     and 226 frames at 25 frames per second.)}
  \label{fig-spattemp-ders-kayak1}
\end{figure*}

\subsection{Scale normalized spatio-temporal derivative expressions}

For regular partial derivatives, normalization with respect to spatial
and temporal scales of a spatio-temporal scale-space derivative of
order $m = (m_1, m_2)$ over space and order $n$ over time
is performed according to
\begin{equation}
  \label{eq-sc-norm-part-der}
  L_{x_1^{m_1} x_2^{m_2} t^n,norm} = s^{(m_1 + m_2)/2} \, \alpha_n(\tau) \, L_{x_1^{m_1} x_2^{m_2} t^n}
\end{equation}
Scale normalization of the spatio-temporal differential expressions in
section~\ref{sec-spat-temp-der-expr} is then
performed by replacing each spatio-temporal partial derivate
by its corresponding scale-normalized expression
(see \cite{Lin15-arXiv-spattemp} for additional details).

For example, for the three quasi quadrature entities in 
equations~(\ref{eq-Q1-not-scalenorm-ders}),
(\ref{eq-Q2-not-scalenorm-ders}) and (\ref{eq-Q3-not-scalenorm-ders}),
their corresponding scale-normalized expressions are of the form:
\begin{align}
  \begin{split}
      & {\cal Q} _{1,(x, y, t),norm} L 
   \end{split}\nonumber\\
  \begin{split}
    & = s \, (L_x^2 + L_y^2) + \alpha_1^2(\tau) \, \varkappa^2 L_t^2 +
  \end{split}\nonumber\\
  \begin{split}
      & \phantom{=} \,
            + C 
                \left( 
                   s^2 (L_{xx}^2 + 2 L_{xy}^2 + L_{yy}^2) 
                \right.
   \end{split}\nonumber\\
  \begin{split}
      \label{eq-Q1-scalenorm-ders}
        & \phantom{= \,+ C \left( \right.}
               \left.
                    + s \, \alpha_1^2(\tau) \, \varkappa^2 (L_{xt}^2 + L_{yt}^2) 
                    + \alpha_2^2(\tau) \, \varkappa^4 L_{tt}^2
               \right),
  \end{split}\\
  \begin{split}
      & {\cal Q} _{2,(x, y, t),norm} L 
  \end{split}\nonumber\\
  \begin{split}
       & = {\cal Q}_{t,norm} L \times {\cal Q}_{(x, y),norm} L 
  \end{split}\nonumber\\
  \begin{split}
      & = \left( \alpha_1^2(\tau) \, L_t^2 + C \, \alpha_2^2(\tau) \, L_{tt}^2\right) \times
  \end{split}\nonumber\\
  \begin{split}
      \label{eq-Q2-scalenorm-ders}
       & \phantom{= \left( \right.}
            \left( s \, (L_x^2 + L_y^2) + C \, s^2 \left( L_{xx}^2 + 2 L_{xy}^2 + L_{yy}^2 \right) \right),
  \end{split}\\
  \begin{split}
     & {\cal Q} _{3,(x, y, t),norm} L 
  \end{split}\nonumber\\
  \begin{split}
     & = {\cal Q}_{(x, y),norm} L_t + C \, {\cal Q}_{(x, y),norm} L_{tt} 
 \end{split}\nonumber\\
  \begin{split}
      & = \alpha_1^2(\tau) 
             \left( 
                 s \, (L_{xt}^2 + L_{yt}^2) 
             + C \, s^2 \left( L_{xxt}^2 + 2 L_{xyt}^2 + L_{yyt}^2 \right)
              \right)
 \end{split}\nonumber\\
  \begin{split}
      & \phantom{=}  \,
           + C \, \alpha_2^2(\tau) 
             \left( 
                 s \, (L_{xtt}^2 + L_{ytt}^2)
             \right.
 \end{split}\nonumber\\
  \begin{split}
      \label{eq-Q3-scalenorm-ders}
& \phantom{= 
           + C \, \alpha_2^2(\tau) \left( \right.}
            \left.
                 + C s^2 (L_{xxtt}^2 + 2 L_{xytt}^2 + L_{yytt}^2) 
            \right).
    \end{split}
\end{align}

\subsection{Experimental results}

Figure~\ref{fig-spattemp-ders-kayak1} shows the result of computing
the above differential expressions for a video sequence of a paddler in a
kayak.

Comparing the spatio-temporal scale-space representation $L$ in the top middle
figure to the original video $f$ in the top left, we can first 
note that a substantial amount of fine scale spatio-temporal textures,
{\em e.g.\/}\ waves of the water surface, is suppressed by the spatio-temporal
smoothing operation.
The illustrations of the spatio-temporal scale-space representation
$L$ in
the top middle figure and its first- and second-order temporal
derivatives $L_{t,norm}$ and $L_{tt,norm}$ in the left and middle figures in the second row do also show the
spatio-temporal traces that are left by a moving object; see in
particular the image structures below the raised paddle that 
respond to spatial points in the image domain where the paddle has
been in the past.

The slight jagginess in the bright response that can be seen below the paddle
in the response to the second-order temporal derivative $L_{tt,norm}$
is a temporal sampling artefact caused by sparse temporal sampling
in the original video. With 25 frames per second there are 40~ms
between adjacent frames, during which there may happen a lot in the
spatial image domain for rapidly moving objects.
This situation can be compared to mammalian vision where many
receptive fields operate continuously over time scales in the range 20-100 ms.
With 40~ms between adjacent frames it is not possible to simulate such 
continuous receptive fields smoothly over time, since such a frame rate corresponds
to either zero, one or at best two images within the effective
time span of the receptive field. 
To simulate rapid continuous time receptive fields more accurately in a
digital implementation, one should therefore preferably aim at
acquiring the input video with a higher temporal frame rate. 
Such higher frame rates are indeed now becoming available,
even in consumer cameras.
Despite this limitation in the input data, we can observe
that the proposed model is able to compute geometrically meaningful
spatio-temporal image features from the raw video.

The illustrations of $\partial_t (\nabla_{(x,y),norm}^2 L)$ and $\partial_{tt} (\nabla_{(x,y),norm}^2 L)$
in the left and middle of 
the third row show the responses of our
idealized model of non-lagged and lagged LGN cells complemented by a
quasi-quadrature energy measure of these responses in the right
column.
These entities correspond to applying a spatial Laplacian operator to
the first- and second-order temporal derivatives in the second row
and it can be seen how this operation enhances spatial variations.
These spatio-temporal entities can also be compared to the purely spatial interest
operators, the Laplacian $\nabla_{(x, y),norm}^2 L$ and the
determinant of the Hessian $\det {\cal H}_{(x, y),norm} L$ in the
first and second rows of the third column.
Note how the genuinely spatio-temporal recursive
fields enhance spatio-temporal structures compared to purely spatial
operators and how static structures, such as the label in the lower
right corner, disappear altogether by genuine spatio-temporal
operators.
The fourth row shows how three other genuine spatio-temporal operators,
the spatio-temporal Hessian $\partial_t (\nabla_{(x,y),norm}^2 L)$,
the rescaled Gaussian curvature ${\cal G}_{(x,y,t),norm} L$ and the
quasi quadrature measure ${\cal Q}_t(\det {\cal H}_{(x, y),norm} L)$,
also respond to points where there are
simultaneously both strong spatial and strong temporal variations.

The bottom row shows three idealized models defined to mimic qualitatively
known properties of complex cells and expressed in terms of
quasi quadrature measures of spatio-temporal scale-space derivatives.
For the first quasi quadrature entity ${\cal Q}_{1,(x,y,t),norm} L$ to respond,
in which time is treated in a largely qualitatively similar manner as space,
it is sufficient if there are strong variations over either space or
time. It can be seen that this measure is therefore not highly selective.
For the second and the third entities ${\cal Q}_{2,(x,y,t),norm} L$
and ${\cal Q}_{3,(x,y,t),norm} L$,
it is necessary that there are simultaneous variations over both space
and time, and it can be seen how these entities are as a consequence more
selective.
For the third entity ${\cal Q}_{3,(x,y,t),norm} L$,
simultaneous selectivity over both space and time is additionally
enforced on each primitive linear receptive field that is then combined
into the non-linear quasi quadrature measure. We can see how this
quasi quadrature entity also responds stronger to the moving paddle
than the two other quasi quadrature measures.

\subsection{Geometric covariance and invariance properties}

\paragraph{Rotations in image space.}

The spatial differential expressions 
$|\nabla_{(x, y)} L|$, $\nabla_{(x, y)}^2 L$, $\det {\cal H} _{(x, y)}$, $\tilde{\kappa}(L)$
and ${\cal Q}_{(x, y)} L$
are all invariant under rotations in the image domain
and so are the spatio-temporal derivative expressions
$\partial_t (\nabla_{(x,y)}^2 L)$, $\partial_{tt} (\nabla_{(x,y)}^2L)$,
${\cal Q}_t(\nabla_{(x,y)}^2 L)$, $\partial_t (\det {\cal H}_{(x,y)} L)$,
$\partial_{tt} (\det {\cal H}_{(x,y)} L)$, 
${\cal Q}_t (\det {\cal H}_{(x,y)} L)$, $\det {\cal H}_{(x, y, t)} L$,
${\cal G}_{(x, y, t)} L$, $\nabla_{(x, y, t)}^2 L$,
${\cal Q} _{1,(x, y, t)} L$, ${\cal Q}_{2,(x, y, t)} L$ and
${\cal Q} _{3,(x, y, t)} L$ as well as their corresponding scale-normalized
expressions.

\paragraph{Uniform rescaling of the spatial domain.}

Under a uniform scaling transformation of image space, the spatial
differential invariants
$|\nabla_{(x, y)} L|$, $\nabla_{(x, y)}^2 L$, $\det {\cal H} _{(x, y)}$ and $\tilde{\kappa}(L)$
are covariant under spatial scaling transformations in the sense that their
magnitude values are multiplied by a power of the scaling factor,
and so are their corresponding scale-normalized expressions.
Also the spatio-temporal differential invariants
$\partial_t (\nabla_{(x,y)}^2 L)$, $\partial_{tt} (\nabla_{(x,y)}^2L)$,
$\partial_t (\det {\cal H}_{(x,y)} L)$, $\partial_{tt} (\det {\cal H}_{(x,y)} L)$, 
$\det {\cal H}_{(x, y, t)} L$ and ${\cal G}_{(x, y, t)} L$
and their corresponding scale-normalized expressions
are covariant under spatial scaling transformations in the sense that
their magnitude values are multiplied by a power of the scaling factor
under such spatial scaling transformations.

The quasi quadrature entity ${\cal Q}_{(x, y),norm} L$ is however not
covariant under spatial scaling transformations and not the
spatio-temporal differential invariants ${\cal Q}_{t,norm}(\nabla_{(x,y)}^2 L)$,\newline
${\cal Q}_{t,norm} (\det {\cal H}_{(x,y)} L)$, ${\cal Q} _{1,(x, y, t),norm} L$,
${\cal Q} _{2,(x, y, t),norm} L$ and ${\cal Q} _{3,(x, y, t),norm} L$  either.
Due to the form of ${\cal Q}_{(x, y),norm} L$, ${\cal Q}_{t,norm}(\nabla_{(x,y)}^2 L)$,
${\cal Q}_{t,norm} (\det {\cal H}_{(x,y)} L)$, ${\cal Q} _{2,(x, y, t),norm} L$ and ${\cal Q} _{3,(x, y, t),norm} L$
as being composed of sums of scale-normalized
derivative expressions for $\gamma = 1$, these derivative expressions
can, however, anyway be made scale invariant when combined with a
spatial scale selection mechanism.

\paragraph{Uniform rescaling of the temporal domain independent of the
  spatial domain.}

Under an independent rescaling of the temporal dimension while keeping
the spatial dimension fixed, the partial derivatives 
$L_{x_1^{m_1} x_2^{m_1} t^n}(x_1, x_2, t;\; s, \tau)$ are covariant
under such temporal rescaling transformations, and so are the directional derivatives
$L_{\varphi^{m_1} \orth \varphi^{m_2} t^n}$ for image velocity $v = 0$.
For non-zero image velocities, the 
image velocity parameters of the receptive field would on the other hand need to be adapted to the
local motion direction of the objects/spatio-temporal events of
interest to enable matching between corresponding spatio-temporal
directional derivative operators.

Under an independent rescaling of the temporal dimension while keeping
the spatial dimension fixed, also the spatio-temporal differential invariants
$\partial_t (\nabla_{(x,y)}^2 L)$, $\partial_{tt} (\nabla_{(x,y)}^2L)$,
$\partial_t (\det {\cal H}_{(x,y)} L)$, $\partial_{tt} (\det {\cal H}_{(x,y)} L)$, 
$\det {\cal H}_{(x, y, t)} L$ and ${\cal G}_{(x, y, t)} L$ are
covariant under independent rescaling of the temporal {\em vs.\/}\
spatial dimensions. The same applies to their corresponding
scale-normalized expressions.

The spatio-temporal differential invariants ${\cal Q}_{t,norm}(\nabla_{(x,y)}^2 L)$,
${\cal Q}_{t,norm} (\det {\cal H}_{(x,y)} L)$, ${\cal Q} _{1,(x, y, t),norm} L$,
${\cal Q} _{2,(x, y, t),norm} L$ and ${\cal Q} _{3,(x, y, t),norm} L$ are
however not covariant under independent rescaling of the temporal {\em
  vs.\/} spatial dimensions and would therefore need a temporal scale
selection mechanism to enable temporal scale invariance.

\begin{figure*}[hbt]
  \begin{center}
    \begin{tabular}{ccc}
      {\small $-\sqrt{{\cal Q}_{1,(x,y,t),norm} L}$}
      & {\small $-\sqrt{{\cal Q}_{2,(x,y,t),norm} L}$}
      & {\small $-\sqrt{{\cal Q}_{3,(x,y,t),norm} L}$}\\
      \includegraphics[width=0.30\textwidth]{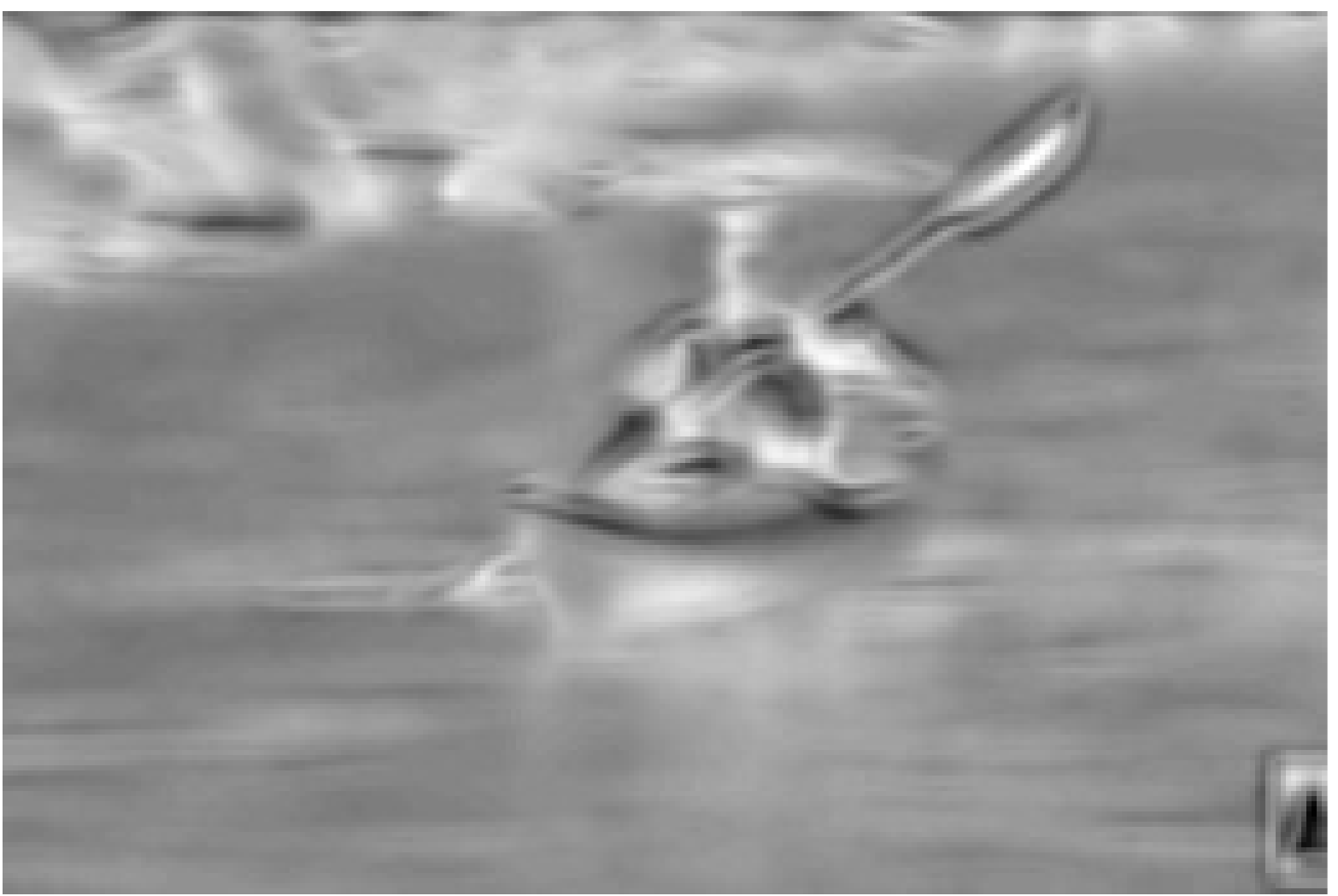} &
      \includegraphics[width=0.30\textwidth]{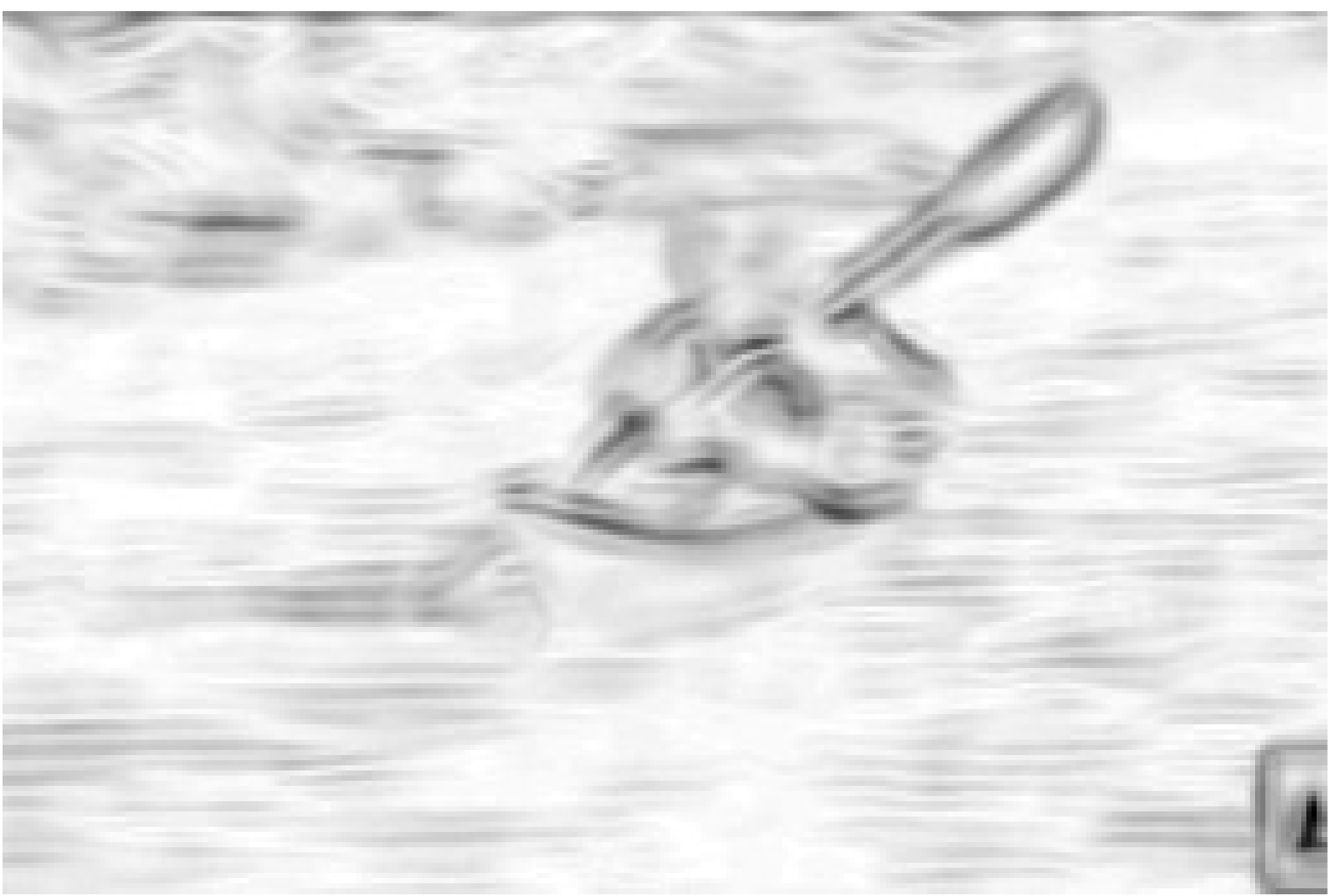} &
      \includegraphics[width=0.30\textwidth]{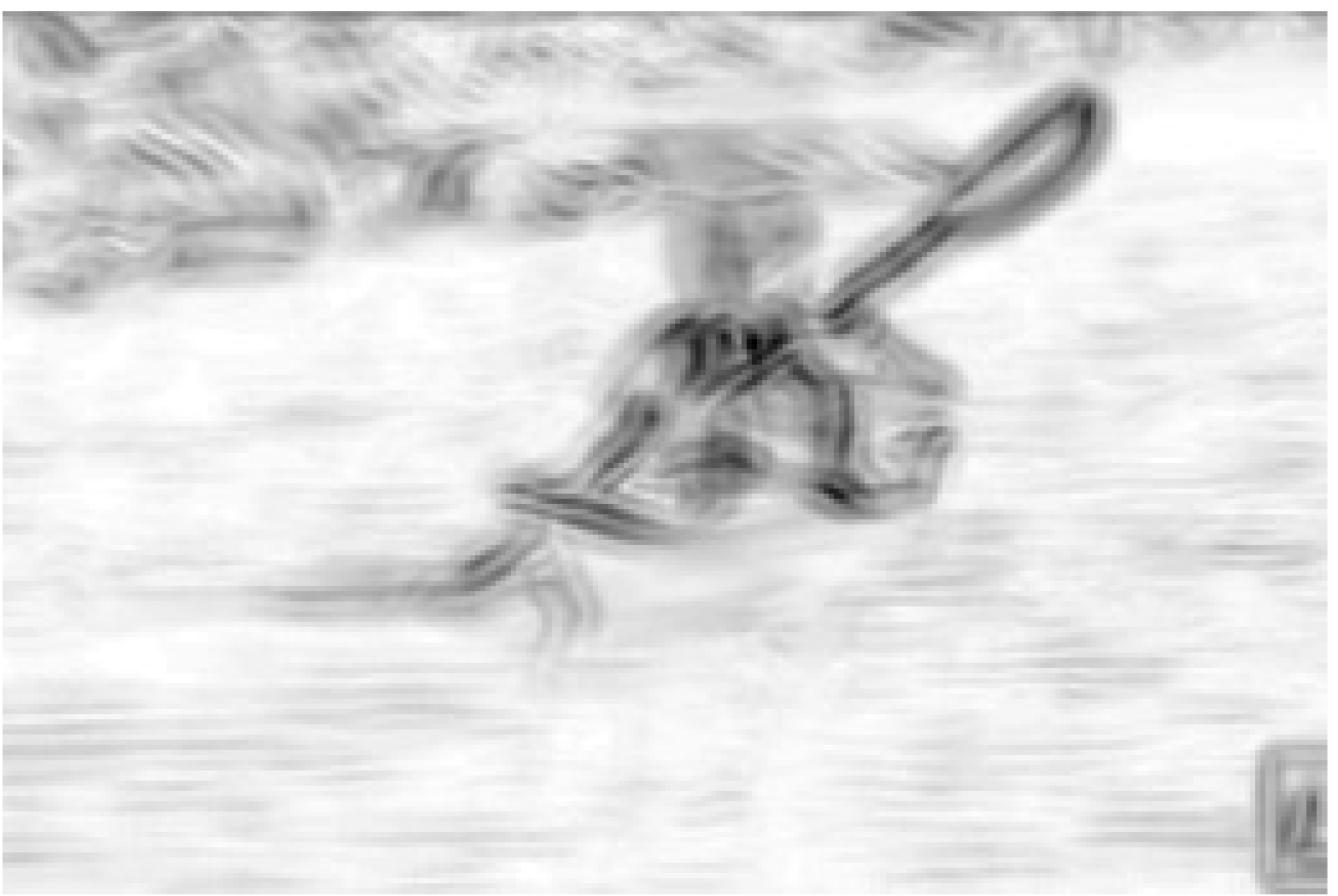} \\
    \end{tabular} 
  \end{center}

   \caption{Illustration of the influence of temporal illumination or exposure
     compensation mechanisms on spatio-temporal receptive field responses, computed from the video
     sequence Kayaking\_g01\_c01.avi (cropped) in the UCF-101 dataset.
     Each figure shows a snapshot at frame 8 for the
     quasi quadrature entity shown above the figure with additional monotone stretching of
     the magnitude values to simplify visual interpretation. Note
     how the time varying illumination or exposure compensation leads
     to a strong overall response in the first quasi quadrature entity
     ${\cal Q}_{1,(x,y,t),norm} L$ caused by strong responses
     in the purely temporal derivatives $L_t$ and $L_{tt}$, whereas the
     responses of second and the third quasi quadrature entities 
     ${\cal Q}_{2,(x,y,t),norm} L$ and ${\cal Q}_{3,(x,y,t),norm} L$ 
     are much less influenced. Indeed, for a
     logarithmic brightness scale the third quasi quadrature entity ${\cal Q}_{3,(x,y,t),norm} L$ is
     invariant under such multiplicative illumination or exposure
     compensation variations.}
  \label{fig-spattemp-ders-kayak1-illumin}
\end{figure*}

\subsection{Invariance to illumination variations and exposure control mechanisms}

Because of all these expressions being composed of spatial, temporal
and spatio-temporal derivatives of non-zero order, it follows that all
these differential expressions are invariant under additive
illumination transformations of the form $L \mapsto L + C$.

This means that if we would take the image values $f$ as representing the
logarithm of the incoming energy $f \sim \log I$ or 
$f \sim \log I^{\gamma} = \gamma \log I$, then all these differential
expressions will be invariant under local multiplicative illumination
transformations of the form $I \mapsto C \, I$ implying
$L \sim \log I + \log C$ or 
$L \sim \log I^{\gamma} = \gamma (\log I + \log C)$.
Thus, these differential expressions will be invariant to local multiplicative variabilities in the
external illumination (with locality defined as over the support region of
the spatio-temporal receptive field) or
multiplicative exposure control parameters such as the aperture of the
lens and the integration time or the sensitivity of the sensor.

More formally, let us assume
 a (i)~perspective camera model extended with 
(ii)~a thin circular lens for gathering incoming light from different
directions and 
(iii)~a Lambertian illumination model extended with 
(iv)~a spatially varying albedo factor for modelling the light that is
reflected from surface patterns in the world.
Then, by theoretical results in (Lindeberg \cite[section~2.3]{Lin13-BICY}) 
a spatio-temporal receptive field response $L_{x^{m_1} y^{m_2}t^n}(\cdot, \cdot;\; s, \tau)$
where ${\cal T}_{s,\tau}$ represents the spatio-temporal
smoothing operator can be expressed as
\begin{align}
  \begin{split}
    & L_{x^{m_1} y^{m_2}t^n}
 \end{split}\nonumber\\
  \begin{split}
     & = \partial_{x^{m_1} y^{m_2} t^n} \, {\cal T}_{s,\tau} \,
        \left(
            \log \rho(x, y, t) + \log i(x, y, t)  \vphantom{\log C_{cam}(\tilde{f}(t))}
        \right.
  \end{split}\nonumber\\
  \begin{split}
     \label{eq-illum-model-spat-rec-field}
    & \phantom{= \partial_{x^{m_1} y^{m_2} t^n} \, {\cal T}_{s,\tau} \,  \left( \right.}
        \left. 
          + \log C_{cam}(\tilde{f}(t)) 
          + V(x, y)
      \right)
  \end{split}
\end{align}
where 
(i)~$\rho(x, y, t)$ is a spatially dependent albedo factor,
(ii)~$i(x, y, t)$ denotes a spatially dependent illumination field,
(iii)~$C_{cam}(\tilde{f}(t)) = \frac{\pi}{4} \frac{d}{f}$ represents
  possibly time-varying internal camera parameters and
(iv)~$V(x, y) = - 2 \log (1 + x^2  + y^2)$ represents a
  geometric natural vignetting effect.

From the structure of equation~(\ref{eq-illum-model-spat-rec-field})
we can note that for any non-zero order of spatial differentiation 
$m_1 + m_2 > 0$, the influence of the internal camera parameters in 
$C_{cam}(\tilde{f}(t))$ will disappear because of the spatial
differentiation with respect to $x_1$ or $x_2$, and so will the effects of any
other multiplicative exposure control mechanism.
Furthermore, for any multiplicative illumination variation
$i'(x, y) = C \, i(x, y)$, where $C$ is a scalar constant,
the logarithmic luminosity will be transformed as
$\log i'(x, y) = \log C + \log i(x, y)$, which implies that
the dependency on $C$ will disappear
after spatial differentiation.
For purely temporal derivative operators, that do not involve any
order of spatial differentiation, such as the first- and second-order
derivative operators, $L_t$ and $L_{tt}$, strong responses may on the
other hand be obtained due to illumination compensation mechanisms
that vary over time as the results of rapid variations in the illumination.
If one wants to design spatio-temporal feature detectors that are
robust to illumination variations and to variations in exposure
compensation mechanisms caused by these, it is therefore essential to
include non-zero orders of spatial differentiation.
The use of Laplacian-like filtering in the first stages of visual
processing in the retina and the LGN can therefore be interpreted as
a highly suitable design to achieve robustness of illumination variations and
adaptive variations in the diameter of the pupil caused by these,
while still being expressed in terms of rotationally symmetric
linear receptive fields over the spatial domain.

If we extend this model to the simplest form of position- and time-dependent
illumination and/or exposure variations as modelled on the form
\begin{equation}
  L \mapsto L + A x + B y + C t
\end{equation}
then we can see that the spatio-temporal differential invariants
$\partial_t (\nabla_{(x,y)}^2 L)$, $\partial_{tt} (\nabla_{(x,y)}^2L)$,
${\cal Q}_t(\nabla_{(x,y)}^2 L)$,
$\partial_t (\det {\cal H}_{(x,y)} L)$, $\partial_{tt} (\det {\cal H}_{(x,y)} L)$, 
${\cal Q}_t (\det {\cal H}_{(x,y)} L)$,
$\det {\cal H}_{(x, y, t)} L$, ${\cal G}_{(x, y, t)} L$
$\nabla_{(x, y, t)}^2 L$ and ${\cal Q}_{3,(x, y, t)} L$
are all invariant under such position- and time-dependent
illumination and/or exposure variations.

The quasi quadrature entities ${\cal Q} _{1,(x, y, t)} L$ and ${\cal Q}_{2,(x, y, t)} L$ 
are however not invariant to such position- and time-dependent
illumination variations.
This property can in particular be noted for the quasi quadrature
entity ${\cal Q} _{1,(x, y, t)} L$, for which what seems as initial
time-varying exposure compensation mechanisms in the camera lead to
large responses in the initial part of the video sequence
(see figure~\ref{fig-spattemp-ders-kayak1-illumin}(left)).
Out of the three quasi quadrature entities ${\cal Q} _{1,(x, y, t)} L$, ${\cal Q}_{2,(x, y, t)} L$ 
and ${\cal Q} _{3,(x, y, t)} L$, the third quasi quadrature entity
does therefore possess the best robustness properties to illumination
variations (see figure~\ref{fig-spattemp-ders-kayak1-illumin}(right)).

\section{Summary and discussion}
\label{sec-sum-disc}

We have presented an improved computational model for spatio-temporal
receptive fields based on time-causal and time-recursive
spatio-temporal scale-space representation defined from a set of
first-order integrators or truncated exponential filters
coupled in cascade over the temporal domain in combination with a Gaussian
scale-space concept over the spatial domain.
This model can be efficiently implemented in terms of recursive
filters over time and we have shown how the continuous model can be
transferred to a discrete implementation while retaining discrete
scale-space properties. Specifically, we have analysed how remaining
design parameters within the theory, in terms of the number of
first-order integrators coupled in cascade and a distribution
parameter of a logarithmic distribution, affect the temporal response
dynamics in terms of temporal delays.

Compared to other spatial and temporal scale-space representations
based on continuous scale parameters, a conceptual difference with the
temporal scale-space representation underlying the proposed
spatio-temporal receptive fields, is that the temporal scale levels
have to be discrete.
Thereby, we sacrifice a continuous scale parameter and full scale
invariance as resulting from the Gaussian
scale-space concepts based on causality or non-enhancement of local
extrema (Koenderink \cite{Koe84}; Lindeberg \cite{Lin10-JMIV}) or 
used as a scale-space axiom in certain axiomatic scale-space formulations
(Iijima \cite{Iij62}; Florack et al.\ \cite{FloRom92-IVC}; 
Pauwels et al.\ \cite{PauFidMooGoo95-PAMI}; 
Weickert et al.\ \cite{WeiIshImi96-ScSp,Wei98-book,WeiIshImi99-JMIV}; 
Duits et al.\ \cite{DuiFelFloPla03-SCSP,DuiFloGraRom04-JMIV}; 
Fagerstr{\"o}m \cite{Fag05-IJCV,Fag07-ScSp});
see also Witkin \cite{Wit83}, 
Babaud {\em et al.\/}\ \cite{BWBD86-PAMI},
Yuille and Poggio \cite{YuiPog86-PAMI},
Koenderink and van Doorn \cite{KoeDoo90-BC,KoeDoo92-PAMI},
Lindeberg \cite{Lin90-PAMI,Lin93-Dis,Lin94-SI,Lin96-ScSp,Lin97-ICSSTCV,Lin13-ImPhys},
Florack et al.\ \cite{FloRomKoeVie92-ECCV,FloRom92-IVC,Flo97-book},
Alvarez {\em et al.\/}\ \cite{Alv92-TR},
Guichard \cite{Gui98-TIP},
ter Haar Romeny {\em et al.\/}\ \cite{RomFloNie01-SCSP,Haa94-GDDbook},
Felsberg and Sommer \cite{FelSom04-JMIV} and 
Tschirsich and Kuijper \cite{TscKui15-JMIV} for other scale-space
approaches closely related to this work, as well as 
Fleet and Langley \cite{FleLan95-PAMI}, Freeman and Adelson \cite{FreAde91-PAMI},
Simoncelli {\em et al.\/}\ \cite{SimFreAdeHee92-IT}
and Perona \cite{per92-ivc} for more filter-oriented approaches,
Miao and Rao \cite{MiaRao07-NeurComp},
Duits and Burgeth \cite{DuiBur07-ScSp},
Cocci {\em et al.\/}\  \cite{CocBarSar12-JOSA}, 
Barbieri {\em et al.\/}\ \cite{BarCitCocSar14-JMIV} and
Sharma and Duits \cite{ShaDui14-HarmAnal}
for Lie group approaches for receptive fields and
Lindeberg and Friberg \cite{LinFri15-PONE,LinFri15-SSVM} for the
application of closely related principles for deriving
idealized computational models of auditory receptive fields.

When using a logarithmic distribution of the intermediate scale
levels, we have however shown that by a limit construction when the
number of intermediate temporal scale levels tends to infinity, we can
achieve true self-similarity and scale invariance over a discrete set of scaling factors.
For a vision system intended to operate in real time using no other explicit
storage of visual data from the past than a compact time-recursive
buffer of spatio-temporal scale-space at different
temporal scales, the loss of a continuous temporal scale parameter may 
however be less of a practical constraint,
since one would anyway have to discretize the temporal scale levels in
advance to be able to register the image data to be able to perform any computations at all.

In the special case when all the time constants of the first-order
integrators are equal, the resulting temporal smoothing kernels 
in the continuous model (\ref{eq-composed-all-mu-equal})
correspond to Laguerre functions (Laguerre polynomials multiplied by a truncated
exponential kernel),
which have been previously used for modelling the temporal response
properties of neurons in the visual system (den Brinker and Roufs \cite{BriRou92-BC})
and for computing spatio-temporal image features in
computer vision (Rivero-Moreno and Bres \cite{RivBre04-ImAnalRec};
Berg et al.\ \cite{BerReyRid14-SensMEMSElOptSyst}). 
Regarding the corresponding discrete model with all time constants
equal, the corresponding discrete temporal smoothing kernels approach
Poisson kernels when the number of temporal smoothing steps increases
while keeping the variance of the composed kernel fixed
(Lindeberg and Fagerstr{\"o}m \cite{LF96-ECCV}).
Such Poisson kernels have also been used for modelling
biological vision (Fourtes and Hodgkin \cite{FouHod64-JPhys}).
Compared to the special case with all time constants equal,
a logarithmic distribution of the intermediate temporal scale levels (\ref{eq-distr-tau-values})
does on the other hand allow for larger flexibility in the trade-off between temporal smoothing and
temporal response characteristics, specifically enabling faster
temporal responses (shorter temporal delays) and higher computational efficiency when computing
multiple temporal or spatio-temporal receptive field responses involving coarser temporal scales.

From the detailed analysis in section~\ref{sec-time-caus-limit-kernel} and
appendix~\ref{app-freq-anal-time-caus-kern} we can conclude that
when the number of first-order integrators that are coupled in cascade
increases while keeping the variance of the composed kernel fixed, the
time-causal kernels obtained by composing truncated exponential
kernels with equal time constants in cascade tend to a limit kernel
with skewness and kurtosis measures zero, or equivalently third- and
fourth-order cumulants equal to zero, whereas the time-causal kernels
obtained by composing truncated exponential kernels having a
logarithmic distribution of the intermediate scale levels tends to a
limit kernel with non-zero skewness and non-zero kurtosis
This property reveals a fundamental difference between the two classes
of time-causal scale-space kernels based on either a logarithmic or a
uniform distribution of the intermediate temporal scale levels.

In a complementary analysis in appendix~\ref{app-comp-koe-model}, we
have also shown how our time-causal kernels can be related to the
temporal kernels in Koenderink's scale-time model \cite{Koe88-BC}.
By identifying the first- and second-order temporal moments of the two
classes of kernels, we have derived closed-form expressions to relate
the parameters between the two models, and showed that although the
two classes of kernels to a large extent share qualitatively similar
properties, the two classes of kernels differ significantly in terms
of their third- and fourth-order skewness and kurtosis measures.

The closed-form expressions for Koenderink's scale-time kernels are
analytically simpler than the explicit expressions for our
kernels, which will be sums of truncated exponential kernels for all
the time constants with the coefficients determined from a partial
fraction expansion. In this respect, the derived mapping between the
parameters of our and Koenderink's models can be used
{\em e.g.\/} for estimation the time of the temporal maximum of our
kernels, which would otherwise have to be determined numerically.
Our kernels do on the other hand have a clear computational advantage
in that they are truly time-recursive, meaning that the primitive
first-order integrators in the model contain sufficient information
for updating the model to new states over time, whereas the kernels in
Koenderink's scale-time model appear to require a complete memory of the past,
since they do not have any known time-recursive formulation.

Regarding the purely temporal scale-space concept used in our
spatio-temporal model, we have notably
replaced the assumption of a semi-group structure over temporal scales
by a weaker Markov property, which however anyway guarantees a
necessary cascade property over temporal scales, to ensure gradual
simplification of the temporal scale-space representation from any finer to any coarser temporal scale.
By this relaxation of the requirement of a semi-group over temporal scales, we have
specifically been able to define a temporal scale-space concept with
much better temporal dynamics than the time-causal semi-groups
derived by Fagerstr{\"o}m \cite{Fag05-IJCV} and Lindeberg
\cite{Lin10-JMIV}. 
Since this new time-causal temporal scale-space concept with a
logarithmic distribution of the intermediate temporal scale levels
would not be found
if one would start from the assumption about a semi-group over
temporal scales as a necessary requirement, we propose that that in
the area of scale-space axiomatics the assumption of a semi-group over
temporal scales should not be regarded as a necessary requirement for
a time-causal temporal scale-space representation.

Recently, and during the development of this article, Mahmoudi \cite{Mah15-soton-preprint} has
presented a very closely related while more neurophysiologically motivated
model for visual receptive fields, based on an electrical circuit
model with spatial smoothing determined by local spatial connections over
a spatial grid and temporal smoothing by first-order temporal
integration. The spatial component in that model is very closely
related to our earlier discrete scale-space models over spatial
and spatio-temporal grids (Lindeberg \cite{Lin90-PAMI,Lin97-ICSSTCV,Lin02-ECCV})
as can be modelled by Z-transforms of the discrete convolution kernels and an
algebra of spatial or spatio-temporal covariance matrices to model
the transformation properties of the receptive
fields under locally linearized geometric image transformations.
The temporal component in that model is in turn similar to our temporal
smoothing model by first-order integrators coupled in cascade as
initially proposed in (Lindeberg \cite{Lin90-PAMI}; Lindeberg and
Fagerstr{\"o}m \cite{LF96-ECCV}), suggested as one of three models for
temporal smoothing in spatio-temporal visual receptive fields in (Lindeberg \cite{Lin13-BICY,Lin13-ImPhys})
and then refined and further developed in (Lindeberg \cite{Lin15-SSVM,Lin15-arXiv-spattemp})
and this article. Our model can also be implemented by 
electric circuits, by combining the temporal electric model in
figure~\ref{fig-first-order-integrators-electric} with the spatial
discretization in section~\ref{app-disc-gauss-smooth} or more general
connectivities between adjacent layers to implement velocity-adapted
receptive fields as can then be described by their resulting
spatio-temporal covariance matrices.
Mahmoudi compares such electrically modelled receptive fields to results of
neurophysiological recordings in the LGN and the primary visual cortex
in a similar way as we compared our theoretically derived receptive
fields to biological receptive fields in (Lindeberg \cite{Lin97-ICSSTCV,Lin10-JMIV,Lin13-BICY,Lin15-SSVM}) and
in this article.

Mahmoudi shows that the resulting transfer function in the layered
electric circuit model approaches a Gaussian when the number of layers
tends to infinity. This result agrees with our earlier results that the
discrete scale-space kernels over a discrete spatial grid approach the
continuous Gaussian when the spatial scale increment tends to zero while the
spatial scale level is held constant \cite{Lin90-PAMI} and that the temporal
smoothing function corresponding to a set of first-order integrators
with equal time constants coupled in cascade tend to the Poisson kernel (which in turn
approaches the Gaussian kernel) when the temporal scale increment
tends to zero while the temporal scale level is held constant
\cite{LF96-ECCV}.

In his article, Mahmoudi \cite{Mah15-soton-preprint} makes a
distinction between our scale-space approach,
which is motivated by the mathematical structure of the environment in
combination with a set of assumptions about the internal structure of
a vision system to guarantee internal consistency between image
representations at different spatial and temporal scales, and his model 
motivated by assumptions about neurophysiology.
One way to reconcile these views is by following the
evolutionary arguments proposed in (Lindeberg
\cite{Lin13-BICY,Lin13-PONE}).
If there is a strong evolutionary pressure on a living organism that
uses vision as a key source of information about its environment
(as there should be for many higher mammals),
then in the competition between two species or two individuals from
the same species, there should be a strong evolutionary advantage for an organism
that as much as possible adapts the structure of its vision system to
be consistent with the structural and transformation properties of its environment.
Hence, there could be an evolutionary pressure for the vision system
of such an organism to develop similar tupes of receptive fields as
can be derived by an idealized mathematical theory,
and specifically develop neurophysiological wetware that permits the
computation of sufficiently good approximations to idealized receptive fields as derived from
mathematical and physical principles.
From such a viewpoint, it is highly interesting to see that the
neurophysiological cell recordings in the LGN and the primary visual
cortex presented by DeAngelis et al.\ \cite{DeAngOhzFre95-TINS,deAngAnz04-VisNeuroSci} 
are in very good qualitative agreement with the predictions
generated by our mathematically and physically motivated normative theory
(see figure~\ref{fig-deang-tins-temp-resp-prof-lagged-nonlagged} and
figure~\ref{fig-biol-model-simple-cells-rec-filters-over-time}).

Given the derived time-causal and time-recursive formulation of our
basic linear spatio-temporal receptive fields, we have described how this
theory can be used for computing different types of both linear and
non-linear scale-normalized spatio-temporal features. Specifically, we have emphasized how scale
normalization by $L_p$-normalization leads to fundamentally different
results compared to more traditional variance-based normalization.  
By the formulation of the corresponding scale normalization factors
for discrete temporal scale space, we have also shown how they permit
the formulation of an operational criterion to estimate how many
intermediate temporal scale levels are needed to approximate true
scale invariance up to a given tolerance.

Finally, we have shown how different types of spatio-temporal features
can defined in terms of spatio-temporal differential invariants built from
spatio-temporal receptive field responses, including their
transformation properties under natural image transformations, with
emphasis on independent scaling transformations over space {\em vs.\/}\ time, 
rotational invariance over the spatial domain and illumination and exposure
control variations. 
We propose that the presented theory can be used for computing features for
generic purposes in computer vision and for computational modelling of 
biological vision for image data over a time-causal spatio-temporal domain, 
in an analogous way as the Gaussian scale-space concept constitutes a 
canonical model for processing image data over a purely spatial domain.

\section*{Acknowledgements}

I would like to thank Michael Felsberg for valuable comments
that improved the presentation of the discrete derivative
approximations in section~\ref{sec-disc-spat-temp-RF}.

\appendix

\section{Frequency analysis of the time-causal kernels}
\label{app-freq-anal-time-caus-kern}

In this appendix, we will perform an in-depth analysis of the proposed 
time-causal scale-space kernels with regard to their frequency properties
and moment descriptors derived via the Fourier transform,
both for the case of a logarithmic distribution of the intermediate temporal scale
levels and a uniform distribution of the intermediate temporal scale levels.
Specifically, the results to be derived will provide a way to characterize
properties of the limit kernel when the number of temporal scale levels $K$ tends to infinity.

\subsection{Logarithmic distribution of the intermediate scale levels}
\label{app-freq-anal-time-caus-kern-log-distr}

In section~\ref{sec-time-caus-limit-kernel}, we gave the following explicit expressions
for the Fourier transform of the time-causal kernels based on a logarithmic distribution
of the intermediate scale levels
\begin{equation}
    \hat{h}_{exp}(\omega;\; \tau, c, K) 
       = \frac{1}{1 + i \, c^{1-K} \sqrt{\tau} \, \omega}
           \prod_{k=2}^{K} \frac{1}{1 + i \, c^{k-K-1} \sqrt{c^2-1} \sqrt{\tau} \, \omega} 
\end{equation}
for which the magnitude and the phase are given by
\begin{align}
  \begin{split}
    & |\hat{h}_{exp}(\omega;\; \tau, c, K) |
 \end{split}\nonumber\\
  \begin{split}
     & = \frac{1}{\sqrt{1 + c^{2(1-K)} \tau \, \omega^2}}
              \prod_{k=2}^{K} \frac{1}{\sqrt{1 + c^{2(k-K-1)} (c^2-1) \tau \, \omega^2}},
 \end{split}\\
  \begin{split}
    & \arg \hat{h}_{exp}(\omega;\; \tau, c, K) 
 \end{split}\nonumber\\
  \begin{split}
     & = \arctan \left( c^{1-K} \sqrt{\tau} \, \omega \right)
             + \sum_{k=2}^{K} \arctan \left( c^{k-K-1} \sqrt{c^2-1} \sqrt{\tau} \, \omega \right).
 \end{split}\nonumber\\
\end{align}
Let us rewrite the magnitude of the Fourier transform on
exponential form
\begin{multline}
    |\hat{h}_{exp}(\omega;\; \tau, c, K) |
      = e^{\log |\hat{h}_{exp}(\omega;\; \tau, c, K) |}
     \\ = e^{-\frac{1}{2} \log(1 + c^{2(1-K)} \tau \, \omega^2) 
                 -\frac{1}{2} \sum_{k=2}^K \log(1 + c^{2(k-K-1)} (c^2-1) \tau \, \omega^2)}
\end{multline}
and compute the Taylor expansion of
\begin{equation}
  \log |\hat{h}_{exp}(\omega;\; \tau, c, K) |
  = C_2 \omega^2 + C_4  \omega^4 
  + {\cal O}(\omega^6)
\end{equation}
where
\begin{align}
  \begin{split}
    C_2 & = - \frac{\tau}{2},
  \end{split}\\
  \begin{split}
    C_4 & 
   = -\frac{\tau ^2 \left(-2 c^{4-4 K}-c^2+1\right)}{4 \left(c^2+1\right)}
   \rightarrow \frac{\left(c^2-1\right) \tau ^2}{4 \left(c^2+1\right)},
  \end{split}
\end{align}
and the rightmost expression for $C_4$ 
shows the limit value when the number
$K$ of first-order integrators coupled in cascade tends to infinity.

Let us next
compute the Taylor expansion of
\begin{equation}
  \arg \hat{h}_{exp}(\omega;\; \tau, c, K) 
  = C_1 \omega + C_3 \omega^3 
   + {\cal O}(\omega^5)
\end{equation}
where the coefficients are given by
\begin{align}
  \begin{split}
    C_1 & 
    = \frac{\sqrt{\tau } c^{-K} \left(-c^2+\sqrt{c^2-1} c-\sqrt{c^2-1} c^K+c\right)}{c-1}
  \end{split}\nonumber\\
  \begin{split}
    \rightarrow - \frac{\sqrt{\left(c^2-1\right)} \, \sqrt{\tau}}{c-1},
  \end{split}\\
  \begin{split}
    C_3 & 
   = \frac{\sqrt{c^2-1} \tau ^{3/2}}{3 \left(c^2+c+1\right)}
      \left(
           \left(c^{3 K}+c^{3K+1}-c^4-c^3\right) c^{-3 K}
      \right.
  \end{split}\nonumber\\
   \begin{split}
       & \phantom{= \frac{\sqrt{c^2-1} \tau ^{3/2}}{3 \left(c^2+c+1\right)} \left( \right.}
           \left.
               + \left(c^5+c^4+c^3\right) \tau ^{3/2} c^{-3 K}
      \right)
  \end{split}\nonumber\\
  \begin{split}
   & \phantom{=} \rightarrow 
          \frac{(c+1) \sqrt{c^2-1} \, \tau^{3/2}}{3 \left(c^2+c+1\right)},
  \end{split}
\end{align}
and again the rightmost expressions for $C_1$ and $C_3$ show the
limit values when the number $K$ of scale levels tends to infinity.

\begin{figure}[hbt]
  \begin{center}
    \begin{tabular}{c}
      {\small\em skewness $\gamma_1(c)$} \\
      \includegraphics[width=0.35\textwidth]{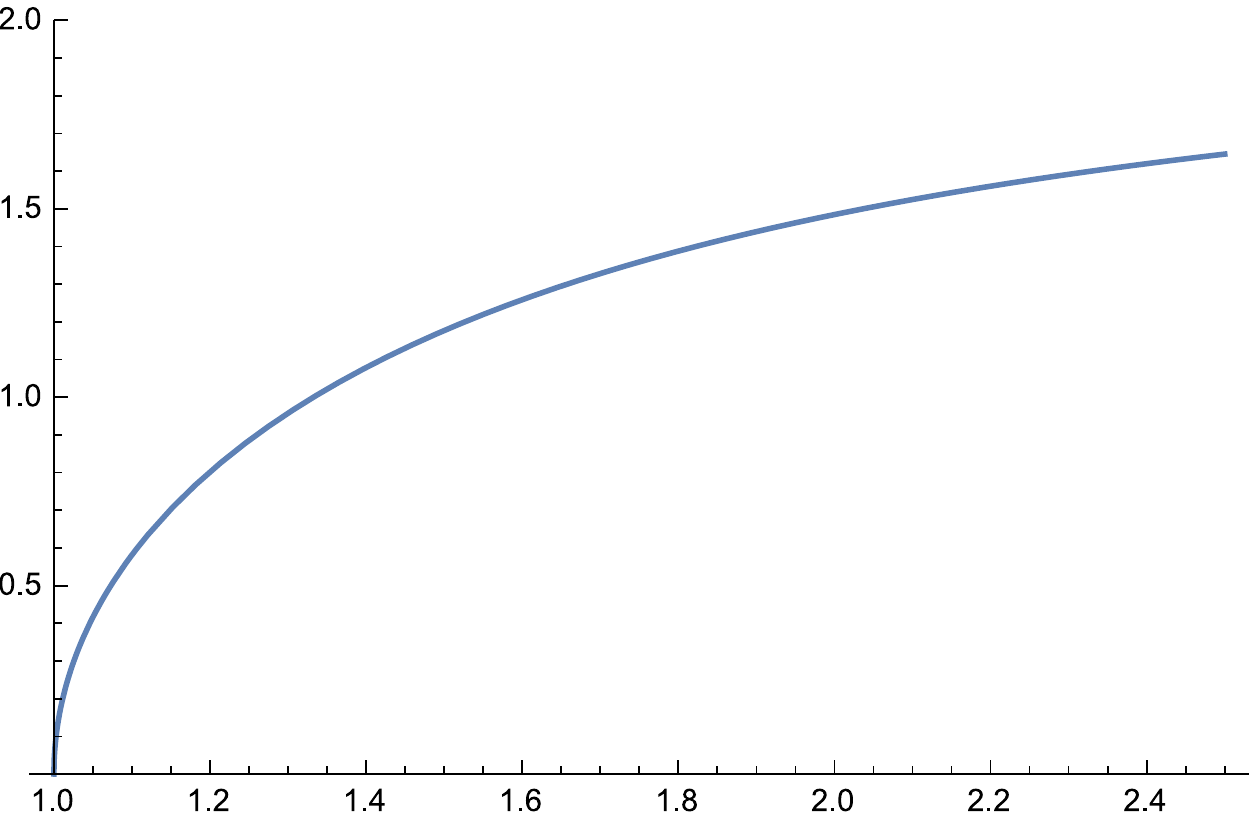}
      \\ \\
     {\small\em kurtosis $\gamma_2(c)$} \\
      \includegraphics[width=0.35\textwidth]{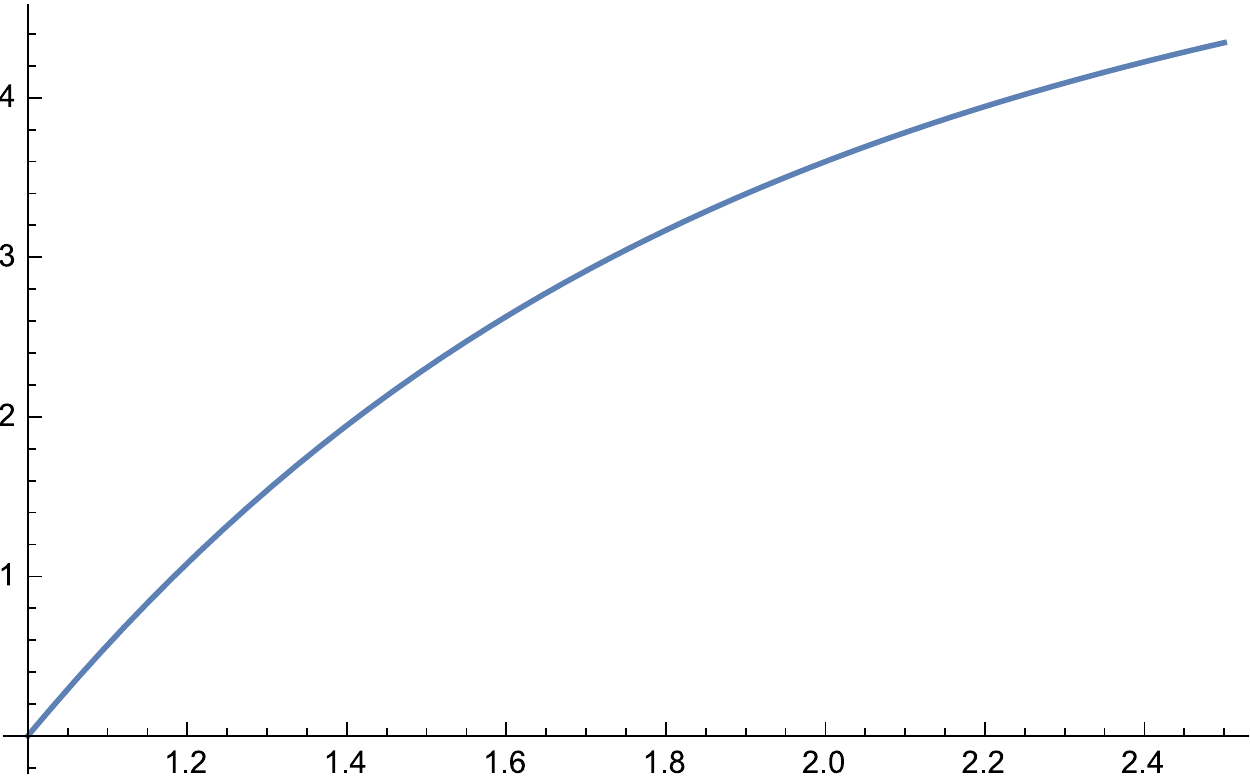} \\
    \end{tabular} 
  \end{center}

   \caption{Graphs of the skewness measure $\gamma_1$
    (\protect\ref{eq-skewness-timecaus-kern-log-distr}) 
     and the kurtosis measure $\gamma_2$
   (\protect\ref{eq-kurtosis-timecaus-kern-log-distr}) 
   as function of the distribution parameter $c$
   for the time-causal scale-space kernels corresponding to limit case
 of $K$ truncated exponential kernels having a logarithmic
 distribution of the intermediate scale levels coupled in cascade in
 the limit case when the number of scale levels $K$ tends to infinity.}
  \label{fig-skew-kurt-explogdistr}
\end{figure}

Following the definition of cumulants $\kappa_n$ defined as the Taylor
coefficients of the logarithm of the Fourier transform
\begin{equation}
  \label{eq-def-cum-from-FT}
  \log h(\omega) 
 = \sum_{n=0}^{\infty} \kappa_n \frac{(-i \omega)^n}{n!},
\end{equation}
we obtain
\begin{align}
  \begin{split}
    & \log \hat{h}_{exp}(\omega;\; \tau, c, K)  
  \end{split}\nonumber\\
  \begin{split}
     & = -C_1 (-i \omega) - C_2  (-i \omega)^2 + C_3 (-i \omega)^3 + C_4  (-i \omega)^4 + {\cal O}(i \omega^5)
  \end{split}\nonumber\\
  \begin{split}
  \label{eq-def-cum-from-series-exp-FT}
    & = \kappa_0 + \frac{\kappa_1}{1!} (-i \omega) + \frac{\kappa_2}{2!} (-i \omega^2) +
    \frac{\kappa_3}{3!} (-i \omega)^3 + \frac{\kappa_4}{4!} (-i \omega)^4 + {\cal O}(i \omega^5)
  \end{split}
\end{align}
and can read the cumulants of the underlying temporal scale-space
kernel as $\kappa_0 = 0$, $\kappa_1 = -C_1$, $\kappa_2 = -2 C_2$,
$\kappa_3 = 6 C_3$ and $\kappa_4 = 24 C_4$.
Specifically, the first-order moment
$M_1$ and the higher-order central moments $M_2$, $M_3$ and $M_4$ are
related to the cumulants according to
\begin{align}
  \begin{split}
      M_1 & = \kappa_1 = -C_1 
      \rightarrow 
        \frac{\sqrt{\left(c^2-1\right)} \, \sqrt{\tau}}{c-1},
  \end{split}\\
  \begin{split}
      M_2 & = \kappa_2 = -2C_2 = \tau,
  \end{split}\\
  \begin{split}
      M_3 & = \kappa_3 = 6 C_3 \rightarrow 
          \frac{2 (c+1) \sqrt{c^2-1} \, \tau^{3/2}}{\left(c^2+c+1\right)},
  \end{split}\\
 \begin{split}
      M_4 & = \kappa_4 + 3 \kappa_2^2  = 24 C_4 + 12 C_2^2
                \rightarrow \frac{3 \left(3 c^2-1\right) \tau ^2}{c^2+1}.
  \end{split}
\end{align}
Thus, the skewness $\gamma_1$ and the kurtosis $\gamma_2$ measures of the
corresponding temporal scale-space kernels are given by
\begin{align}
  \begin{split}
     \label{eq-skewness-timecaus-kern-log-distr}
     \gamma_1 
    & = \frac{\kappa_3}{\kappa_2^{3/2}} = \frac{M_3}{M_2^{3/2}} = \frac{3 C_3}{\sqrt{2} (-C_2)^{3/2}}
    \rightarrow 
          \frac{2 (c+1) \sqrt{c^2-1}}{\left(c^2+c+1\right)},
  \end{split}\\
 \begin{split}
     \label{eq-kurtosis-timecaus-kern-log-distr}
     \gamma_2 
     & = \frac{\kappa_4}{\kappa_2^2} = \frac{M_4}{M_2^2} - 3 = 6 \, \frac{C_4}{C_2^2}
     \rightarrow \frac{6 \left(c^2-1\right)}{c^2+1}.
  \end{split}
\end{align}
Figure~\ref{fig-skew-kurt-explogdistr} shows graphs these skewness and
kurtosis measures as function of the distribution parameter $c$ for
the limit case when the number of scale levels $K$ tends to infinity.
As can be seen, both the skewness and the kurtosis measures of the
temporal scale-space kernels increase with increasing values of the distribution parameter
$c$.

\subsection{Uniform distribution of the intermediate scale levels}

When using a uniform distribution of the intermediate scale levels (\ref{eq-distr-tau-values-uni}),
the time constants of the individual first-order integrators are given by (\ref{eq-mu-k-uni})
and the explicit expression for the Fourier transform (\ref{eq-FT-composed-kern-casc-truncexp}) 
is 
\begin{equation}
    \hat{h}_{exp}(\omega;\; \tau, K) 
       = \frac{1}{\left( 1 + i \, \sqrt{\frac{\tau}{K}} \, \omega \right)^K}.
\end{equation}
Specifically, the magnitude and the phase of the Fourier transform are given by
\begin{align}
  \begin{split}
    |\hat{h}_{exp}(\omega;\; \tau, K) |
     & = \frac{1}{\left( 1 + \frac{\tau}{K} \, \omega^2 \right)^{K/2}},
 \end{split}\\
  \begin{split}
    \arg \hat{h}_{exp}(\omega;\; \tau, K) 
     & = - K \arctan \left( \sqrt{\frac{\tau}{K}} \, \omega \right).
 \end{split}
\end{align}
Let us rewrite the magnitude of the Fourier transform on
exponential form
\begin{align}
  \begin{split}
    |\hat{h}_{exp}(\omega;\; \tau, K) |
     & = e^{\log |\hat{h}_{exp}(\omega;\; \tau, K) |}
      = e^{-\frac{K}{2} \log(1 + \frac{\tau}{K} \, \omega^2)}
 \end{split}\nonumber\\
\end{align}
and compute the Taylor expansion of
\begin{equation}
  \log |\hat{h}_{exp}(\omega;\; \tau, K) |
  = C_2 \omega^2 + C_4  \omega^4 + {\cal O}(\omega^6)
\end{equation}
where
\begin{align}
  \begin{split}
    C_2 & = - \frac{\tau}{2},
  \end{split}\\
  \begin{split}
    C_4 & = \frac{\tau ^2}{4 K}.
  \end{split}
\end{align}
Next, let us compute the Taylor expansion of
\begin{equation}
  \arg \hat{h}_{exp}(\omega;\; \tau, K) 
  = C_1 \omega + C_3 \omega^3 
   + {\cal O}(\omega^5)
\end{equation}
where the coefficients are given by
\begin{align}
  \begin{split}
    C_1 & 
    = - \sqrt{K \tau},
  \end{split}\\
  \begin{split}
    C_3 & 
   = \frac{\tau ^{3/2}}{3 \sqrt{K}}.
  \end{split}
\end{align}
Following the definition of cumulants $\kappa_n$ according to
(\ref{eq-def-cum-from-FT}),
we can in an analogous way to (\ref{eq-def-cum-from-series-exp-FT}) in
previous section read
$\kappa_0 = 0$, $\kappa_1 = -C_1$, $\kappa_2 = -2 C_2$,
$\kappa_3 = 6 C_3$ and $\kappa_4 = 24 C_4$,
and relate the first-order moment
$M_1$ and the higher-order central moments $M_2$, $M_3$ and $M_4$ to the cumulants according to
\begin{align}
  \begin{split}
      M_1 & = \kappa_1 = -C_1 = \sqrt{K \tau},
  \end{split}\\
  \begin{split}
      M_2 & = \kappa_2 = -2C_2 = \tau,
  \end{split}\\
  \begin{split}
      M_3 & = \kappa_3 = 6 C_3 = \frac{2 \tau ^{3/2}}{\sqrt{K}},
  \end{split}\\
 \begin{split}
      M_4 & = \kappa_4 + 3 \kappa_2^2  = 24 C_4 + 12 C_2^2
                = 3 \tau ^2 + \frac{6 \tau ^2}{K}.
  \end{split}
\end{align}
Thus, the skewness $\gamma_1$ and the kurtosis $\gamma_2$ of the
corresponding temporal scale-space kernels are given by
\begin{align}
  \begin{split}
     \label{eq-skewness-timecaus-kern-uni-distr}
     \gamma_1 
    & = \frac{\kappa_3}{\kappa_2^{3/2}} = \frac{M_3}{M_2^{3/2}} = \frac{2}{\sqrt{K}},
  \end{split}\\
 \begin{split}
     \label{eq-kurtosis-timecaus-kern-uni-distr}
     \gamma_2 
     & = \frac{\kappa_4}{\kappa_2^2} = \frac{M_4}{M_2^2} - 3 = \frac{6}{K}.
  \end{split}
\end{align}
From these expressions we can note that when the number $K$ of first-order integrators that are coupled in
cascade increases, these skewness and kurtosis
measures tend to zero for the temporal scale-space kernels 
having a uniform distribution of the intermediate temporal scale
levels. The corresponding skewness and kurtosis measures
(\ref{eq-skewness-timecaus-kern-log-distr}) and
(\ref{eq-kurtosis-timecaus-kern-log-distr}) for the
kernels having a logarithmic distribution of the intermediate temporal
scale levels do on the other hand remain strictly positive. 
These properties reveal a fundamental difference between
the two classes of time-causal kernels obtained by distributing the
intermediate scale levels of first-order integrators coupled in
cascade according to a logarithmic {\em vs.\/}\ a
uniform distribution.

\section{Comparison with Koenderink's scale-time model}
\label{app-comp-koe-model}

In his scale-time model, Koenderink \cite{Koe88-BC} proposed to perform a
logarithmic mapping of the past via a time delay $\delta$ and then
applying Gaussian smoothing on the transformed domain, leading to a
time-causal kernel of the form, here largely following the notation in
Florack \cite[result~4.6, page 116]{Flo97-book}
\begin{equation}
  h_{log}(t;\; \sigma, \delta, a) 
  = \frac{1}{\sqrt{2 \pi } \sigma  (\delta -a)}
e^{-\frac{\log ^2\left(\frac{t-a}{\delta -a}\right)}{2 \sigma ^2}}
\end{equation}
with $a$ denoting the present moment, $\delta$ denoting the time delay and
$\sigma$ is a dimensionless temporal scale parameter relative to the
logarithmic time axis.
For simplicity, we will henceforth assume $a = 0$ leading to kernels
of the form
\begin{equation}
  h_{log}(t;\; \sigma, \delta) 
  =\frac{1}{\sqrt{2 \pi } \sigma \,\delta}
  e^{-\frac{\log ^2\left(\frac{t}{\delta }\right)}{2 \sigma ^2}}
 \end{equation}
and with convolution reversal of the time axis
such that causality implies $h_{log}(t;\; \sigma, \delta) = 0$ 
for $t < 0$.
By integrating this kernel symbolically in Mathematica, we find
\begin{equation}
  \int_{t=-\infty}^{\infty} h_{log}(t;\; \sigma, \delta) \, dt
  = e^{\frac{\sigma^2}{2}}
 \end{equation}
implying that the corresponding time-causal kernel normalized to unit
$L_1$-norm should be
\begin{equation}
  h_{Koe}(t;\; \sigma, \delta) 
  =\frac{1}{\sqrt{2 \pi } \sigma \,\delta}
  e^{-\frac{\log ^2\left(\frac{t}{\delta }\right)}{2 \sigma ^2} -\frac{\sigma^2}{2}}.
 \end{equation}
The temporal mean of this kernel is
\begin{equation}
  M_1 = \bar{t} = \int_{t=-\infty}^{\infty} t \, h_{Koe}(t;\; \sigma, \delta) \, dt
  = \delta \, e^{\frac{3 \sigma ^2}{2}} 
 \end{equation}
and the higher-order central moments
\begin{align}
  \begin{split}
     M_2 
     & = \int_{t=-\infty}^{\infty} (t - \bar{t})^2 \, h_{Koe}(t;\; \sigma, \delta) \, dt
  \end{split}\nonumber\\
  \begin{split}
      & = \delta ^2 e^{3 \sigma ^2} \left(e^{\sigma ^2}-1\right),
  \end{split}\\
  \begin{split}
      M_3 
       & = \int_{t=-\infty}^{\infty} (t - \bar{t})^3 \, h_{Koe}(t;\; \sigma, \delta) \, dt
  \end{split}\nonumber\\
  \begin{split}
        & = \delta ^3 e^{\frac{9 \sigma ^2}{2}} \left(e^{\sigma ^2}-1\right)^2 \left(e^{\sigma
   ^2}+2\right),
  \end{split}\\
  \begin{split}
    M_4 
       & = \int_{t=-\infty}^{\infty} (t - \bar{t})^4 \, h_{Koe}(t;\; \sigma, \delta) \, dt
   \end{split}\\
  \begin{split}
       & = \delta ^4 e^{6 \sigma ^2} \left(e^{\sigma ^2}-1\right)^2 \left(3 e^{2 \sigma ^2}+2 e^{3
   \sigma ^2}+e^{4 \sigma ^2}-3\right).
  \end{split}
\end{align}

\begin{figure}[hbt]
  \begin{center}
    \begin{tabular}{c}
      {\small\em skewness $\gamma_1(\sigma)$} \\
           \includegraphics[width=0.35\textwidth]{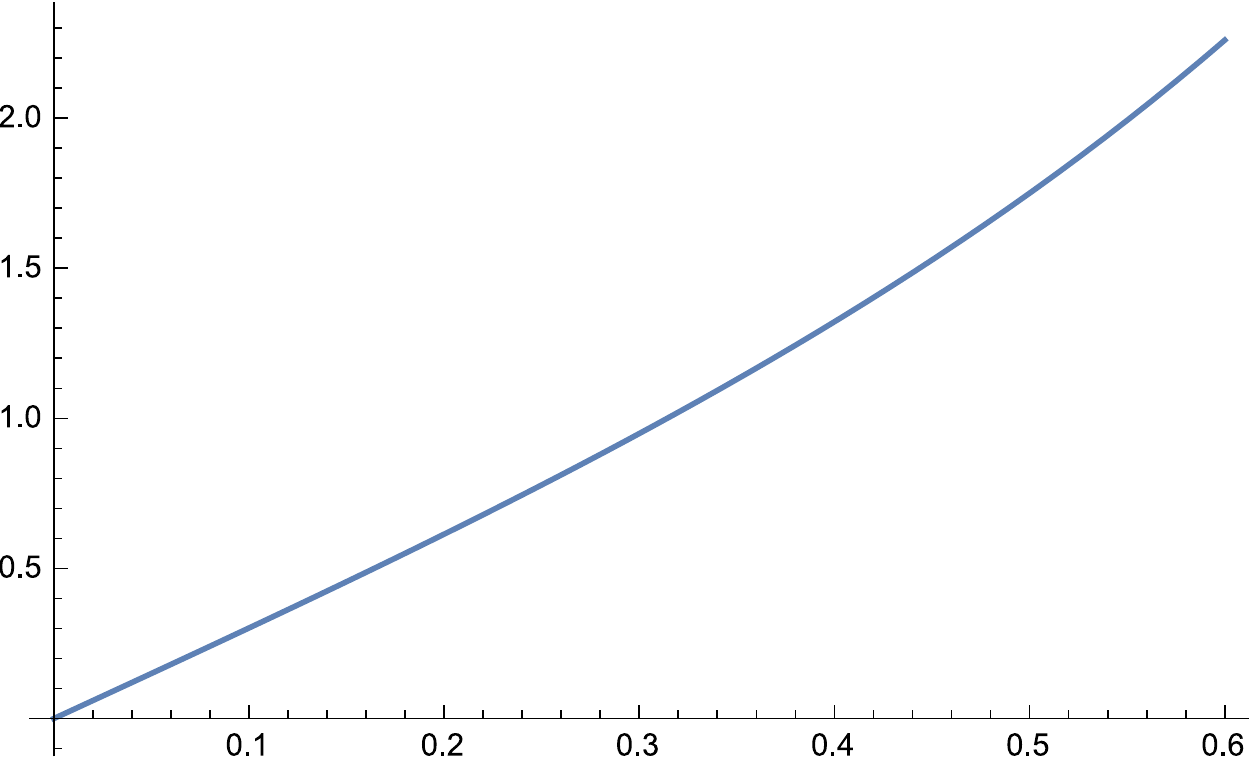}
           \\ \\
      {\small\em kurtosis $\gamma_2(\sigma)$} \\
      \includegraphics[width=0.35\textwidth]{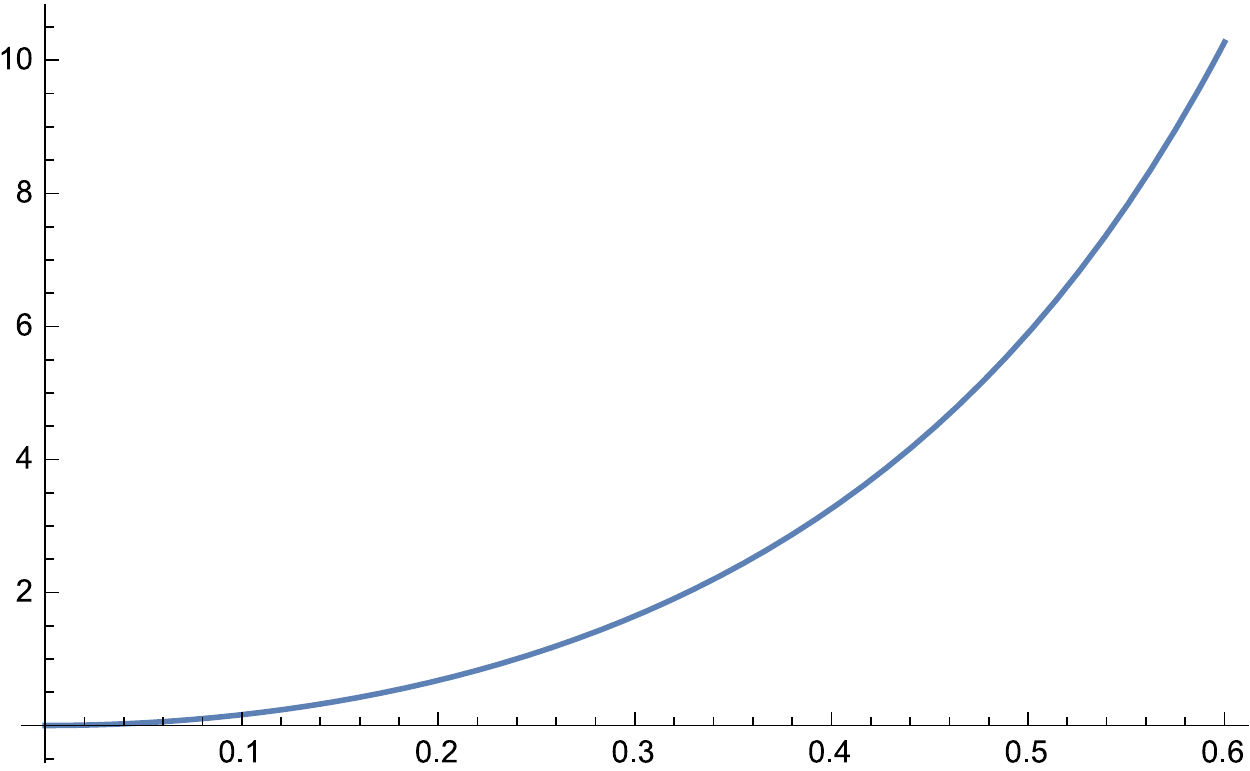} \\
    \end{tabular} 
  \end{center}

   \caption{Graphs of the skewness measure $\gamma_1$
    (\protect\ref{eq-skewness-scaletime}) 
     and the kurtosis measure $\gamma_2$
   (\protect\ref{eq-kurtosis-scaletime}) 
   as function of the dimensionless temporal scale parameter $\sigma$
   relative to the logarithmic transformation of the past for the
   time-causal kernels in
   Koenderink's scale-time model.}
  \label{fig-skew-kurt-scaletime}
\end{figure}

\noindent
Thus, the skewness $\gamma_1$ and the kurtosis $\gamma_2$ of the
temporal kernels in Koenderink's scale-time model
are given by (see figure~\ref{fig-skew-kurt-scaletime} for graphs)
\begin{align}
  \begin{split}
     \label{eq-skewness-scaletime}
     \gamma_1 = \frac{M_3}{M_2^{3/2}} = \sqrt{e^{\sigma ^2}-1} \left(e^{\sigma ^2}+2\right),
  \end{split}\\
 \begin{split}
     \label{eq-kurtosis-scaletime}
     \gamma_2 = \frac{M_4}{M_2^2} - 3 = 3 e^{2 \sigma ^2}+2 e^{3 \sigma ^2}+e^{4 \sigma ^2}-6.
  \end{split}
\end{align}

\begin{figure*}[hbtp]
  \begin{center}
   \begin{tabular}{ccc}
        {\small $h(t;\; K=7, c = \sqrt{2})$} 
      & {\small $h_{t}(t;\; K=7, c = \sqrt{2}))$} 
      & {\small $h_{tt}(t;\; K=7, c = \sqrt{2}))$} \\
      \includegraphics[width=0.23\textwidth]{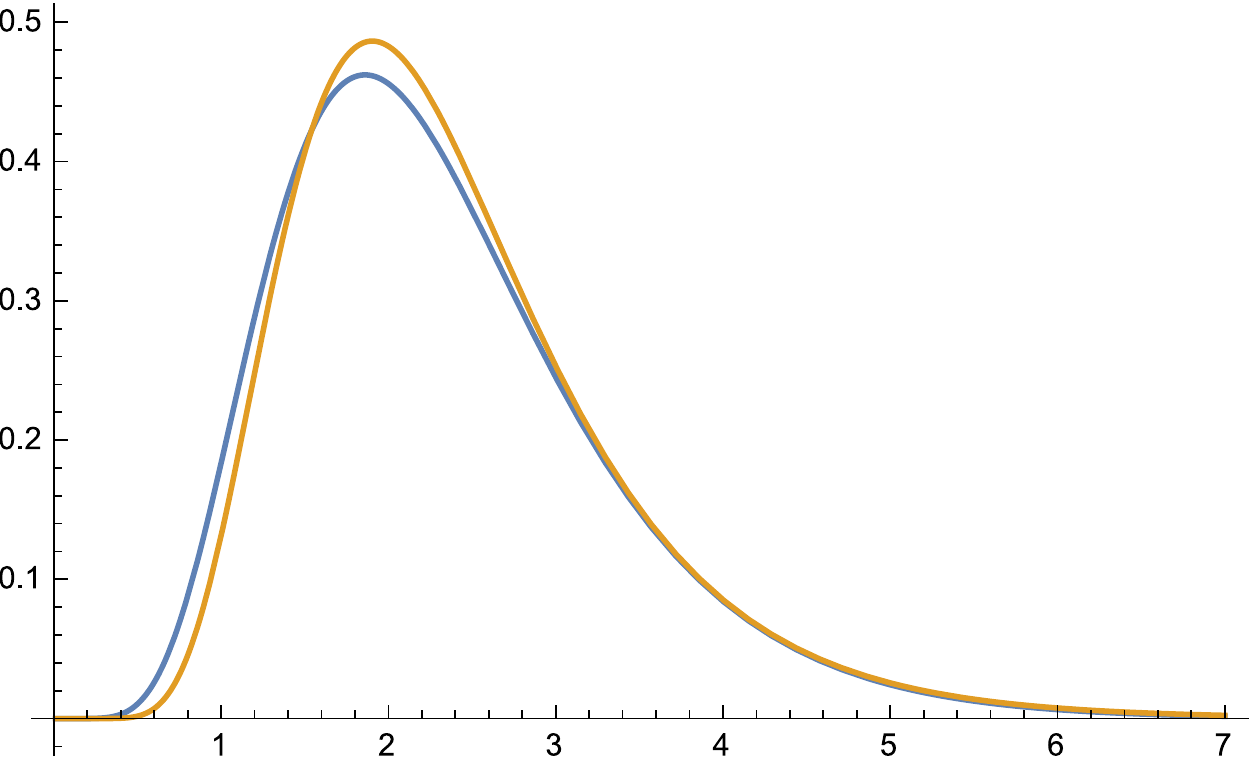} &
      \includegraphics[width=0.23\textwidth]{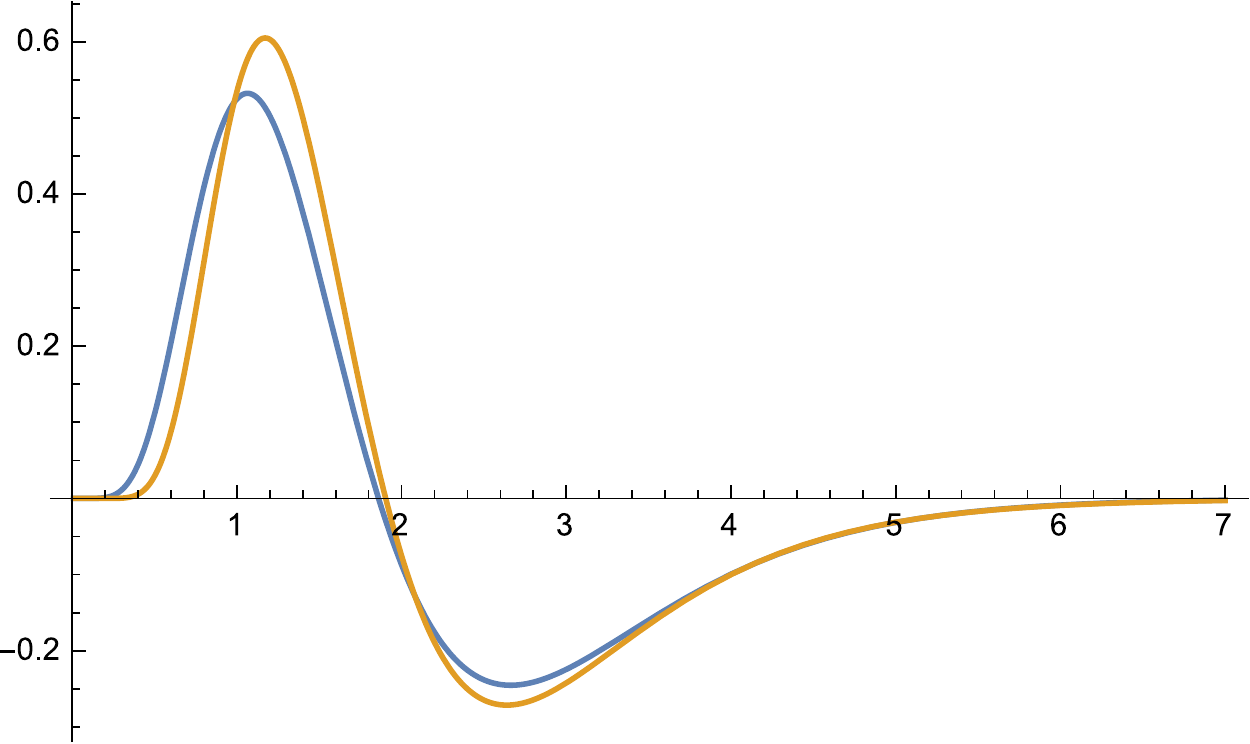} &
      \includegraphics[width=0.23\textwidth]{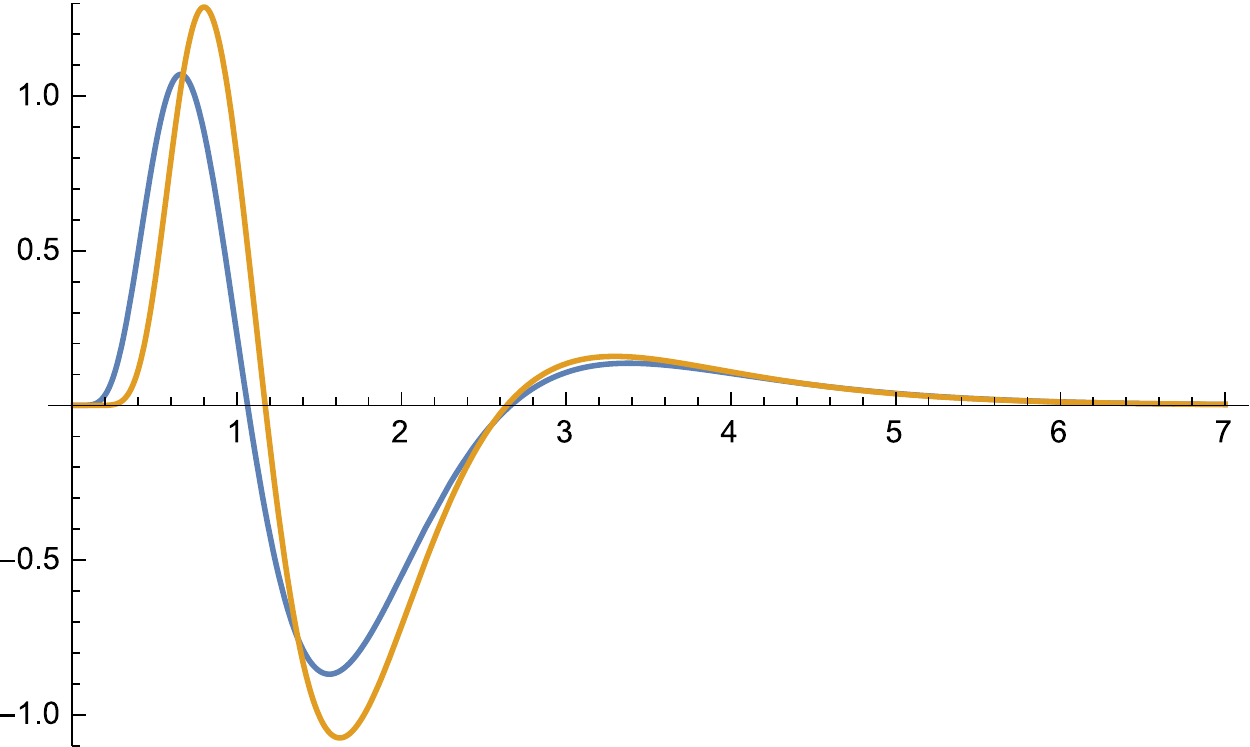} \\
\\
      {\small $h(t;\; K=7, c = 2^{3/4})$} 
      & {\small $h_{t}(t;\; K=7, c = 2^{3/4}))$} 
      & {\small $h_{tt}(t;\; K=7, c = 2^{3/4}))$} \\
      \includegraphics[width=0.23\textwidth]{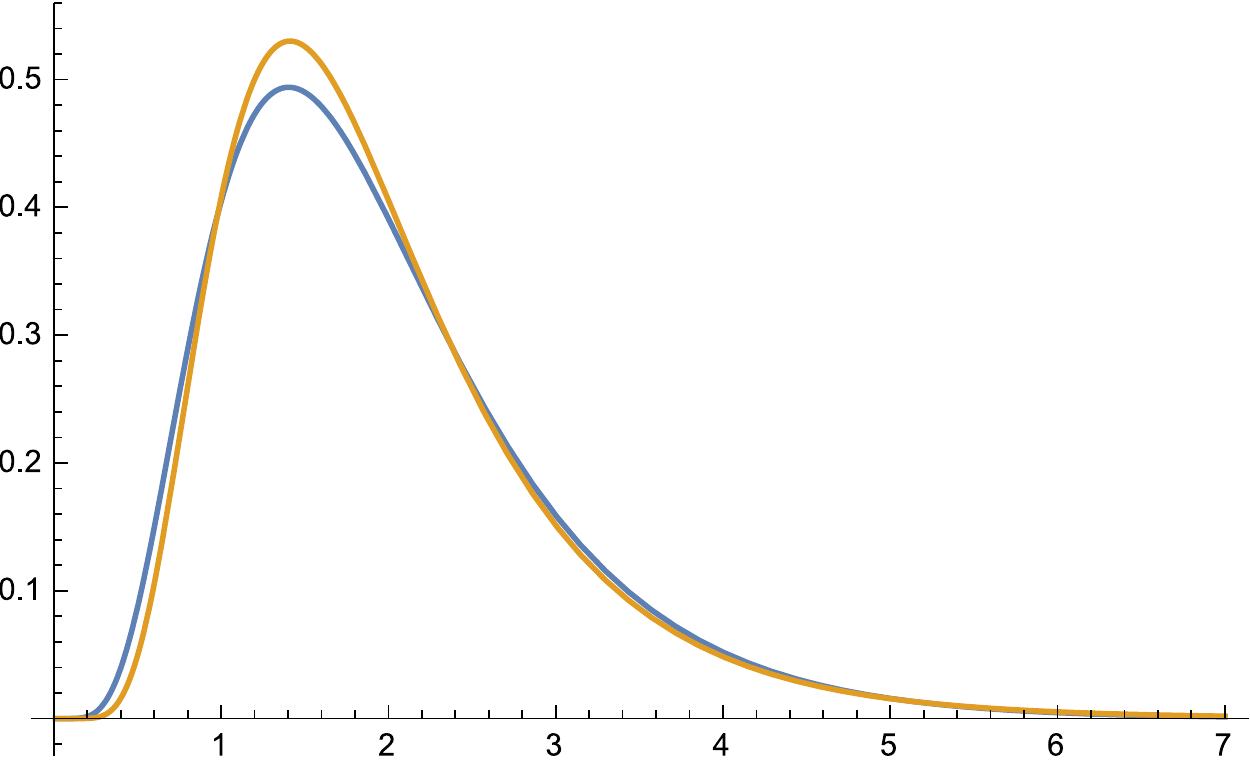} &
      \includegraphics[width=0.23\textwidth]{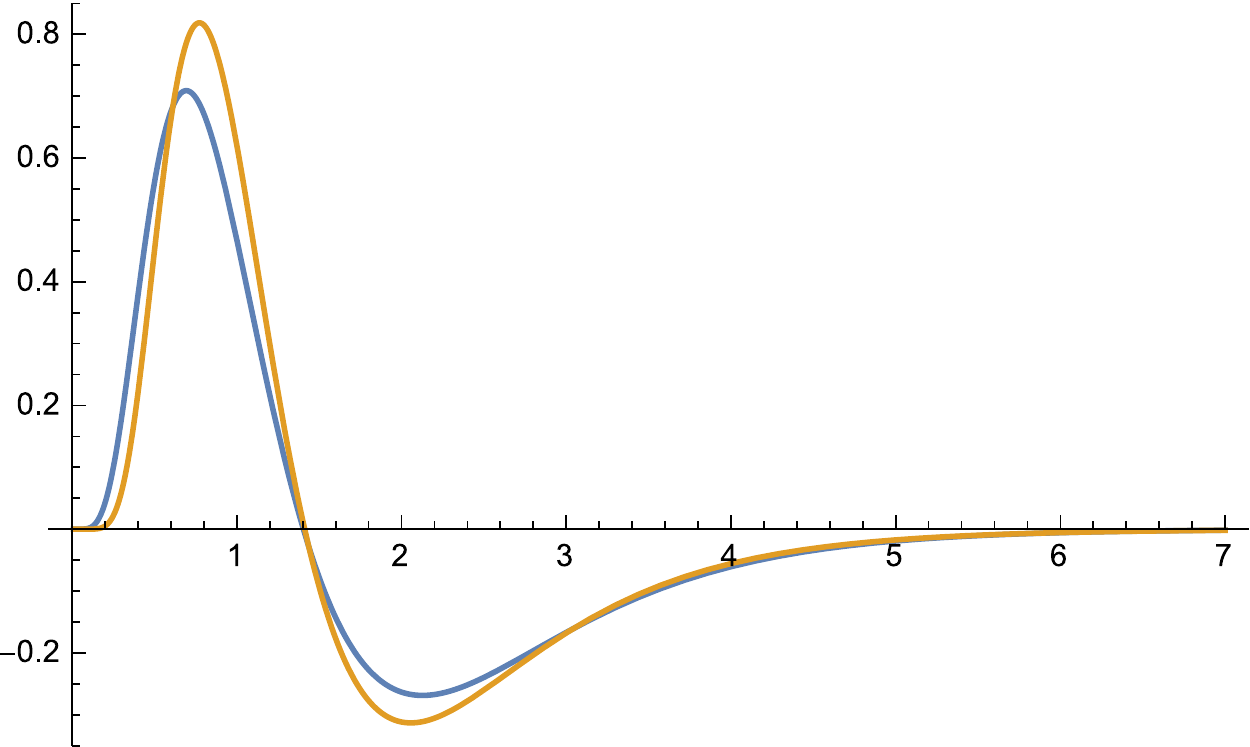} &
      \includegraphics[width=0.23\textwidth]{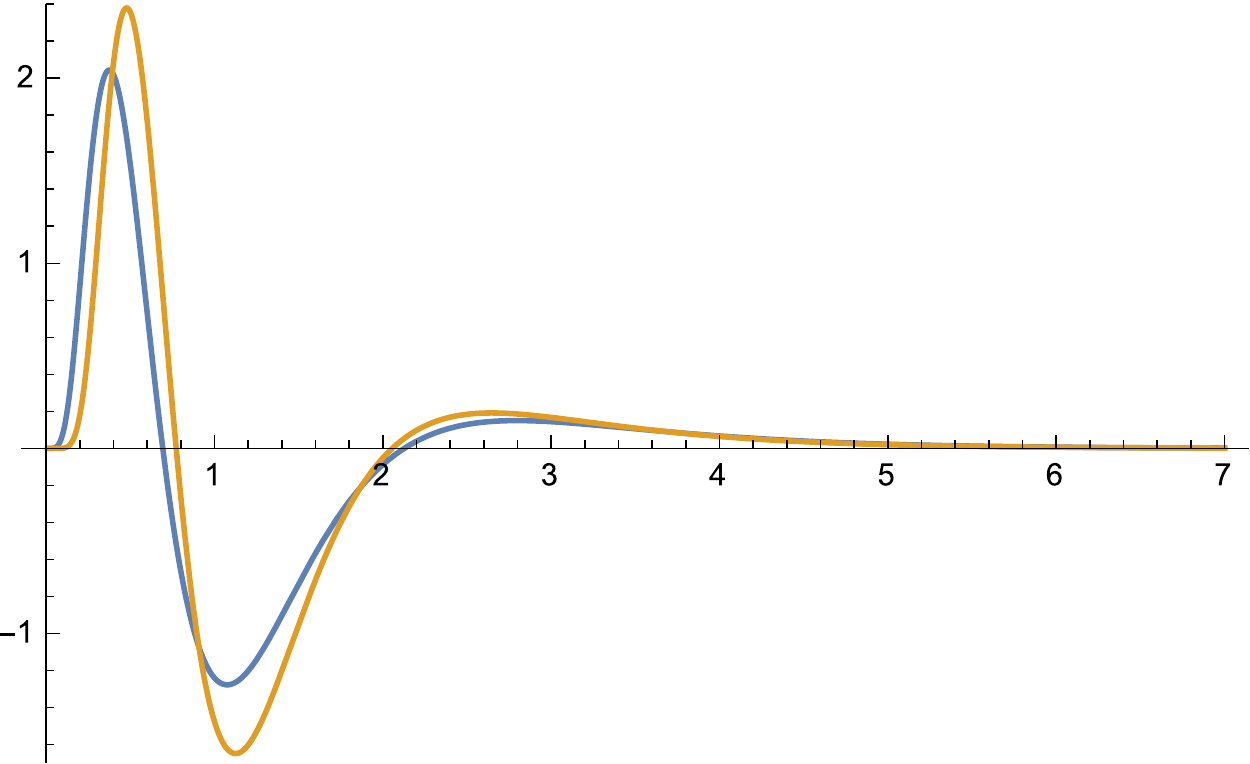} \\
\\
      {\small $h(t;\; K=7, c = 2)$} 
      & {\small $h_{t}(t;\; K=7, c = 2))$} 
      & {\small $h_{tt}(t;\; K=7, c = 2))$} \\
      \includegraphics[width=0.23\textwidth]{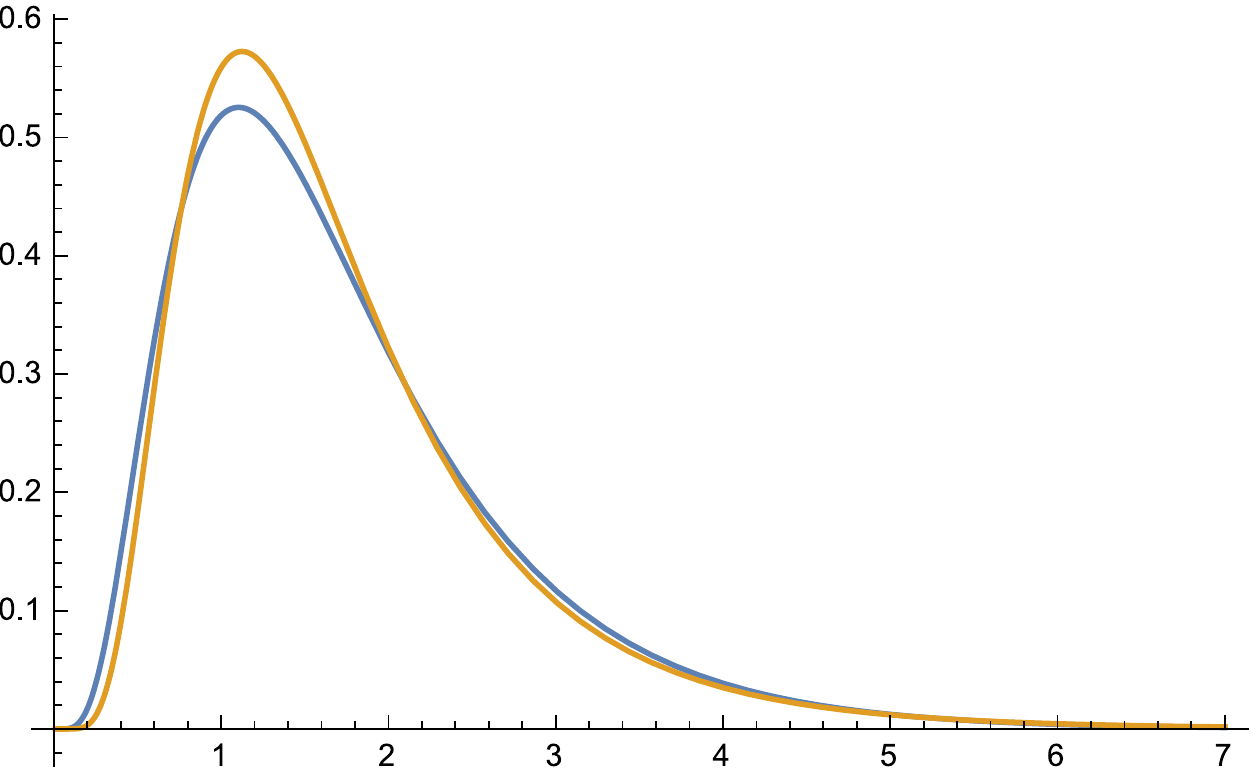} &
      \includegraphics[width=0.23\textwidth]{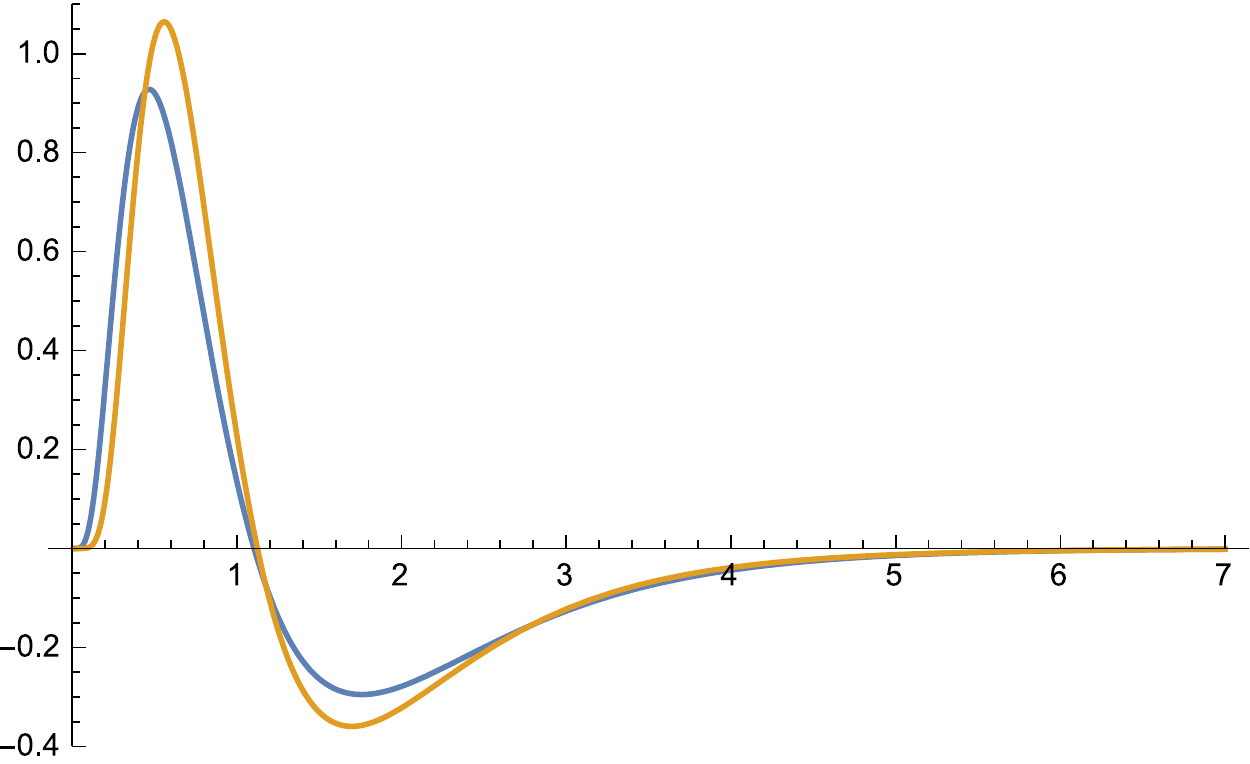} &
      \includegraphics[width=0.23\textwidth]{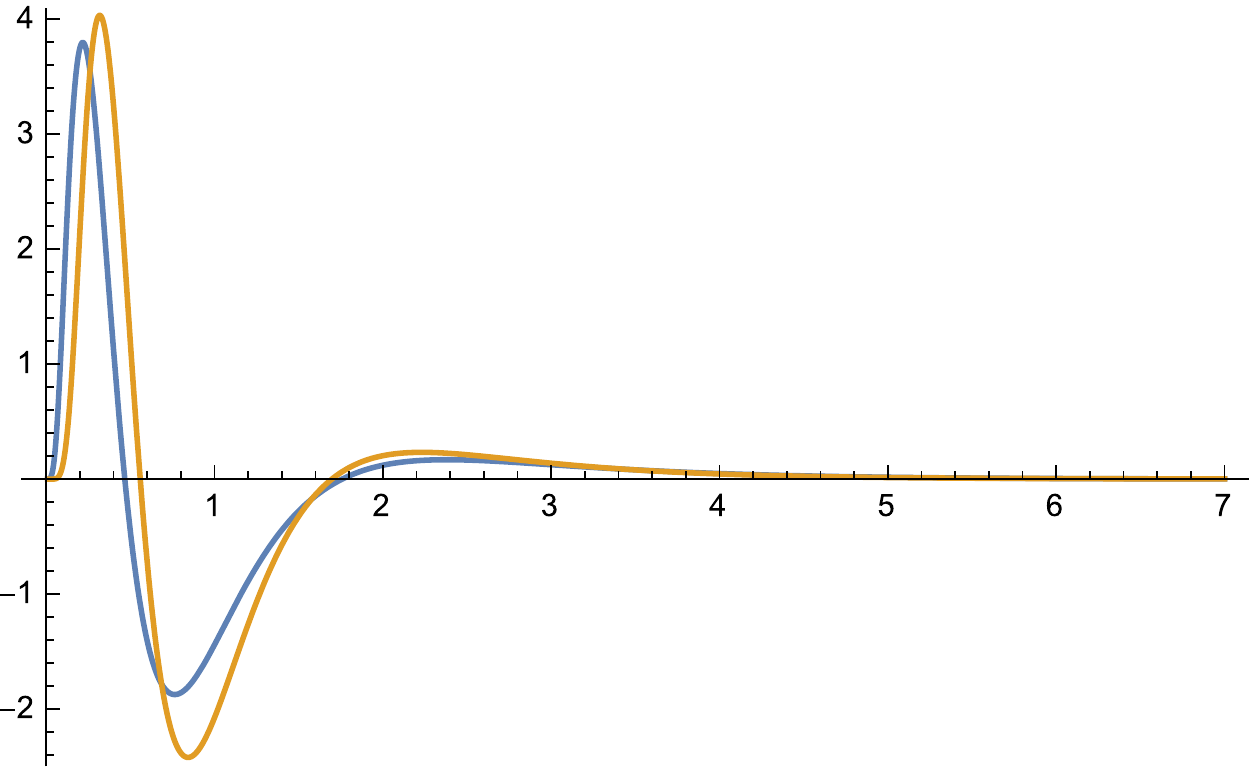} \\
    \end{tabular} 
  \end{center}
   \caption{Comparison between the proposed time-causal kernels corresponding to the composition of
           truncated exponential kernels in cascade (blue curves) for a
           logarithmic distribution of the intermediate scale levels
           and the temporal kernels in Koenderink's scale-time model
           (brown curves) shown for both the original smoothing kernels and their
           first- and second-order temporal derivatives.
           All kernels correspond to temporal scale (variance) $\tau = 1$ with
           the additional parameters determined such that the
           temporal mean values (the first-order temporal moments) become equal in
           the limit case when the number of temporal scale levels $K$
           tends to infinity
           (equation~(\protect\ref{eq-var-transf-par-tau-c-sigma-delta-explogdistr-scaletime-models})).
           (top row) Logarithmic distribution of the temporal scale levels
           for $c = \sqrt{2}$ and $K = 10$.
          (middle row) Corresponding results
           for $c = 2^{3/4}$ and $K = 10$.
          (bottom row) Corresponding results 
           for $c = 2$ and $K = 10$.}
  \label{fig-trunc-exp-kernels-1D+scaletime}
\end{figure*}

\noindent
If we want to relate these kernels in Koenderink's scale-time model to our
time-causal scale-space kernels, a natural starting point is to
require that the total amount of temporal
smoothing as measured by the variances $M_2$ of the two kernels should be equal.
Then, this implies the relation
\begin{equation}
  \label{eq-c2mom-eq-rel}
  \tau = \delta ^2 e^{3 \sigma ^2} \left(e^{\sigma ^2}-1\right).
\end{equation}
If we additionally relate the kernels by enforcing the temporal delays
as measured by the first-order temporal moments to be equal, 
then we obtain for the limit case when $K \rightarrow \infty$
\begin{equation}
  \label{eq-1mom-eq-rel}
  {\bar t} = \sqrt{\frac{c+1}{c-1}} \sqrt{\tau} = \delta \,  e^{\frac{3 \sigma ^2}{2}}.
\end{equation}
Solving the system of equations (\ref{eq-c2mom-eq-rel}) and
(\ref{eq-1mom-eq-rel}) then gives the following mappings between the
parameters in the two temporal scale-space models
\begin{equation}
  \label{eq-var-transf-par-tau-c-sigma-delta-explogdistr-scaletime-models}
  \left\{
    \begin{array}{l}
        \tau = \delta^2 \, e^{3 \sigma ^2} \left(e^{\sigma ^2}-1\right)  \\
      c = \frac{e^{\sigma ^2}}{2-e^{\sigma ^2}}
    \end{array}
  \right.
  \quad\quad
\left\{
    \begin{array}{l}
      \sigma = \sqrt{\log \left(\frac{2 c}{c+1}\right)}  \\
      \delta = \frac{(c+1)^2 \sqrt{\tau}}{2 \sqrt{2} \sqrt{(c-1) c^3}}
    \end{array}
  \right.
\end{equation}
which hold as long as $c > 1$ and $\sigma < \sqrt{\log 2} \approx 0.832$. 
Specifically, for small values of $\sigma$ a series expansion of the
relations to the left gives
\begin{equation}
  \left\{
    \begin{array}{l}
      \tau = \delta^2 \sigma^2 
                 \left( 
                     1 + \frac{7 \sigma^2}{2} + \frac{37 \sigma^4}{6}
                     + \frac{175 \sigma^6}{24} + {\cal O}(\sigma^8)
                 \right), \\
       c = 1 + 2 \sigma^2 + 3 \sigma^4 + \frac{13 \sigma^6}{3} + {\cal O}(\sigma^8).
   \end{array}
   \right.
\end{equation}
If we additionally reparameterize the distribution parameter $c$ such
that $c = 2^a$ for some $a > 0$ and perform a series expansion, we obtain
\begin{equation}
  a = \frac{\sigma ^2-\log \left(2-e^{\sigma ^2}\right)}{\log (2)}
     = \frac{2 \sigma^2}{\log 2}
        \left( 
           1 + \frac{\sigma ^2}{2} + \frac{\sigma ^4}{2}
           + \frac{13 \sigma ^6}{24} + {\cal O}(\sigma^8)
        \right)
\end{equation}
and with $b = a \log 2$ to simplify the following expressions
\begin{equation}
  \left\{
    \begin{array}{l}
      \sigma 
      = \frac{\sqrt{b}}{\sqrt{2}}
          \left(
             1-\frac{b}{8}-\frac{b^2}{128}+\frac{13 b^3}{3072}+\frac{49 b^4}{98304}
            +{\cal O}(b^5)
          \right),\\
      \delta 
      = \frac{\sqrt{2} \sqrt{\tau}}{\sqrt{b}}
      \left(
      1-\frac{3 b}{4}+\frac{49 b^2}{96}-\frac{31 b^3}{128}+\frac{959
   b^4}{10240}+{\cal O}(b^5)
      \right).
   \end{array}
   \right.
\end{equation}
These expressions relate the
parameters in the two temporal scale-space models in the limit case
when the number of temporal scale levels tends to infinity for the
time-causal model based on first-order integrators coupled in cascade
and with a logarithmic distribution of the intermediate temporal scale levels.

For a general finite number of $K$, the corresponding relation to
(\ref{eq-1mom-eq-rel}) that identifies the first-order temporal moments does
instead read
\begin{equation}
  \label{eq-1mom-eq-rel-general-K}
  {\bar t} 
  =\frac{c^{-K} \left(c^2-\left(\sqrt{c^2-1}+1\right) c+\sqrt{c^2-1} \, c^K\right)}{c-1} \, \sqrt{\tau }
  = \delta \,  e^{\frac{3 \sigma ^2}{2}}.
\end{equation}
Solving the system of equations  (\ref{eq-c2mom-eq-rel}) and
(\ref{eq-1mom-eq-rel-general-K}) then gives
\begin{equation}
  \label{eq-var-transf-par-tau-c-sigma-delta-explogdistr-scaletime-models-finite-K}
\left\{
    \begin{array}{l}
      \sigma = \sqrt{\log\left(\frac{A}{B}\right)} \\
      \delta = 
         \frac{C \sqrt{\tau}} 
                {2 \sqrt{2} (c-1) c^K \left( \frac{D}{E} \right)^{3/2}}
    \end{array}
  \right.
\end{equation}
where
\begin{align}
  \begin{split}
    A & = 2 c 
              \left( \vphantom{\sqrt{c^2-1}}
                  c^{4 K}-4 c^{K+2}-4 c^{K+3}+3 c^{2 K+3}
              \right.
 \end{split}\nonumber\\
   \begin{split}
    & \phantom{= 2 c \left( \right.}
        \left.
              -3 c^{3 K+2}+c^{4 K+1}+2 c^3
              \right.
  \end{split}\nonumber\\
   \begin{split}
     & \phantom{= 2 c \left( \right.}
        \left.
+\left(\sqrt{c^2-1}-1\right) c^{3 K}-\left(\sqrt{c^2-1}-4\right) c^{2
   K+1}
      \right.
  \end{split}\nonumber\\
   \begin{split}
     & \phantom{= 2 c \left( \right.}
        \left.
       +\left(\sqrt{c^2-1}+5\right) c^{2 K+2}-\left(\sqrt{c^2-1}+4\right) c^{3
   K+1}
       \right),
  \end{split}\\
  \begin{split}
    B & = \left(c^{2 K}-2 c^{K+1}-2 c^{K+2}+c^{2 K+1}+2 c^2\right)^2,
  \end{split}\\
   \begin{split}
     C & = \left(c^2-\left(\sqrt{c^2-1}+1\right) c+\sqrt{c^2-1} c^K\right),
  \end{split}\\
  \begin{split}
    D & = c 
              \left( \vphantom{\sqrt{c^2-1}}
             c^{4 K}-4 c^{K+2}-4 c^{K+3}+3 c^{2 K+3}-3 c^{3 K+2}+c^{4 K+1}+2
   c^3
              \right.
  \end{split}\nonumber\\
 \begin{split}
      & \phantom{= c \left( \right.}
         \left.
+\left(\sqrt{c^2-1}-1\right) c^{3 K}-\left(\sqrt{c^2-1}-4\right) c^{2
  K+1}
         \right.
  \end{split}\nonumber\\
   \begin{split}
      & \phantom{= c \left( \right.}
         \left.
+\left(\sqrt{c^2-1}+5\right) c^{2 K+2}-\left(\sqrt{c^2-1}+4\right) c^{3
   K+1}\right),
  \end{split}\\
  \begin{split}
    E & = \left(c^{2 K}-2 c^{K+1}-2 c^{K+2}+c^{2 K+1}+2 c^2\right)^2.
  \end{split}
\end{align}
Unfortunately, it is harder to derive a closed-form expression for
$c$ as function of $\sigma$ for a general (non-infinite) value of $K$.

Figure~\ref{fig-trunc-exp-kernels-1D+scaletime} shows examples of
kernels from the two families generated for this mapping 
between the parameters in the two families of temporal smoothing
kernels
for the limit case (\ref{eq-var-transf-par-tau-c-sigma-delta-explogdistr-scaletime-models}) 
when the number of temporal scale levels tends to infinity.
As can be seen from the graphs, the kernels from the two families do
to a first approximation share qualitatively largely similar properties.
From a more detailed inspection, we can, however, note that the two
families of kernels differ more in their temporal
derivative responses in that: (i)~the temporal derivative responses are
lower and temporally more spread out (less peaky) in the time-causal
scale-space model based on first-order integrators coupled in cascade 
compared to Koenderink's scale-time model and 
(ii)~the temporal derivative responses are somewhat faster 
in the temporal scale-space model based on
first-order integrators coupled in cascade.

A side effect of this analysis is that if we take the liberty of
approximating the limit case of the time-causal kernels corresponding to a
logarithmic distribution of the intermediate scale levels by the
kernels in Koenderink's scale-time model with the parameters
determined such that the first- and second-order temporal moments are equal,
then we obtain the following approximate expression for the temporal
location of the maximum point of the limit kernel
\begin{equation}
  \label{eq-approx-temp-pos-time-caus-log-distr}
  t_{max} \approx \frac{(c+1)^2 \, \sqrt{\tau}}{2 \sqrt{2} \sqrt{(c-1) c^3}} = \delta.
\end{equation}
From the discussion above, it follows that this estimate can be
expected to be an overestimate of the temporal location of the maximum
point of our time-causal kernels. This overestimate
will, however, be better than the previously mentioned
overestimate in terms of the temporal mean.
For finite values of $K$ not corresponding to the limit case, we can
for higher accuracy alternatively estimate the position of the local
maximum from $\delta$ in 
(\ref{eq-var-transf-par-tau-c-sigma-delta-explogdistr-scaletime-models-finite-K}).


\begin{figure}[hbt]
  \begin{center}
    \begin{tabular}{c}
      {\small\em skewness $\gamma_1(c)$} \\
            \includegraphics[width=0.35\textwidth]{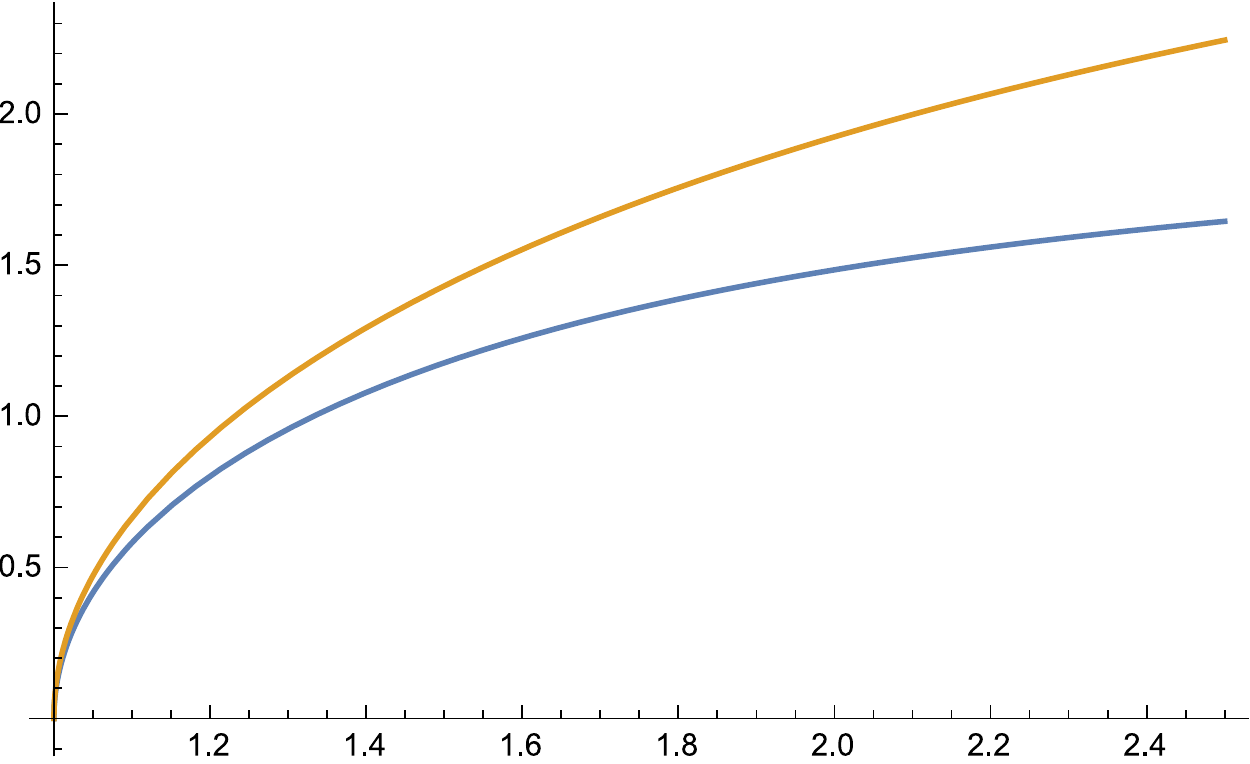}
            \\ \\
      {\small\em kurtosis $\gamma_2(c)$} \\
      \includegraphics[width=0.35\textwidth]{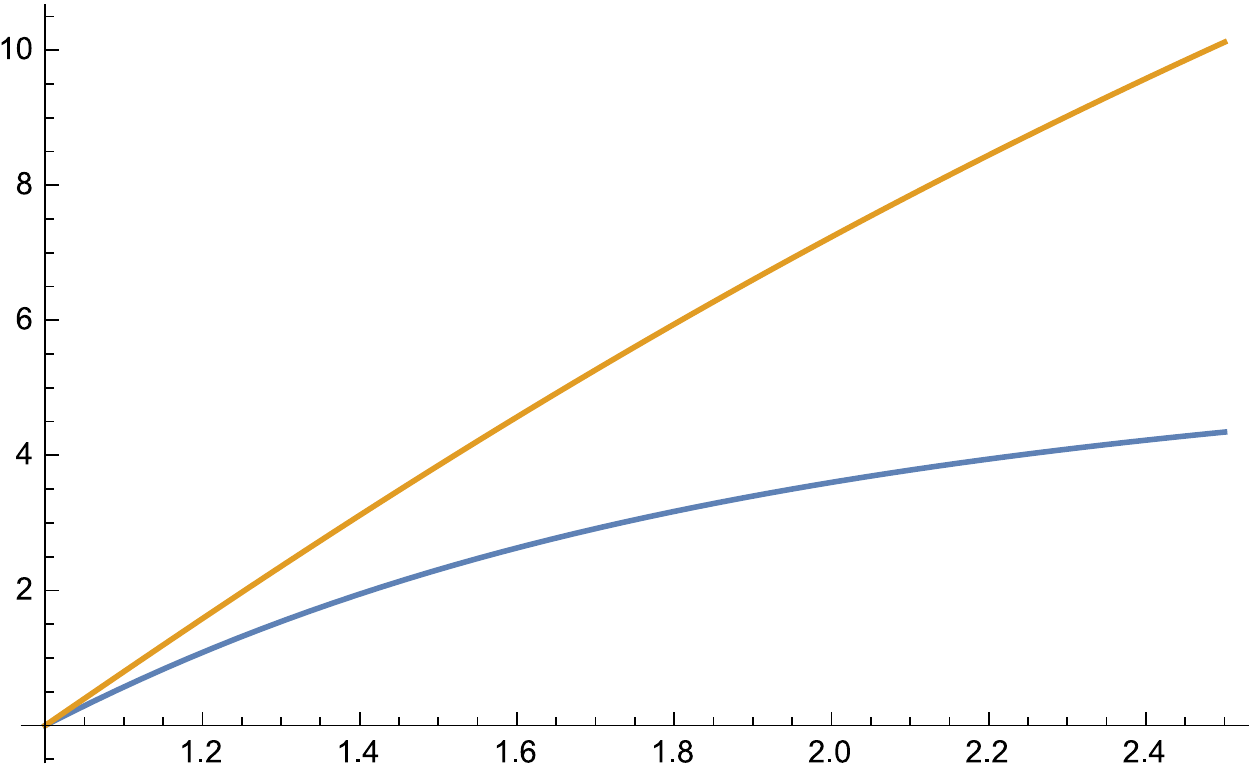} \\
    \end{tabular} 
  \end{center}

   \caption{Comparison between the skewness and the kurtosis measures for
     the time-causal kernels corresponding to the limit case of $K$ first-order
     integrators coupled in cascade when the number of temporal scale levels
     $K$ tends to infinity (blue curves) and the
     corresponding temporal kernels in Koenderink's scale-time model
     (brown curves) with the parameter values determined such that the first- and
     second-order temporal moments are equal 
    (equation~(\protect\ref{eq-var-transf-par-tau-c-sigma-delta-explogdistr-scaletime-models})).}
  \label{fig-skew-kurt-compare-truncexpcasc-scaletime}
\end{figure}

Figure~\ref{fig-skew-kurt-compare-truncexpcasc-scaletime} shows an
additional quantification of the differences between these two classes
of temporal smoothing kernels by showing how the
skewness and the kurtosis measures vary as function of the distribution
parameter $c$ for the same mapping 
(\ref{eq-var-transf-par-tau-c-sigma-delta-explogdistr-scaletime-models})  
between the parameters in the two families of temporal smoothing kernels.
As can be seen from the graphs, both the skewness and the
kurtosis measures are higher for the kernels in Koenderink's
scale-time model compared to our time-causal kernels corresponding to
first-order integrators coupled in cascade and do in these respect
correspond to a larger deviation from a Gaussian behaviour over the
temporal domain. (Recall that for a purely Gaussian temporal model all the cumulants of higher order
than two are zero, including the skewness and the kurtosis measures.)

\section{Scale invariance and covariance of scale-normalized temporal derivatives
  based on the limit kernel}
\label{app-sc-inv-sc-norm-temp-der-limit-kern}

In this appendix we will show that in the special case when the
temporal scale-space concept is given by convolution with the limit
kernel according to (\ref{eq-temp-scsp-conv-limit-kernel}) and
(\ref{eq-FT-comp-kern-log-distr-limit}), the corresponding
scale-normalized derivatives by either variance-based normalization
(\ref{eq-sc-norm-der-var-norm}) or $L_p$-normalization
(\ref{eq-sc-norm-der-Lp-norm-1}) are perfectly scale invariant for
temporal scaling transformations with temporal scaling factors $S$
that are integer powers of the distribution parameter $c$.
As a pre-requisite for this result, we start by deriving the transformation
property of scale-normalized derivatives by $L_p$-normalization
(\ref{eq-sc-norm-der-Lp-norm-1}) under temporal scaling
transformations.

\subsection{Transformation property of $L_p$-norms of scale-normalized
  temporal derivative kernels under
  temporal scaling transformations}

By differentiating the transformation property
(\ref{eq-sc-transf-limit-kernel}) of the limit kernel under scaling
transformations for $S = c^j$
\begin{equation}
   \Psi(t;\; \tau, c) = c^j \Psi(c^j \, t;\; c^2 j \tau, c) 
\end{equation}
we obtain
\begin{equation}
  \label{eq-sc-transf-limit-kernel-nth-order-der}
   \Psi_{t^n}(t;\; \tau, c) = c^j c^{nj} \Psi_{t^n}(c^j \, t;\; c^2 j
   \tau, c) 
= c^{j(n+1)} \Psi_{t^n}(c^j \, t;\; c^2 j \tau, c) .
\end{equation}
The $L_p$-norm of the $n$:th-order derivative of the limit kernel at
temporal scale $\tau = c^{2j}$ 
\begin{equation}
  \| \Psi_{t^n}(\cdot;\; c^2j, c) \|_p^p
  = \int_{u=0}^{\infty}
          \left| \psi_{t^n}(u;\; c^2j, c) \right|^p
\end{equation}
can then by the change of variables $u = c^j z$ with $du = c^j dz$ and using the
transformation property
(\ref{eq-sc-transf-limit-kernel-nth-order-der})
be transformed to the
$L_p$-norm at temporal scale $\tau = 1$ according to
\begin{align}
  \begin{split}
     \| \psi_{t^n}(\cdot;\; c^2j, c) \|_p^p
     & = \left(
               \int_{z=0}^{\infty}
                 \left|
                     c^{-j(n+1)} 
                     \Psi_{t^n}(z;\; 1, c)
                 \right|^p
               dz
            \right)
            c^j
  \end{split}\nonumber\\
  \begin{split}
     & = c^{-j(n+1)p+j} \| \psi_{t^n}(\cdot;\; 1, c) \|_p^p
  \end{split}
\end{align}
thus implying the following transformation property over scale
\begin{equation}
  \| \psi_{t^n}(\cdot;\; c^2j, c) \|_p 
  = c^{-j(n+1)+j/p}
      \| \psi_{t^n}(\cdot;\; 1, c) \|_p.
\end{equation}
Thereby, the scale normalization factors for temporal derivatives
in equation (\ref{eq-sc-norm-der-Lp-norm-2}) 
\begin{align}
  \begin{split}
     \alpha_{n,\gamma}(c^{2j}) 
     & = \frac{G_{n,\gamma}}{\| h_{t^n}(\cdot;\; c^{2j}) \|_p }
     = c^{j(n+1)-j/p} \, \frac{G_{n,\gamma}}{\| \psi_{t^n}(\cdot;\; 1, c) \|_p}
  \end{split}\nonumber\\
  \begin{split}
     \label{eq-sc-norm-factor-Lp-norm-limit-kernel}
     & = c^{j(n+1)-j/p} \, N_{n,\gamma}
  \end{split}
\end{align}
evolve in a similar way over temporal scales as the scaling factors of
variance-based normalization (\ref{eq-sc-norm-der-var-norm}) 
for $\tau = c^{2j}$ 
\begin{equation}
  \label{eq-sc-norm-factor-var-norm-limit-kernel}
  \tau^{n \gamma/2} = c^{jn\gamma}
\end{equation}
if and only if 
\begin{equation}
  \label{eq-rel-p-gamma-temp-scsp-limit-kernel}
  p = \frac{1}{1 + n(1-\gamma)}.
\end{equation}

\subsection{Transformation property of scale-normalized temporal
  derivatives under temporal scaling transformations}

Consider two signals $f$ and $f'$ that are related by a temporal
scaling transform $f'(t') = f(t)$ for $t' = c^{j'-j} t$ 
according to (\ref{eq-scale-cov-limit-kernel-scsp})
\begin{equation}
  \label{eq-scale-cov-limit-kernel-scsp-app}
  L'(t';\; \tau', c) = L(t;\; \tau, c)
\end{equation} 
between corresponding temporal scale levels $\tau' = c^{2(j'-j)} \tau$.
By differentiating (\ref{eq-scale-cov-limit-kernel-scsp-app}) and with
$\partial_t = c^{j'-j} \partial_{t'}$ we obtain
\begin{equation}
  c^{n(j' - j)} L_{t'^n}(t';\; \tau', c) = L_{t^n}(t;\; \tau, c).
\end{equation}
Specifically, for any temporal scales $\tau' = c^{2j'}$ and $\tau =
c^{2j}$ we have
\begin{equation}
  \label{eq-sc-cov-sc-norm-limit-ker-der-p1-gamma1-spec}
  c^{n j'} L_{t'^n}(t';\; c^{2j'}, c) = c^{n j} L_{t^n}(t;\; c^{2j}, c).
\end{equation}
This implies that {\em for the temporal scale-space concept defined by
convolution with the limit kernel, scale-normalized derivatives computed with scale normalization
factors defined by either $L_p$-normalization
(\ref{eq-sc-norm-factor-Lp-norm-limit-kernel}) for $p = 1$ or
variance-based normalization
(\ref{eq-sc-norm-factor-var-norm-limit-kernel}) for $\gamma = 1$
will be equal\/}
\begin{equation}
  L'_{\zeta'^n}(t';\; \tau', c) = L_{\zeta^n}(t;\; \tau, c)
\end{equation}
{\em between matching scale levels under temporal scaling transformations
with temporal scaling factors $S =c^{j'-j}$ that are integer powers of
the distribution parameter $c$.}

More generally, for $L_P$-normalization for any value of $p$ with a
corresponding $\gamma$-value according to
(\ref{eq-rel-p-gamma-temp-scsp-limit-kernel}) it holds that
\begin{align}
  \begin{split}
     L'_{\zeta'^n}(t';\, \tau', c)
     & = \alpha_{n,\gamma}(\tau') \, L'_{t'^n}(t';\; \tau', c)
        = \left\{
          \mbox{eq. (\ref{eq-sc-norm-factor-Lp-norm-limit-kernel})} \right\} 
  \end{split}\nonumber\\
  \begin{split}
    & = c^{j'(n+1)-j'/p} \, N_{n,\gamma} \, L'_{t'^n}(t';\, c^{nj'}, c)
       = \left\{
          \mbox{eq. (\ref{eq-rel-p-gamma-temp-scsp-limit-kernel})} \right\} 
  \end{split}\nonumber\\
  \begin{split}
     & = c^{j'n\gamma} \, N_{n,\gamma} \, L'_{t'^n}(t';\, c^{nj'}, c)
  \end{split}\nonumber\\
  \begin{split}
     & = c^{j'n (\gamma-1)} \, N_{n,\gamma} \, c^{j'n} L'_{t'^n}(t';\, c^{nj'}, c)
     = \left\{
          \mbox{eq. (\ref{eq-sc-cov-sc-norm-limit-ker-der-p1-gamma1-spec})} \right\} 
  \end{split}\nonumber\\
  \begin{split}
     & = c^{j'n (\gamma-1)} \, N_{n,\gamma} \, c^{jn} L_{t^n}(t;\, c^{nj}, c)
  \end{split}\nonumber\\
  \begin{split}
     & = c^{(j'-j)n (\gamma-1)} \, c^{jn\gamma} \, N_{n,\gamma} \, L_{t^n}(t;\, c^{nj}, c) 
  \end{split}\nonumber\\
  \begin{split}
     & = c^{(j'-j)n (\gamma-1)} \, c^{j(n+1)-j/p} \, N_{n,\gamma} \, L_{t^n}(t;\, c^{nj}, c) 
  \end{split}\nonumber\\
  \begin{split}
      & = c^{(j'-j)n (\gamma-1)} \, \alpha_{n,\gamma}(\tau) \, L_{t^n}(t;\; \tau, c) 
  \end{split}\nonumber\\
  \begin{split}
     & = c^{(j'-j)n (\gamma-1)} \, L_{\zeta^n}(t;\, \tau, c)
= \left\{
          \mbox{eq. (\ref{eq-rel-p-gamma-temp-scsp-limit-kernel})} \right\} 
  \end{split}\nonumber\\
  \begin{split}
     & = c^{(j'-j) (1 - 1/p)} \, L_{\zeta^n}(t;\, \tau, c)
  \end{split}
\end{align}
In the proof above, we have for the purpose of calculations related
the evolution properties over scale relative to the temporal scale
$\tau = 1$ and normalized the relative strengths between temporal
derivatives of different order to the corresponding strengths
$G_{n,\gamma}$ of $L_p$-norms of Gaussian derivates. 
These assumptions are however not essential for the scaling properties
and corresponding 
scaling transformations can be derived relative to any other temporal
base level $\tau_0$ as well as for other ways of normalizing the
relative strengths of scale-normalized derivatives between different
orders $n$ and distribution parameters $c$.

{\footnotesize
\bibliographystyle{spmpsci}
\bibliography{defs,tlmac}}

\end{document}